\documentclass[10pt,journal,compsoc]{IEEEtran}
\pdfoutput=1
\newif\ifpeerreview

\peerreviewtrue

\usepackage{algorithm}
\usepackage{algpseudocode}
\usepackage{amsmath,amssymb}
\usepackage{booktabs}
\usepackage{caption}
\usepackage{comment}
\usepackage{graphicx}
\usepackage{hyperref}
\usepackage{hyphenat}
\usepackage[switch]{lineno}
\usepackage{multirow}
\usepackage[numbers,sort&compress]{natbib}
\usepackage{overpic}
\usepackage{printlen}
\usepackage{subcaption}
\usepackage{url}
\usepackage[usenames, dvipsnames]{xcolor}
\usepackage{listings}

\usepackage[capitalize]{cleveref} %
\crefname{section}{Sec.}{Secs.}
\Crefname{section}{Section}{Sections}
\crefname{table}{Tab.}{Tabs.}
\Crefname{table}{Table}{Tables}

\usepackage{pifont}
\newcommand{\cmark}{\ding{51}}
\newcommand{\xmark}{\ding{55}}

\definecolor{uoftblue}{RGB}{30, 55, 101} %

\definecolor{uoftsecondaryblue}{RGB}{0,127,163} %

\definecolor{uoftpurple}{RGB}{109,36,122} %
\definecolor{uoftwarmred}{RGB}{220,70,51} %
\definecolor{uoftcoolblue}{RGB}{111,199,234} %
\definecolor{uoftteal}{RGB}{0,161,137} %
\definecolor{uoftfuchsia}{RGB}{171,19,104} %
\definecolor{uoftdarkgreen}{RGB}{13,83,77} %
\definecolor{uoftyellow}{RGB}{241,197,0} %
\definecolor{uoftlightgreen}{RGB}{141,191,46} %

\definecolor{uoftcoolgray}{RGB}{208,209,201} %

\hypersetup{colorlinks,allcolors=uoftsecondaryblue}

\newcommand{\psnrssim}[2]{#1 / #2}
\newcommand{\ssim}[2]{#2}
\newcommand{\psnr}[2]{#1}

\newcommand{\raw}{raw} %
\newcommand{\newSIDD}{SIDD-CC}
\newcommand{\etal}{\textit{et al.}}

\newcommand{\blem}{\mathcal{E}_{\text{bl}}}
\newcommand{\ble}{\mathcal{M}_{\text{bl}}}
\newcommand{\bl}{{\ell}_{\text{bl}}}
\newcommand{\wl}{{\ell}_{\text{wl}}}

\title{Why Low-Light Cameras Go Color Blind: Removing Color Bias in Raw
Denoising}

\author{
Mohammad~Mohammadi$^{1,2,3}$,
Sina~Honari$^{3}$,
Stavros~Tsogkas$^{3}$,
Tristan~Aumentado-Armstrong$^{3}$,
Michael~S.~Brown$^{3,4}$,
Iqbal~Mohomed$^{3}$,
Konstantinos~G.~Derpanis$^{1,2,3,4}$,
Alex~Levinshtein$^{3,*}$,
and~Igor~Gilitschenski$^{1,2,*}$
\thanks{$^{1}$University of Toronto, $^{2}$Vector Institute, $^{3}$AI-Center Toronto, Samsung Electronics, $^{4}$York University. $^*$Joint Supervision. E-mail: \{mohammadi, gilitschenski\}@cs.toronto.edu;
kosta@yorku.ca;
\{sina.honari, stavros.t, tristan.a, michael.b1, i.mohomed, alex.lev\}@samsung.com.
}
}

\begin{document}

\IEEEtitleabstractindextext{%
\vspace{-1pt}
\newcommand{\insertfig}{
\begin{center}\setcounter{figure}{0}
\vspace{-5mm}
\includegraphics[width=1.0\linewidth]{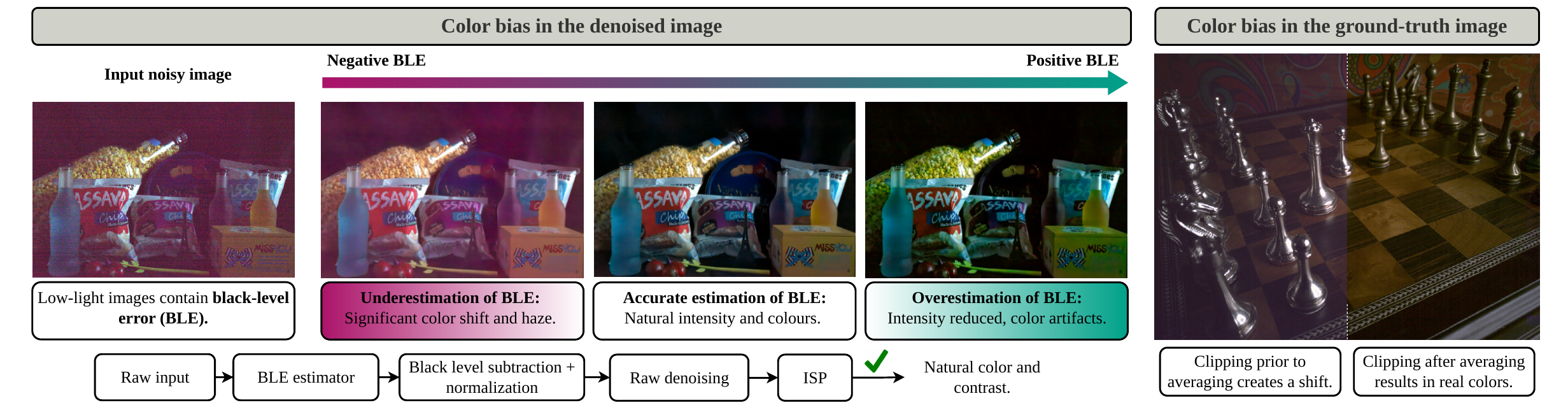}
\vspace{-16pt}
\captionof{figure}{
\textbf{Identification and mitigation of color bias (CB) in low-light denoising.}
\textit{Left:} CB induced by black-level error (BLE) in the denoising pipeline.
\textit{Right:} CB due to \raw{} processing of GT images, based on multi-frame fusion.
\textit{Bottom:} Diagram of our proposed CB-correcting denoising method, including the \raw{} BLE estimator.
}
\label{fig:teaser}
\end{center}
\vspace{9pt}
}
\makeatletter
\insertfig

\begin{abstract}
Raw images inherently suffer from noise due to the stochastic nature of light and sensor hardware imperfections. 
As real photon counts fall, the ratio of this noise to the signal degrades; consequently, for low-light conditions, robust denoising is especially vital for high-quality results. 
While recent data-driven methods achieve strong performance, they typically rely on large-scale noisy-clean image pairs that are costly and difficult to collect. 
Alternatively, parametric noise models can generate synthetic training data, but this necessitates precise camera calibration, which is often impractical for unknown devices.
In this work, we propose a camera-agnostic, calibration-free paradigm for low-light \raw{} denoising.
We identify that color bias from black-level error is a primary source of performance degradation and causes severe color shifts. 
To mitigate this, we introduce a bias estimator network that predicts the black-level error as a global feature of the noisy input. 
We evaluate our approach across the ELD, SID, and LRID datasets, demonstrating superior performance among blind denoisers, particularly in terms of color correction. In many cases, we are competitive with---or can even surpass---methods with stronger supervision. 
Furthermore, we reveal that the widely used SIDD dataset contains significant color bias in its ground-truth images, which yields unrealistic color reproduction in trained models. 
We introduce a new ground-truth extraction framework to resolve this issue and provide a benchmark of existing methods on the corrected dataset. 
\end{abstract}

}

\maketitle

\IEEEraisesectionheading{
  \section{Introduction}\label{sec:introduction}
}
\label{sec:intro}

\IEEEPARstart{R}aw image formation in digital cameras~\cite{maitre2017photon} is characterized by noise arising from both the stochastic nature of photons arriving on the camera sensor and the subsequent hardware processes of electron-to-voltage and analog-to-digital conversion. 
These effects become particularly pronounced in low-light conditions, where photon scarcity necessitates higher ISO settings, which further amplify noise.

To fix such degradations, low-light \raw{} image denoisers~\cite{gharbi2016deep, zhang2023practical, liu2023joint, qian2022rethinking, jin2023dnf} are essential; yet, their strong performance often relies on having synthetic noise with matched distributions (no domain gap)~\cite{zhang2023practical, kim2025idf, jiang2022fast, zhang2021plug, zamir2022restormer, chen2022simple}.
Importantly, such models often fail to generalize to real-world scenarios, where noise characteristics are more complex and camera-dependent. A common approach to address this gap is to collect paired clean-noisy images from the target camera~\cite{abdelhamed2019noise, lu2025dark, cao2023physics, zhang2023towards, zou2025calibration}, enabling supervised training. Alternatively, one can physically calibrate camera-specific noise parameters~\cite{zhang2021rethinking, wei2021physics, feng2022learnability, jiang2025msfa, feng2026learning, lu20252}, enabling realistic noise synthesis for training a dedicated denoiser. Both approaches, however, impose substantial practical burdens for each new camera, requiring either labor-intensive paired data acquisition or careful per-device calibration.

Unsupervised denoising approaches~\cite{lehtinen2018noise2noise, batson2019noise2self, song2021noise2void, kim2021noise2score, xie2020noise2same, wang2023noise2info, quan2020self2self, wang2022blind2unblind} alleviate part of this burden by requiring only noisy images from the target camera. 
There has also been the ``zero-shot'' extension of these models, in which a denoiser is trained on a single noisy image~\cite{liu2025zero, mansour2023zero}. 
Most existing methods, however, are developed for standard illumination conditions in the sRGB domain and typically assume pixel-independent~\cite{lehtinen2018noise2noise, batson2019noise2self, song2021noise2void, kim2021noise2score, xie2020noise2same, wang2023noise2info, liu2025zero, mansour2023zero}, zero-mean noise~\cite{lehtinen2018noise2noise, batson2019noise2self, song2021noise2void, kim2021noise2score, xie2020noise2same, liu2025zero, mansour2023zero, lee2022ap, neshatavar2022cvf}.
Moreover, these methods still require retraining or adapting the trained model for each target noise distribution.

In contrast, YOND~\cite{feng2025yond} is a recent denoising framework that is both calibration-free and camera-agnostic. However, while YOND is effective under standard lighting conditions, it relies on parameter estimation techniques that struggle in low-light scenarios, where the complex noise distribution violates its simplifying assumptions~\cite{feng2025yond}. Eschewing camera-specific processing, 
while achieving strong low-light denoising performance, thus remains an open problem.

In this work, we advance the direction of blind (calibration-free, camera-agnostic) denoising, with a model capable of handling the complex noise characteristics of low-light \raw{} imagery. 
We particularly focus on the unnatural color tones and chromatic artifacts in denoised low-light images (especially in dark regions), a phenomenon we term \textit{color bias} (see Fig.~\ref{fig:teaser}), which severely impacts perceptual quality.
Through a systematic analysis of noise formulations and denoising strategies, we identify \textit{black-level error} (BLE) as a primary source of color bias. 
BLE measures the discrepancy between the recorded metadata black-level and the ideal one for that sensor.
To mitigate it, we therefore introduce a black-level bias estimator (BLBE) module, which corrects the BLE in the \raw{} noisy input image \textit{before} any other processing.
This corrected image is subsequently denoised by a generic, camera-agnostic model trained specifically for low-light conditions.
We show that this decoupled approach is far more effective than attempting to have the denoiser directly address BLE; further, it provides state-of-the-art performance in blind denoising.
Surprisingly, %
it is often even competitive with 
camera-specific methods, 
likely because they do not explicitly handle BLE.

In addition, we find that color bias can be induced by its presence in ground-truth (GT) training images. Specifically, we identify color distortions in the GT of SIDD~\cite{abdelhamed2018high}, a widely used denoising benchmark. We therefore provide a \underline{c}olor-\underline{c}orrected version of it, denoted SIDD-CC, to the community, which shows markedly reduced color bias.
Finally, we benchmark several state-of-the-art denoisers on SIDD-CC, establishing a more reliable baseline for future research. %

Our main contributions are as follows:
\begin{itemize}
    \item We provide a systematic analysis of how different noise formulations and training design choices influence the emergence of color bias in denoising.
    \item We propose a method to mitigate color bias induced by BLE within a camera-agnostic, calibration-free denoising framework for low-light \raw{} images, demonstrating state-of-the-art performance on multiple low-light benchmarks in blind denoising.
    \item We uncover color bias in the GT of SIDD~\cite{abdelhamed2018high} and introduce a color-corrected version of the dataset (SIDD-CC), on which we %
    evaluate several denoisers as an initial benchmark
    for the community.
\end{itemize}

\section{Related Work}
\label{sec:related-work}

\noindent \textbf{Calibration-based noise synthesis.}
Noise can be synthesized by collecting data from target cameras and physically modeling its underlying components. In general, image noise is decomposed into signal-dependent (SD) and signal-independent (SI) sources. SD noise correlates with the number of photons accumulated at each pixel and is commonly modeled as a Poisson distribution with a calibrated gain term~\cite{zhang2021rethinking, wei2021physics, feng2022learnability}.
In contrast, SI noise is unrelated to incident photons, is more challenging to model, and becomes increasingly significant in low-light photography; in its simplest form, it is approximated by a Gaussian distribution~\cite{feng2025yond}.
For realistic low-light modeling, additional components such as dark shading noise, banding noise, and quantization effects must also be considered~\cite{wei2021physics, jiang2025msfa}. Fitting these parameters requires multiple calibration frames, including bias frames (no incident light) and flat-field frames (uniform illumination), per ISO. To fit these calibration frames more accurately for low light photography, one needs to consider exposure time, sensor temperature, aperture value, and lens and filter combination, in addition to ISO setting.

Several camera-specific calibration approaches bypass explicit modeling of SI noise via real captured dark frames~\cite{zhang2021rethinking, feng2022learnability, feng2026learning,li2025noise}. One recent method~\cite{lu20252} reduces this requirement to two captures per ISO: a dark frame and a noisy frame of a scene, to estimate SI and SD noise, respectively. Nevertheless, this approach is still camera-specific, requiring not only per-ISO calibration, but also data generation and retraining for every new camera.

\noindent \textbf{Learning-based noise synthesis.}
With sufficient paired data from a target camera, models can effectively learn to suppress noise~\cite{abdelhamed2019noise, lu2025dark, cao2023physics, zhang2023towards, zou2025calibration}. However, acquiring such datasets is costly and labor-intensive, motivating approaches that instead synthesize realistic target-domain noise (e.g., via
normalizing flows \cite{abdelhamed2019noise} or diffusion-based
generation \cite{lu2025dark}).

Calibration can also be combined with learning.
For instance, calibrated noise can be sampled via generative models~\cite{cao2023physics,zhang2023towards},
while Zou~\etal~\cite{zou2025calibration} leverage multiple calibrated source cameras to infer noise parameters for an uncalibrated target.
Despite these advances, such approaches still require calibration in either the source~\cite{zou2025calibration} or target domains~\cite{cao2023physics, zhang2023towards}. 
In addition, they rely on either paired data~\cite{abdelhamed2019noise, lu2025dark, cao2023physics, zhang2023towards} or collections of noisy images~\cite{zou2025calibration} from the target domain. 
More importantly, they necessitate domain-specific noisy image synthesis and denoiser training for \textit{every} new target camera.
In contrast, we focus on the ``blind'' scenario, 
without calibrations or target data.

\begin{figure*}[t]
\centering
\includegraphics[width=\textwidth]{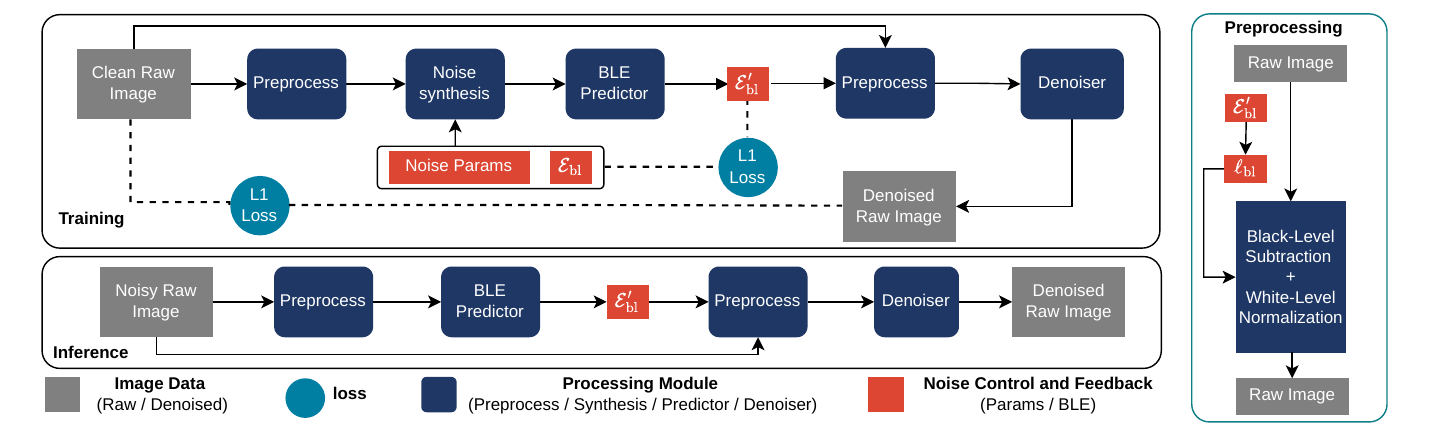}
\caption{\textbf{Training and inference pipeline.} During training (top), synthetic noise and black-level error (BLE), ${\mathcal{E}_{\text{bl}}}$, are added to the clean image, which is then passed to the BLE predictor. The estimated BLE, ${\mathcal{E}_{\text{bl}}}\text{'}$, is then used to update the black-level, $\ell_\text{bl}$, in the preprocessing before the denoising step (right). The model is trained end-to-end. For inference (bottom) on real noisy images, the BLE predictor is used to correct the black-level of the noisy \raw{} image prior to the denoising step.}
\label{fig:overview}
\end{figure*}

\noindent \textbf{Blind denoising.}
Our work considers \emph{blind denoising}, a variant of the denoising problem that does \textit{not} assume the availability of camera-specific calibration, paired clean-noisy data, or even collections of noisy images from the target device. For example, YOND~\cite{feng2025yond} adopts a coarse-to-fine framework with a variance-stabilization transform that normalizes the noise to approximately additive white Gaussian noise. At inference time, the method estimates the noise variance parameter and feeds it, together with the transformed noisy input, to the denoiser. However, this formulation is limited to modeling shot noise and a simple Gaussian read noise, which is insufficient for realistic low-light imaging. Moreover, the variance-stabilization step assumes zero-mean noise and neglects black-level offsets, reducing its applicability to more complex noise characteristics. In contrast, our approach accommodates richer noise models and is better suited to low-light photography.

\noindent \textbf{Color bias.}
In low-light photography, color bias is a prominent source of artifacts, often manifesting as purplish tones in dark regions or as chromatic distortions across the image. A primary contributor to this effect is dark-shading noise, which comprises fixed-pattern noise (FPN), black-level error (BLE), and current shot noise induced by thermal effects~\cite{cao2023physics}. FPN is spatially variant and can lead to a non-injective mapping between noisy and clean color patches~\cite{feng2022learnability}. 
The black-level represents the minimum sensor value interpreted as absolute black, below which pixel values are clipped; it is typically estimated from dark frames captured without incident light. Although this value mainly varies with ISO~\cite{feng2022learnability}, it also exhibits fluctuations at a fixed ISO~\cite{jiang2025msfa} due to exposure time and sensor temperature, making the black-level recorded by the camera's meta-data prone to error~\cite{wei2021physics}. In low-intensity conditions, such inaccuracies are further amplified by high sensor gains (e.g., $100\times$), resulting in pronounced color artifacts~\cite{wei2021physics}.

Several prior works~\cite{wei2021physics, cao2023physics} mitigate color bias in Bayer filter arrays by explicitly modeling dark-current noise, while others~\cite{jiang2025msfa} extend this modeling to multi-spectral filter arrays as a function of light intensity, image column, and spectral band. 
Dark-shading–induced color bias has also been addressed through a range of strategies, including physically calibrating and subtracting it from noisy observations during both training and inference~\cite{feng2022learnability, feng2026learning}, sampling dark frames directly from target cameras~\cite{zhang2021rethinking, li2025noise}, incorporating calibrated black-level error into Gaussian read-noise models~\cite{wei2021physics}, and leveraging paired real data from specific cameras to train either noise generators~\cite{zhang2023towards} or denoising networks~\cite{jin2023lighting}. 
Although effective at reducing color bias, these approaches rely on either physical calibration~\cite{wei2021physics, cao2023physics, jiang2025msfa, feng2022learnability, feng2026learning, zhang2021rethinking, li2025noise} or paired data from the target domain~\cite{zhang2023towards, jin2023lighting}. In contrast, our work aims to eliminate both requirements by addressing color bias in the blind denoising setting.

\section{Methodology}
\label{sec:method}

    Let  $\mathcal{I}=\{I_i\}_{i=1}^N$ be a dataset of clean \raw{} images $I_i\in\mathbb{R}^{h\times w}$. 
Our goal is to train a denoiser which, given a noisy \raw{} image, $D$, can produce an estimate $\hat{I}$ of the underlying clean image $I$, with $\{D, I, \hat{I}\} \in \mathbb{R}^{h\times w}$. 
$\mathcal{I}$ does not contain paired data and we do not assume that noisy images are captured with the same sensor as those in the training set.

Instead, we opt for synthetically generated noisy images, obtained by manually applying realistic degradations. 
These degradations follow a parametric noise model (\cref{sec:noise_model}) and \emph{color biases} induced by dark-shading noise in low-light scenarios, particularly due to black-level error -- BLE (\cref{sec:dark_shading_noise}). 
We find that simply adding such color biases in  our noise synthesis protocol does not resolve the color shift issues (\cref{sec:llnm}). 
Instead, we propose leveraging a learnable black-level bias estimator (BLBE), followed by a black-level correction step at test time, before applying our denoising model (\cref{sec:actualmethod}). 
Our full pipeline is shown in \cref{fig:overview}.

\subsection{Noise Formation Model}
\label{sec:noise_model}
In physics-based noise modeling for CMOS image sensors, the recorded image, $D$, is typically expressed as the sum of the underlying clean signal $I$ and an additive noise $N$:
\begin{equation}
D = I + N.
\end{equation}
Following standard noise formation models for low-light imaging~\cite{wei2021physics, jin2023lighting, cao2023physics}, we decompose the noise term as

\begin{equation}
    \label{eq:noise_sum}
    N = N_\text{shot} + N_\text{read} + N_\text{band} + N_{q},
\end{equation}
where the terms denote shot, read, banding-pattern, and quantization noise, respectively. 
$N_\text{shot}$ is signal-dependent, while the others constitute the signal-independent noise.  
We briefly describe each component below.

\textbf{Shot Noise.} 
When incident photons strike the photo-sensor, they generate photoelectrons whose count follows a Poisson distribution, denoted by $\mathcal{P}(\cdot)$, with a rate proportional to the underlying irradiance. Consequently, shot noise is signal-dependent, with an expected value equal to the clean signal. This process can be modeled as
\begin{equation}
    {K}^{-1}(I + N_\text{shot}) \sim \mathcal{P}\left(\frac{I}{K}\right),
\end{equation}
where $K$ denotes the overall system gain, defined as the ratio between digital intensity units and the number of detected photo-electrons.

\textbf{Read Noise.}
Signal-independent noise arises during the conversion of photoelectrons to voltage and subsequent sensor readout. 
It comprises multiple components, including thermal noise (e.g., reset noise and source-follower noise) and gain-dependent amplification noise. 
Zero-mean Gaussian noise is commonly used as an approximation~\cite{jiang2025msfa, feng2025yond}:
\begin{equation}
    \label{eq:n_gauss}
    N_\text{read} \sim \mathcal{N}(0,\,\sigma^{2}).
\end{equation}

\textbf{Banding-pattern Noise.}
Banding-pattern noise manifests as horizontal or vertical stripe artifacts caused by non-uniform high-speed readout, which introduces row- or column-wise variations in the output voltage. Since vertical banding is typically negligible in our setting, we focus on row-wise banding noise. This component can be modeled as a zero-mean Gaussian process in which all pixels within the same row share the same sampled noise realization:
\begin{equation}
    N_\text{band} \approx N_\text{row} \sim \mathcal{N}(0,\,\sigma_{\text{row}}^2).
\end{equation}

\textbf{Quantization Noise.}
During analog-to-digital conversion, the continuous voltage signal is discretized into finite digital levels. The resulting quantization error is commonly modeled as a zero-mean uniform random variable:
\begin{equation}
    N_{q} \sim {U}\left(-\frac{1}{2}q,\,\frac{1}{2}q\right),
\end{equation}
where $q$ denotes the quantization step size.

Together, these components form the basis of common noise synthesis configurations: PG (shot + read), PGR (+row), and PGRQ (+quantization). %

\subsection{Dark-shading Noise}
\label{sec:dark_shading_noise}
The noise model in Sec.\ \ref{sec:noise_model} assumes zero-mean noise. 
This assumption holds under normal lighting but breaks down at low light and high ISO, where dark-shading noise becomes non-negligible.
Dark-shading noise consists mainly of \textbf{fixed-pattern noise (FPN)} and \textbf{black-level error (BLE)}~\cite{feng2022learnability, feng2026learning, cao2023physics}. 
The black-level (BL) is a global offset corresponding to the sensor response under zero illumination; pixel values below this threshold are treated as noise and clipped. 
BLE arises from discrepancies between the true sensor BL and the value recorded in metadata or used during pre-processing. 
While BL mainly depends on ISO, other factors such as exposure time, and sensor temperature can cause errors, especially in low light photography.
Moreover, low-light denoising models are typically trained to map extremely low-intensity \raw{} values to properly exposed outputs, often under substantial amplification factors (e.g., $100\times$-$200\times$). 
Under such high gains, even minor inaccuracies in BL estimation can induce significant output shifts. 
These shifts manifest most prominently as chromatic artifacts in dark regions (i.e., black colors), where small offsets in low-intensity color channels are amplified into visible color distortions (see Fig.\ref{fig:teaser}).

Previous work has modeled BLE by changing the mean of the read noise distribution~\cite{wei2021physics, zou2025calibration}, which is equivalent to an additional independent uniform distribution, where all pixels in the image share the same sampled offset:

\begin{equation}
    \label{eq:ble}
    \blem %
    \sim {U}(-\ble,\, \ble).
\end{equation}
$\ble$ denotes the maximum magnitude of the BLE. 
This term captures the uncertainty in black-level by applying a global image-level shift from zero-mean noise.

FPN refers to spatially varying pixel responses under uniform illumination, whose impact on perceptual quality is particularly severe in low-light settings with low signal-to-noise ratio. Denoising it requires learning a non-injective mapping from heterogeneous noisy observations to a spatially homogeneous clean signal. In this work, we focus on correcting the global BLE component; we characterize dark-shading structure across sensors in \cref{sec:supp_dark_shading_analysis} and discuss a proof-of-concept extension to FPN in \cref{sec:supp_fpn}.

\begin{figure*}[t]
\centering

\subfloat[Noisy]{%
    \includegraphics[width=0.235\textwidth]{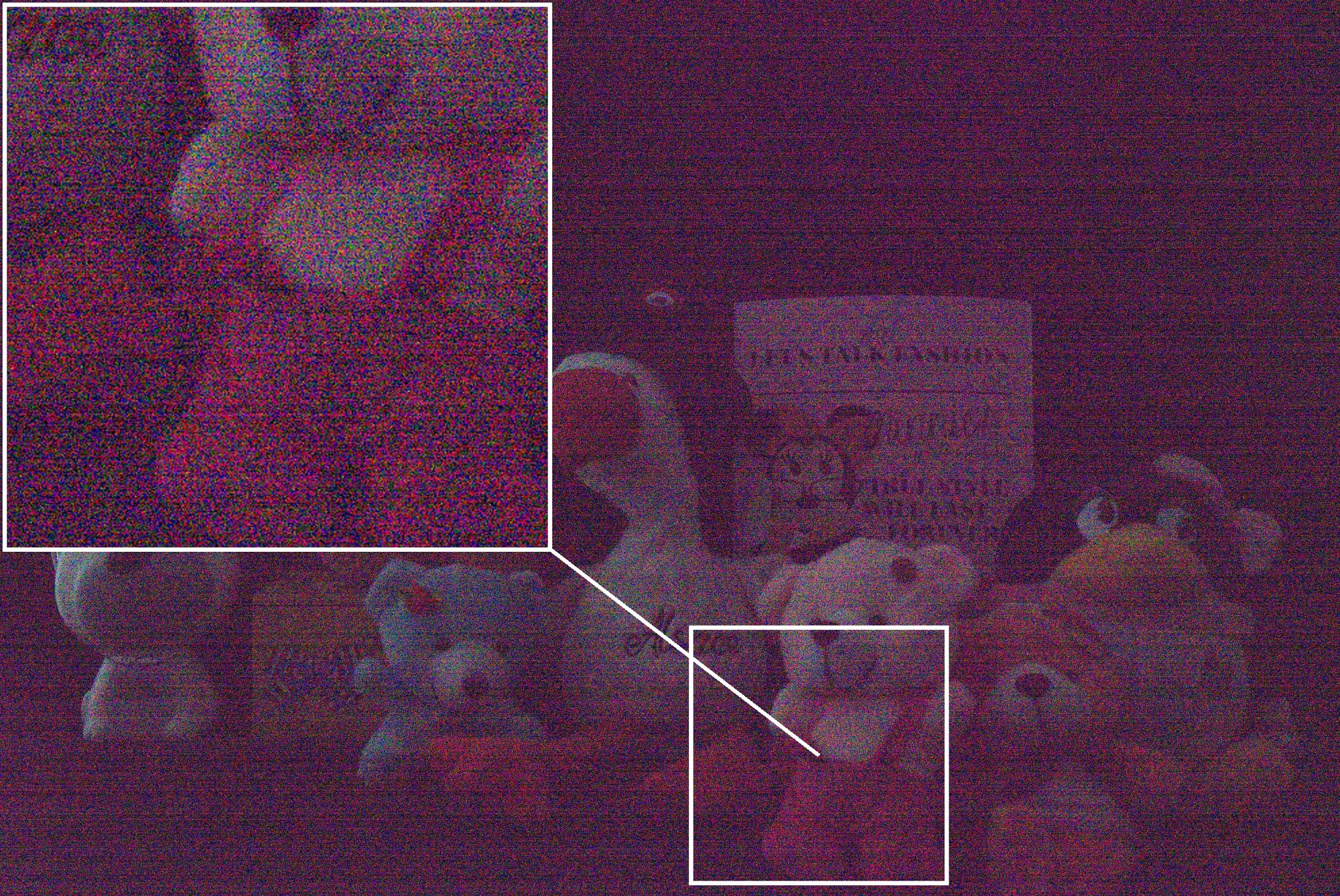}
    \label{fig:results_eld_sony_noisy}
}
\subfloat[PG]{%
    \includegraphics[width=0.235\textwidth]{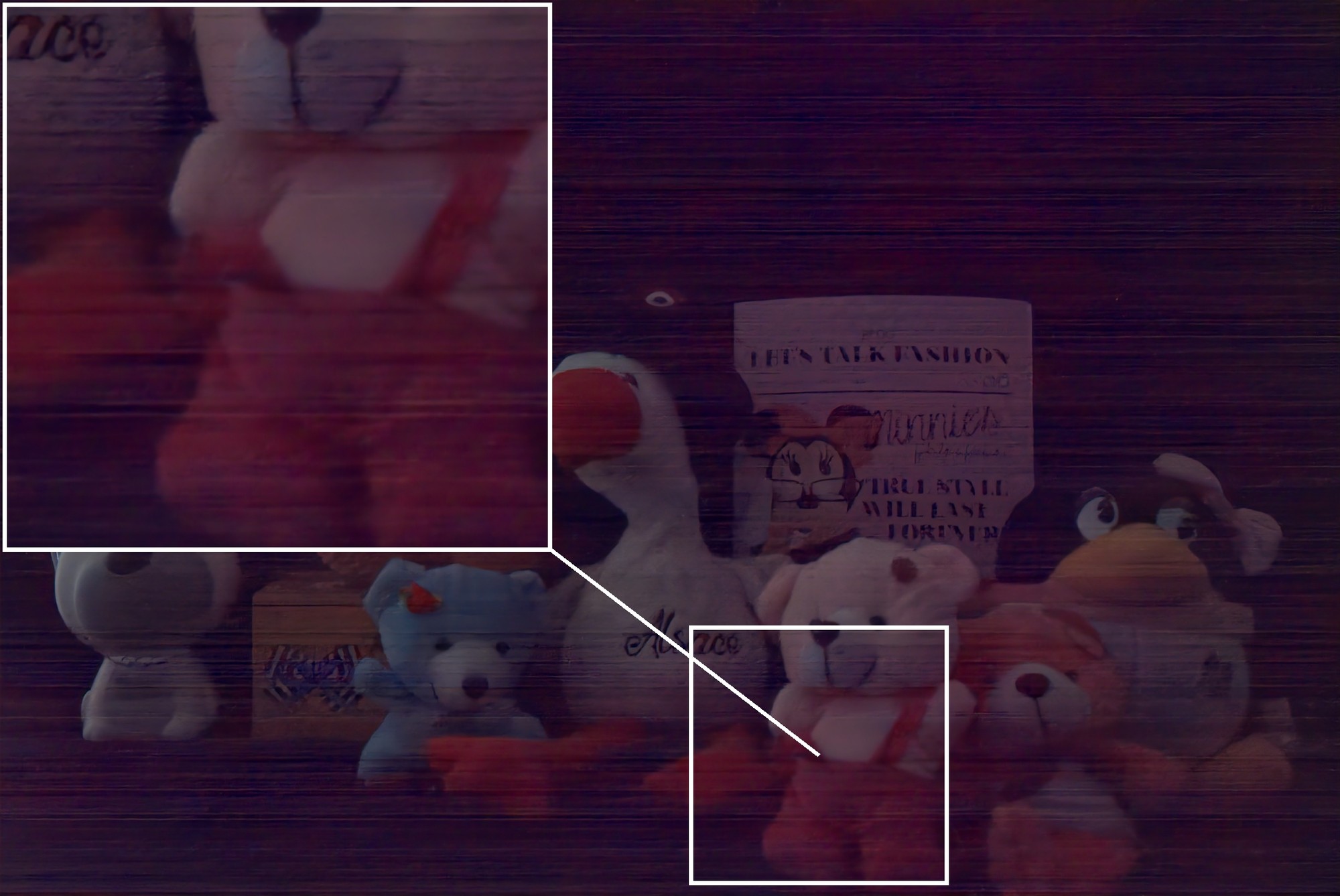}
    \label{fig:PG}
}
\subfloat[PGRQ]{%
    \includegraphics[width=0.235\textwidth]{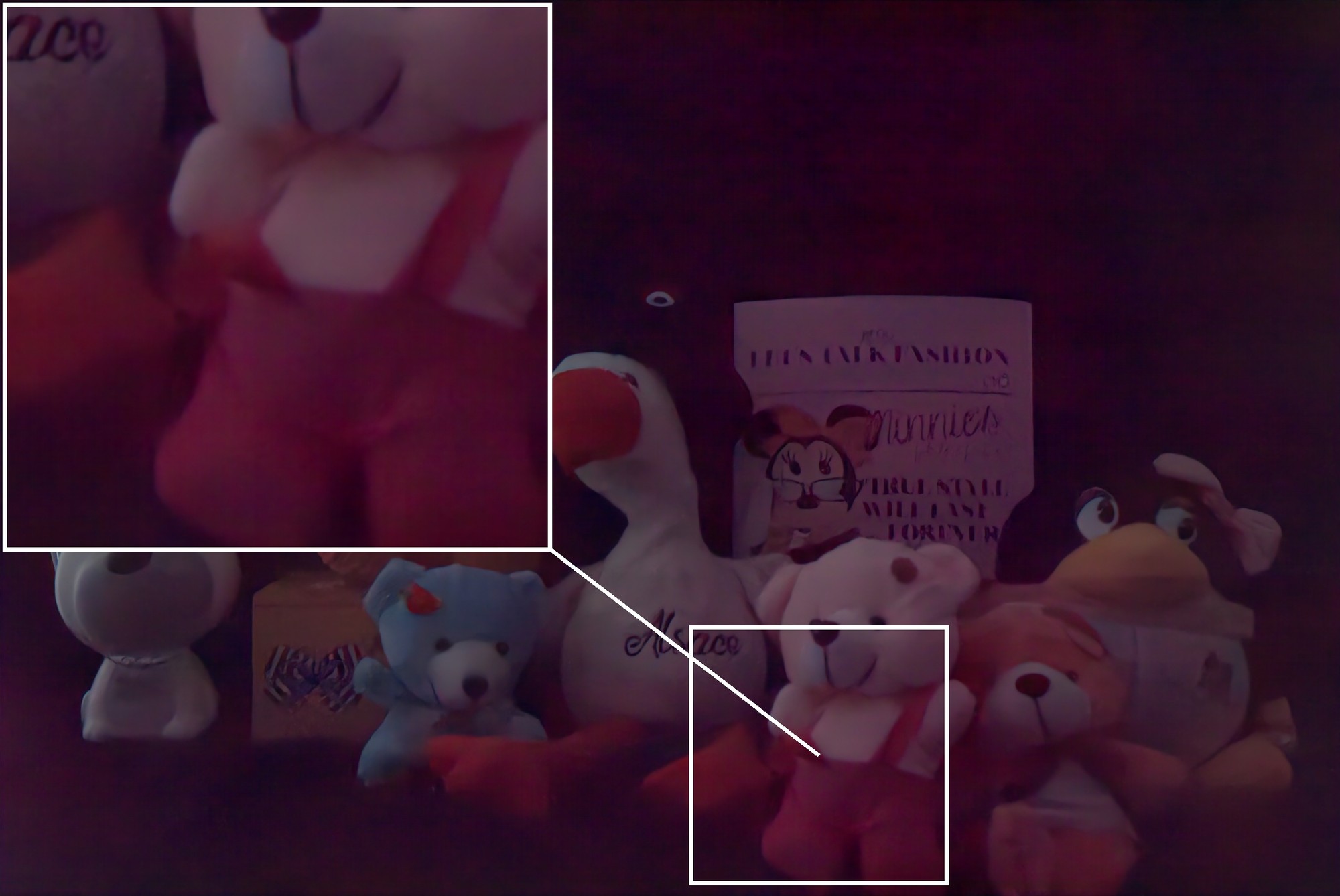}
    \label{fig:PGRQ}
}
\subfloat[PGRQB]{%
    \includegraphics[width=0.235\textwidth]{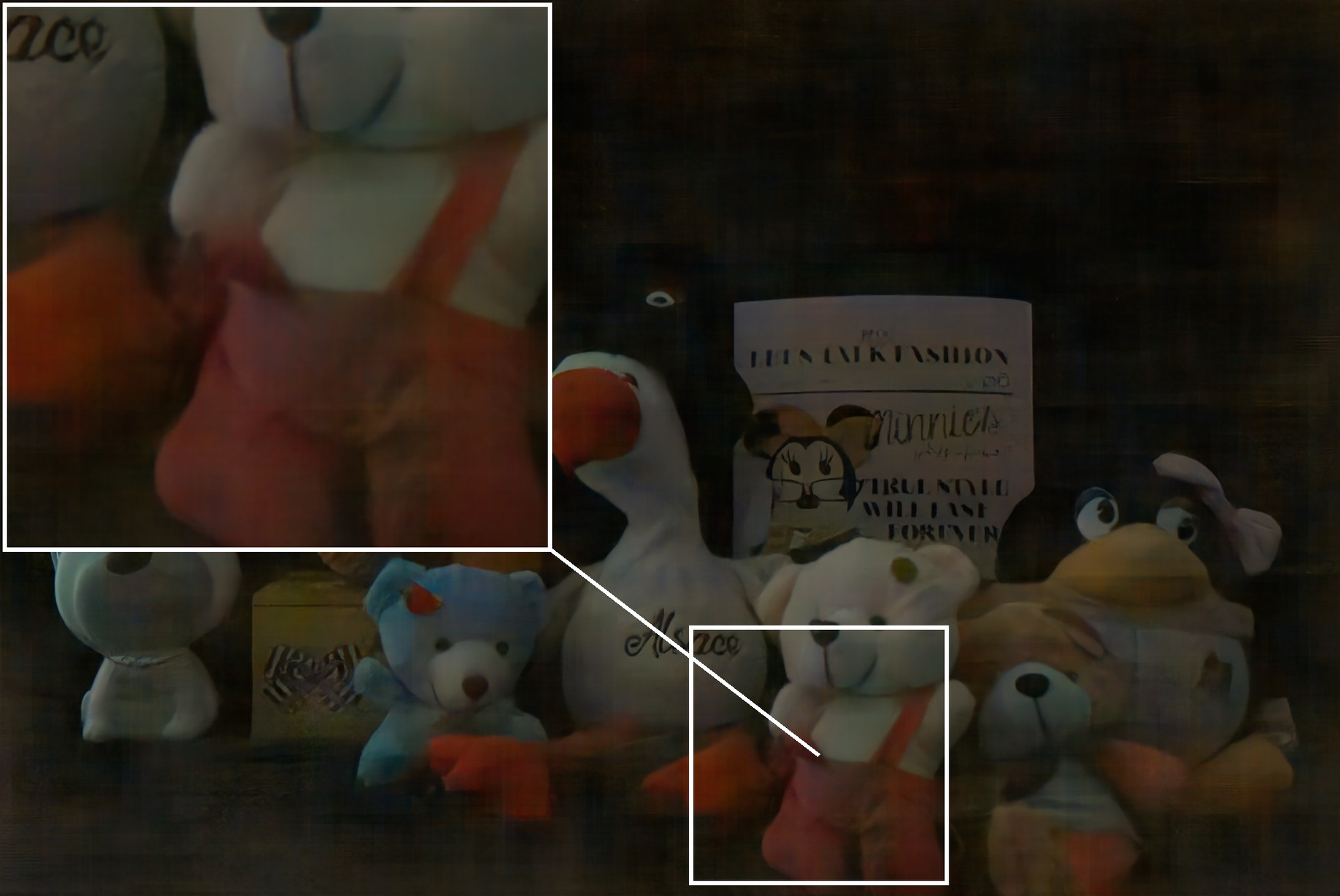}
    \label{fig:PGRQB}
}\\

\vspace{2pt}

\subfloat[ELD~\cite{wei2021physics}]{%
    \includegraphics[width=0.235\textwidth]{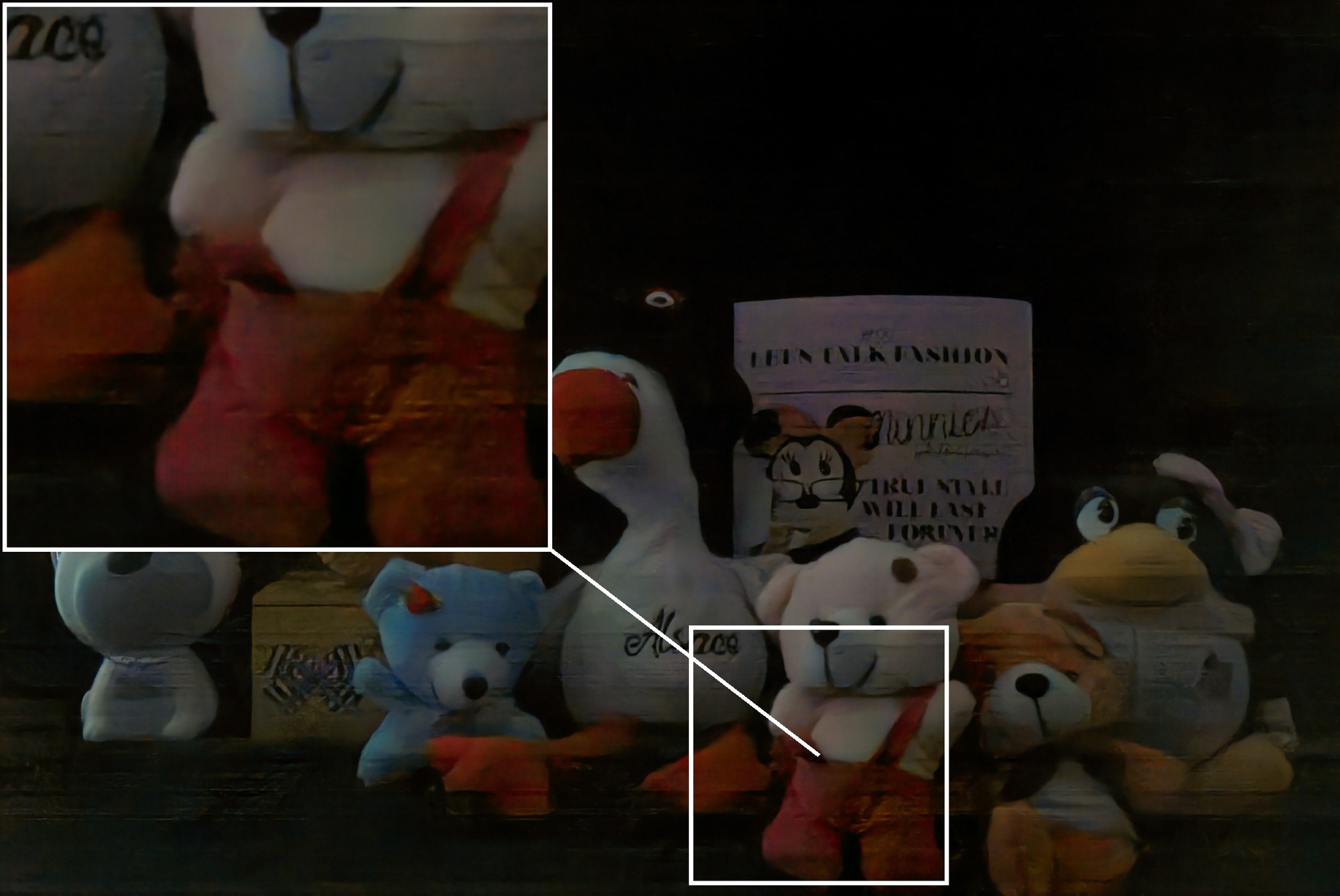}
    \label{fig:ELD}
}
\subfloat[NoiseDiff~\cite{lu2025dark}]{%
    \includegraphics[width=0.235\textwidth]{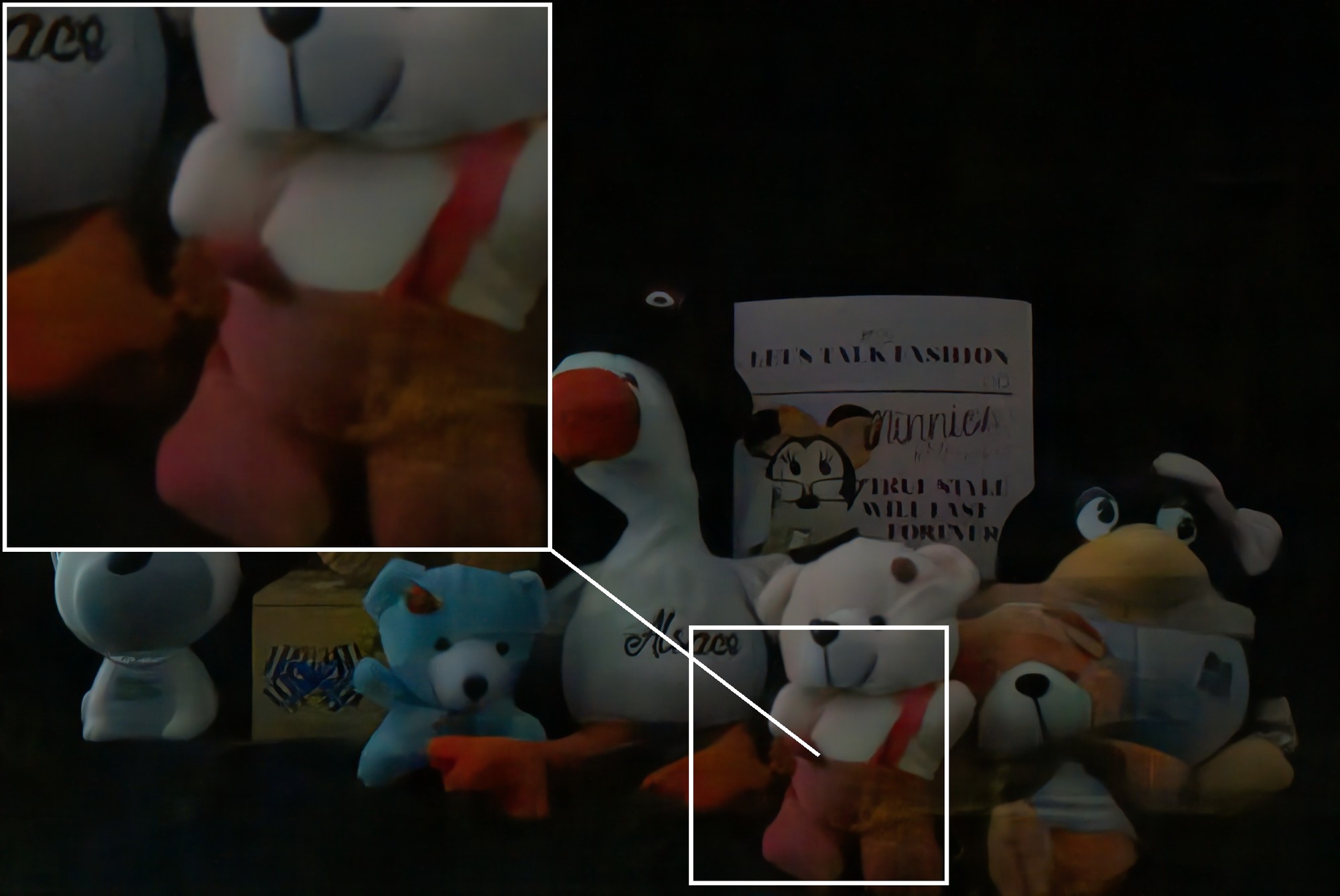}
    \label{fig:NoiseDiff}
}
\subfloat[Ours]{%
    \includegraphics[width=0.235\textwidth]{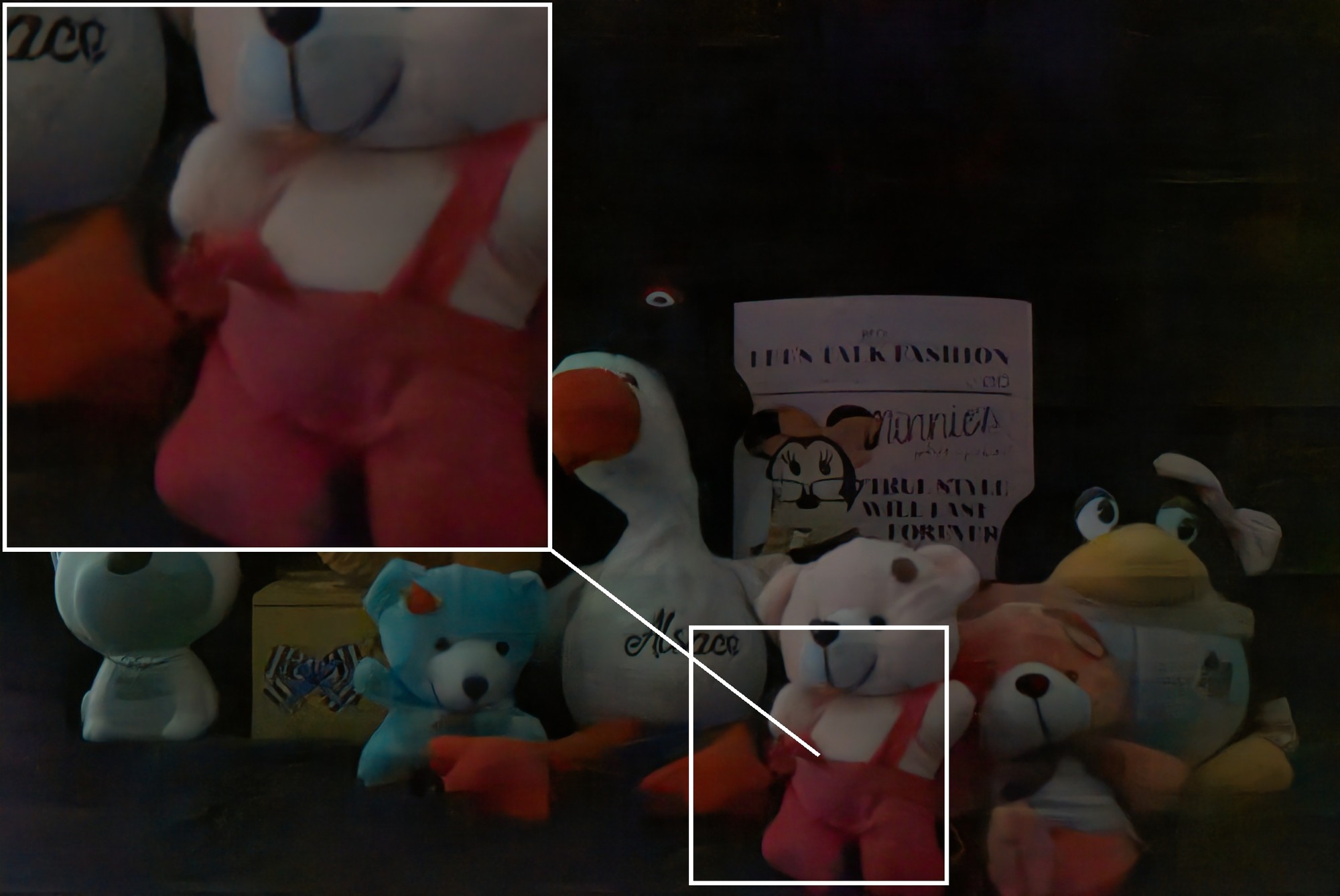}
    \label{fig:results_eld_sony_ours}
}
\subfloat[GT]{%
    \includegraphics[width=0.235\textwidth]{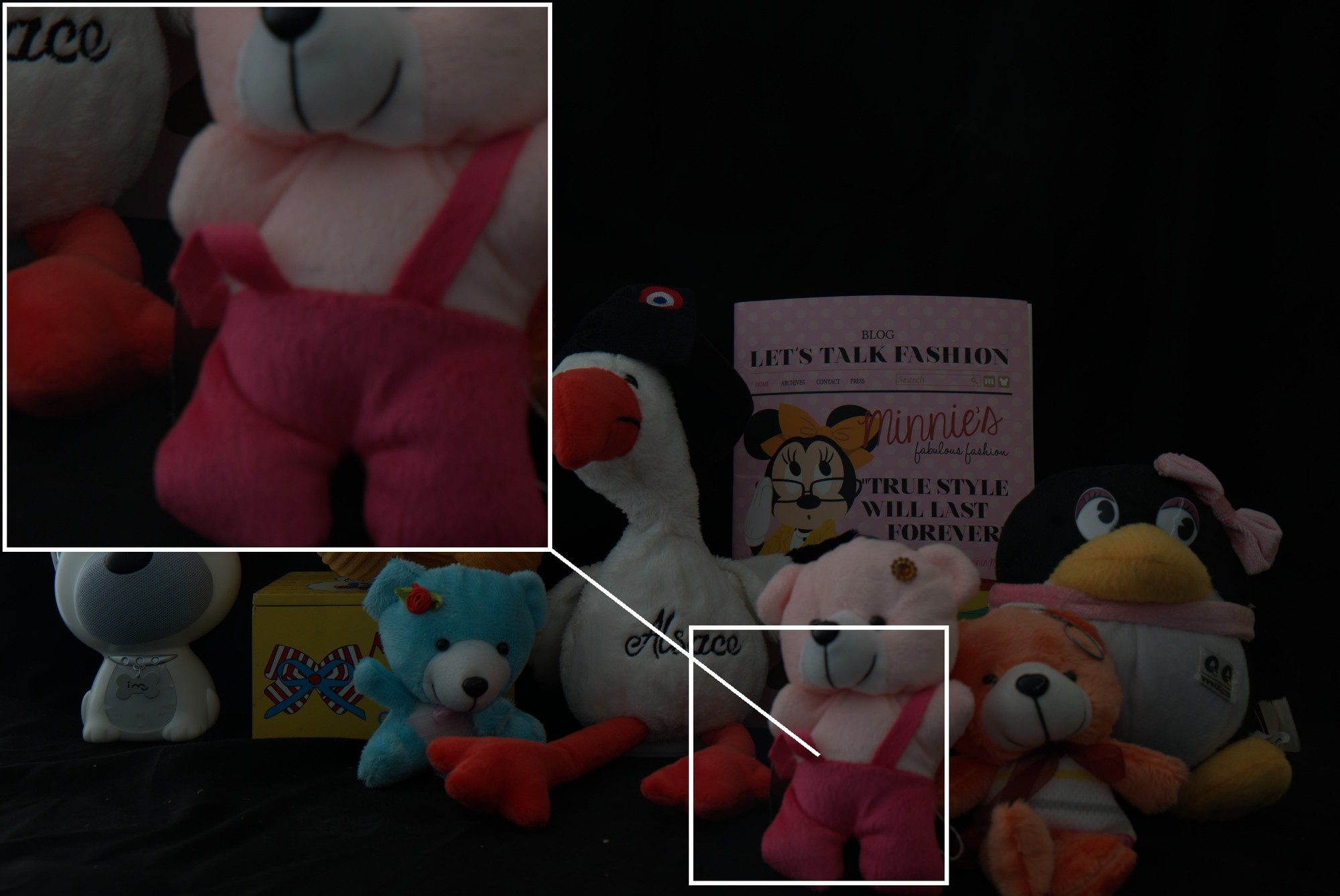}
    \label{fig:results_eld_sony_gt}
}

\caption{\textbf{Black-level-induced color bias handling of different noise models on the ELD-Sony~\cite{wei2021physics} dataset.} Progressive noise modeling with PG (Poisson-Gaussian), PGRQ (+Row, Quantization), and PGRQB (+Black-Level Error) show incremental improvement, yet color bias persists. The calibrated ELD noise model, and the paired-data diffusion-based noise synthesis model of NoiseDiff also exhibit color bias. Our noise modeling approach most closely matches the ground-truth color.}
\label{fig:noise_progression}
\end{figure*}

\subsection{Low-light Noise Modeling}
\label{sec:llnm}
To analyze denoising performance under varying low-light noise assumptions, we train uncalibrated models with progressively richer noise formulations.
\cref{fig:noise_progression} compares \textbf{PG} (Poisson-Gaussian), \textbf{PGRQ} (+Row, Quantization), and \textbf{PGRQB} (+Black-Level Error), alongside the calibrated \textbf{ELD} model~\cite{wei2021physics} and the diffusion-based noise model \textbf{NoiseDiff}~\cite{lu2025dark} trained on  paired samples of the target sensor. 
Noise in image restoration has commonly been synthesized using PG~\cite{luisier2010image, zhang2023practical, feng2025yond} or heteroscedastic Gaussian models~\cite{foi2009clipped, foi2008practical}, which jointly capture signal-dependent shot noise and a simplified form of signal-independent noise. 
YOND~\cite{feng2025yond} demonstrated that the PG model is effective for noise synthesis within a calibration-free denoising framework for normal-light; however, the performance of these models in low-light settings is still lacking, as shown in \cref{fig:PG}. 

While the PG model focuses on pixel-wise noise, PGRQ additionally captures spatially correlated row noise and provides a more faithful modeling of the analog-to-digital conversion process, resulting in improved denoising performance. However, as illustrated in \cref{fig:PGRQ}, dark current-related artifacts persist in the restored images, leading to unrealistic color reproduction.
Incorporating BLE (\cref{eq:ble}) removes a large portion of the color bias (see \cref{fig:PGRQB}). 
Nevertheless, the recovered colors remain inconsistent with the ground-truth (GT) image. 
A similar behavior is observed for the physically calibrated ELD model in \cref{fig:ELD}, which also incorporates BLE in its noise formulation. 
These observations suggest that simply introducing BLE into the noise model does not fully resolve color bias in low-light denoising. 
The paired-data diffusion-based model of NoiseDiff also exhibits a persistent color bias (\cref{fig:NoiseDiff}), demonstrating that the issue may still arise even when utilizing paired data sourced from the target sensor. 
The same trends hold in \cref{fig:results_eld_nikon} and in Appendix~\cref{fig:supp_results_eld_sony,fig:supp_results_eld_nikon,fig:supp_results_sid}.

\subsection{Training and Denoising Pipeline}
\label{sec:actualmethod}

To address these artifacts, we design a color-bias-aware denoising framework, illustrated in Fig.~\ref{fig:overview}. 
The training pipeline starts by subtracting the black-level, $\bl$, provided in the \raw{} image metadata, followed by  normalization, resulting in the adjusted image $I_\text{adjusted}$:
\begin{equation}
    \label{eq:preprocess}
    I_\text{adjusted} = (I - \bl) / (\wl - \bl),
\end{equation}
where $\wl$ is the white-level. We then add synthetically generated corruptions, obtained by applying the noise model described in \cref{sec:noise_model} and the BLE of \cref{sec:dark_shading_noise} to  $I_\text{adjusted}$; i.e.,
$D_\text{noisy} := I_\text{adjusted} + N + \blem$, where $N$ is obtained by sampling noise according to \cref{eq:noise_sum} and $\blem$ according to \cref{eq:ble}. Each noise component is sampled from a broad parameter range without relying on camera-specific calibration. 
This training strategy removes any correlation that may exist among different noise parameters (e.g., those tied to a specific ISO setting), effectively making denoising harder, but also exposes the model to a wider range of settings, enhancing its ability to generalize across different cameras.

During training, the BLBE predicts the magnitude of the BLE that is synthetically injected to the   input. 
We train with an $\mathcal{L}_1$ loss between the prediction ($\blem\text{'}$) and the target ($\blem$):
\begin{equation}
    \mathcal{L}_\text{ble} =  \|\blem\text{'} - \blem\|_1.
    \label{eq:loss_ble}
\end{equation}
The black-level, $\bl$, is then updated as ${\bl}\text{'}= \bl + \blem\text{'}$, before re-applying \cref{eq:preprocess}. The denoising network then removes any remaining noise. 
We also use an $\mathcal{L}_1$ loss between the predicted ($I\text{'}$) and the clean ($I$) \raw{} images:
\begin{equation}
    \mathcal{L}_{I} =  \|I\text{'} - I\|_1.
\label{eq:denoiser_loss}
\end{equation}
The denoiser is trained end-to-end with a total loss combining both $\mathcal{L}_1$ terms, balanced with coefficient $\alpha$:
\begin{equation}
    \mathcal{L}_\text{denoise} =  \mathcal{L}_{I} + \alpha \mathcal{L}_{\text{ble}}.
\label{eq:E2E_loss}
\end{equation}

At inference, given a noisy \raw{} image, BLBE first predicts the black-level error, $\blem\text{'}$. 
This estimate is used to adjust the black-level, $\bl$, recorded in the image metadata. 
The corrected black-level is then applied as in \cref{eq:preprocess} before passing $I_\text{adjusted}$ to the denoiser for restoration.
Implementation details and pseudocode are provided in \cref{sec:supp_implementation}. 

\section{Smartphone Image Denoising Dataset (SIDD) Color Bias Correction}
\label{sec:SIDD}

SIDD~\cite{abdelhamed2018high} is a widely used smartphone denoising dataset, 
which captures a burst of noisy images and fuses them to construct ground-truth (GT) as follows:
(i) black-level subtraction (BLS) and clipping to $[0,1]$, 
(ii) aligning captures,
and 
(iii) performing robust estimation of the mean. %
However, in low-light conditions, %
clipping after BLS biases the GT estimate,
as noisy dark pixels with non-zero true signal are truncated, distorting the mean.
After rendering, this manifests as a purple tone, due to unequal color amplification during white balance (see Fig.~\ref{fig:teaser}).

To address this, we instead average the aligned noisy frames first, then apply BLS and clipping. 
As shown in \cref{fig:sidd_fix}, the original ordering 
introduces chromatic bias, whereas our revised pipeline 
maintains color accuracy (see \cref{sec:supp_siddcc}). 
This is particularly relevant when evaluating models 
trained on data (real or synthetic) \textit{without} this color bias, to avoid penalizing them for correct colors.
We refer to this dataset as \textbf{SIDD-Color Corrected (\newSIDD{})}.

\section{Experiments}
\label{sec:main_experiments}

\subsection{Implementation Details}
\label{sec:main_implmentation_detail}
Following prior work~\cite{chen2018learning, wei2021physics, lu2025dark}, we use a U-Net denoiser~\cite{ronneberger2015u}. The BLE prediction network shares an identical encoder with the denoiser, followed by an MLP head. Both networks are trained end-to-end for $750$ epochs on a single RTX~5090 GPU with a batch size of four, base learning rate of $2\times10^{-4}$, and cosine scheduling using Adam~\cite{kingma2015adam}. We set $\alpha=1.0$ (\cref{eq:E2E_loss}) and sample the BLE independently per RGBG channel. Additional details are in \cref{sec:supp_implementation}. 

\begin{figure}[t]
\centering
\includegraphics[width=\linewidth]{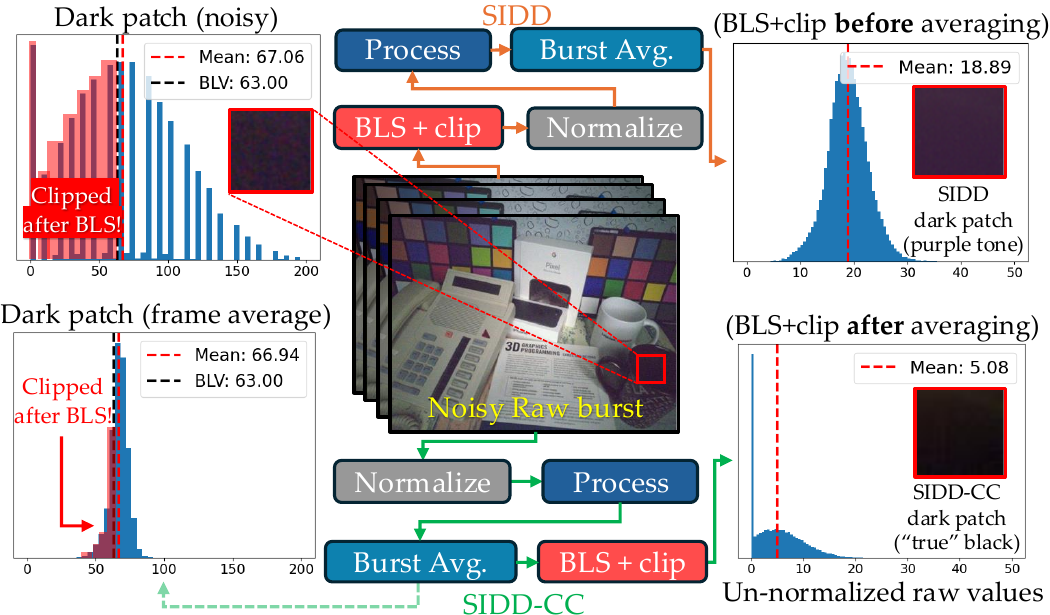}
\caption{
\textbf{SIDD vs.\ SIDD-CC.}
{Black-level (BL)} denotes the lowest intensity recorded by a camera sensor. 
In practice, however, \raw{} image values do not zero-out below BL, due to dark noise (top-left histogram). 
As a result,
when generating GT from noisy bursts, the order of processing steps, such as \textbf{black-level subtraction (BLS)}, \textbf{clipping}, and \textbf{averaging}, can impact color quality.
The original SIDD pipeline (top flow diagram) applies BLS and clipping to noisy frames \emph{before} burst averaging.
Due to the wide spread of intensity values around BL, many pixels are clipped, shifting the intensity distribution mean and introducing a purple tone, especially visible in dark areas (top-right histogram).
Instead, we apply BLS and clipping \emph{after} burst averaging (bottom flow diagram), which drastically reduces dark noise: 
despite an identical mean, the intensity distribution for the same dark patch becomes much tighter (bottom-left histogram).
This means fewer pixels are clipped, avoiding a distributional shift and preserving the true color (bottom-right histogram).
}
\label{fig:sidd_fix}
\end{figure}

\subsection{Datasets and Metrics}
\label{sec:main_datasets_metrics}
We train the models using clean images from the SID~\cite{chen2018learning} Sony \raw{} training set. 
We extract crops from each Bayer image and convert them to four-channel (RGBG) \raw{} images. For noise synthesis, we subtract the black-level, normalize by the white level (\cref{eq:preprocess}), then divide by an exposure ratio sampled uniformly from $[100, 300]$ to simulate low-light conditions. 
We sample noise parameters from broad ranges encompassing all calibrated values reported in~\cite{wei2021physics} across four sensors. 
Since we assume no access to the target sensor, parameters are sampled independently, although in practice they are often correlated and ISO-dependent~\cite{wei2021physics}. 
This broad sampling enables generalization across diverse camera configurations.
The synthesized noise is added to the ground truth, then rescaled by the same exposure ratio.

Following~\cite{wei2021physics, lu2025dark}, we evaluate all methods using PSNR on real noisy datasets: the combined SID validation and test sets, ELD~\cite{wei2021physics} (Sony and Nikon), and LRID~\cite{feng2022learnability} (see \cref{sec:supp_quantitative_results} for SSIM and CIEDE2000~\cite{sharma2005ciede2000}). 
As in prior work~\cite{wei2021physics, lu2025dark, lu20252}, we apply illumination correction for intensity mismatches in SID, ELD, and LRID.\footnote{Illumination mismatch stems from intensity differences between low- and high-quality captures and is corrected by a multiplicative factor on the denoised output. Black-level error, by contrast, is an offset error in the zero-level intensity of the image.} 
Finally, we benchmark \newSIDD{} and show the impact of color bias on SIDD evaluation.%

\subsection{Comparison With Baselines}
\label{sec:baselines}
\begin{table*}
\centering
\footnotesize

\resizebox{\textwidth}{!}{%
\begin{tabular}{ccccccccccccc}
\toprule
\multirow{2}{*}{Dataset} &
\multirow{2}{*}{Ratio} &
Real-data-based &
\multicolumn{4}{c}{Blind} &
\multicolumn{6}{c}{W/ Calibration} \\
\cmidrule(lr){3-3}\cmidrule(lr){4-7}\cmidrule(lr){8-13}

& &
Supervised U-Net &
PG &
PGRQ & 
PGRQB &
Ours &
ELD~\cite{wei2021physics} & 
SFRN~\cite{zhang2021rethinking} & 
PMN~\cite{feng2022learnability} & 
LRD~\cite{zhang2023towards} &
NoiseDiff~\cite{lu2025dark} & 
2-Shots~\cite{lu20252} \\

\midrule

\multirow{2}{*}{ELD - Sony}
& $\times100$
& \psnr{45.52}{0.977}
& \psnr{42.09}{0.870}
& \psnr{44.22}{0.934}
& \psnr{\underline{44.75}}{\underline{0.963}}
& \textbf{\psnr{46.18}{0.972}}
& \psnr{45.44}{0.975}
& \psnr{46.38}{0.979}
& \psnr{\underline{46.99}}{\underline{0.984}}
& \psnr{46.16}{0.983}
& \psnr{46.95}{0.978}
& \textbf{\psnr{47.13}{0.986}}
\\

& $\times200$
& \psnr{42.45}{0.945}
& \psnr{38.20}{0.782}
& \psnr{40.80}{0.863}
& \psnr{\underline{42.60}}{\underline{0.945}}
& \textbf{\psnr{43.85}{0.953}}
& \psnr{43.42}{0.954}
& \psnr{44.38}{0.941}
& \psnr{44.85}{\underline{0.969}}
& \psnr{43.91}{0.968}
& \textbf{\psnr{45.11}{0.971}}
& \psnr{\underline{44.89}}{\underline{0.969}}
\\

\midrule

\multirow{2}{*}{ELD - Nikon}
& $\times100$
& \psnr{41.28}{0.938}
& \psnr{42.84}{0.906}
& \psnr{\underline{43.55}}{\underline{0.949}}
& \psnr{41.70}{0.922}
& \textbf{\psnr{44.59}{0.962}}
& \underline{\psnr{42.27}{0.936}}
& \psnr{41.72}{\underline{0.942}}$^{\dagger\ddagger}$
& \psnr{\textbf{42.42}}{0.937}$^{\dagger\ddagger}$
& \psnr{41.70}{0.909}$^{\dagger\ddagger}$
& \psnr{41.65}{0.884}$^{\dagger\ddagger}$
& \psnr{{42.26}}{0.895}$^{\dagger\ddagger}$
\\

& $\times200$
& \psnr{39.44}{0.910}
& \psnr{39.71}{0.836}
& \psnr{\underline{40.81}}{\underline{0.909}}
& \psnr{40.20}{\underline{0.911}}
& \textbf{\psnr{42.82}{0.948}}
& \textbf{\psnr{40.36}{0.908}}
& \psnr{39.60}{\underline{0.886}}$^{\dagger\ddagger}$
& \psnr{39.75}{0.882}$^{\dagger\ddagger}$
& \psnr{39.41}{0.863}$^{\dagger\ddagger}$
& \psnr{39.08}{0.862}$^{\dagger\ddagger}$
& \psnr{\underline{39.86}}{0.853}$^{\dagger\ddagger}$
\\

\midrule

\multirow{3}{*}{SID}
& $\times100$
& \psnr{42.95}{0.958} 
& \psnr{39.89}{0.899}
& \psnr{\underline{40.66}}{0.923}
& \psnr{39.81}{\underline{0.934}}
& \textbf{\psnr{42.38}{0.945}}
& \psnr{42.75}{0.949}
& \psnr{42.81}{0.957}
& \psnr{43.47}{\textbf{0.961}}
& \psnr{42.81}{0.957}
& \textbf{\psnr{43.92}{0.961}}
& \psnr{\underline{43.57}}{\textbf{0.961}}
\\

& $\times250$
& \psnr{40.27}{0.943}
& \psnr{34.54}{0.777}
& \psnr{\underline{36.38}}{0.866}
& \psnr{36.22}{\underline{0.901}}
& \textbf{\psnr{38.71}{0.912}}
& \psnr{40.60}{0.932}
& \psnr{40.18}{0.934}
& \psnr{41.04}{\textbf{0.947}}
& \psnr{40.69}{0.941}
& \psnr{\textbf{41.28}}{\underline{0.946}}
& \psnr{\underline{41.24}}{0.945}
\\

& $\times300$
& \psnr{37.32}{0.928}
& \psnr{31.12}{0.669}
& {\psnr{33.19}{0.812}}
& \psnr{\underline{33.75}}{\underline{0.867}}
& \textbf{\psnr{35.23}{0.884}}
& \psnr{36.46}{0.916}
& \psnr{37.09}{0.918}
& \psnr{\underline{37.87}}{\textbf{0.934}}
& \psnr{37.48}{0.919}
& \psnr{\textbf{37.90}}{\underline{0.929}}
& \psnr{37.77}{\underline{0.929}}
\\

\midrule
\multirow{2}{*}{LRID - Indoor}

&
$\times128$
& \psnr{47.10}{0.986}
& \psnr{\textbf{46.79}}{\underline{0.983}}
& \psnr{\underline{46.73}}{\textbf{0.984}}
& \psnr{43.76}{0.973}
& \psnr{46.34}{0.977}
& \psnr{46.69}{0.984}
& \psnr{46.75}{0.986}
& \underline{\psnr{47.60}{0.987}}
& \psnr{45.75}{0.982}
& \psnr{46.62}{0.980}
& \textbf{\psnr{47.72}{0.987}}
\\

&
$\times256$
& \psnr{44.89}{0.979}
& \psnr{\underline{44.12}}{\underline{0.968}}
& \textbf{\psnr{44.18}{0.971}}
& \psnr{42.26}{0.960}
& \psnr{43.91}{0.962}
& \psnr{44.47}{0.974}
& \psnr{44.84}{0.979}
& \underline{\psnr{45.41}{0.981}}
& \psnr{43.29}{0.970}
& \psnr{44.26}{0.967}
& \psnr{\textbf{45.50}}{0.979}
\\

\midrule
\multirow{2}{*}{LRID - Outdoor}

& $\times128$
& \psnr{44.52}{0.982}
& \psnr{\underline{43.98}}{0.976}
& \psnr{43.89}{\textbf{0.977}}
& \psnr{41.96}{0.963}
& \psnr{\textbf{44.02}}{\underline{0.976}}
& \psnr{43.63}{0.974}
& \psnr{43.83}{0.977}
& \textbf{\psnr{44.90}{0.983}}
& \psnr{43.44}{0.976}$^{\dagger}$
& \psnr{43.81}{0.967}$^{\dagger}$
& \underline{\psnr{44.76}{0.980}}
\\

& $\times256$
& \psnr{42.71}{0.971}
& \psnr{\underline{42.05}}{\underline{0.957}}
& \textbf{\psnr{42.11}{0.962}}
& \psnr{40.64}{0.949}
& \psnr{41.92}{0.956}
& \psnr{41.52}{0.948}
& \psnr{42.08}{0.961}
& \textbf{\psnr{43.01}{0.970}}
& \psnr{41.43}{0.959}$^{\dagger}$
& \psnr{41.57}{0.947}$^{\dagger}$
& \underline{\psnr{42.72}{0.958}}
\\
\bottomrule
\end{tabular}%
}

\caption{\textbf{PSNR results on ELD~\cite{wei2021physics}, SID~\cite{chen2018learning}, and LRID~\cite{feng2022learnability}.} Our method outperforms blind baselines on ELD-Sony and SID, achieves state-of-the-art performance on ELD-Nikon, and remains competitive with the blind baselines on LRID. Bold and underlined values denote the best and second-best results within each group, respectively. $^{\dagger}$Results obtained by evaluating the model trained on the SID-Sony sensor. $^{\ddagger}$Calibrated dark shading noise subtraction is unavailable.}

\label{tab:main_results}
\end{table*}

We compare five classes of methods on the SID, ELD and LRID datasets.
\textbf{(i) Real-noise}: a U-Net trained on real noisy--clean pairs from the target sensor.
\textbf{(ii) Physics-based blind}: PG, PGRQ, PGRQB (U-Nets trained on synthetic noise without sensor-specific calibration) and our approach.
\textbf{(iii) Physics-based calibrated}: ELD~\cite{wei2021physics} and PMN~\cite{feng2022learnability}, which model noise with parametric distributions calibrated for the target sensor.
\textbf{(iv) Learning-based generators}: LRD~\cite{zhang2023towards} and NoiseDiff~\cite{lu2025dark}, which learn to generate sensor-specific noise.
\textbf{(v) Sampling-based generators}: SFRN~\cite{zhang2021rethinking} and 2-Shots~\cite{lu20252}, which use dark frame and Poisson sampling for signal-independent and signal-dependent noise, respectively.

PMN, LRD, NoiseDiff, and 2-Shots apply dark shading 
correction (DSC) before denoising: dark shading noise, including fixed-pattern noise and BLE, is estimated from dark frames per ISO and subtracted from the input, so these methods do not learn to remove it but rely on pre-calibrated statistics. 
ELD models BLE as a calibrated per-ISO offset during training. 
SFRN samples synthetic noise from a dark frame database rather than applying DSC at inference, but the noise remains sensor-specific.

\textbf{ELD-Sony.}
As shown in \cref{tab:main_results}, enriching the physics-based model from PG to PGRQB consistently improves performance, but PGRQB often introduces local color distortions due to the difficulty of handling BLE (\cref{fig:PGRQ}). 
Our model instead represents BLE as a global degradation and achieves the best performance among blind baselines with consistent color restoration (\cref{fig:results_eld_sony_ours}). 
Notably, it also outperforms ELD in PSNR despite no access to the target sensor during training, which highlights the strength of our BLE prediction. 
ELD and NoiseDiff also struggle with accurate color and texture (\cref{fig:ELD,fig:NoiseDiff}). 
As expected, our method remains below most calibrated baselines, which benefit from real target sensor noise and DSC.

\textbf{ELD-Nikon.}
\begin{figure*}[t]
\centering

\subfloat[Noisy]{%
    \includegraphics[width=0.235\textwidth]{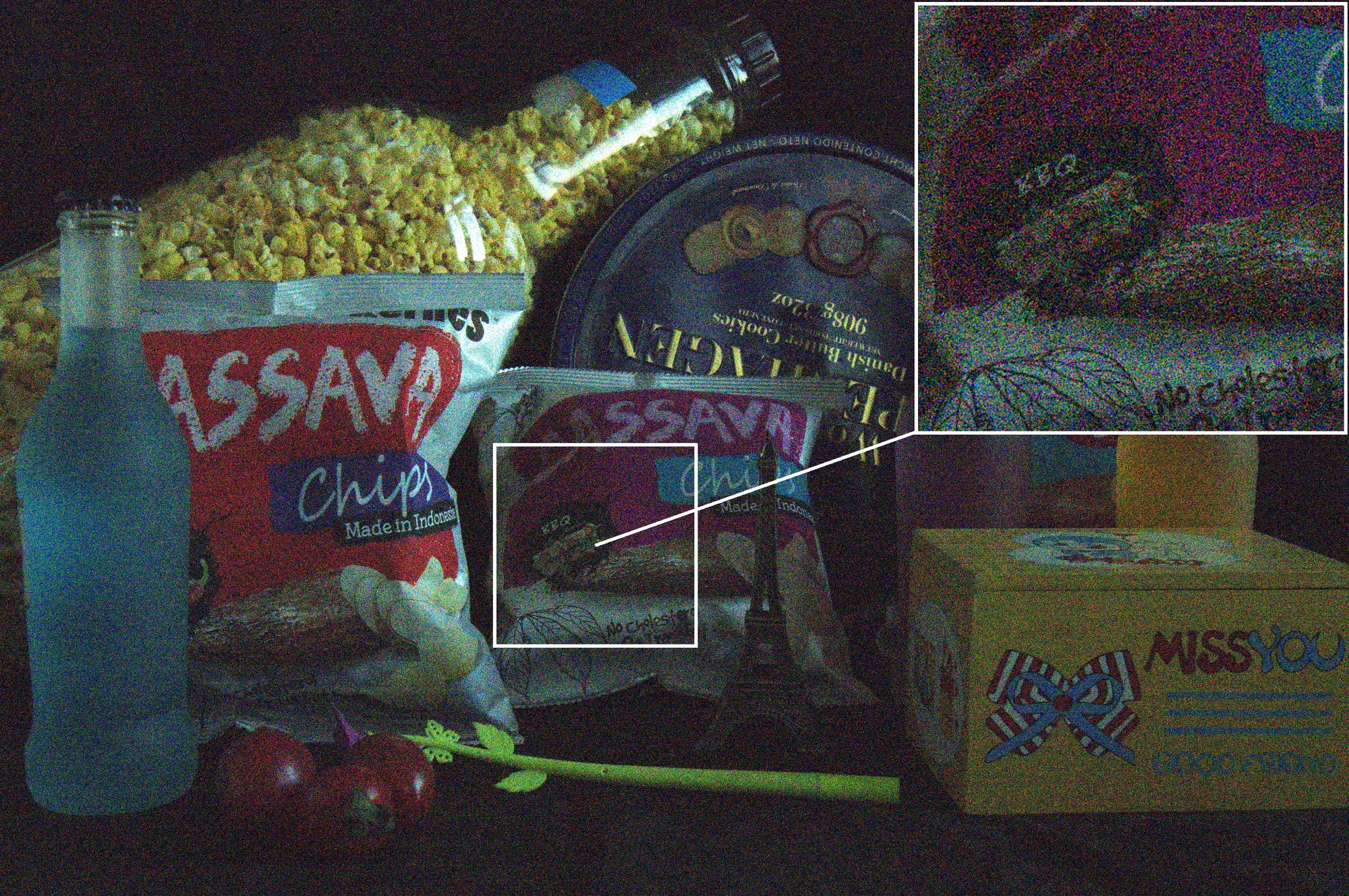}
    \label{fig:results_eld_nikon_noisy}
}
\subfloat[PG]{%
    \includegraphics[width=0.235\textwidth]{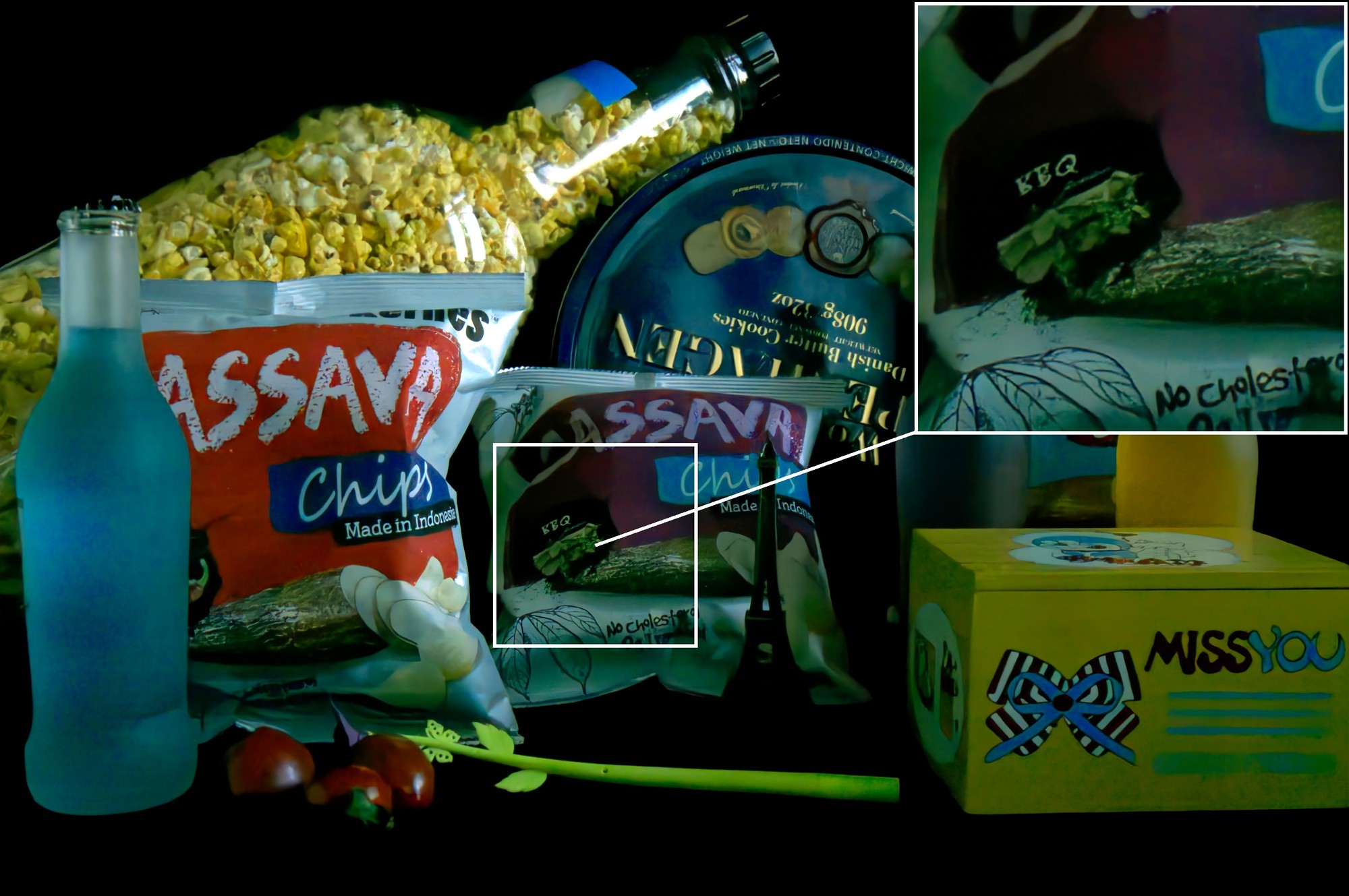}
    \label{fig:results_eld_nikon_pg}
}
\subfloat[PGRQ]{%
    \includegraphics[width=0.235\textwidth]{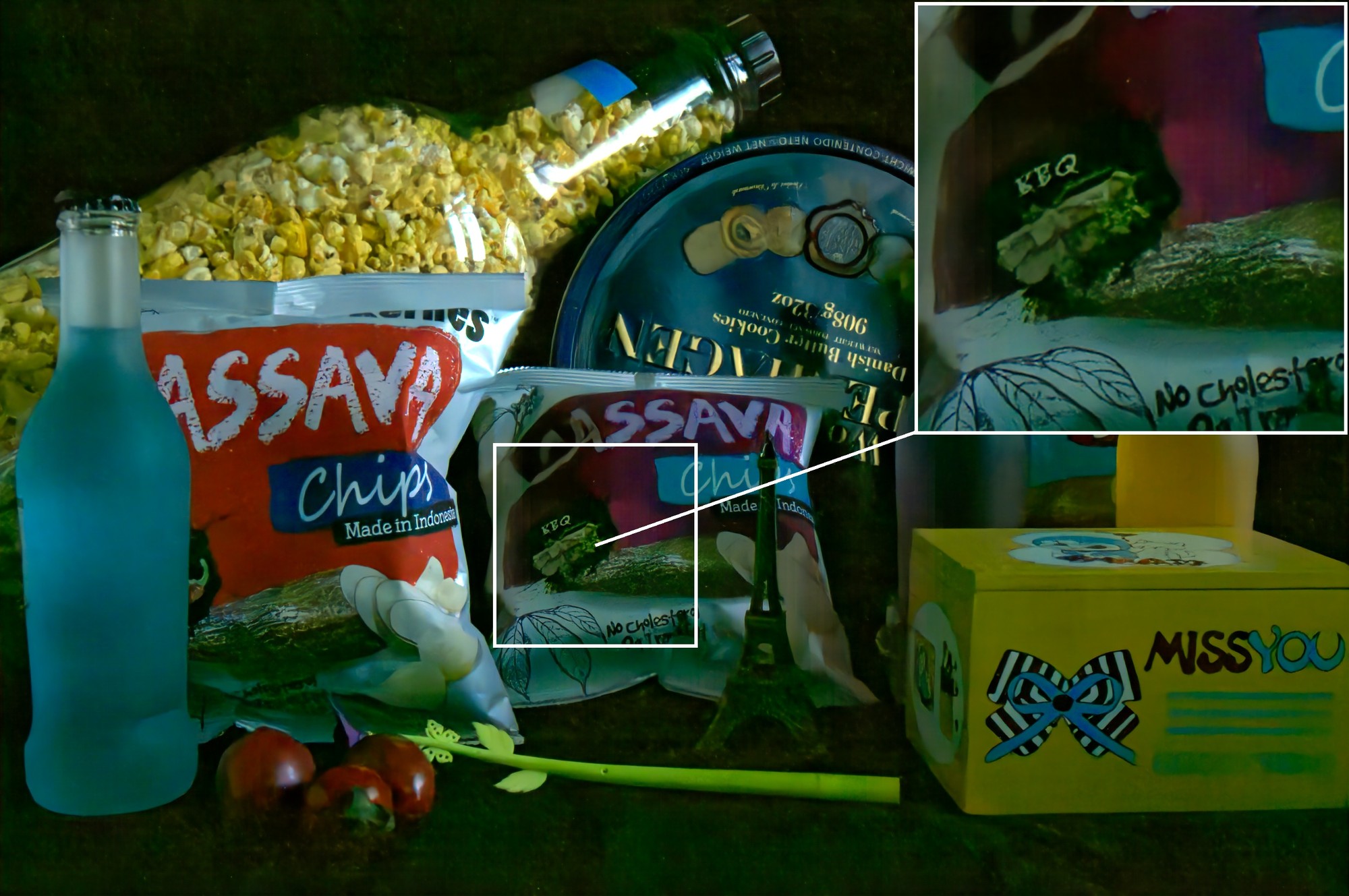}
    \label{fig:results_eld_nikon_pgrq}
}
\subfloat[PGRQB]{%
    \includegraphics[width=0.235\textwidth]{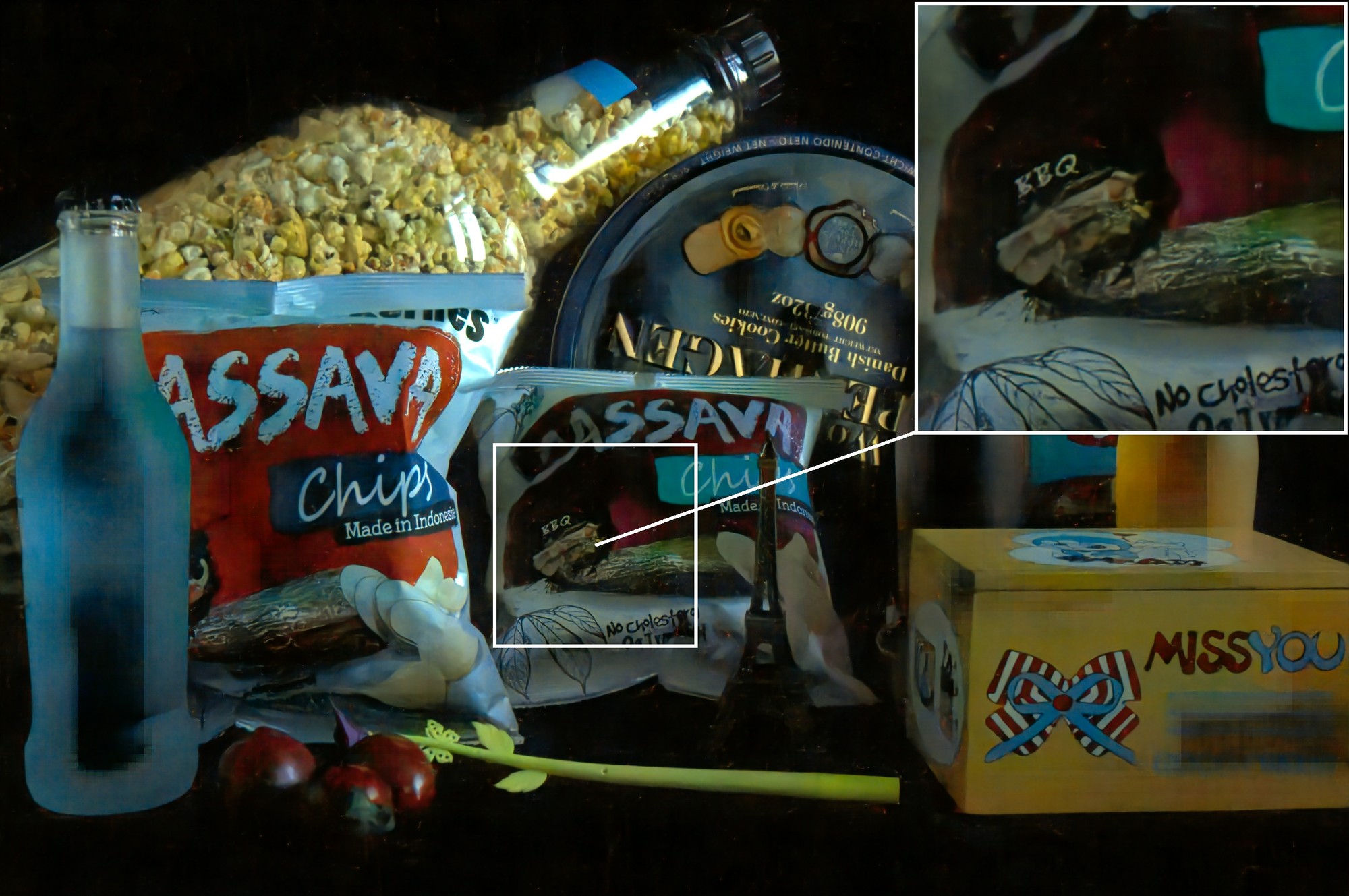}
    \label{fig:results_eld_nikon_pgrqb}
}\\

\vspace{2pt}

\subfloat[ELD~\cite{wei2021physics}]{%
    \includegraphics[width=0.235\textwidth]{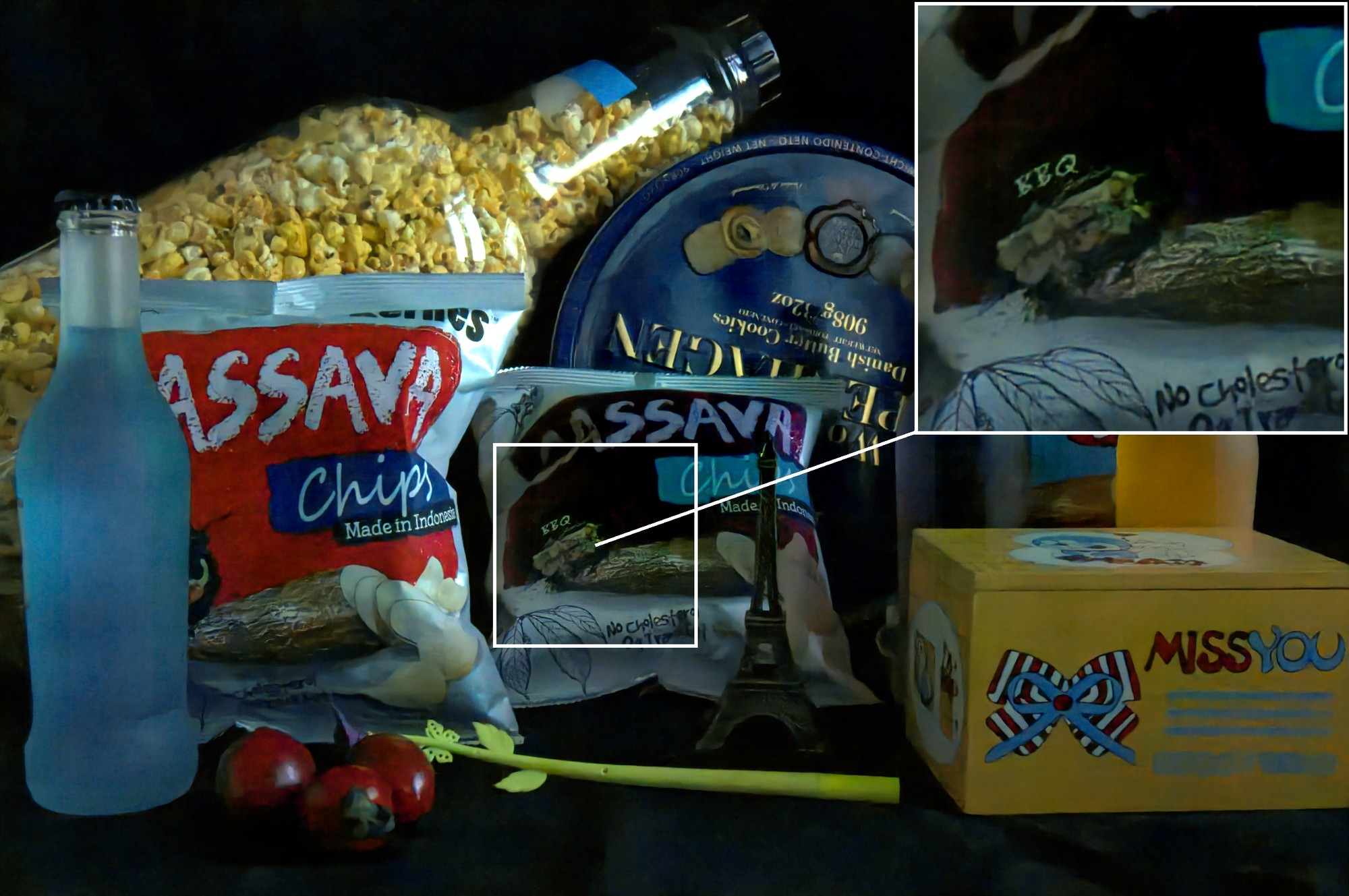}
    \label{fig:results_eld_nikon_eld}
}
\subfloat[2-Shots~\cite{lu20252}]{%
    \includegraphics[width=0.235\textwidth]{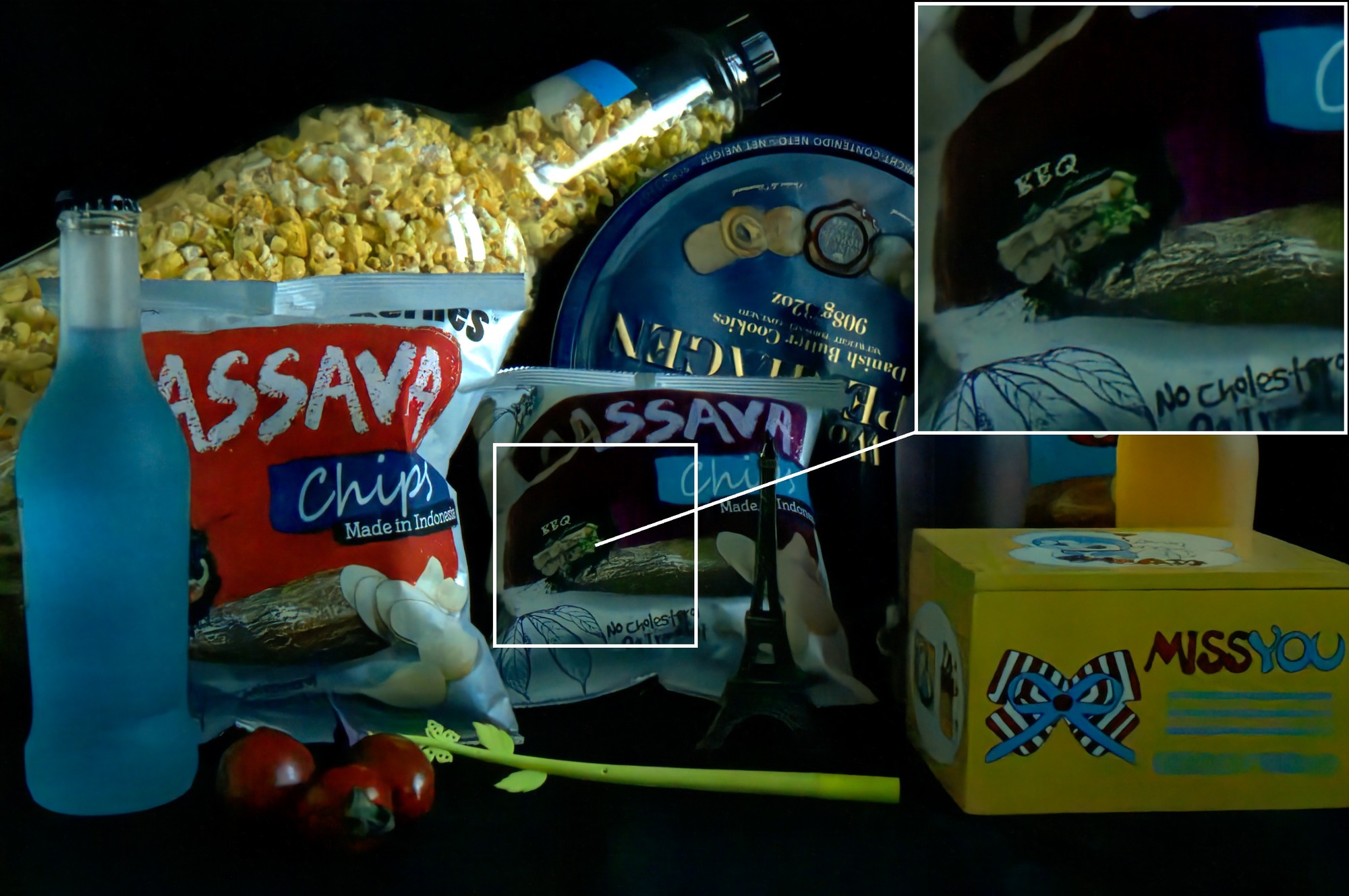}
    \label{fig:results_eld_nikon_2shots}
}
\subfloat[Ours]{%
    \includegraphics[width=0.235\textwidth]{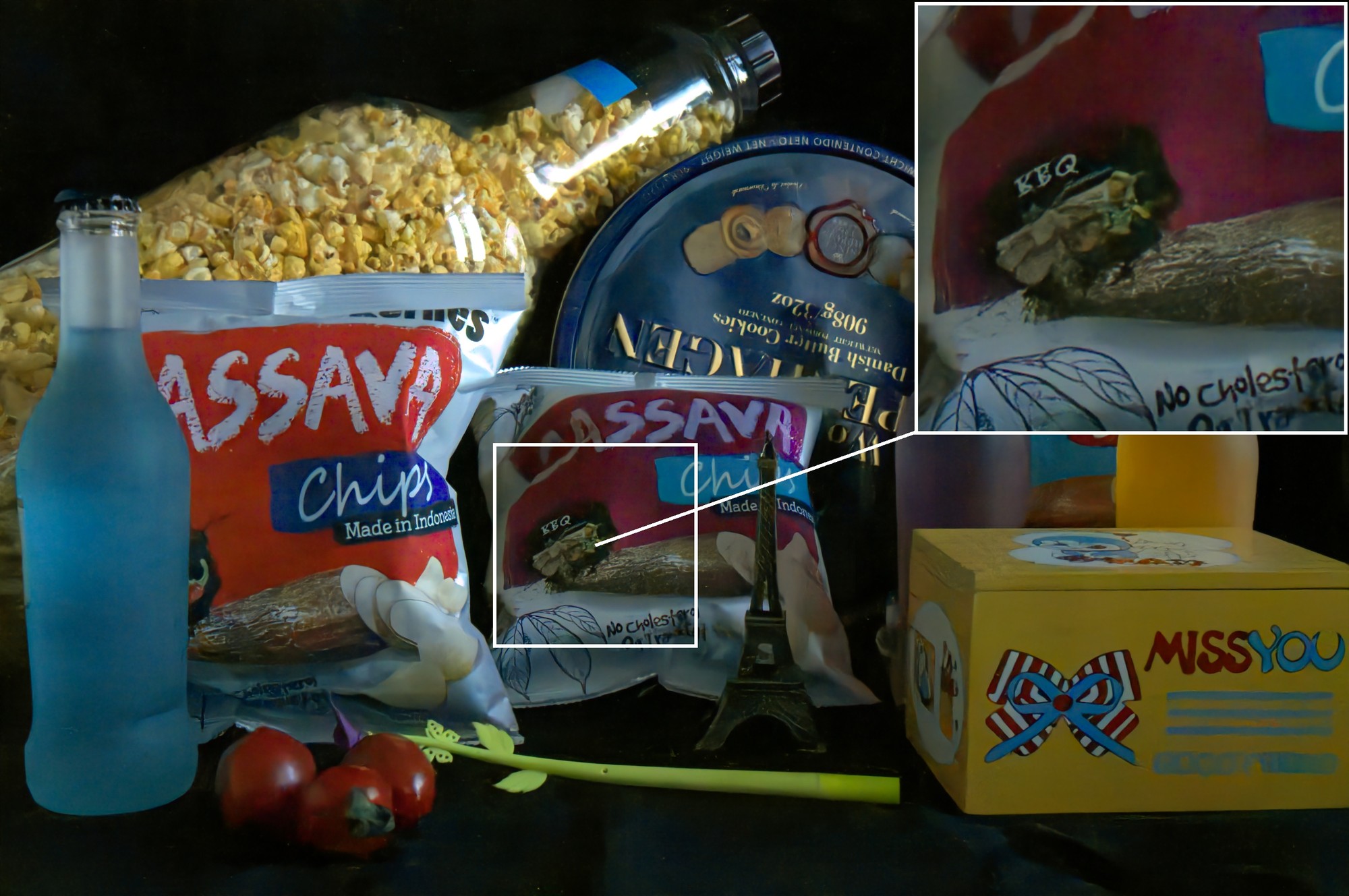}
    \label{fig:results_eld_nikon_ours}
}
\subfloat[GT]{%
    \includegraphics[width=0.235\textwidth]{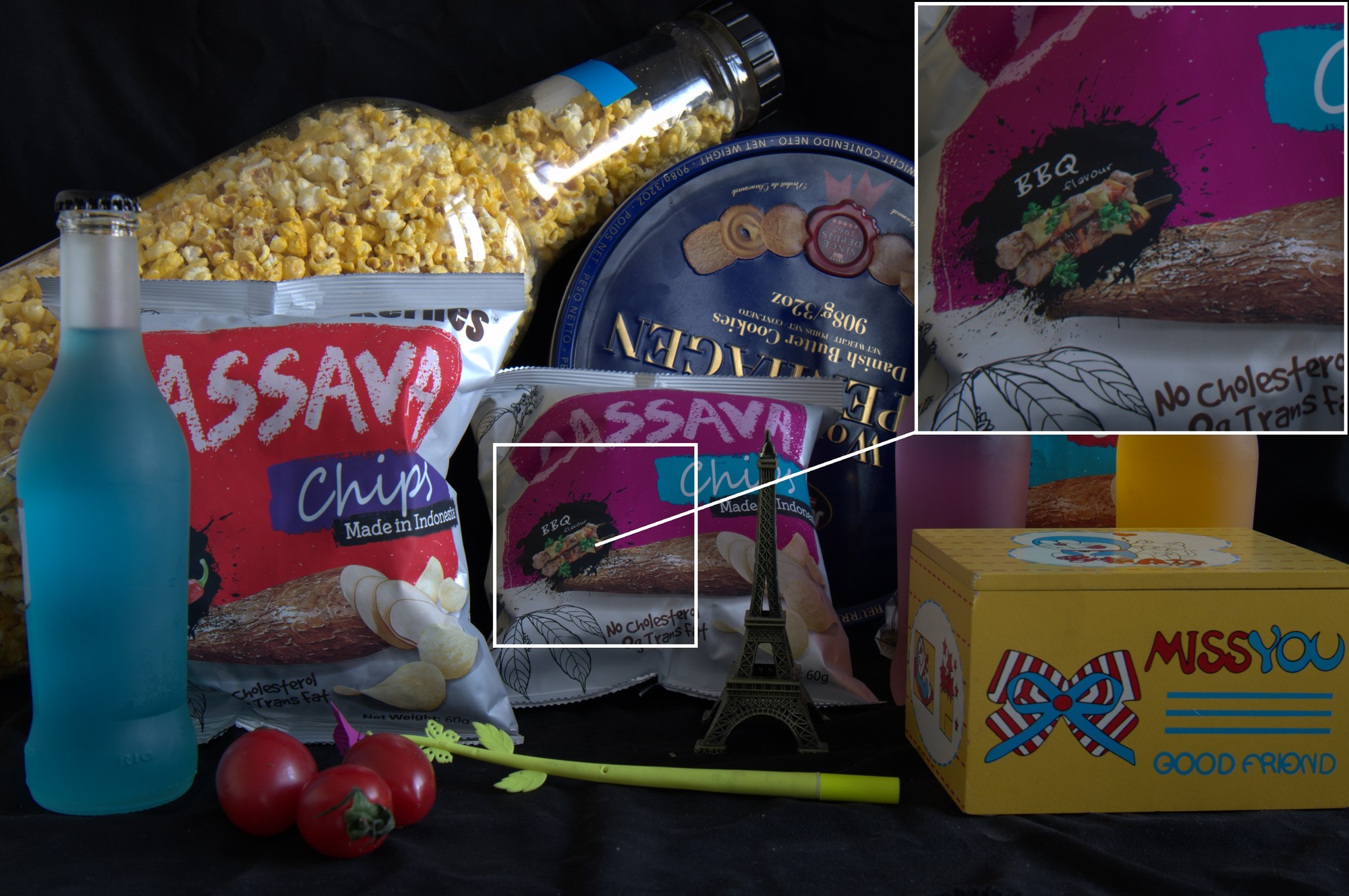}
    \label{fig:results_eld_nikon_gt}
}

\caption{
\textbf{Comparison on ELD-Nikon images~\cite{wei2021physics}.} Progressive noise modeling with PG (Poisson-Gaussian), PGRQ (+Row, Quantization), and PGRQB (+Black-Level Error) all exhibit color bias. The calibrated ELD and 2-Shots noise models also show it. In contrast, our noise modeling approach has much more accurate colors (see zoomed inset in particular).
}
\label{fig:results_eld_nikon}
\vspace{-5pt}
\end{figure*}

\cref{tab:main_results} reveals several key observations. First, PGRQ outperforms PGRQB, which confirms that BLE remains difficult to handle---PGRQB exhibits noticeable local color distortions (\cref{fig:results_eld_nikon_pgrqb}). 
Second, our method achieves the most faithful color restoration relative to ground truth (\cref{fig:results_eld_nikon_ours}) and yields state-of-the-art performance. 
Third, our model surpasses ELD, which suggests that calibrated sensor parameters alone are insufficient without explicit BLE modeling. 
The remaining calibrated baselines are evaluated by direct application of their models trained for ELD-Sony to ELD-Nikon without DSC; PMN, LRD, NoiseDiff, and 2-Shots all degrade, which highlights limited generalization to unseen sensors. SFRN also fails to recover accurate colors, as it relies on Sony-specific synthetic noise that does not generalize to Nikon dark shading characteristics.

\textbf{SID.}
Our model consistently outperforms PGRQB across all ratios and achieves the best performance among blind methods. 
Compared to ELD dataset, SID contains strong BLE and substantial fixed-pattern noise. As noise levels increase, fixed-pattern noise becomes more dominant, which widens the gap between our method and calibration-based baselines that subtract pre-calibrated fixed-pattern noise before denoising. 
Nevertheless, our method remains competitive without any target-sensor retraining or calibration. 

\textbf{LRID.}
Noisy images in LRID suffer only from subtle dark shading noise and BLE, which makes it a less suitable benchmark for methods that specifically address BLE (see Appendix \cref{sec:supp_ds_discussion}). 
\cref{tab:main_results} shows that PGRQB drops relative to PGRQ, which is due to color artifacts in PGRQB. 
Our model substantially improves over PGRQB but remains on par with PGRQ, consistent with the fact that explicit black-level correction matters less when BLE is negligible. 
Additional qualitative results and analysis for SID and LRID are provided in Appendix~\cref{sec:supp_qualitative_results}.

\textbf{Summary.}
When the target camera is unknown, our approach either achieves the best performance or is on-par with the best performing blind model, while other baselines suffer a noticeable drop on some sensors. 
We consistently outperform PGRQB, showing that modeling the black-level error (BLE) as a global offset is far more effective than treating it as a local additive corruption.
Moreover, the significant gap between PGRQ and our method highlights the importance of BLE modeling.
The results on ELD-Nikon are particularly notable: unlike calibration-based alternatives, we require no target-sensor access, dark frames, or per-sensor retraining.
Instead, a single denoiser obtains strong performance across all sensors.
Overall, these results show that BLE is a critical factor in blind \raw{} denoising.

\subsection{Ablation and Analysis}
\begin{table}[t]
\centering
\footnotesize
\renewcommand{\arraystretch}{1.05}

\begin{subtable}[t]{0.45\columnwidth}
\centering
\setlength{\tabcolsep}{3pt}
\footnotesize 
\begin{tabular}{l c}
\hline
Auxiliary Loss & Ours \\
\hline
No  & \psnr{40.16}{0.911} \\
Yes & \textbf{\psnr{42.82}{0.948}} \\
\hline
\end{tabular}
\caption{\textbf{BLE loss.} The auxiliary loss on predicted black-level error (BLE) is essential for accurate offset prediction.}
\label{tab:ablations_aux}
\end{subtable}
\hspace{0.05\columnwidth}
\begin{subtable}[t]{0.45\columnwidth}
\centering
\setlength{\tabcolsep}{3pt}
\footnotesize 
\begin{tabular}{l c c}
\hline
Clipping & PGRQB & Ours \\
\hline
No  & \textbf{\psnr{40.20}{0.911}} & \textbf{\psnr{42.82}{0.948}} \\
Yes & \psnr{38.78}{0.889} & \psnr{40.05}{0.881} \\
\hline
\end{tabular}
\caption{\textbf{Clipping.} Models struggle with producing accurate colors if the input is clipped between $[0, 1]$ after BLS.}
\label{tab:ablations_clipping}
\end{subtable}

\vspace{1em}

\begin{subtable}[t]{0.49\columnwidth}
\centering
\setlength{\tabcolsep}{3pt}
\footnotesize  
\begin{tabular}{l c c}
\hline
Architecture & Param. & PSNR \\
\hline
U-Net (default) & 7.79M  & \psnr{40.20}{0.911} \\
U-Net (large) & 31.03M & \psnr{40.31}{0.917} \\
\hline
Ours  & 12.54M & \textbf{\psnr{42.82}{0.948}} \\
\hline
\end{tabular}
\caption{\textbf{Model size.} Main gain of our paradigm is due to explicit modeling of black-level error and not the added size.}
\label{tab:ablations_model}
\end{subtable}
\hspace{0.04\columnwidth}
\begin{subtable}[t]{0.45\columnwidth}
\centering
\setlength{\tabcolsep}{3pt}
\footnotesize 
\begin{tabular}{l c c}
\hline
Infer. Res. & PGRQB & Ours \\
\hline
$256^2$  & \psnr{39.58}{0.909} & \psnr{37.54}{0.838} \\
$512^2$  & \psnr{39.93}{0.910} & \psnr{40.74}{0.919} \\
$1024^2$  & \psnr{40.06}{0.910} & \psnr{42.28}{0.937} \\
$2048^2$ & \psnr{40.17}{0.911} & \psnr{42.74}{0.946} \\
Full & \textbf{\psnr{40.20}{0.911}} & \textbf{\psnr{42.82}{0.948}} \\
\hline
\end{tabular}
\caption{\textbf{Inference resolution.} Increasing the resolution provides more visual context for accurate BLE estimation.}
\label{tab:ablations_infer}
\end{subtable}

\vspace{1em}

\begin{subtable}[t]{0.47\columnwidth}
\centering
\setlength{\tabcolsep}{3pt}
\footnotesize 
\begin{tabular}{l c c}
\hline
Train. Res. & PGRQB & Ours \\
\hline
$512^2$ & \psnr{39.91}{0.899} & \psnr{40.95}{0.885} \\
$1024^2$ & \textbf{\psnr{40.20}{0.911}} & \psnr{41.96}{0.914} \\
$1536^2$ & \psnr{40.12}{0.909} & \textbf{\psnr{42.82}{0.948}} \\
\hline
\end{tabular}
\caption{\textbf{Train resolution.} Larger input size in training provides a stronger signal for accurate BLE estimation. Full resolution is used at inference. 
}
\label{tab:ablations_input}
\end{subtable}
\hspace{0.04\columnwidth}
\begin{subtable}[t]{0.47\columnwidth}
\centering
\setlength{\tabcolsep}{3pt}
\footnotesize 
\begin{tabular}{l c c}
\hline
PGRQ Param. & PGRQB & Ours \\
\hline
Blind & \psnr{40.20}{0.911} & \psnr{42.82}{0.948} \\
ELD-Sony~\cite{wei2021physics} & \psnr{\textbf{40.49}}{0.929} & \psnr{\textbf{43.22}}{0.961} \\
\hline
\end{tabular}
\caption{\textbf{Sensor adaptation.} Using PGRQ parameter ranges calibrated for ELD-Sony improves our model over fully blind sampling of all components in PGRQB.}
\label{tab:ablations_noise_calibration}
\end{subtable}

\caption{
\textbf{Ablation studies.}
Best results are in bold.
}
\label{tab:ablations_v2}
\end{table}

We analyze the contribution of each component in our method. All ablations are reported on ELD-Nikon $\times200$.

\textbf{BLE Loss.}
As shown in \cref{tab:ablations_aux}, auxiliary supervision of \cref{eq:loss_ble} on the BLE predictor is essential. %
Optimizing only the final denoising loss (\cref{eq:denoiser_loss}) sharply degrades performance, on par with PGRQB, demonstrating that architectural changes alone are not sufficient to learn the BLE.

\textbf{Clipping.}
Prior work has explored both clipping~\cite{wei2021physics, lu2025dark} and non-clipping~\cite{feng2022learnability, lu20252} of the noisy input to the denoiser after BLS.
As shown in \cref{tab:ablations_clipping}, clipping to $[0,1]$ degrades performance for both our model and PGRQB, supporting our intuition that truncation causes information loss and makes the BLE estimation ambiguous.

\textbf{Model Size.}
The BLE predictor increases the overall model size. 
For fair comparison, we also double the number of channels of the baseline U-Net; however, as shown in \cref{tab:ablations_model}, this yields only marginal improvements, which indicates that the gains from our approach stem from explicitly modeling BLE rather than added capacity.

\textbf{Inference Resolution.}
\cref{tab:ablations_infer} reports performance of PGRQB and our model at different inference resolutions, where the \raw{} input is divided into non-overlapping patches processed independently. 
Increasing the patch size from $256^2$ to $2048^2$ significantly improves our model, which highlights that BLE prediction relies on global context---larger regions provide more information for accurate estimation. 
PGRQB is less sensitive to resolution, suggesting it treats color restoration as a local operation. 
For both methods, processing the full image yields the best results.

\textbf{Training Resolution.}
Training patch size has a substantial impact on our method (\cref{tab:ablations_input}): larger patches improve performance by providing more context for BLE estimation and stabilizing training. 
For PGRQB, increased patch size yields only marginal improvements.

\textbf{Sensor Adaptation.}
So far we adopted a fully blind setting, sampling noise parameters independently rather than jointly, as their joint distribution is sensor-specific and requires calibration. While dark-shading noise is sensor-specific, other components learned from calibrated data may generalize to unseen sensors. We thus take the PGRQ parameters calibrated on ELD Sony and combine them with our blind BLE estimation and modeling of \cref{eq:ble}, as in PGRQB. As shown in \cref{tab:ablations_noise_calibration}, these jointly calibrated parameters improve over fully blind independent sampling, despite calibration from a different sensor. As before, our framework outperforms PGRQB.

\subsection{Results on \newSIDD{}}

\begin{table}[t]
\centering
\small
\setlength{\tabcolsep}{4pt}
\begin{minipage}[t]{0.98\columnwidth}
\centering
\resizebox{0.9\linewidth}{!}{%
\begin{tabular}{lccc}
\hline
Category & Non-Learning & Self-Supervised  & Supervised \\
Method & BM3D~\cite{bm3d} & AT-BSN~\cite{atbsn} & NAFNet~\cite{chen2022simple} \\
\hline
PSNR & 30.03 & 36.78 &  44.67 \\
SSIM & 0.692 & 0.957 & 0.962 \\
\hline
\end{tabular}%
}
\caption*{(a) sRGB denoising on the \newSIDD{} validation set.}
\label{tab:siddcc_results_rgb}
\end{minipage}
\hfill
\begin{minipage}[t]{0.98\columnwidth}
\centering
\resizebox{\linewidth}{!}{%
\begin{tabular}{lcccc}
\hline
Category & Non-Learning & Self-Supervised & Blind & Supervised \\
Method & BM3D~\cite{bm3d} & B2U~\cite{wang2022blind2unblind} & YOND~\cite{feng2025yond} & NAFNet~\cite{chen2022simple} \\
\hline
PSNR & 46.99 & 49.41 & 50.15 & 51.63 \\
SSIM & 0.967 & 0.988 & 0.991 & 0.995 \\
\hline
\end{tabular}%
}
\caption*{(b) Raw %
denoising on the \newSIDD{} validation set.}
\label{tab:siddcc_results_raw}
\end{minipage}
\caption{\textbf{Benchmark on \newSIDD{}: sRGB (a) and \raw{} (b).}}
\label{tab:rgb_raw}
\vspace{-5pt}
\end{table}

\begin{figure}[t]
\centering
\captionsetup[subfloat]{font=footnotesize, labelfont=footnotesize}
\setlength{\tabcolsep}{1pt}

\begin{tabular}{cccc}
\subfloat[Noisy\label{fig:sidd_noisy}]{%
\includegraphics[width=0.24\linewidth]{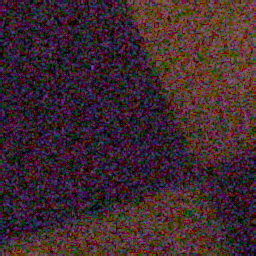}} &
\subfloat[SIDD GT\label{fig:sidd_gt}]{%
\includegraphics[width=0.24\linewidth]{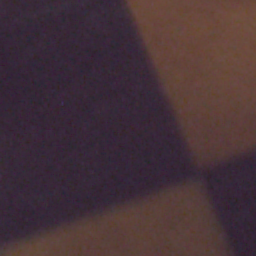}} &
\subfloat[SIDD-CC GT\label{fig:siddcc_gt}]{%
\includegraphics[width=0.24\linewidth]{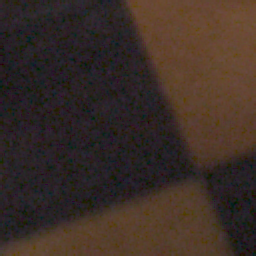}} &
\subfloat[YOND~\cite{feng2025yond}\label{fig:sidd_yond}]{%
\includegraphics[width=0.24\linewidth]{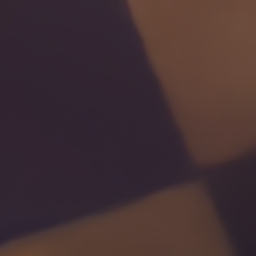}} \\

\subfloat[SIDD Sup.\label{fig:sidd_sup}]{%
\includegraphics[width=0.24\linewidth]{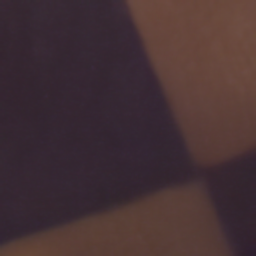}} &
\subfloat[SIDD-CC Sup.\label{fig:siddcc_sup}]{%
\includegraphics[width=0.24\linewidth]{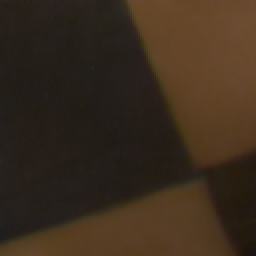}} &
\subfloat[PG\label{fig:sidd_pg}]{%
\includegraphics[width=0.24\linewidth]{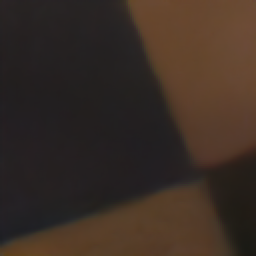}} &
\subfloat[B2U~\cite{wang2022blind2unblind}\label{fig:sidd_b2u}]{%
\includegraphics[width=0.24\linewidth]{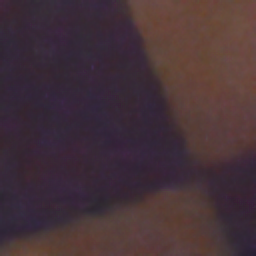}}
\end{tabular}

\caption{\textbf{Qualitative comparison on SIDD~\cite{abdelhamed2018high} and \newSIDD{}.} (d), (e), and (h) contain the purple tint seen in (b), while (f) and (g) match the dark regions of (c).}
\label{fig:sidd_comparison}
\end{figure}

We provide quantitative benchmark of state-of-the-art methods on the \newSIDD{} in \cref{tab:rgb_raw}. We show qualitative comparisons on \raw{} validation patches from \newSIDD{} in \cref{fig:sidd_comparison}, where \raw{} outputs are rendered for visualization. 
The noisy inputs are identical between SIDD and \newSIDD{}; only the ground truth differs. 
As shown in \cref{fig:sidd_gt,fig:siddcc_gt}, the original SIDD ground truth exhibits a noticeable purple tint compared to \newSIDD{}. 
Comparing two supervised NAFNet models trained on SIDD and \newSIDD{} (\cref{fig:sidd_sup,fig:siddcc_sup}), the SIDD-trained model inherits the color bias, while the \newSIDD{}-trained model produces accurate dark regions without this artifact. 
The blind PG baseline (\cref{fig:sidd_pg}, defined in \cref{sec:baselines}) avoids the purple bias, consistent with its independence from the SIDD GT. These results suggest that supervised models can learn the dataset’s color bias, whereas independently trained noise models can avoid such a bias. However, if evaluated on a color-biased dataset, they would be wrongly penalized.
YOND (\cref{fig:sidd_yond}) and B2U (\cref{fig:sidd_b2u}) also exhibit color shifts despite being self-supervised and not trained on SIDD ground truth, indicating limitations in their noise modeling. See \cref{sec:supp_siddcc} for additional qualitative results and benchmarking details.

\section{Conclusion}
\label{sec:conclusion}

In advancing models for low-light blind denoising, 
we identify \textit{color bias} as an insidious yet prevalent issue that manifests in two forms:
(a) color distortions in denoised results, specifically for low-light inputs, and
(b) color inaccuracies in the ground-truth from the widely used denoising dataset, SIDD~\cite{abdelhamed2018high}.
Our analysis points to two reasons for these color artifacts: 
erroneous black-level estimates in the captured \raw{} metadata,
in the case of (a),
and pre-processing issues 
with black-level clipping,
in the case of (b). 
In this work, we have addressed both problems, first by introducing a black-level error prediction module, which estimates black-level discrepancies and corrects them before denoising, 
and also by proposing an alternative processing pipeline for SIDD, which removes GT color artifacts. 
Our contributions improve low-light denoising both visually and quantitatively, compared to other camera-agnostic baselines, while in some cases even performing competitively with methods that require calibration.
We also hope that our proposed SIDD-CC can serve as a more reliable benchmark, especially when evaluating models trained on data that do not contain similar color biases.

\section*{Acknowledgements}
We acknowledge the support of the Natural Sciences and
Engineering Research Council of Canada (NSERC).

\bibliographystyle{IEEEtran}
\bibliography{references}

\appendices
\crefalias{section}{appendix}
\clearpage
\setcounter{page}{1}

\section{Implementation Details}
\label{sec:supp_implementation}

We provide additional implementation details for our model and the baselines---PG (Poisson--Gaussian), PGRQ (+Row Quantization), and PGRQB (+Black-Level Error)---complementing \cref{sec:main_experiments}. We first present pseudocode of our algorithm and then describe the implementation details needed to reproduce our results. Training hyperparameters are listed in \cref{tab:impl_hparams} and noise synthesis parameters in \cref{tab:noise_params}.

\begin{algorithm}
\caption{Denoising Pipeline}
\label{alg:supp_denoising}
\begin{algorithmic}[1]
\Procedure{Denoise}{$D$, $\bl$, $\wl$}
    \State $I_\text{adjusted} \gets \textsc{Preprocess}(D, \bl, \wl)$
    \State $\blem\text{'} \gets \text{BLEPredictor}(I_\text{adjusted})$
    \State $I_\text{corrected} \gets \textsc{Preprocess}(D, \bl + \blem\text{'}, \wl)$
    \State $\hat{I} \gets \text{Denoiser}(I_\text{corrected})$
    \State \textbf{return} $\hat{I}$
\EndProcedure
\end{algorithmic}
\end{algorithm}

\cref{alg:supp_denoising} summarizes the inference pipeline of our method. Given a noisy RAW image $D$, we first read the black-level $\bl$ and white-level $\wl$ from the metadata. The \textsc{Preprocess} routine subtracts the black-level, normalizes by $\wl - \bl$, and clips the result to $[-1, 1]$ for robustness to outliers while preserving information in dark regions. The preprocessed image is passed to the black-level error (BLE) predictor, which estimates a residual black-level error $\blem\text{'}$. We then re-run \textsc{Preprocess} with the corrected black-level $\bl + \blem\text{'}$ to produce $I_\text{corrected}$, which is finally fed to the denoiser to obtain the clean estimate $\hat{I}$.

\textbf{Dataset.}
As described in \cref{sec:main_datasets_metrics}, we train on clean images from SID~\cite{chen2018learning} and synthesize noise using parametric distributions. In each epoch, we randomly crop one patch per clean image, synthesize the corresponding noise map, and convert the single-channel RAW Bayer patch into a four-channel RGBG image. The crop size (in the single-channel Bayer representation) is $1536\times1536$ for PG, PGRQ, and our model, and $1024\times1024$ for PGRQB, following the results in \cref{tab:ablations_input}.

\begin{table}[b]
\centering
\small
\setlength{\tabcolsep}{6pt}
\begin{tabular}{l|cccc}
\toprule
\textbf{Hyperparameter} & \textbf{PG} & \textbf{PGRQ} & \textbf{PGRQB} & \textbf{Ours} \\
\midrule
Dataset & \multicolumn{4}{c}{SID Sony (RAW)~\cite{chen2018learning}} \\
Batch size & \multicolumn{4}{c}{4} \\
Epochs & \multicolumn{4}{c}{750} \\
Optimizer & \multicolumn{4}{c}{Adam~\cite{kingma2015adam}} \\
Learning rate & \multicolumn{4}{c}{$2\times10^{-4}$} \\
Scheduler & \multicolumn{4}{c}{Cosine} \\
GPU & \multicolumn{4}{c}{1$\times$ NVIDIA RTX 5090} \\
\midrule
Crop size & $1536^2$ & $1536^2$ & $1024^2$ & $1536^2$ \\
$\alpha$ & -- & -- & -- & 1.0 \\
\bottomrule
\end{tabular}
\caption{Implementation hyperparameters for training baselines and our method.}
\label{tab:impl_hparams}
\end{table}

\begin{table}[t]
\centering
\small
\setlength{\tabcolsep}{6pt}
\begin{tabular}{llc}
\toprule
\textbf{Component} & \textbf{Parameter} & \textbf{Range} \\
\midrule
Shot noise & $K_{\min}$ / $K_{\max}$ & $0.05$ / $30$ \\
\midrule
Read noise & $\log\sigma_{\min}$ / $\log\sigma_{\max}$ & $-2.0$ / $+3.0$ \\
\midrule
Row-wise banding & $\log\sigma_{\min}$ / $\log\sigma_{\max}$ & $-3.0$ / $+2.0$ \\
\midrule
Quantization & $q$ & \text{constant} $2$ \\
\midrule
Exposure ratio & $\min$ / $\max$ & $100$ / $300$ \\
\midrule
BLE Max.~Magnitude & $\mathcal{M}_\text{bl}$ & \text{constant} $2$ \\
\bottomrule
\end{tabular}
\caption{Noise synthesis parameter ranges, shared across PG, PGRQ, PGRQB, and our method. Each baseline uses the subset of components relevant to its noise model. The ranges encompass the calibrated parameters of four sensors from~\cite{wei2021physics}, widened by more than a factor of two to cover a broader variety of sensors. Clean images are divided by an exposure ratio sampled from the listed range to simulate low-light conditions prior to noise injection.}
\label{tab:noise_params}
\end{table}

\textbf{Noise Synthesis.}
We assume no information about the target sensor, which imposes two constraints: (i) we discard correlations between parameters of different noise components, since these correlations are sensor-specific~\cite{wei2021physics}; and (ii) we choose a broad range for each noise component to cover diverse sensors. Starting from the calibrated parameter ranges of four sensors reported in~\cite{wei2021physics}, we widen each range by more than a factor of two to ensure coverage across a broader set of sensors (see \cref{tab:noise_params}). A new set of noise parameters is sampled per training patch. Following~\cite{wei2021physics}, shot, read, and row-wise banding noise parameters are sampled log-uniformly within the listed ranges (the table reports the linear bounds for shot noise and the log-space bounds for the others), whereas quantization and black-level error (BLE) are sampled uniformly in linear space. Noise parameters are shared across the RGBG channels, while noise values---including the BLE offset---are sampled independently per channel, yielding distinct per-channel BLE offsets. These parameters are applied after dividing the clean image by an exposure ratio sampled uniformly from $[100, 300]$ to simulate low-light conditions.

\textbf{Architecture.}
We adopt the U-Net from~\cite{chen2018learning} as the denoiser. The BLE predictor reuses the encoder of the same U-Net,  followed by the prediction head defined below:
\begin{lstlisting}[mathescape=true]
head: AdaptiveAvgPool(1) $\to$ Flatten
    $\to$ Linear(512, 128) $\to$ SiLU
    $\to$ Linear(128, $C$) $\to$ Tanh
\end{lstlisting}
To stabilize training, the predictor outputs a normalized value rather than the raw BLE. At inference, the predicted value is rescaled to the original range and added to the noisy input, which is then passed to the denoiser.

\textbf{Evaluation.}
Given a noisy image, we first subtract the black-level recorded in the metadata from both the noisy and clean images and normalize by the white-level. We avoid clipping to $[0,1]$ at this stage, as it causes information loss in dark regions and degrades performance (\cref{tab:ablations_clipping}); instead, we clip to $[-1, 1]$ to improve robustness to outliers while preserving information in dark regions. After denoising, we apply illumination correction following~\cite{wei2021physics, lu2025dark, lu20252} to compensate for intensity mismatches between the noisy and clean images. Finally, we clip both the denoiser output and the clean image to $[0,1]$ and report PSNR and SSIM. We additionally report CIEDE2000~\cite{sharma2005ciede2000} on the post-ISP rendered images to assess color accuracy.

\section{Additional Results}
\subsection{Pipeline Walkthrough}
\begin{figure}[t]
\centering

\subfloat[Noisy \raw{} input]{%
    \includegraphics[width=0.44\linewidth]{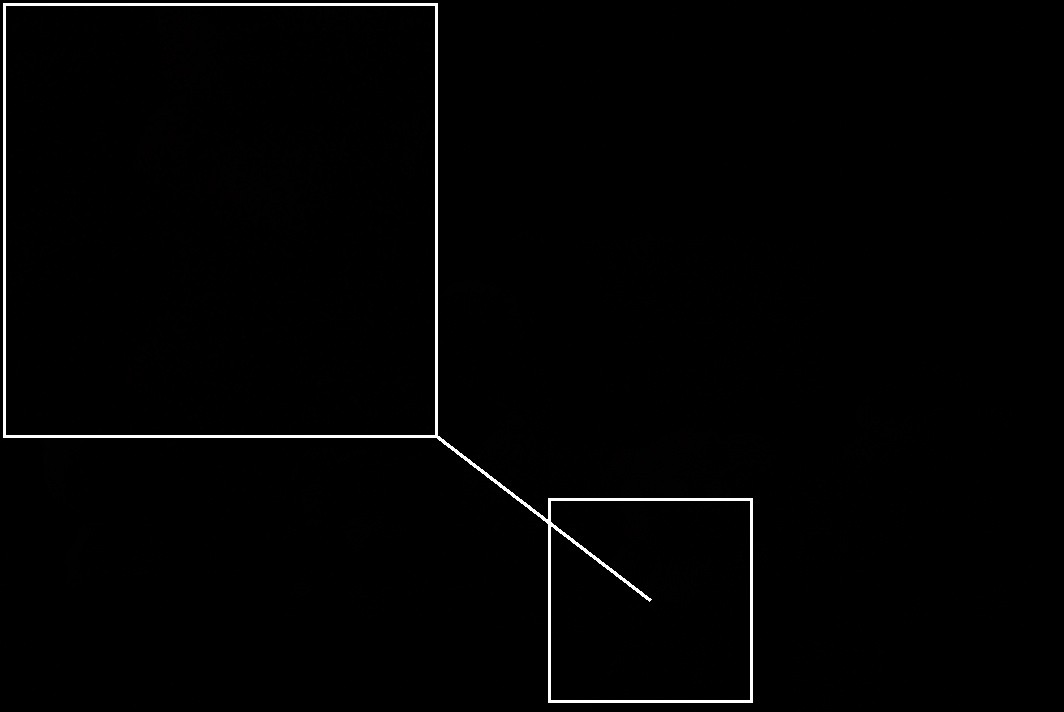}
    \label{fig:supp_pipeline_darknoisy}
}
\hspace{0.01mm}
\subfloat[After $\times200$ gain]{%
    \includegraphics[width=0.44\linewidth]{Figures/ELD_Sony/5-10-Sony-noisy.jpg}
    \label{fig:supp_pipeline_noisy}
}

\vspace{0.5mm}

\subfloat[After BLE correction]{%
    \includegraphics[width=0.44\linewidth]{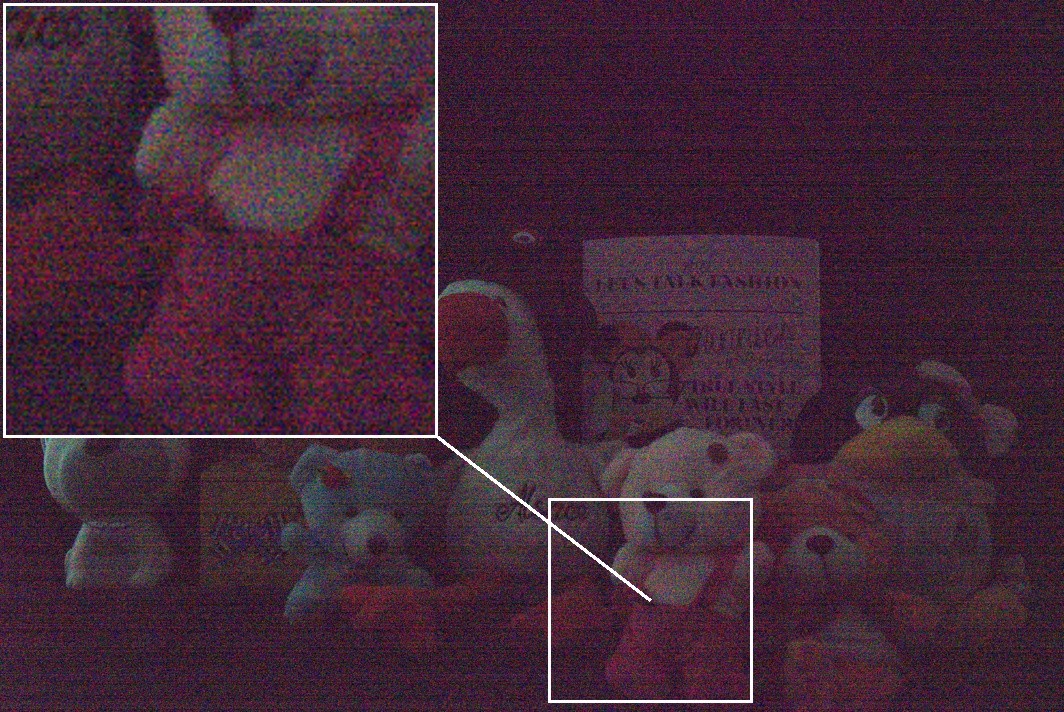}
    \label{fig:supp_pipeline_blbe}
}
\hspace{0.01mm}
\subfloat[Our prediction]{%
    \includegraphics[width=0.44\linewidth]{Figures/ELD_Sony/5-10-Sony-ours.jpg}
    \label{fig:supp_pipeline_prediction}
}

\caption{\textbf{Stage-by-stage visualization of our denoising pipeline} on an ELD-Sony~\cite{wei2021physics} $\times200$ sample. \textbf{(a)} The captured low-light \raw{} input is severely underexposed and barely visible. \textbf{(b)} Multiplying by the dataset-provided $\times200$ gain matches the ground-truth exposure but reveals strong noise and a pronounced color shift. \textbf{(c)} The BLBE module predicts the four per-channel black-level error offsets and subtracts each from its corresponding RGBG channel, correcting the global color shift. \textbf{(d)} The denoiser removes the remaining noise to produce the final clean estimate. All stages are processed in \raw{} and rendered through the ISP for visualization. Zoom for details.}
\label{fig:supp_pipeline_walkthrough}
\end{figure}

\label{sec:supp_pipeline_walkthrough}

To complement the abstract procedure in Algorithm~\ref{alg:supp_denoising}, we provide a per-stage visualization of our pipeline on a representative ELD~\cite{wei2021physics} example in \cref{fig:supp_pipeline_walkthrough}. Noisy images in ELD~\cite{wei2021physics} and SID~\cite{chen2018learning} at exposure ratios of $\times100$ and above are captured under high ISO and short exposure, yielding extremely low-light inputs in which the scene is barely visible (\cref{fig:supp_pipeline_darknoisy}). To match the ground-truth exposure, each input is multiplied by a per-sample gain factor provided by the dataset, producing a normal-light image that nonetheless retains strong noise and a pronounced color shift (\cref{fig:supp_pipeline_noisy}). The BLBE module then predicts the four per-channel black-level error (BLE) offsets and subtracts each from all pixels of its corresponding RGBG channel, which corrects the global color shift (\cref{fig:supp_pipeline_blbe}). Finally, the denoiser removes the remaining noise to produce the clean estimate (\cref{fig:supp_pipeline_prediction}).

\subsection{Quantitative Results}
\label{sec:supp_quantitative_results}
\begin{table*}[t]
\centering
\footnotesize

\resizebox{\textwidth}{!}{%
\begin{tabular}{ccccccccccccc}
\toprule
\multirow{2}{*}{Dataset} &
\multirow{2}{*}{Ratio} &
Real-data-based &
\multicolumn{4}{c}{Blind} &
\multicolumn{6}{c}{W/ Calibration} \\
\cmidrule(lr){3-3}\cmidrule(lr){4-7}\cmidrule(lr){8-13}

& &
Sup.~U-Net & %
PG &
PGRQ & 
PGRQB &
Ours &
ELD~\cite{wei2021physics} & 
SFRN~\cite{zhang2021rethinking} & 
PMN~\cite{feng2022learnability} & 
LRD~\cite{zhang2023towards} &
NoiseDiff~\cite{lu2025dark} & 
2-Shots~\cite{lu20252} \\

\midrule

\multirow{2}{*}{ELD - Sony}
& $\times100$
& \ssim{45.52}{0.977}
& \ssim{42.09}{0.870}
& \ssim{44.22}{0.934}
& \ssim{\underline{44.75}}{\underline{0.963}}
& \textbf{\ssim{46.18}{0.972}}
& \ssim{45.44}{0.975}
& \ssim{46.38}{0.979}
& \ssim{\underline{46.99}}{\underline{0.984}}
& \ssim{46.16}{0.983}
& \ssim{46.95}{0.978}
& \textbf{\ssim{47.13}{0.986}}
\\

& $\times200$
& \ssim{42.45}{0.945}
& \ssim{38.20}{0.782}
& \ssim{40.80}{0.863}
& \ssim{\underline{42.60}}{\underline{0.945}}
& \textbf{\ssim{43.85}{0.953}}
& \ssim{43.42}{0.954}
& \ssim{44.38}{0.965}
& \ssim{44.85}{\underline{0.969}}
& \ssim{43.91}{0.968}
& \textbf{\ssim{45.11}{0.971}}
& \ssim{\underline{44.89}}{\underline{0.969}}
\\

\midrule

\multirow{2}{*}{ELD - Nikon}
& $\times100$
& \ssim{41.28}{0.938}
& \ssim{42.84}{0.906}
& \ssim{\underline{43.55}}{\underline{0.949}}
& \ssim{41.70}{0.922}
& \textbf{\ssim{44.59}{0.962}}
& {\ssim{42.27}{0.936}}
& \ssim{41.72}{\textbf{0.942}}$^{\dagger\ddagger}$
& \ssim{42.42}{\underline{0.937}}$^{\dagger\ddagger}$
& \ssim{41.70}{0.909}$^{\dagger\ddagger}$
& \ssim{41.65}{0.884}$^{\dagger\ddagger}$
& \ssim{\underline{42.26}}{0.895}$^{\dagger\ddagger}$
\\

& $\times200$
& \ssim{39.44}{0.910}
& \ssim{39.71}{0.836}
& \ssim{\underline{40.81}}{{0.909}}
& \ssim{40.20}{\underline{0.911}}
& \textbf{\ssim{42.82}{0.948}}
& \textbf{\ssim{40.36}{0.908}}
& \ssim{39.60}{\underline{0.886}}$^{\dagger\ddagger}$
& \ssim{39.75}{0.882}$^{\dagger\ddagger}$
& \ssim{39.41}{0.863}$^{\dagger\ddagger}$
& \ssim{39.08}{0.862}$^{\dagger\ddagger}$
& \ssim{\underline{39.86}}{0.853}$^{\dagger\ddagger}$
\\

\midrule

\multirow{3}{*}{SID}
& $\times100$
& \ssim{42.95}{0.958} 
& \ssim{39.89}{0.899}
& \ssim{\underline{40.66}}{0.923}
& \ssim{39.81}{\underline{0.934}}
& \textbf{\ssim{42.38}{0.945}}
& \ssim{42.75}{0.949}
& \underline{\ssim{42.81}{0.957}}
& \ssim{43.47}{\textbf{0.961}}
& \underline{\ssim{42.81}{0.957}}
& \textbf{\ssim{43.92}{0.961}}
& \ssim{\underline{43.57}}{\textbf{0.961}}
\\

& $\times250$
& \ssim{40.27}{0.943}
& \ssim{34.54}{0.777}
& \ssim{\underline{36.38}}{0.866}
& \ssim{36.22}{\underline{0.898}}
& \textbf{\ssim{38.71}{0.912}}
& \ssim{40.60}{0.932}
& \ssim{40.18}{0.934}
& \ssim{41.04}{\textbf{0.947}}
& \ssim{40.69}{0.941}
& \ssim{\textbf{41.28}}{\underline{0.946}}
& \ssim{\underline{41.24}}{0.945}
\\

& $\times300$
& \ssim{37.32}{0.928}
& \ssim{31.12}{0.669}
& {\ssim{33.19}{0.812}}
& \ssim{\underline{33.75}}{\underline{0.862}}
& \textbf{\ssim{35.23}{0.884}}
& \ssim{36.46}{0.916}
& \ssim{37.09}{0.918}
& \ssim{\underline{37.87}}{\textbf{0.934}}
& \ssim{37.48}{0.919}
& \ssim{\textbf{37.90}}{\underline{0.929}}
& \ssim{37.77}{\underline{0.929}}
\\

\midrule
\multirow{2}{*}{LRID - Indoor}

&
$\times128$
& \ssim{47.10}{0.986}
& \ssim{\textbf{46.79}}{\underline{0.983}}
& \ssim{\underline{46.73}}{\textbf{0.984}}
& \ssim{43.76}{0.973}
& \ssim{46.34}{0.977}
& \ssim{46.69}{0.984}
& \underline{\ssim{46.75}{0.986}}
& \textbf{\ssim{47.60}{0.987}}
& \ssim{45.75}{0.982}$^{\dagger}$
& \ssim{46.62}{0.980}$^{\dagger}$
& \textbf{\ssim{47.72}{0.987}}
\\

&
$\times256$
& \ssim{44.89}{0.979}
& \ssim{\underline{44.12}}{\underline{0.968}}
& \textbf{\ssim{44.18}{0.971}}
& \ssim{42.26}{0.960}
& \ssim{43.91}{0.962}
& \ssim{44.47}{0.974}
& \underline{\ssim{44.84}{0.979}}
& \textbf{\ssim{45.41}{0.981}}
& \ssim{43.29}{0.970}$^{\dagger}$
& \ssim{44.26}{0.967}$^{\dagger}$
& \underline{\ssim{45.50}{0.979}}
\\

\midrule
\multirow{2}{*}{LRID - Outdoor}

& $\times128$
& \ssim{44.52}{0.982}
& \ssim{43.98}{\underline{0.976}}
& \ssim{43.89}{\textbf{0.977}}
& \ssim{41.96}{0.963}
& \ssim{\textbf{44.02}}{\underline{0.976}}
& \ssim{43.63}{0.974}
& \ssim{43.83}{0.977}
& \textbf{\ssim{44.90}{0.983}}
& \ssim{43.44}{0.976}$^{\dagger}$
& \ssim{43.81}{0.967}$^{\dagger}$
& \underline{\ssim{44.76}{0.980}}
\\

& $\times256$
& \ssim{42.71}{0.971}
& \ssim{\underline{42.05}}{\underline{0.957}}
& \textbf{\ssim{42.11}{0.962}}
& \ssim{40.65}{0.949}
& \ssim{41.92}{0.956}
& \ssim{41.52}{0.948}
& \underline{\ssim{42.08}{0.961}}
& \textbf{\ssim{43.01}{0.970}}
& \ssim{41.43}{0.959}$^{\dagger}$
& \ssim{41.57}{0.947}$^{\dagger}$
& {\ssim{42.72}{0.958}}
\\
\bottomrule
\end{tabular}%
}

\caption{\textbf{SSIM results on ELD~\cite{wei2021physics}, SID~\cite{chen2018learning}, and LRID~\cite{feng2022learnability}.} Complementary to the PSNR results reported in \cref{tab:main_results}. Our method outperforms all blind baselines on ELD-Sony and SID, achieves state-of-the-art performance on ELD-Nikon, and remains competitive with the blind baselines on LRID. Bold and underlined values denote the best and second-best results within each group, respectively. $^{\dagger}$Results obtained by evaluating the model trained on the SID-Sony sensor. $^{\ddagger}$Calibrated dark shading noise subtraction is unavailable.}
\label{tab:supp_ssim_results}
\end{table*}

\begin{table*}[t]
\centering
\footnotesize

\resizebox{\textwidth}{!}{%
\begin{tabular}{ccccccccccccc}
\toprule
\multirow{2}{*}{Dataset} &
\multirow{2}{*}{Ratio} &
Real-data-based &
\multicolumn{4}{c}{Blind} &
\multicolumn{6}{c}{W/ Calibration} \\
\cmidrule(lr){3-3}\cmidrule(lr){4-7}\cmidrule(lr){8-13}

& &
Sup.~U-Net &
PG &
PGRQ &
PGRQB &
Ours &
ELD~\cite{wei2021physics} &
SFRN~\cite{zhang2021rethinking} &
PMN~\cite{feng2022learnability} &
LRD~\cite{zhang2023towards} &
NoiseDiff~\cite{lu2025dark} &
2-Shots~\cite{lu20252} \\

\midrule

\multirow{2}{*}{ELD - Sony}
& $\times100$ & 3.24 & 8.75 & 7.39 & \underline{4.42} & \textbf{3.36} & 3.39 & 2.79 & 3.00  & 3.32 & \underline{2.67} & \textbf{2.50} \\
& $\times200$ & 5.00 & 11.59 & 10.33 & \underline{5.60} & \textbf{4.16} & 4.32 & 3.36 & 4.34 & 4.08 & \underline{3.34} & \textbf{3.24} \\

\midrule

\multirow{2}{*}{ELD - Nikon}
& $\times100$ & 5.22 & \underline{4.45} & 5.07 & 4.75 & \textbf{3.67} & \underline{4.26} & 4.78$^{\dagger}$ & 4.93$^{\dagger}$ & 4.53$^{\dagger}$ & 4.45$^{\dagger}$ & \textbf{4.04}$^{\dagger}$ \\
& $\times200$ & 6.33 & 6.16 & 7.41 & \underline{5.42} & \textbf{4.15} & \underline{5.06} & 6.74$^{\dagger}$ & 6.74$^{\dagger}$ & 5.42$^{\dagger}$ & 5.42$^{\dagger}$ & \textbf{4.91}$^{\dagger}$ \\

\midrule

\multirow{3}{*}{SID}
& $\times100$ & 5.82 & 9.26 & 9.12 & \underline{7.38} & \textbf{6.40} & 6.67 & 6.03 & 6.39 & 5.82 & \textbf{5.62} & \underline{5.74} \\
& $\times250$ & 6.64 & 12.43 & 12.31 & \underline{8.66} & \textbf{7.88} & 7.34 & 6.93 & 7.08 & \underline{6.65} & 7.40 & \textbf{6.34} \\
& $\times300$ & 7.11 & 15.43 & 15.41 & \underline{9.29} & \textbf{8.97} & 8.01 & 7.77 & \underline{7.62} & 7.63 & 7.73 & \textbf{7.21} \\

\midrule

\multirow{2}{*}{LRID - Indoor}
& $\times128$ & 3.23 & \underline{3.49} & \textbf{3.09} & 5.69 & 4.09 & 5.18 & 4.92 & 4.35 & 4.60 & \underline{3.97} & \textbf{2.93} \\
& $\times256$ & 3.88 & \underline{4.68} & \textbf{3.79} & 6.45 & 5.09 & 6.13 & 6.69 & \underline{4.24} & 5.20 & 4.64 & \textbf{3.70} \\

\midrule

\multirow{2}{*}{LRID - Outdoor}
& $\times128$ & 3.24 & \underline{3.94} & \textbf{3.59} & 6.26 & 4.44 & 5.36 & 4.80 & \underline{4.07} & 4.35 & 4.12 & \textbf{3.14} \\
& $\times256$ & 4.00 & \underline{5.55} & \textbf{4.55} & 7.18 & 5.73 & 6.08 & 6.31 & \underline{4.99} & 5.25 & 5.53 & \textbf{4.72} \\

\bottomrule
\end{tabular}%
}

\caption{\textbf{CIEDE2000~\cite{sharma2005ciede2000} ($\downarrow$) results on ELD~\cite{wei2021physics}, SID~\cite{chen2018learning}, and LRID~\cite{feng2022learnability}.} Complementary to the PSNR results reported in \cref{tab:main_results}.  Bold and underlined values denote the best and second-best results within each group, respectively. Results for calibration-based models are obtained by evaluating their checkpoints on the corresponding dataset. $^{\dagger}$Calibrated darkshading noise subtraction is unavailable.}
\label{tab:supp_ciede2000_results}
\end{table*}

\cref{tab:main_results} reports PSNR results for our method and the baselines. Due to space constraints, the corresponding SSIM values are provided separately in \cref{tab:supp_ssim_results}. The trends observed under SSIM closely follow those under PSNR, and the conclusions drawn in the main paper remain consistent. In particular, our method outperforms all blind baselines on ELD-Sony, ELD-Nikon, and SID across all exposure ratios. On ELD-Nikon, it further achieves state-of-the-art performance among all evaluated methods, including the calibrated baselines, as dark shading correction is unavailable for this sensor and the calibrated models do not explicitly account for black-level error. On LRID, our method surpasses PGRQB and performs on par with PG and PGRQ.

Because our method specifically targets the color-shift problem, we additionally report CIEDE2000~\cite{sharma2005ciede2000} in \cref{tab:supp_ciede2000_results}, which measures the perceptual color accuracy between the predicted and ground-truth images (computed post-ISP). The trends observed under CIEDE2000 closely follow those under PSNR and SSIM, and the conclusions drawn in the main paper remain consistent. In particular, our method consistently outperforms PGRQB across all datasets. On SID, ELD-Sony, and ELD-Nikon, it further achieves state-of-the-art performance among all blind approaches, mirroring the trends observed under PSNR and SSIM.

\subsection{Qualitative Results}
\label{sec:supp_qualitative_results}
\begin{figure*}[t]
\centering

\subfloat[Noisy]{%
    \includegraphics[width=0.24\textwidth]{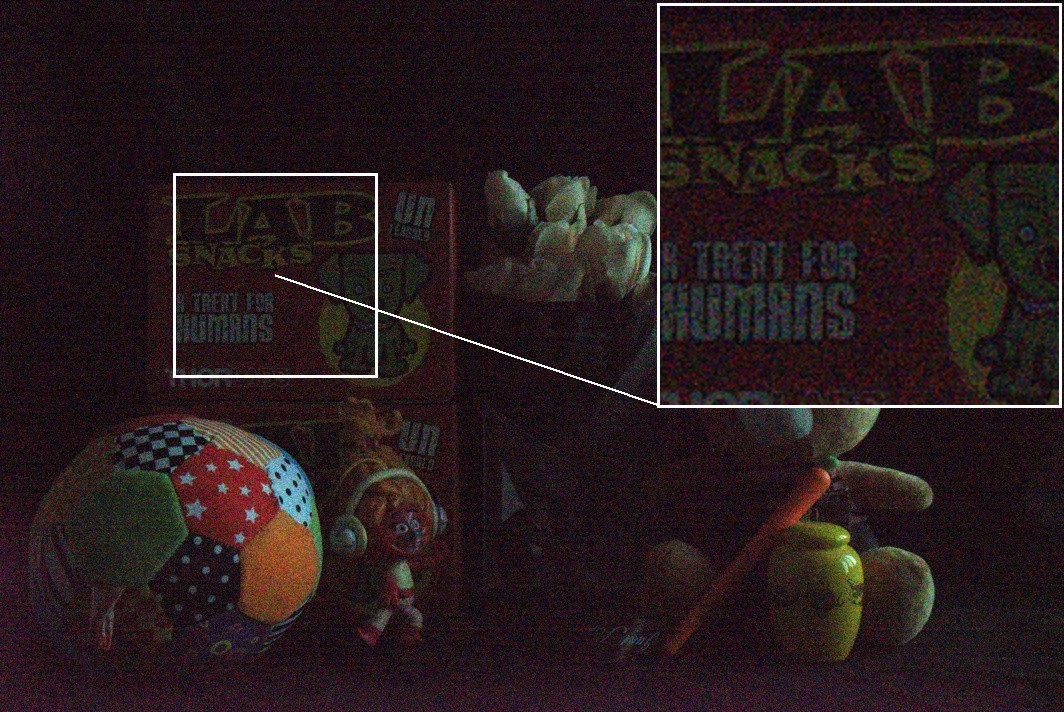}
}
\subfloat[PG]{%
    \includegraphics[width=0.24\textwidth]{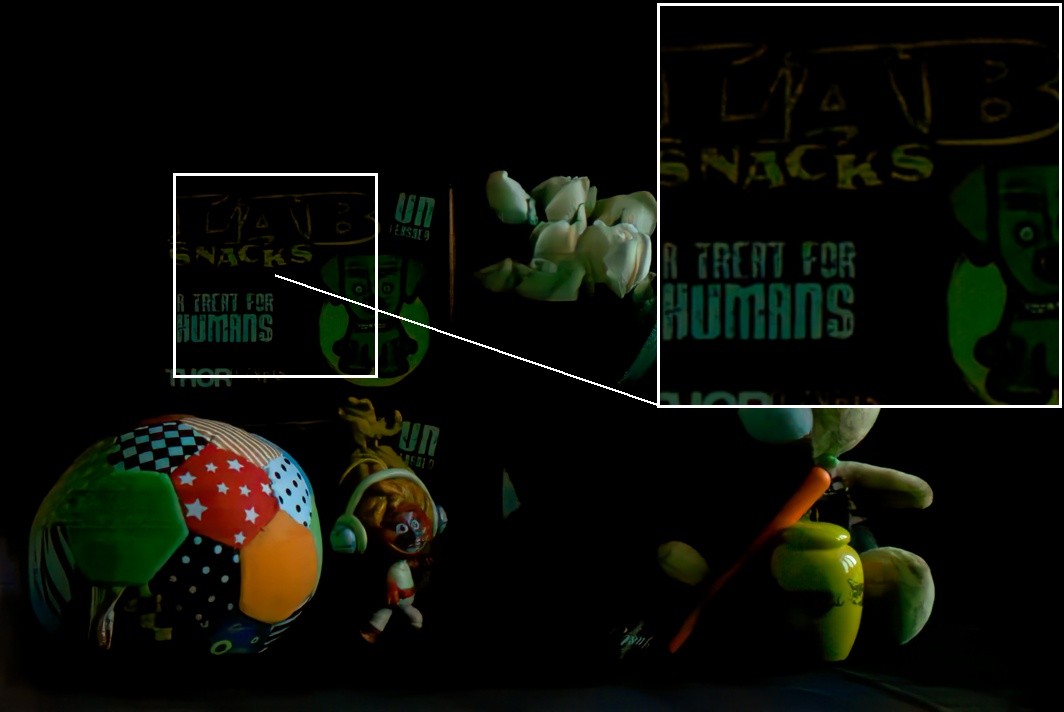}
}
\subfloat[PGRQ]{%
    \includegraphics[width=0.24\textwidth]{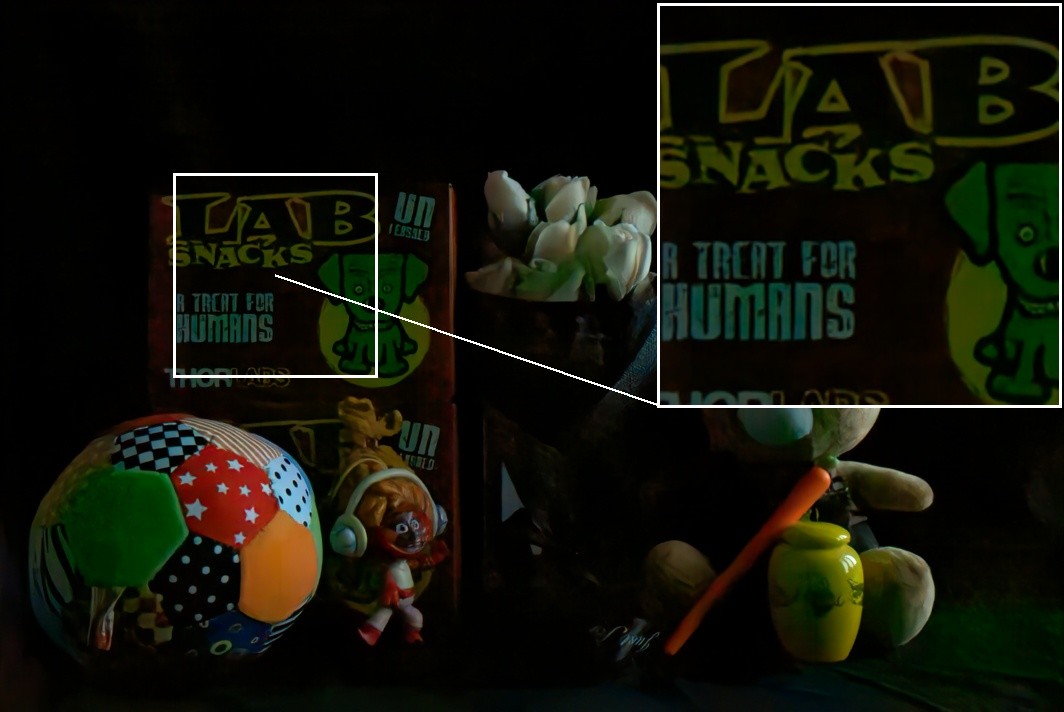}
}
\subfloat[PGRQB]{%
    \includegraphics[width=0.24\textwidth]{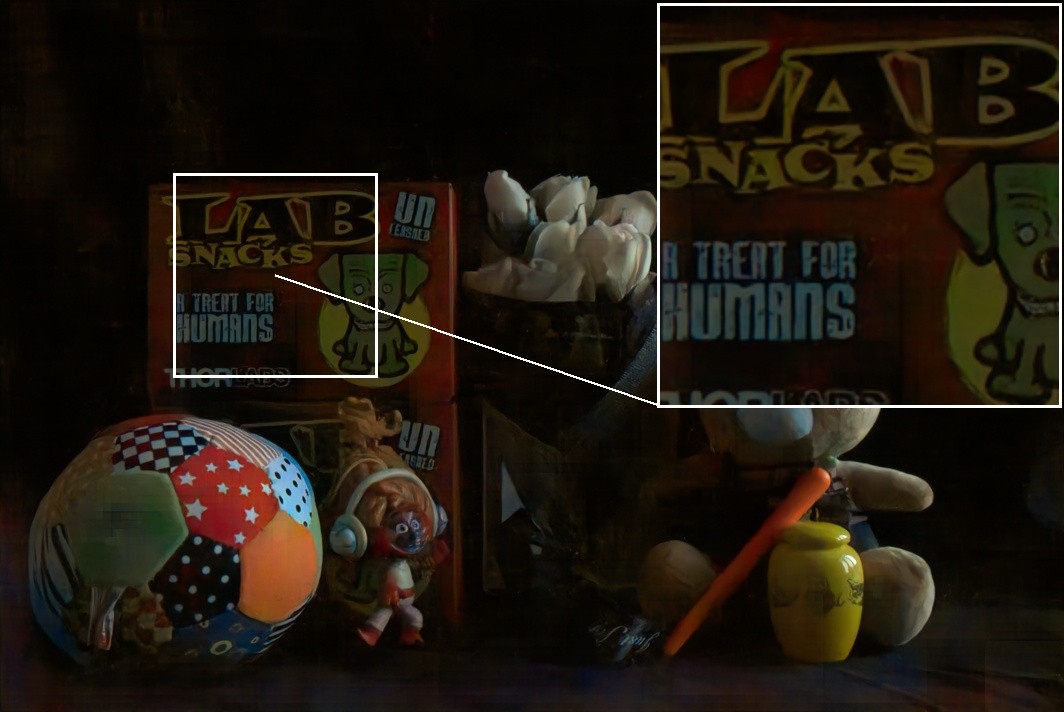}
}\\

\vspace{2pt}

\subfloat[ELD~\cite{wei2021physics}]{%
    \includegraphics[width=0.24\textwidth]{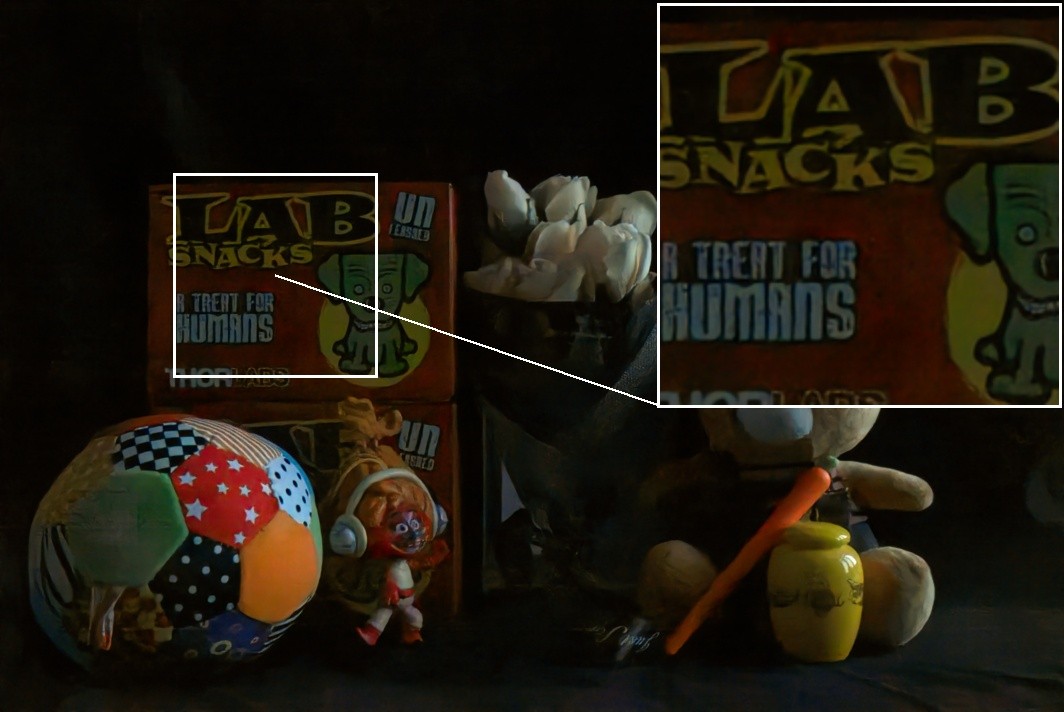}
}
\subfloat[NoiseDiff~\cite{lu2025dark}]{%
    \includegraphics[width=0.24\textwidth]{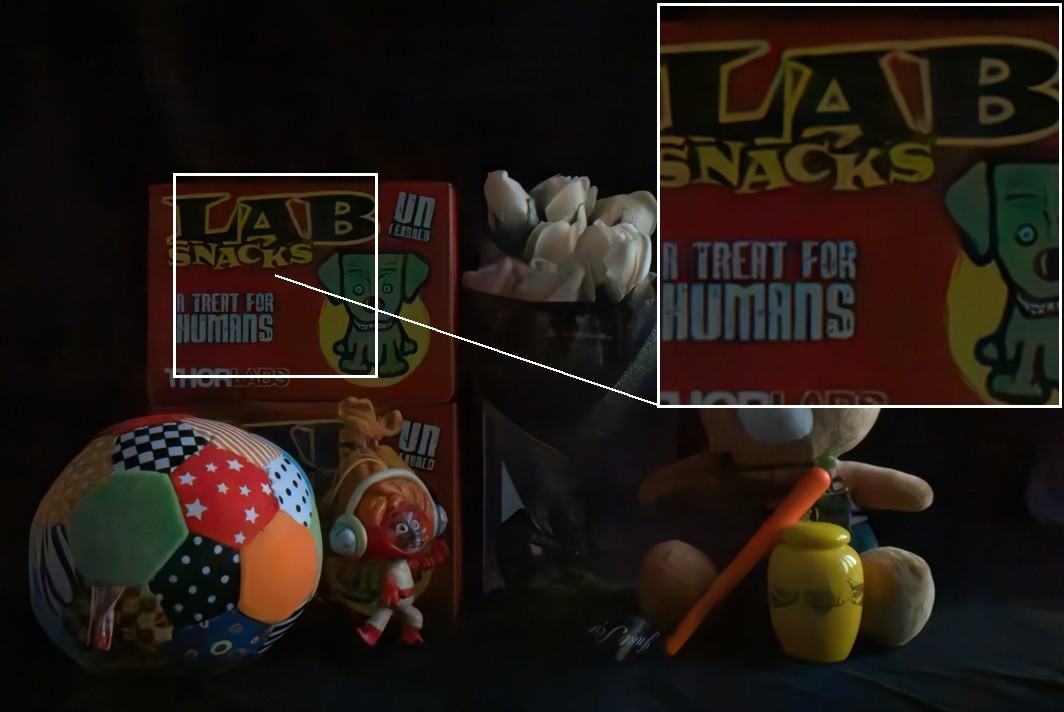}
}
\subfloat[Ours]{%
    \includegraphics[width=0.24\textwidth]{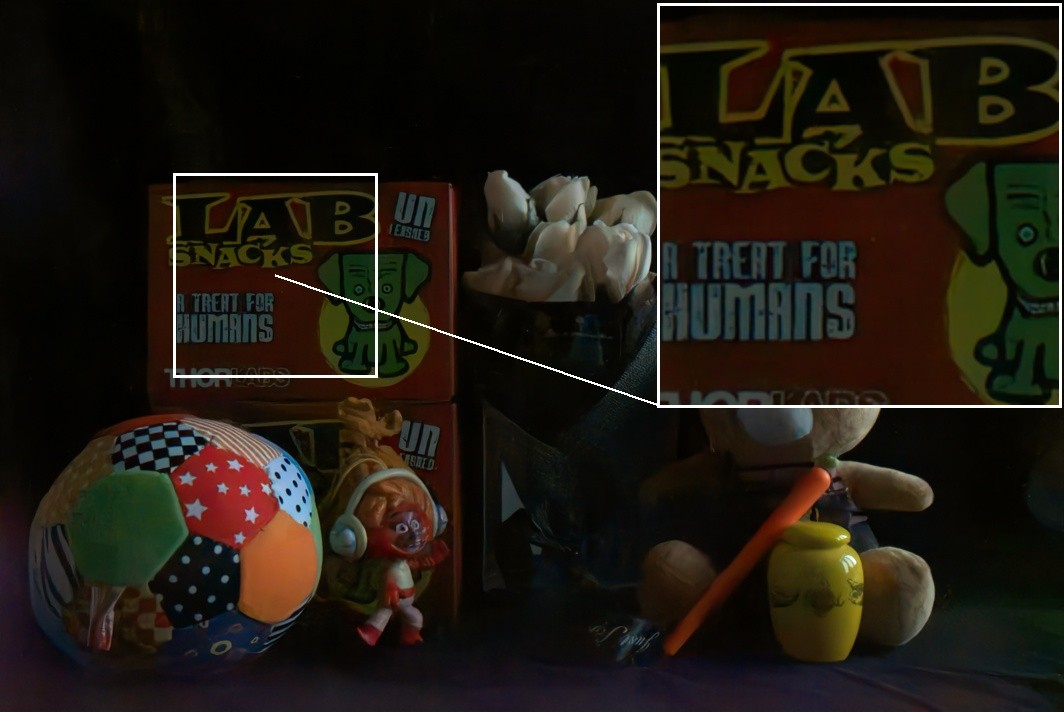}
}
\subfloat[GT]{%
    \includegraphics[width=0.24\textwidth]{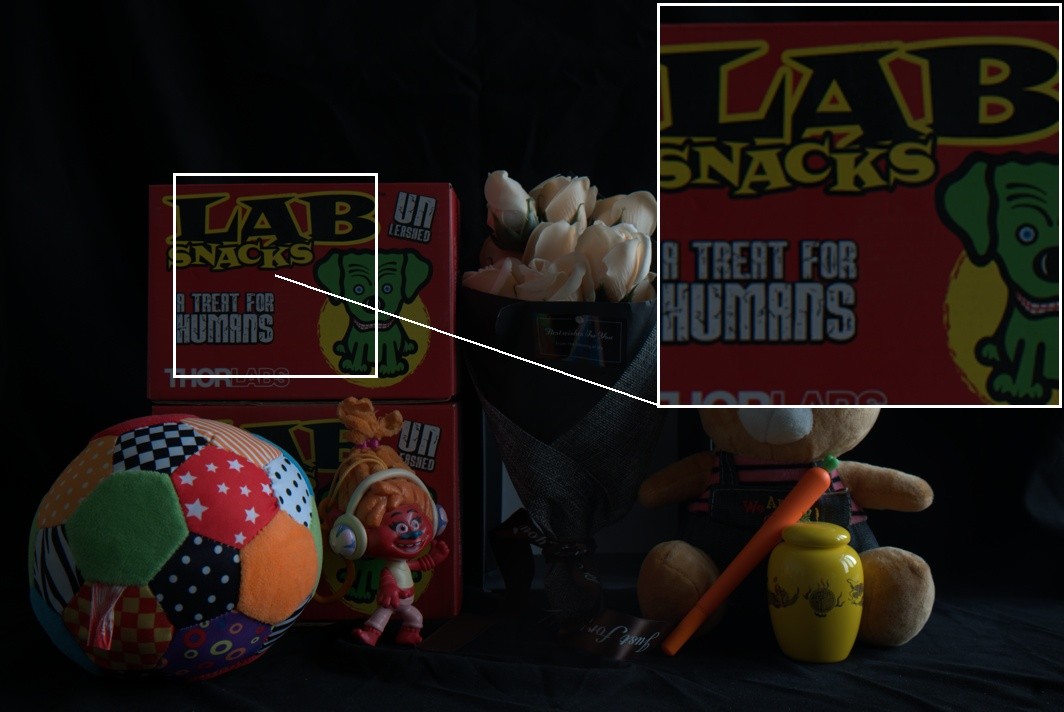}
}\\

\vspace{4pt}

\subfloat[Noisy]{%
    \includegraphics[width=0.24\textwidth]{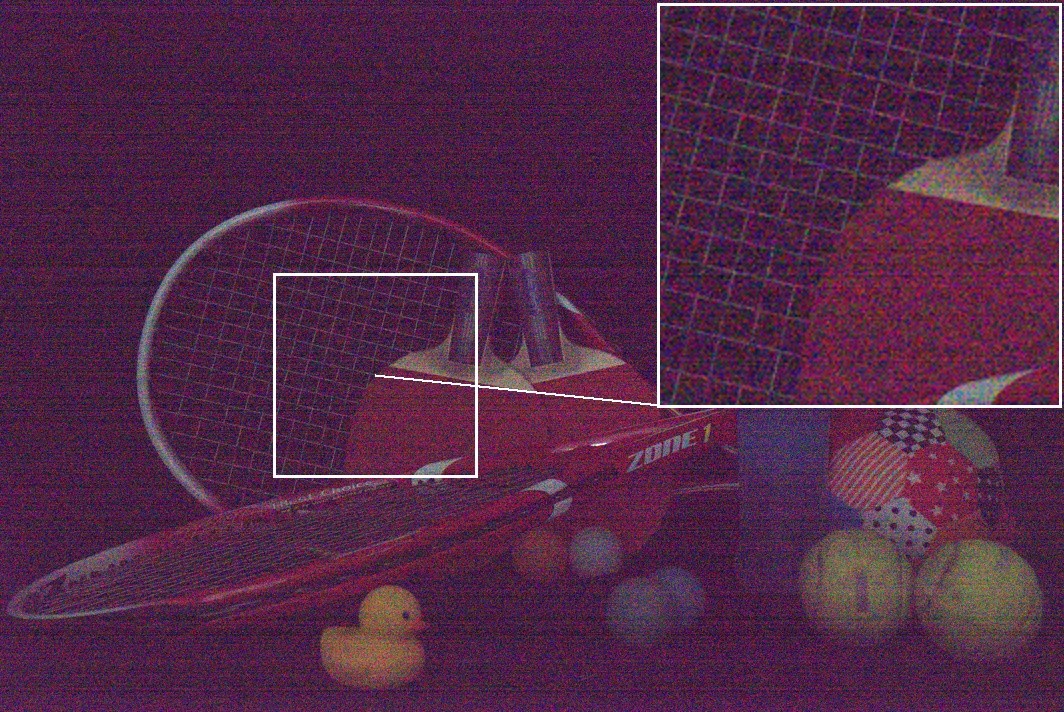}
}
\subfloat[PG]{%
    \includegraphics[width=0.24\textwidth]{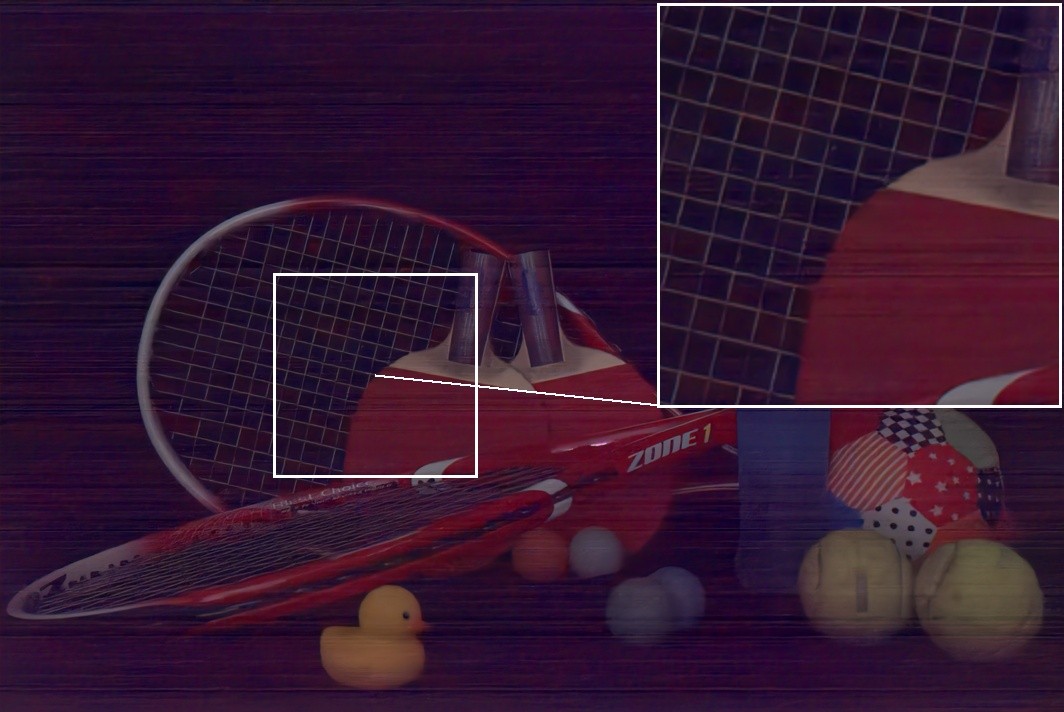}
}
\subfloat[PGRQ]{%
    \includegraphics[width=0.24\textwidth]{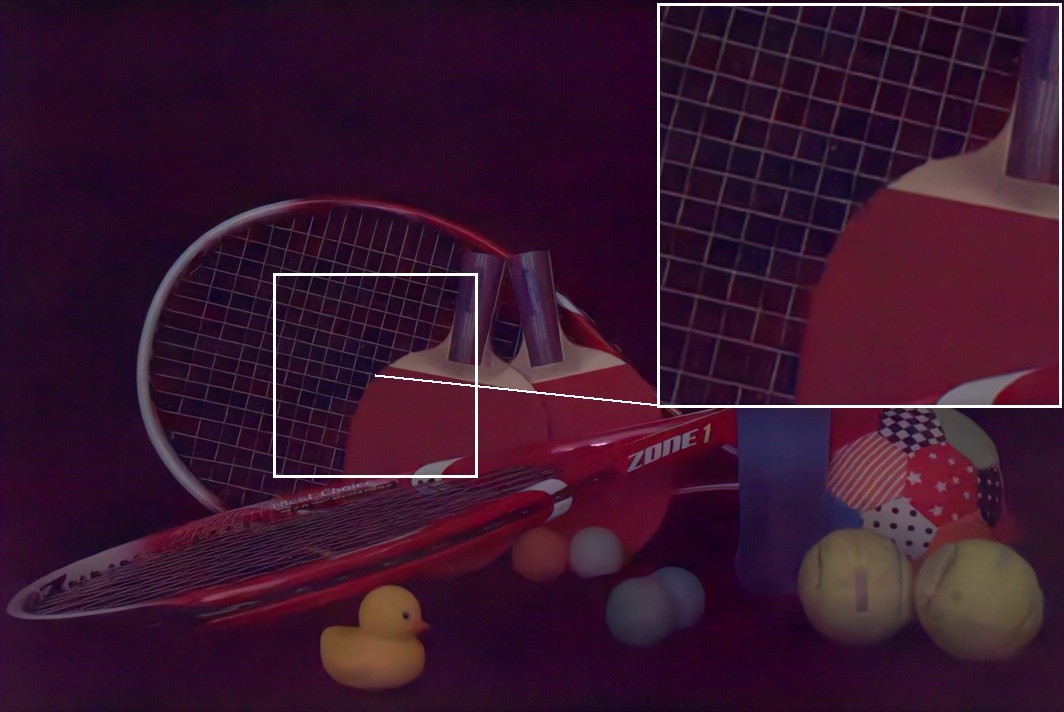}
    }
\subfloat[PGRQB]{%
    \includegraphics[width=0.24\textwidth]{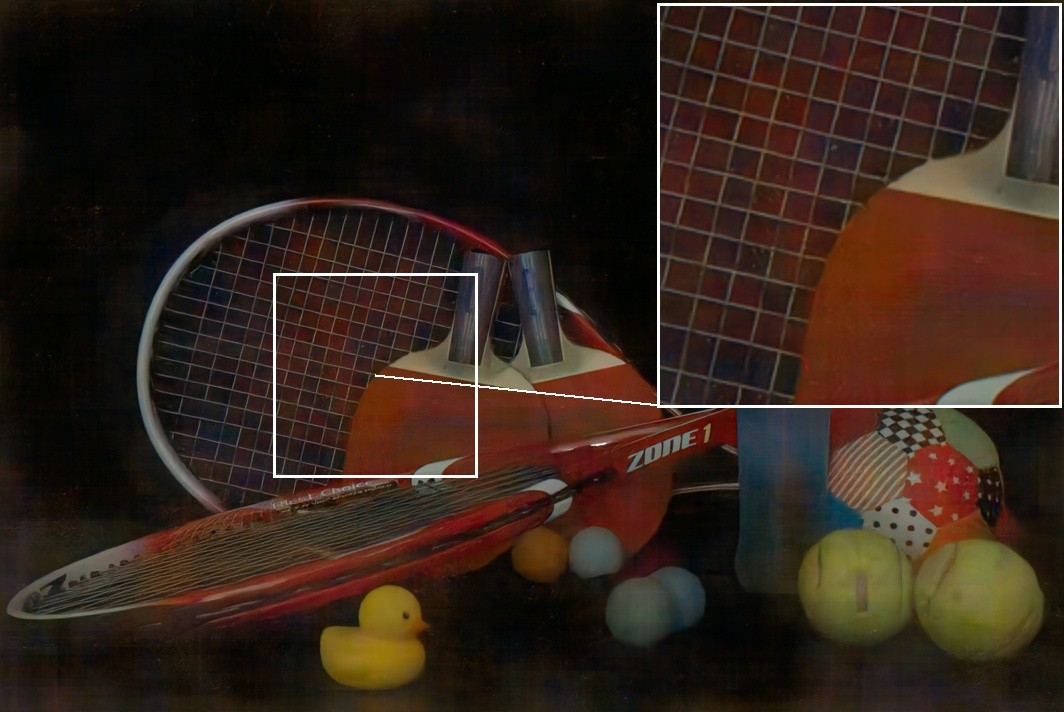}
}\\

\vspace{2pt}

\subfloat[ELD~\cite{wei2021physics}]{%
    \includegraphics[width=0.24\textwidth]{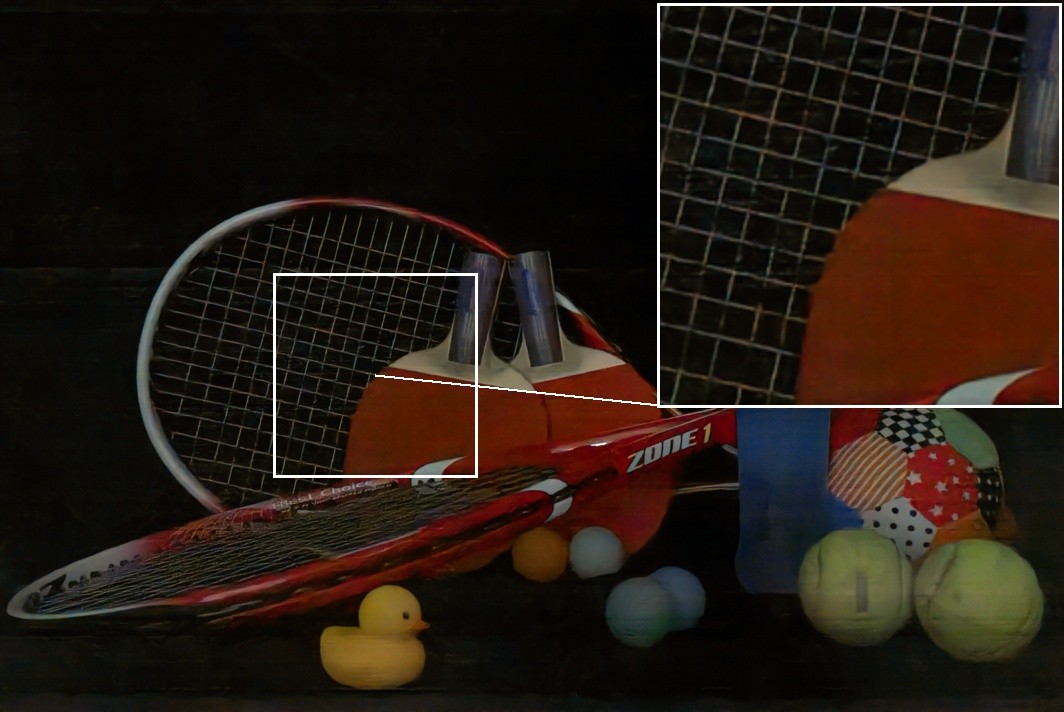}
}
\subfloat[NoiseDiff~\cite{lu2025dark}]{%
    \includegraphics[width=0.24\textwidth]{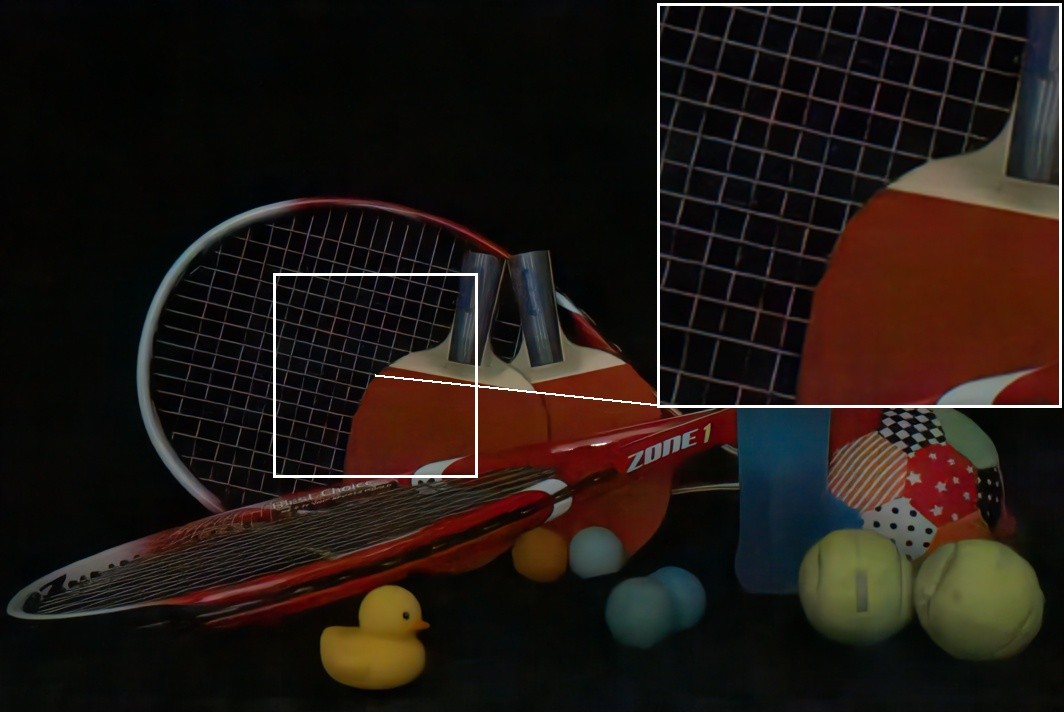}
}
\subfloat[Ours]{%
    \includegraphics[width=0.24\textwidth]{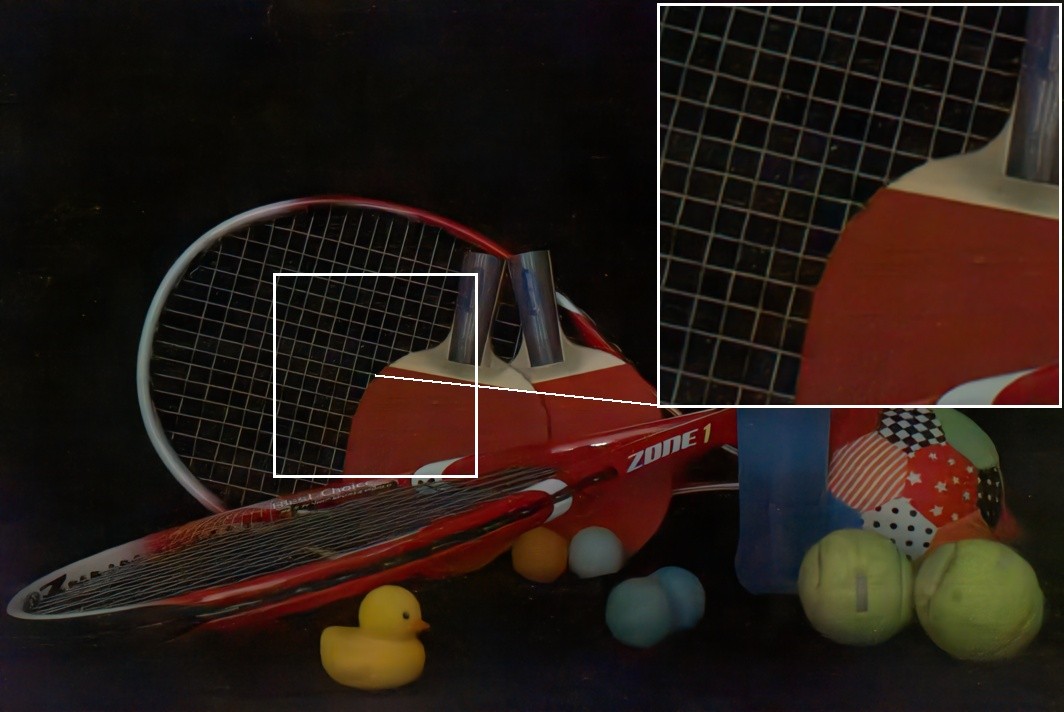}
}
\subfloat[GT]{%
    \includegraphics[width=0.24\textwidth]{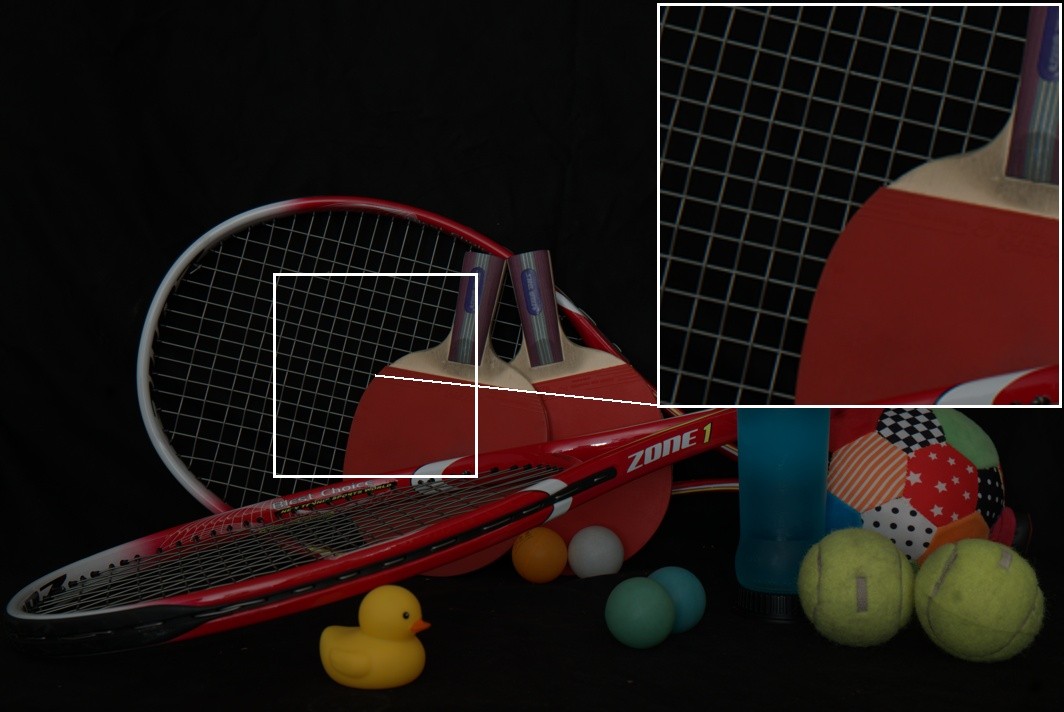}
}

\caption{\textbf{Additional qualitative comparisons on ELD-Sony~\cite{wei2021physics}.} Two scenes are shown across four rows (blind baselines on top, calibrated baselines and ours on the bottom of each scene). PG and PGRQ exhibit residual color bias, and PGRQB introduces local color distortions, whereas our method closely matches the ground truth without target-sensor data (see zoomed insets).}
\label{fig:supp_results_eld_sony}

\end{figure*}

\begin{figure*}[t]
\centering

\subfloat[Noisy]{%
    \includegraphics[width=0.24\textwidth]{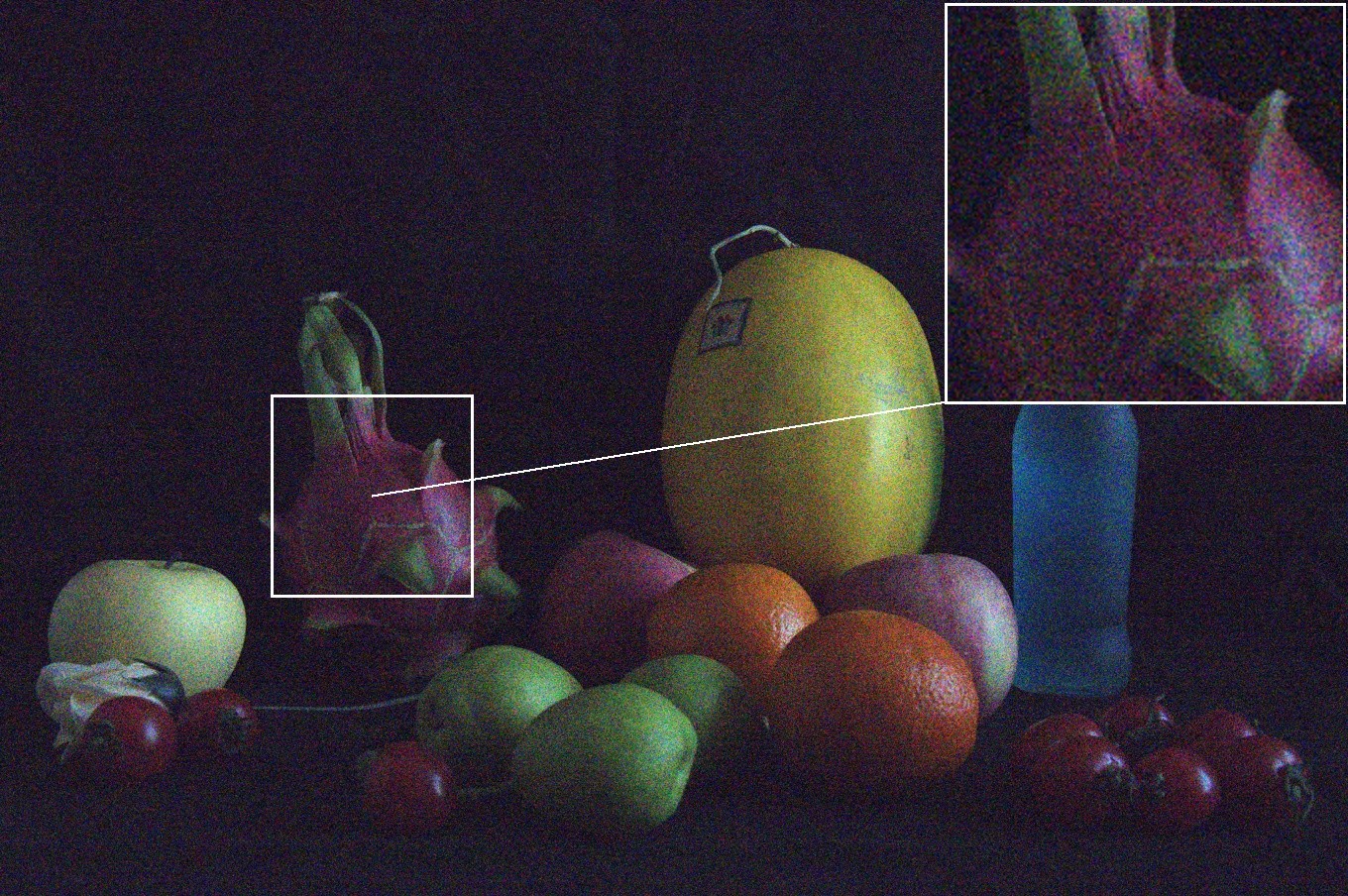}
}
\subfloat[PG]{%
    \includegraphics[width=0.24\textwidth]{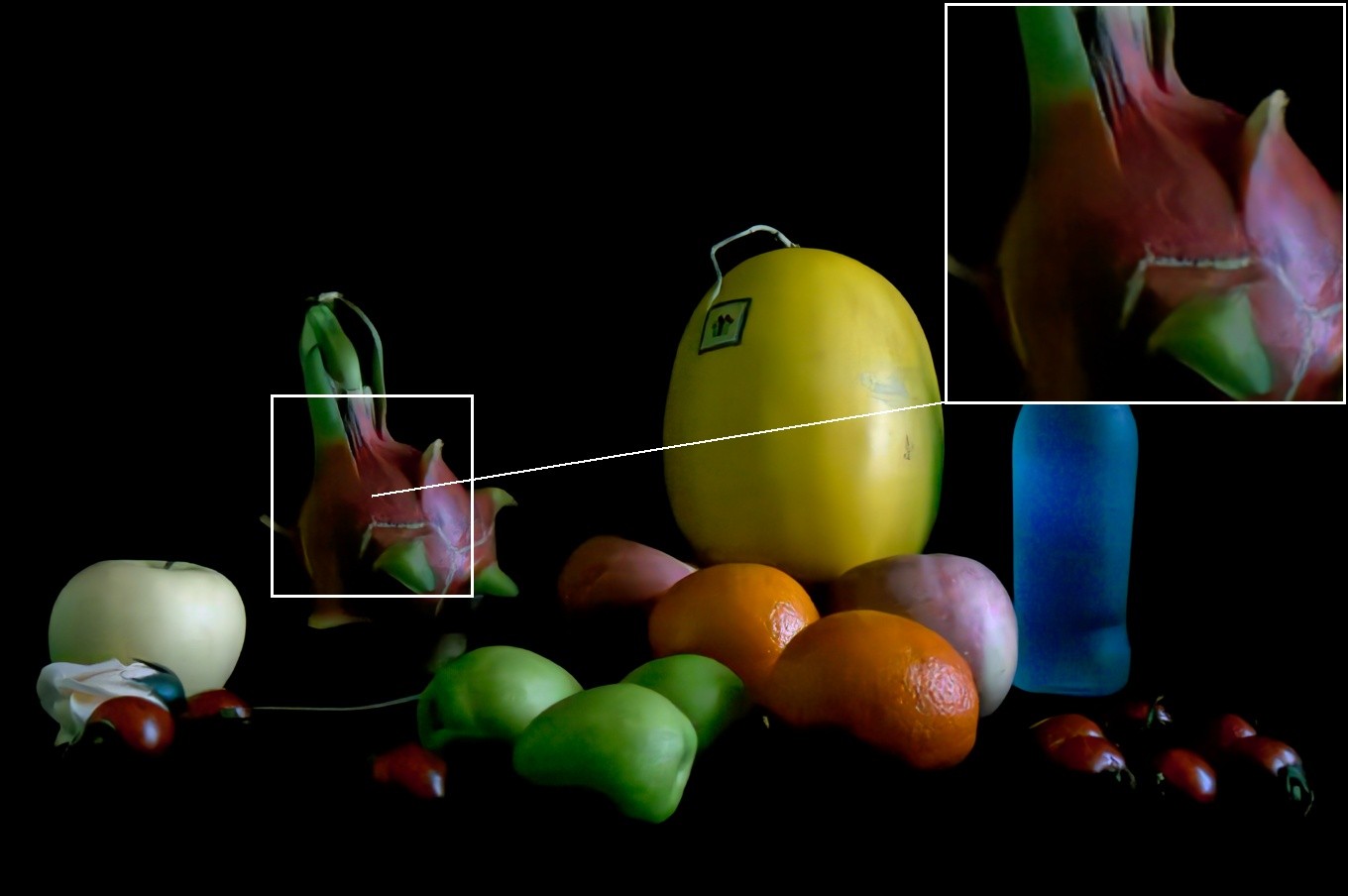}
}
\subfloat[PGRQ]{%
    \includegraphics[width=0.24\textwidth]{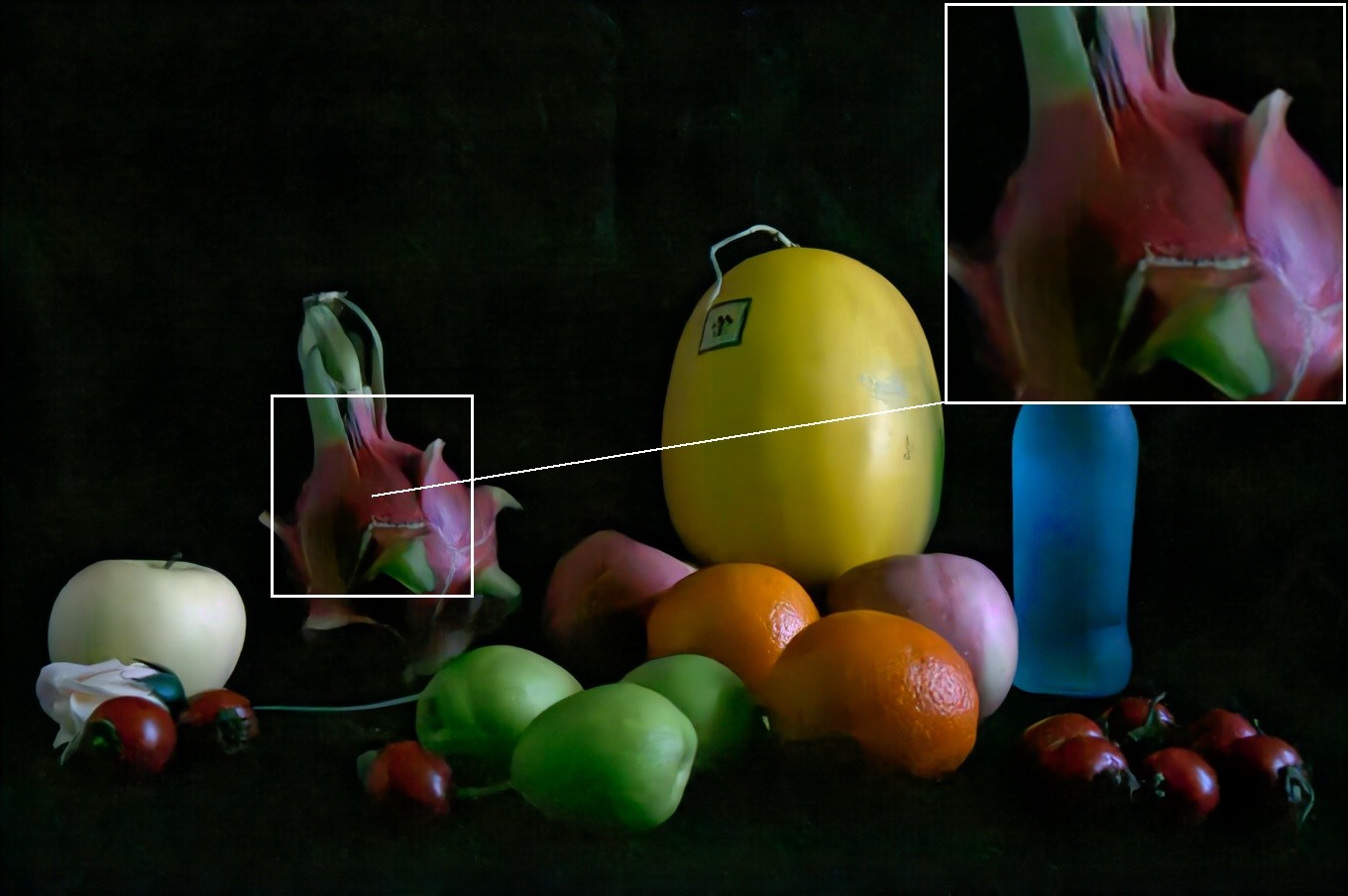}
}
\subfloat[PGRQB]{%
    \includegraphics[width=0.24\textwidth]{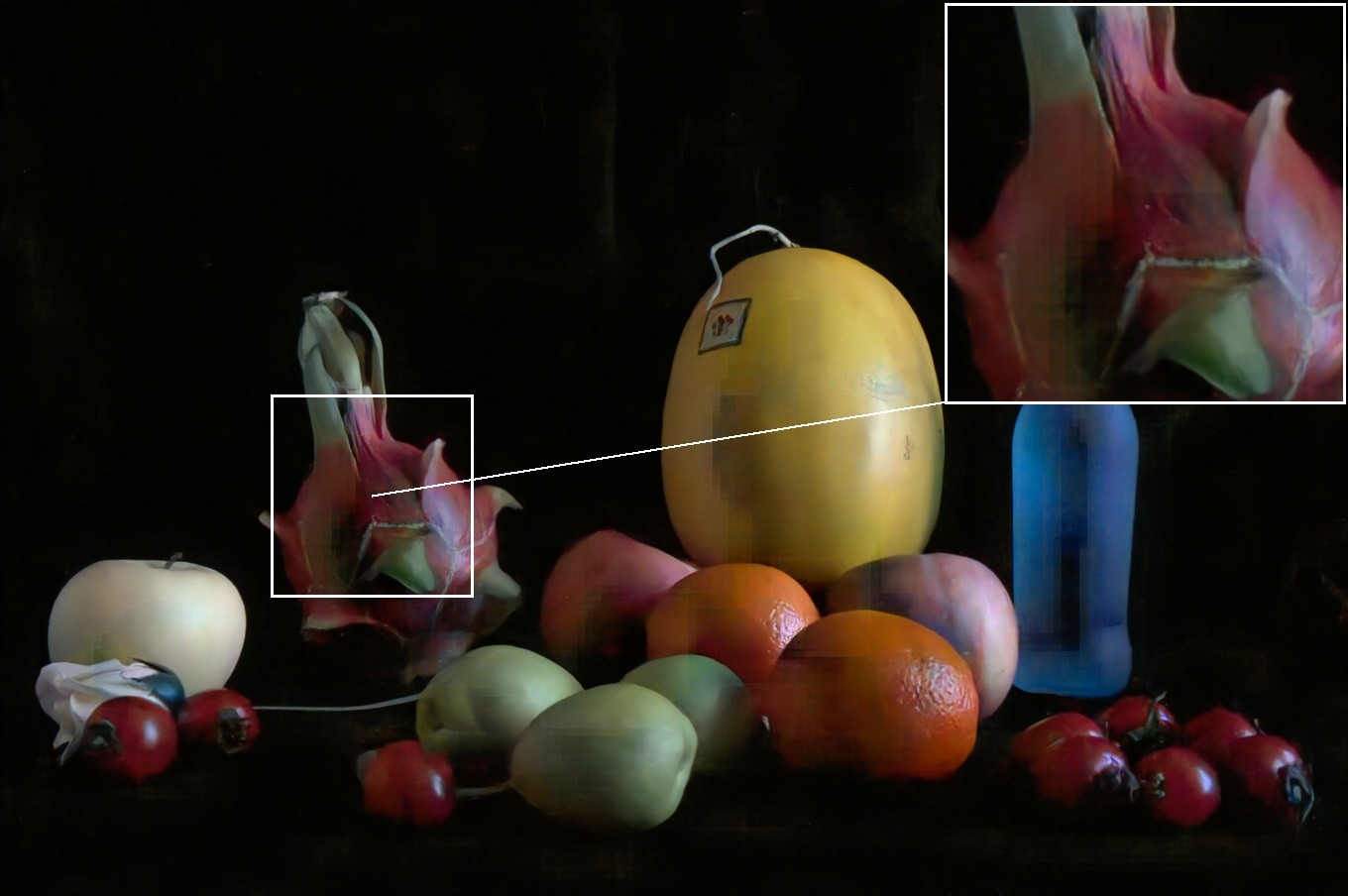}
}\\

\vspace{2pt}

\subfloat[ELD~\cite{wei2021physics}]{%
    \includegraphics[width=0.24\textwidth]{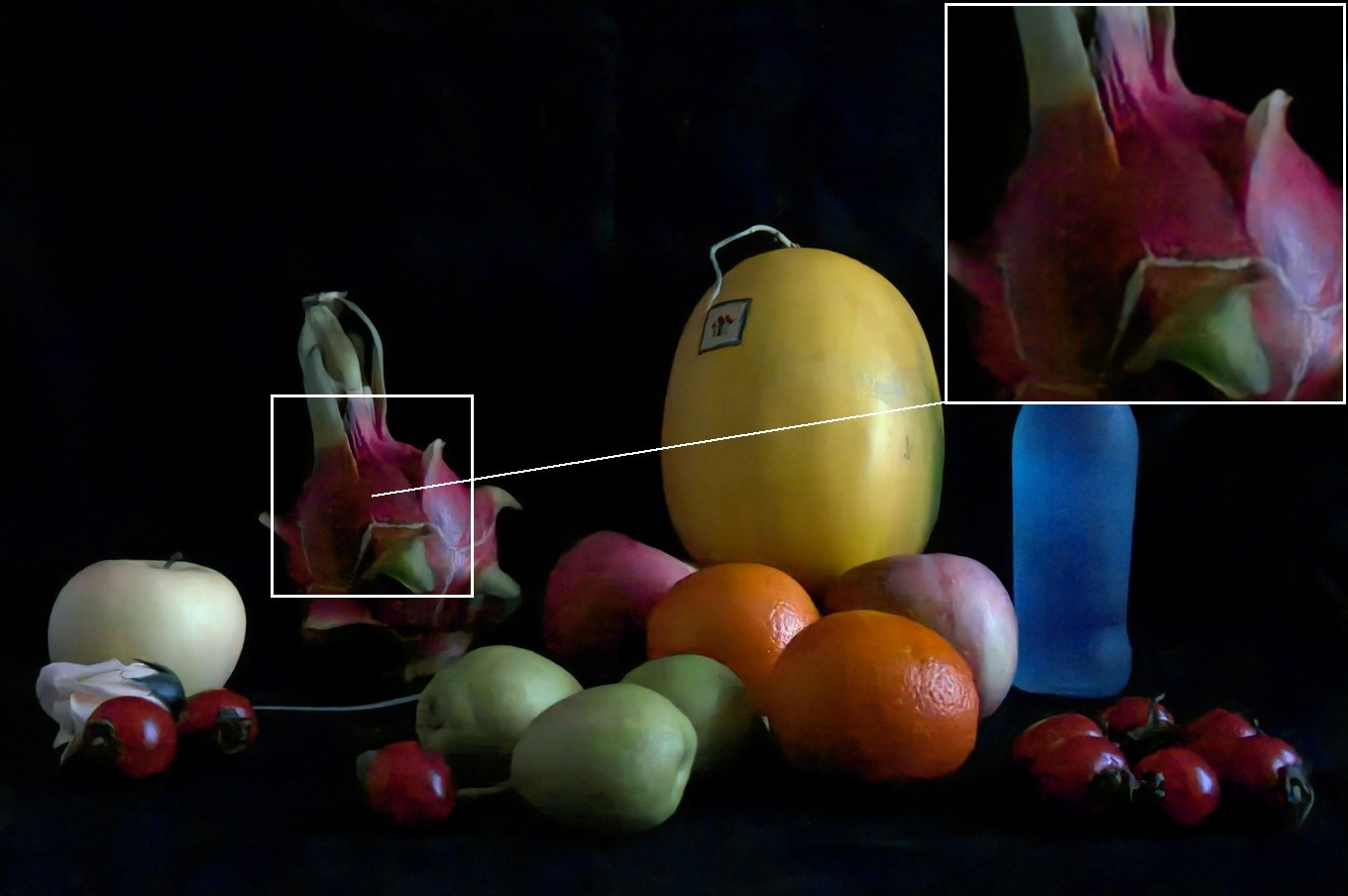}
}
\subfloat[2-Shots~\cite{lu20252}]{%
    \includegraphics[width=0.24\textwidth]{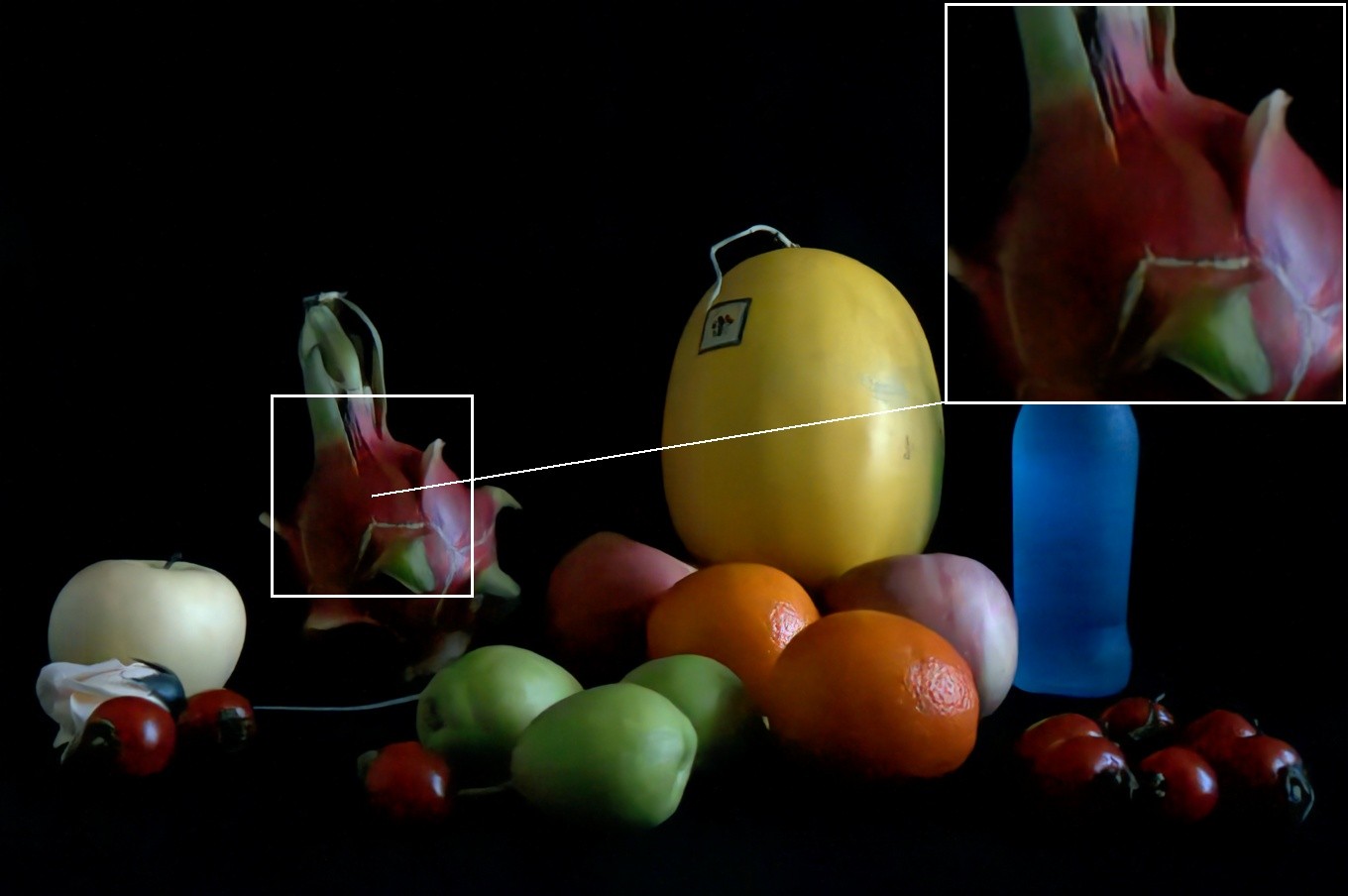}
}
\subfloat[Ours]{%
    \includegraphics[width=0.24\textwidth]{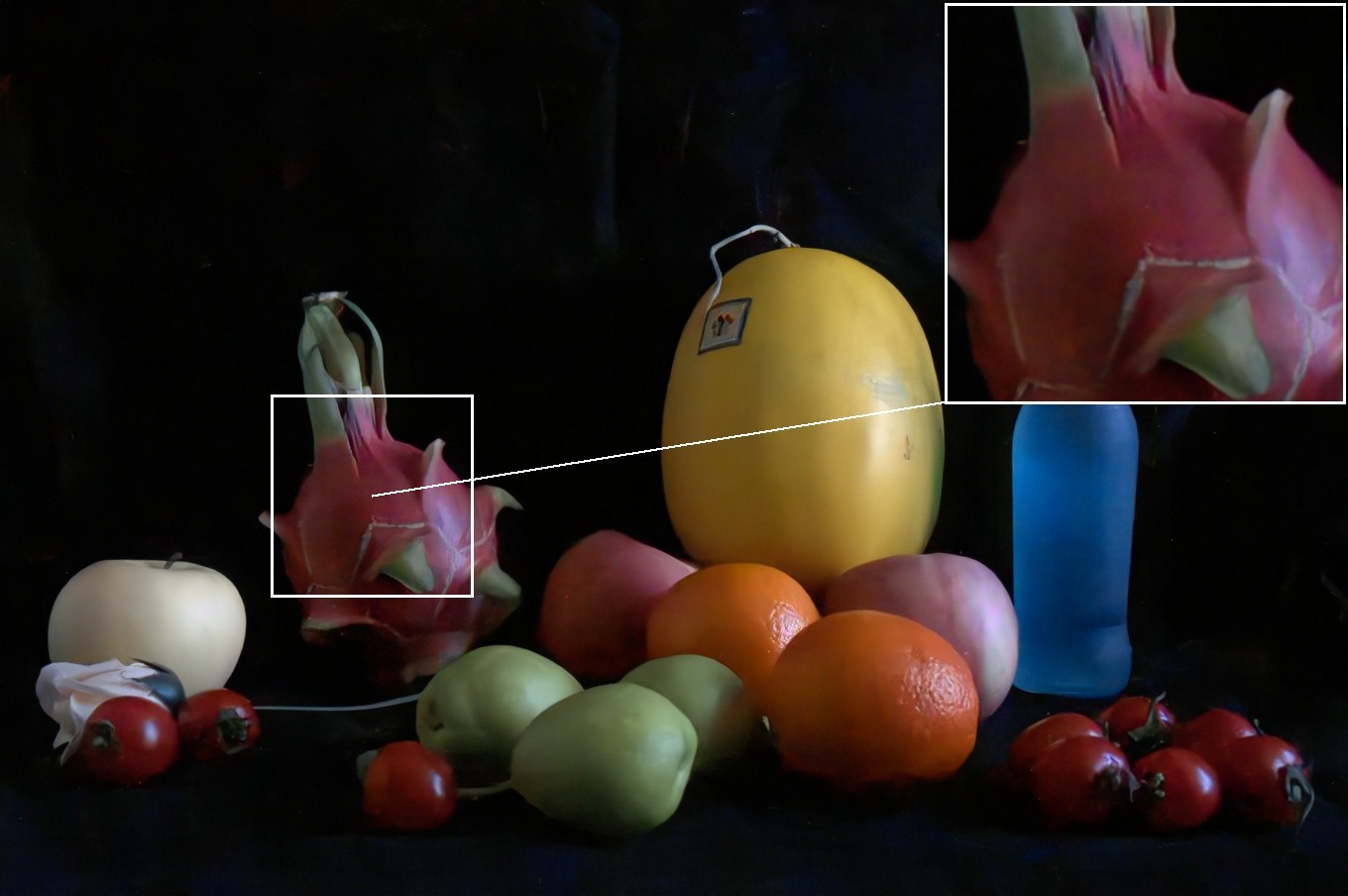}
}
\subfloat[GT]{%
    \includegraphics[width=0.24\textwidth]{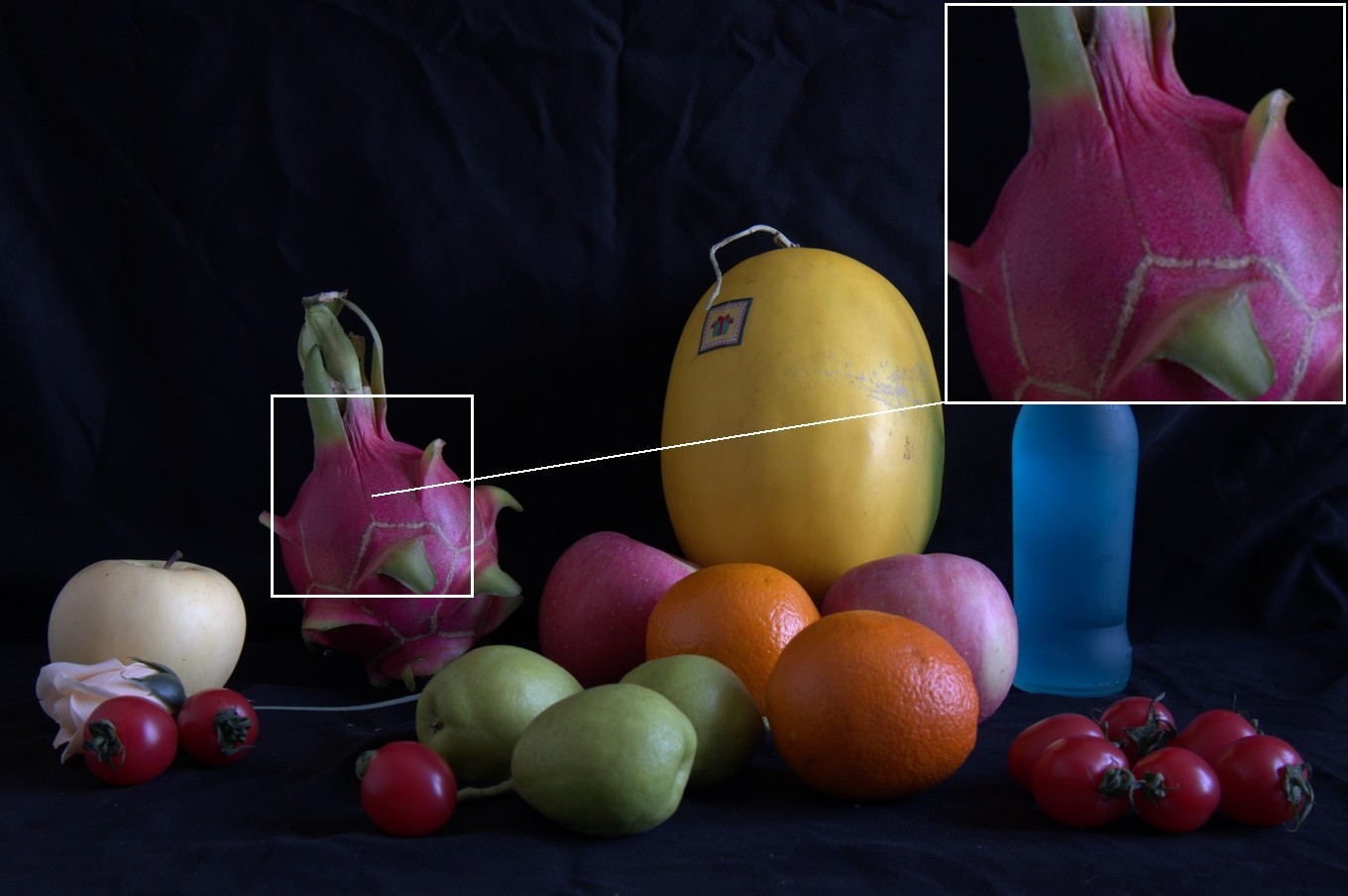}
}\\

\vspace{4pt}

\subfloat[Noisy]{%
    \includegraphics[width=0.24\textwidth]{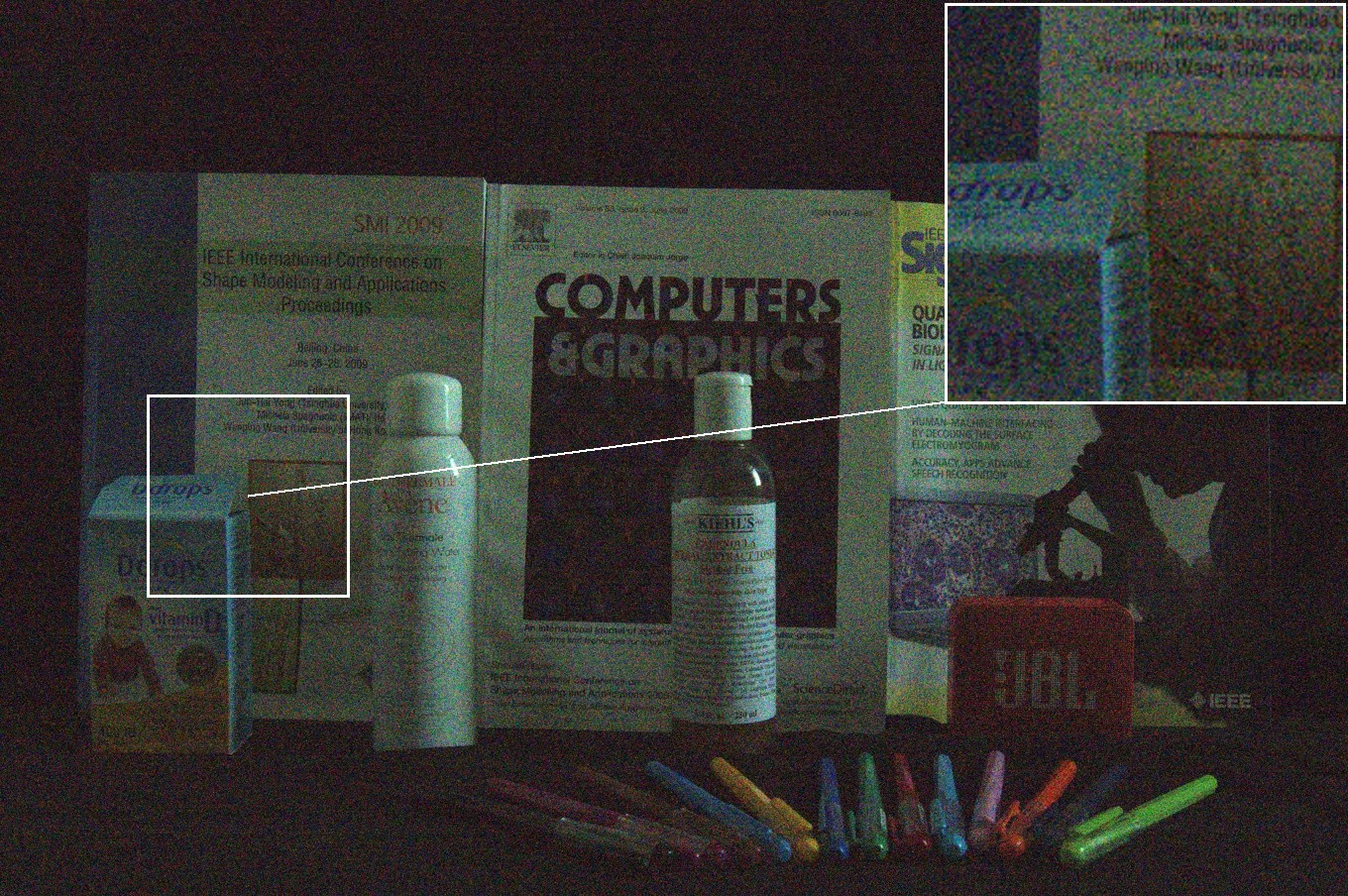}
}
\subfloat[PG]{%
    \includegraphics[width=0.24\textwidth]{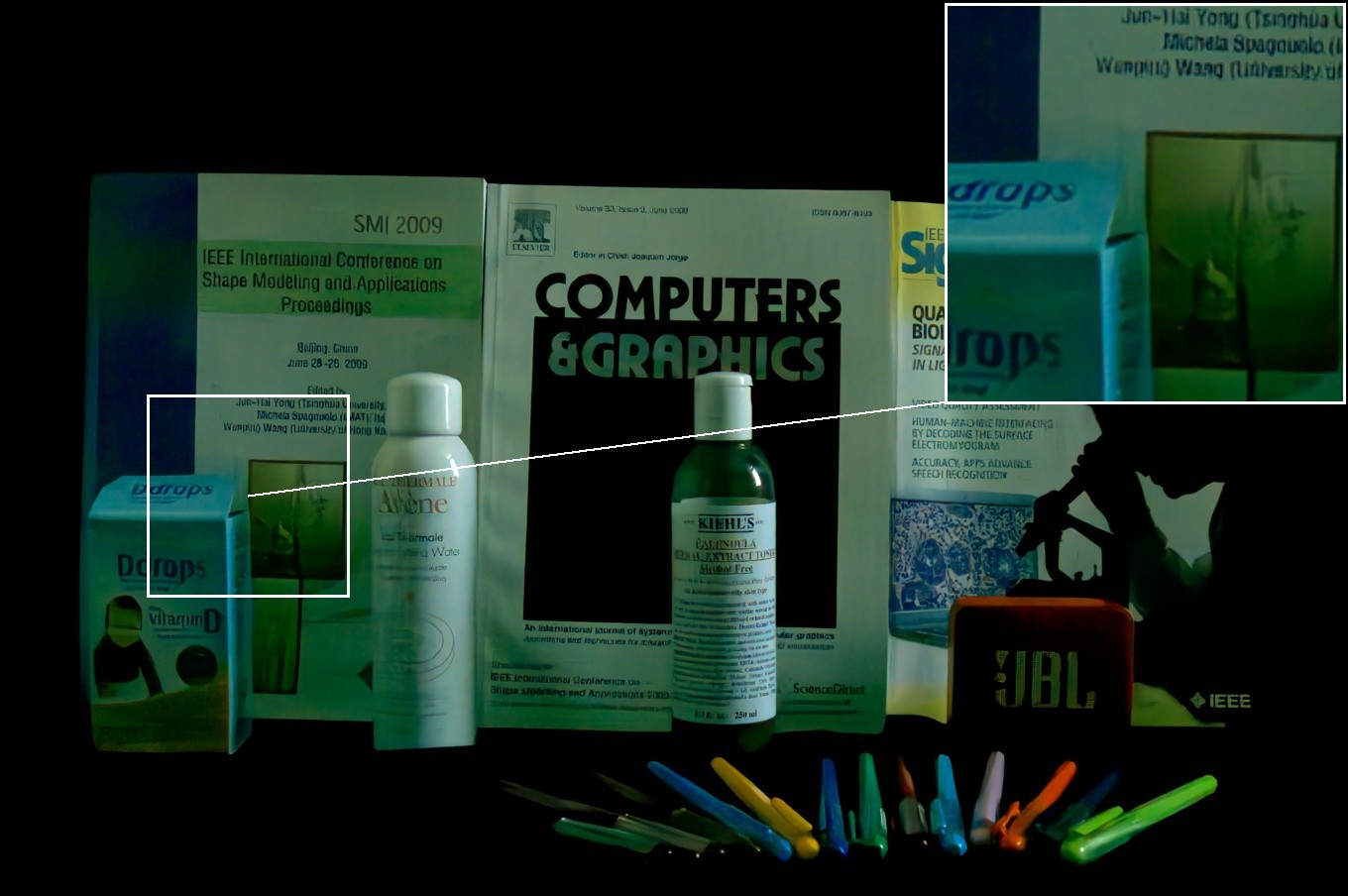}
}
\subfloat[PGRQ]{%
    \includegraphics[width=0.24\textwidth]{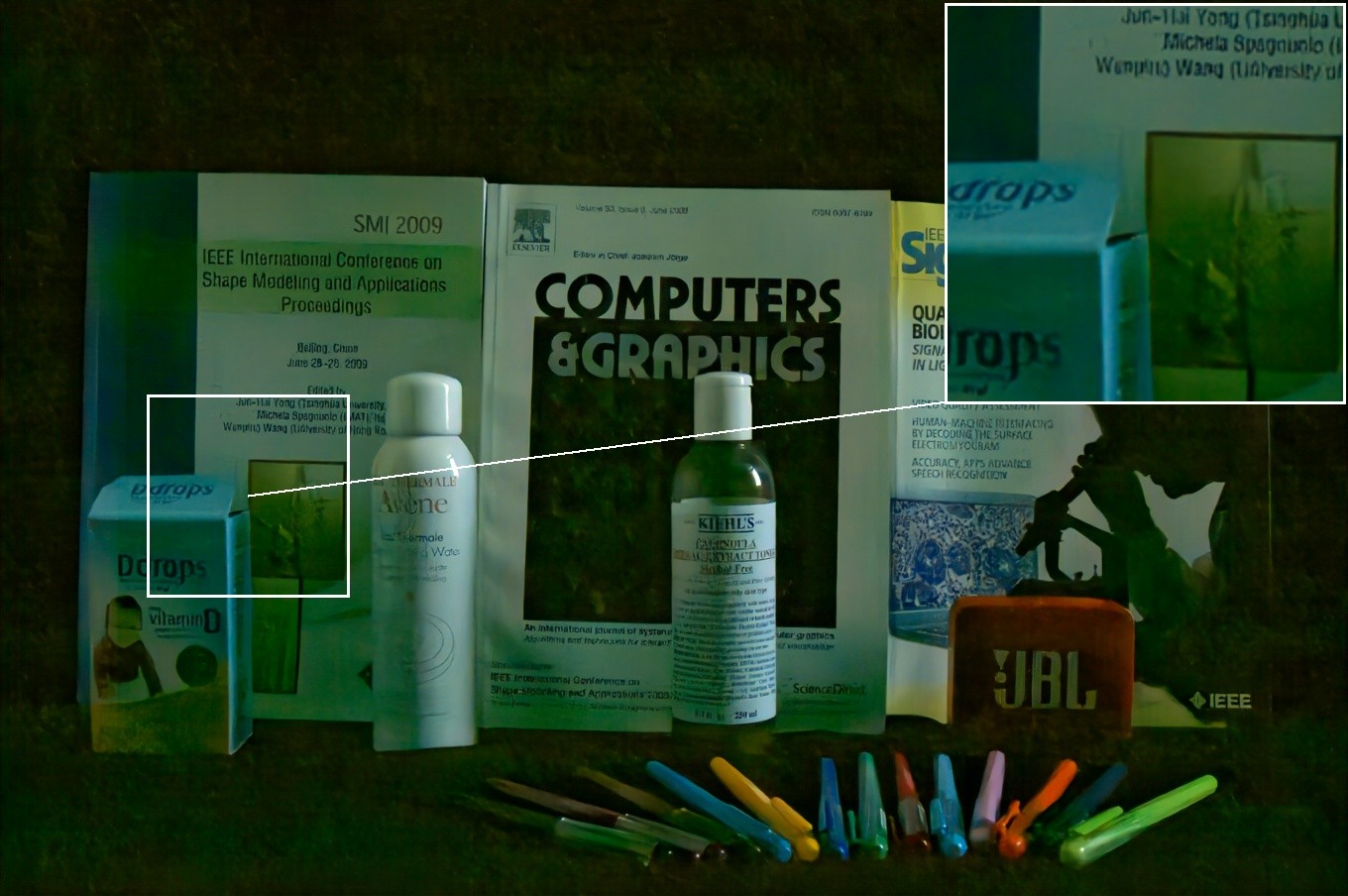}
}
\subfloat[PGRQB]{%
    \includegraphics[width=0.24\textwidth]{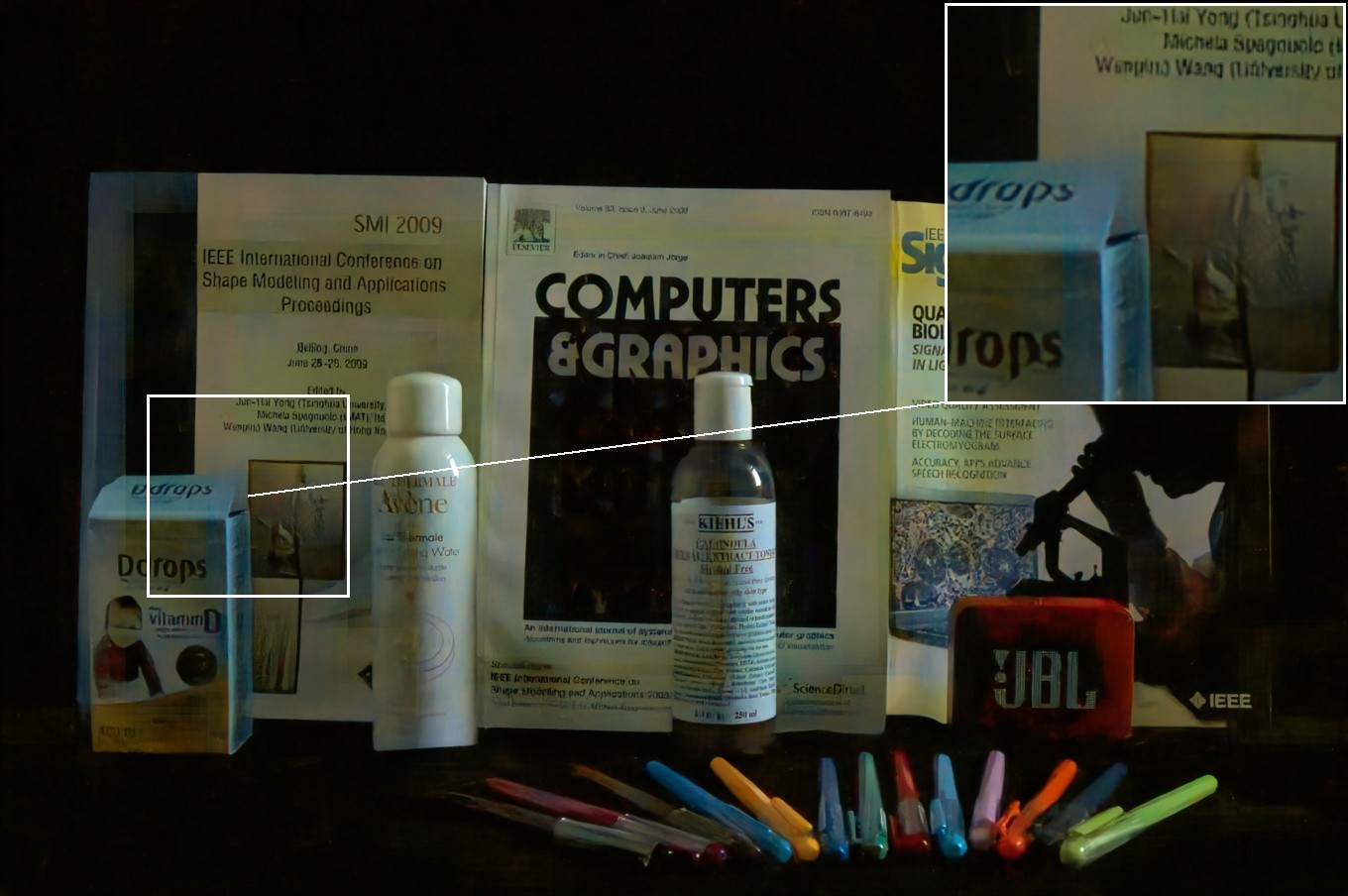}
}\\

\vspace{2pt}

\subfloat[ELD~\cite{wei2021physics}]{%
    \includegraphics[width=0.24\textwidth]{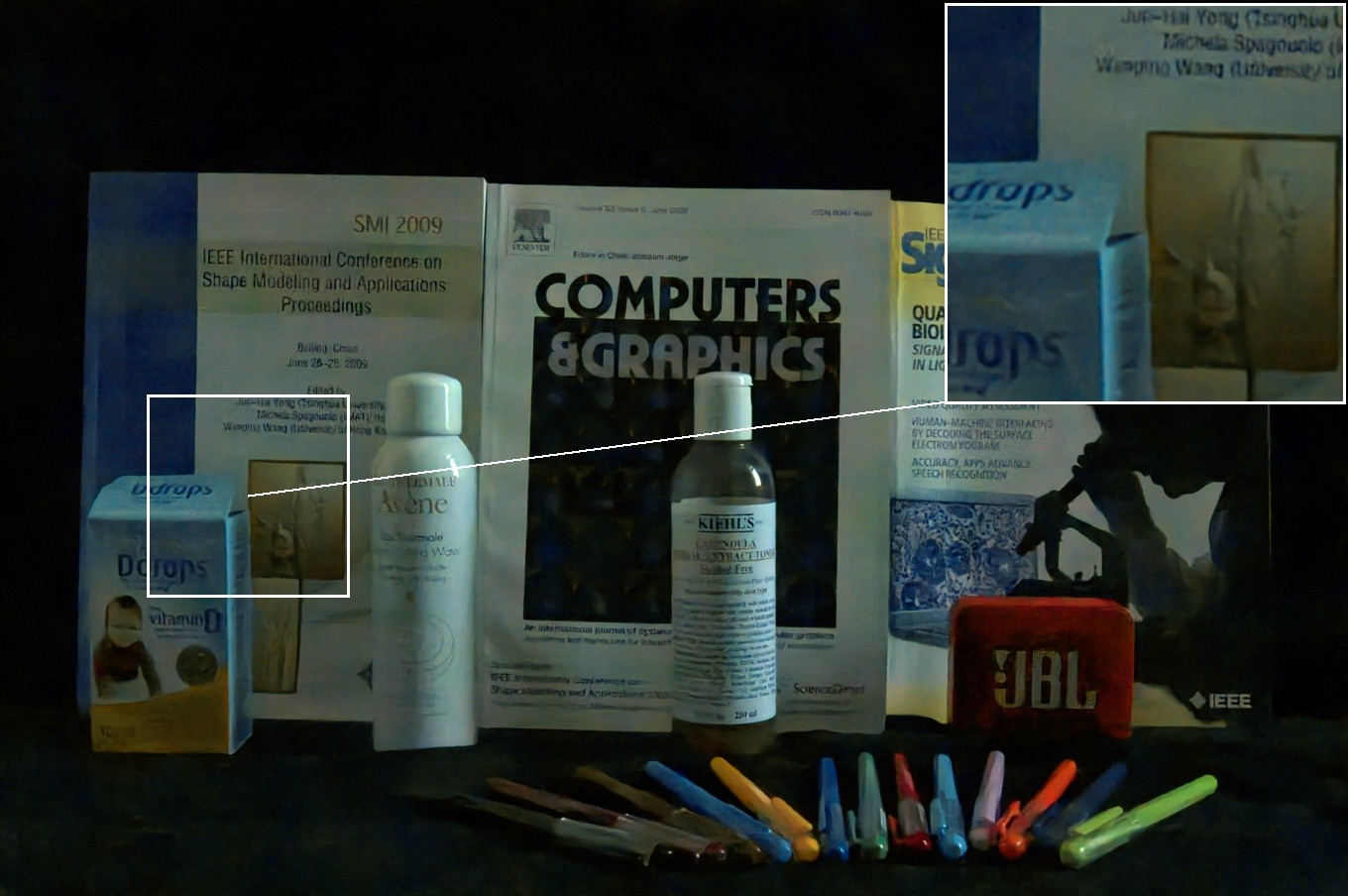}
}
\subfloat[2-Shots~\cite{lu20252}]{%
    \includegraphics[width=0.24\textwidth]{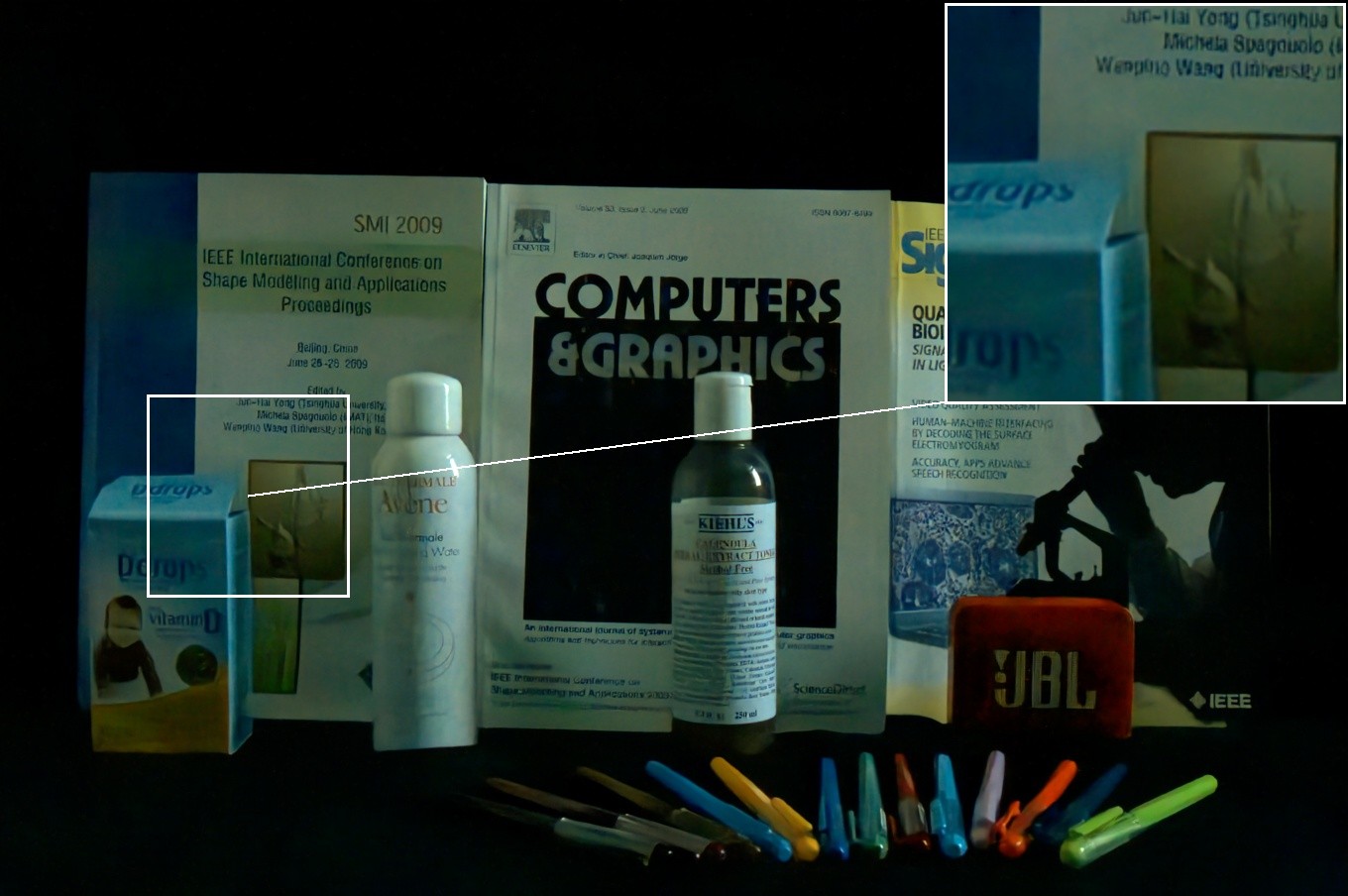}
}
\subfloat[Ours]{%
    \includegraphics[width=0.24\textwidth]{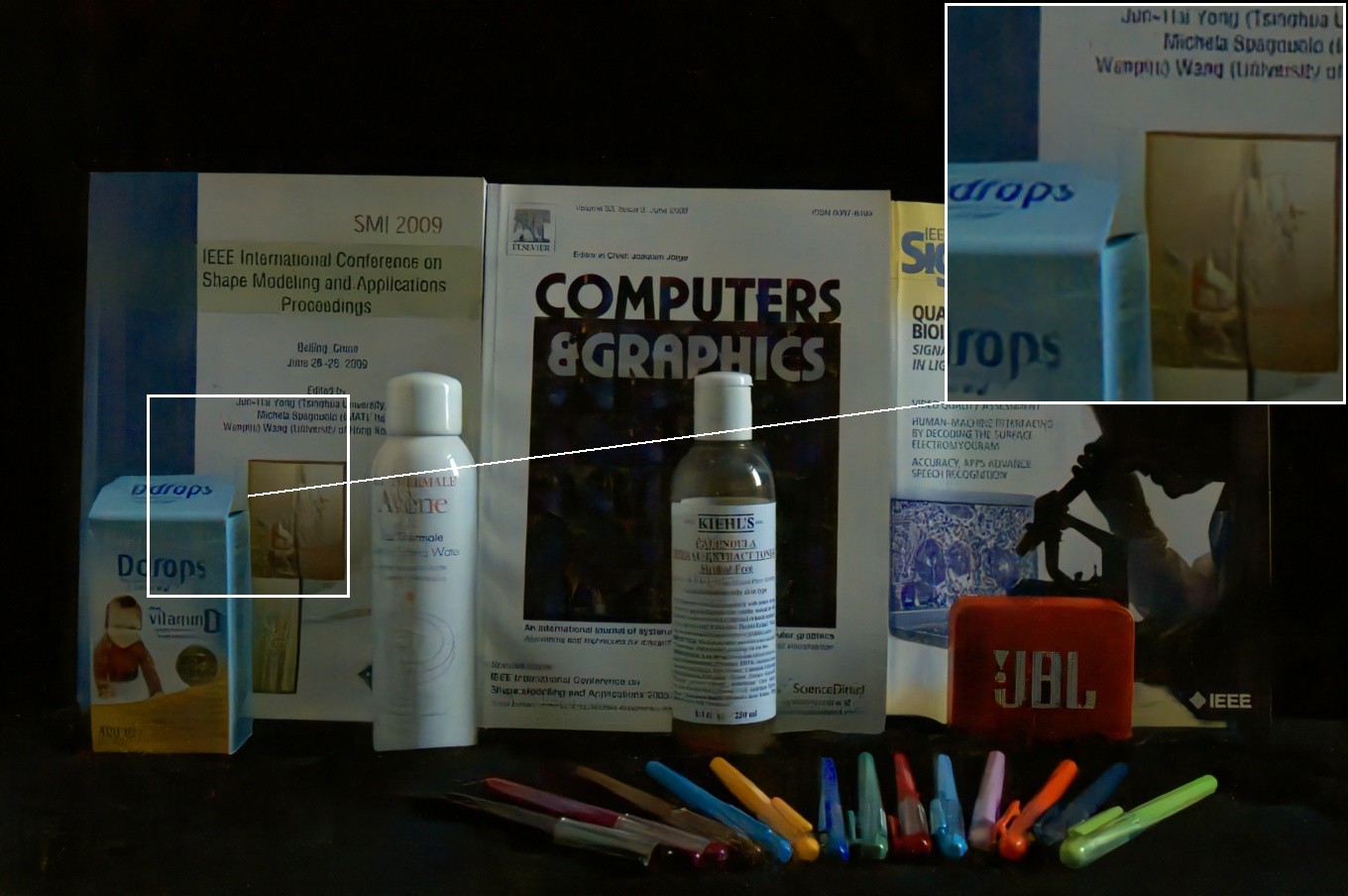}
}
\subfloat[GT]{%
    \includegraphics[width=0.24\textwidth]{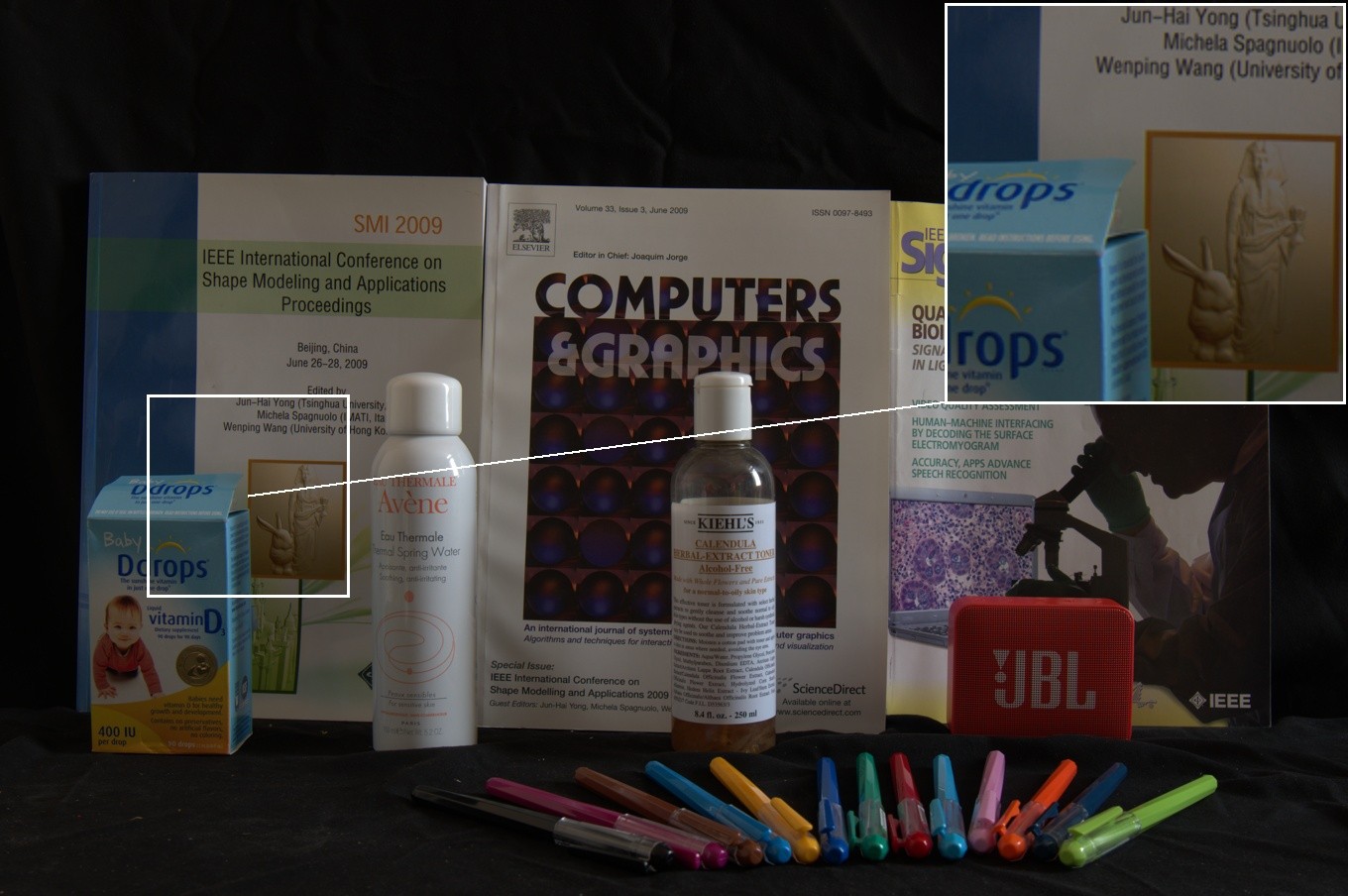}
}

\caption{\textbf{Additional qualitative comparisons on ELD-Nikon~\cite{wei2021physics}.} Two scenes are shown across four rows (blind baselines on top, calibrated baselines and ours on the bottom of each scene). PG and PGRQ exhibit residual color bias, and PGRQB introduces local color distortions, whereas our method closely matches the ground truth without target-sensor data (see zoomed insets).}
\label{fig:supp_results_eld_nikon}

\end{figure*}

\begin{figure*}[t]
\centering

\subfloat[Noisy]{%
    \includegraphics[width=0.24\textwidth]{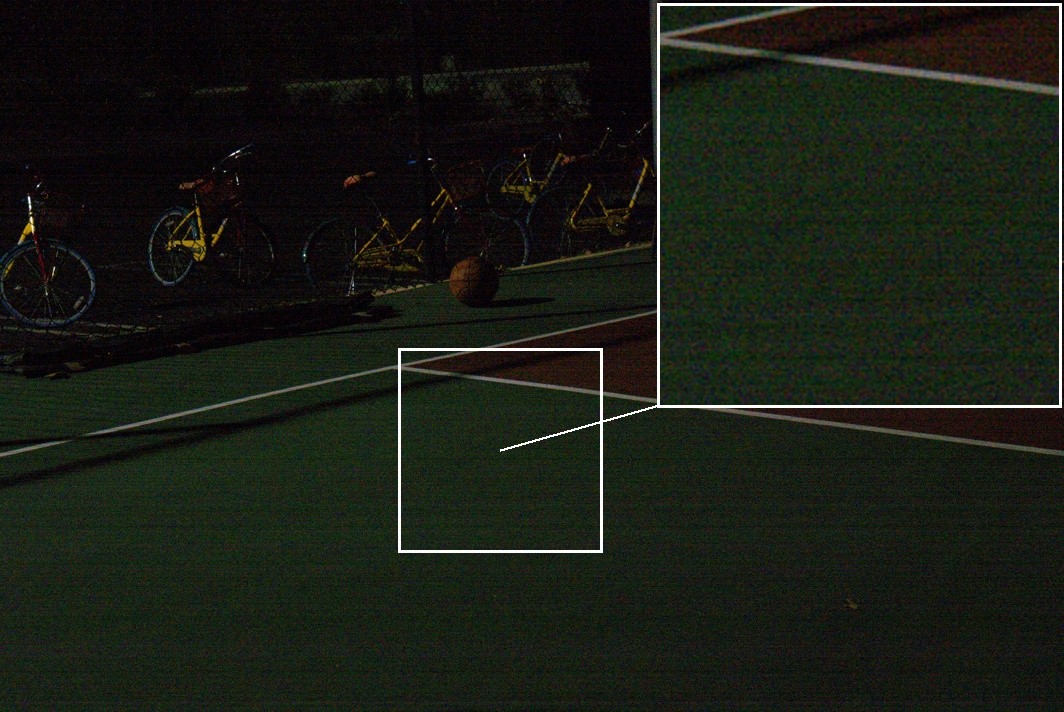}
}
\subfloat[PG]{%
    \includegraphics[width=0.24\textwidth]{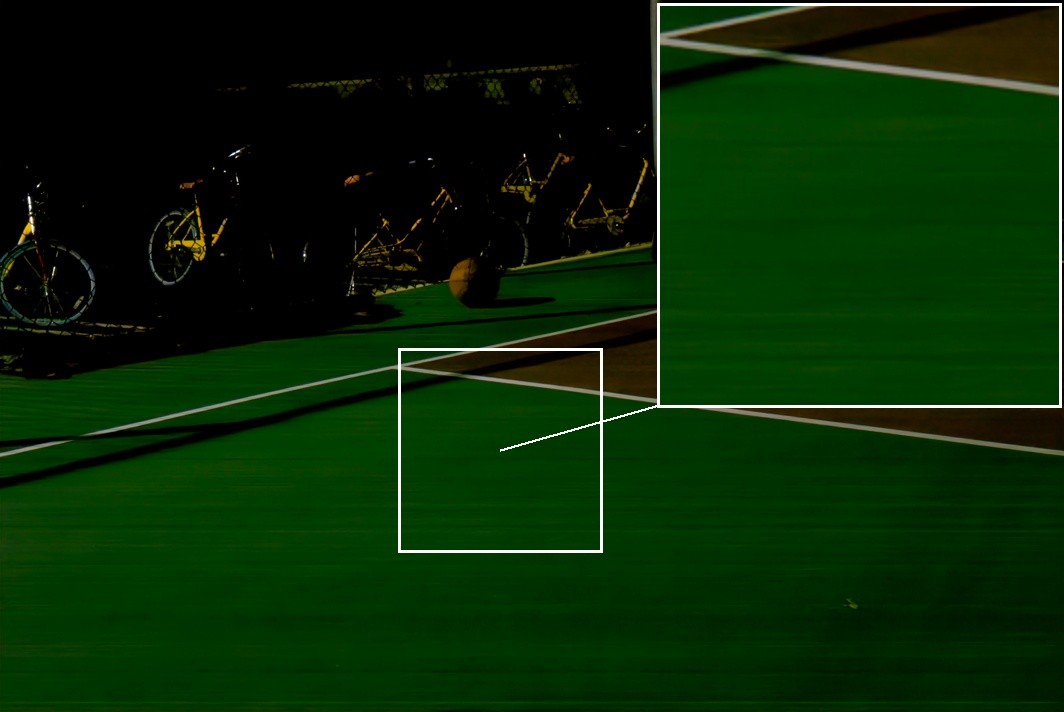}
}
\subfloat[PGRQ]{%
    \includegraphics[width=0.24\textwidth]{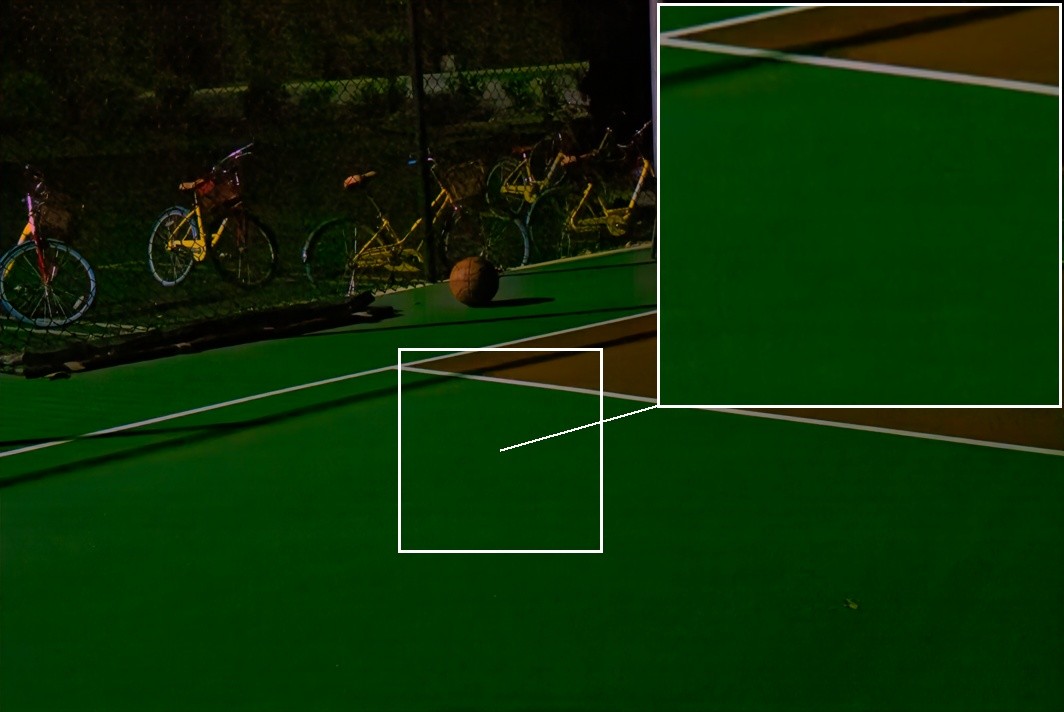}
}
\subfloat[PGRQB]{%
    \includegraphics[width=0.24\textwidth]{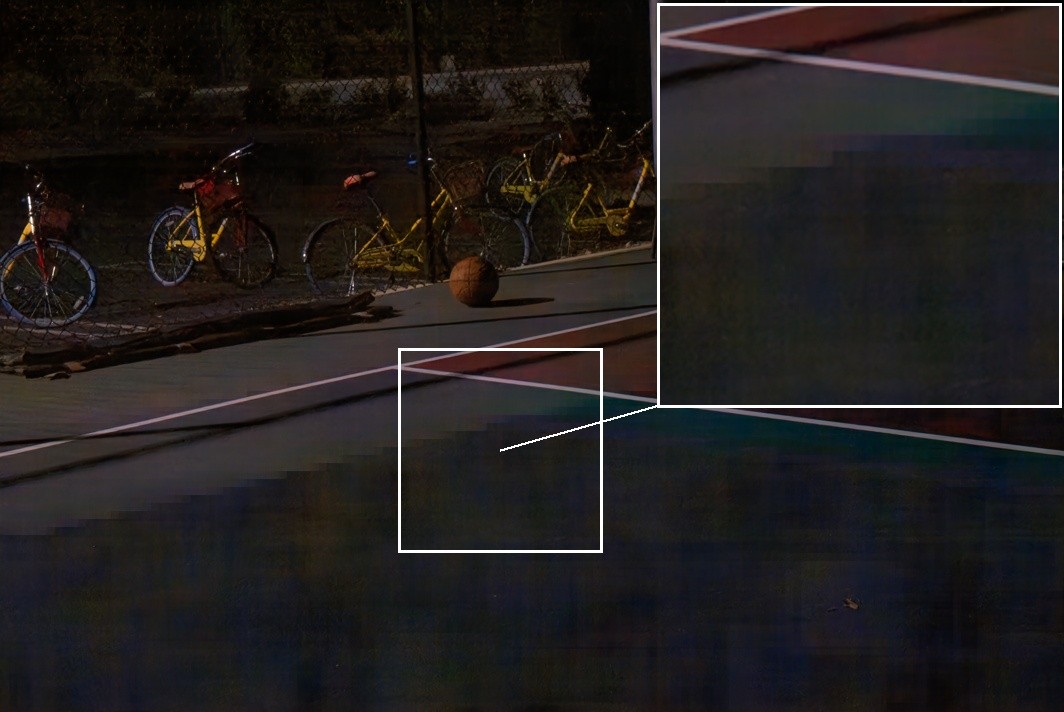}
}\\

\vspace{2pt}

\subfloat[ELD~\cite{wei2021physics}]{%
    \includegraphics[width=0.24\textwidth]{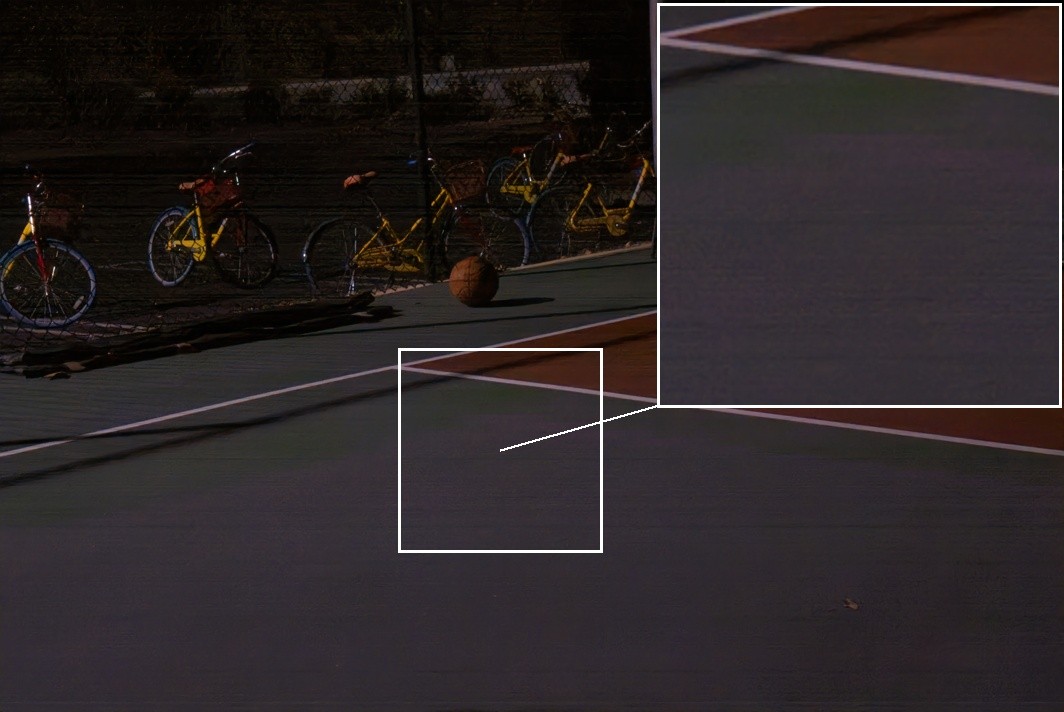}
}
\subfloat[2-Shots~\cite{lu20252}]{%
    \includegraphics[width=0.24\textwidth]{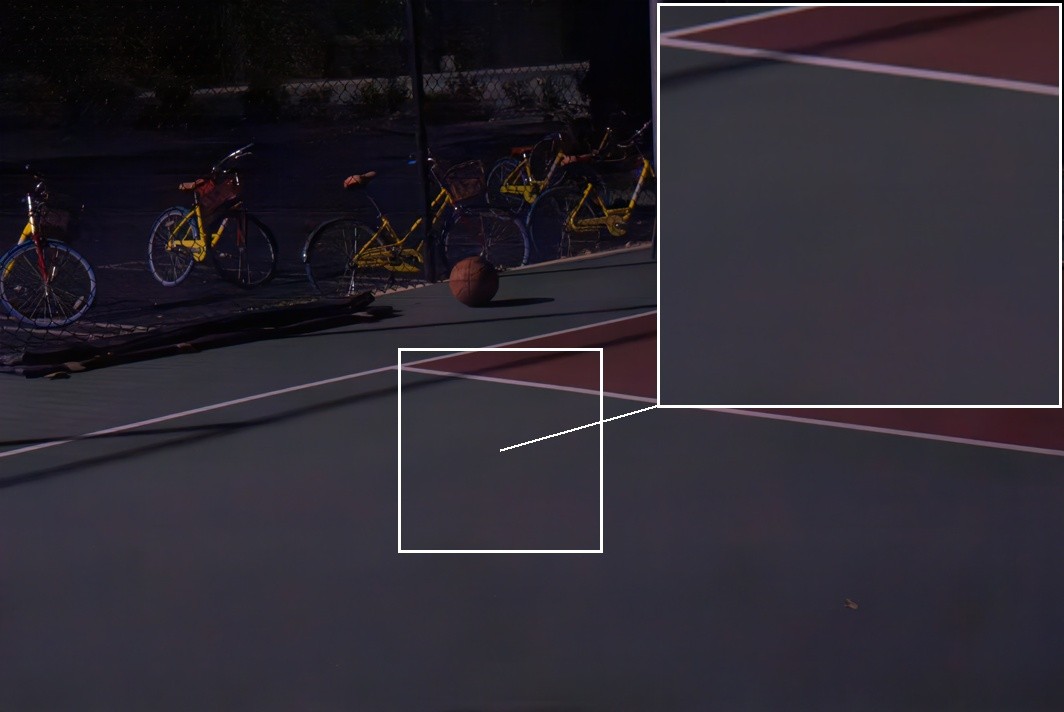}
}
\subfloat[Ours]{%
    \includegraphics[width=0.24\textwidth]{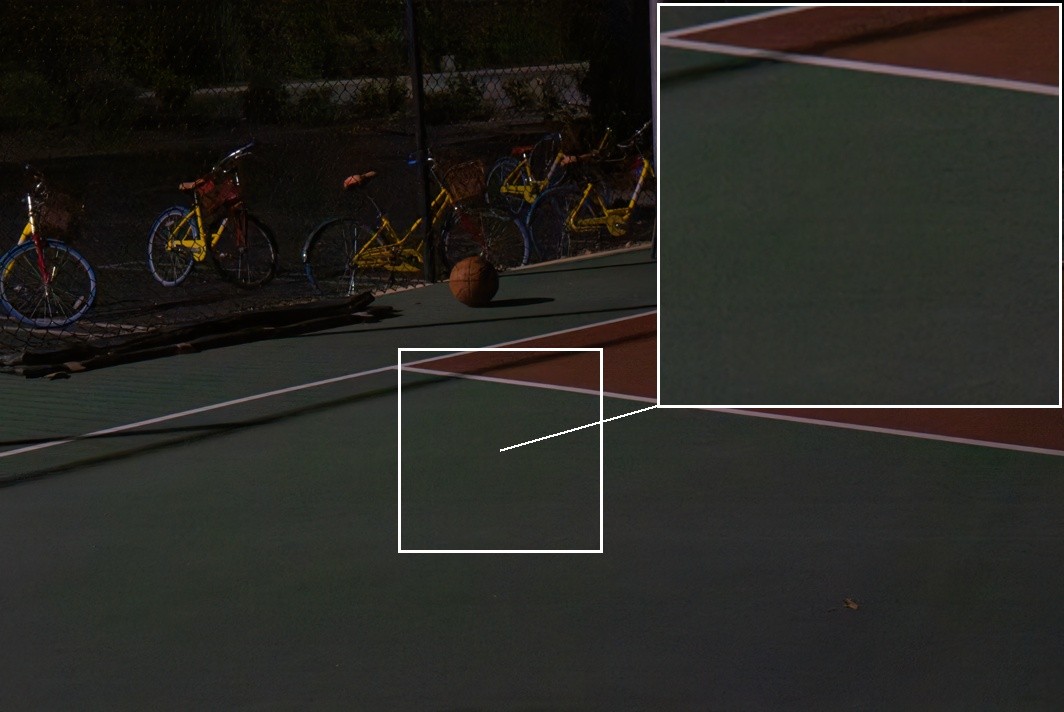}
}
\subfloat[GT]{%
    \includegraphics[width=0.24\textwidth]{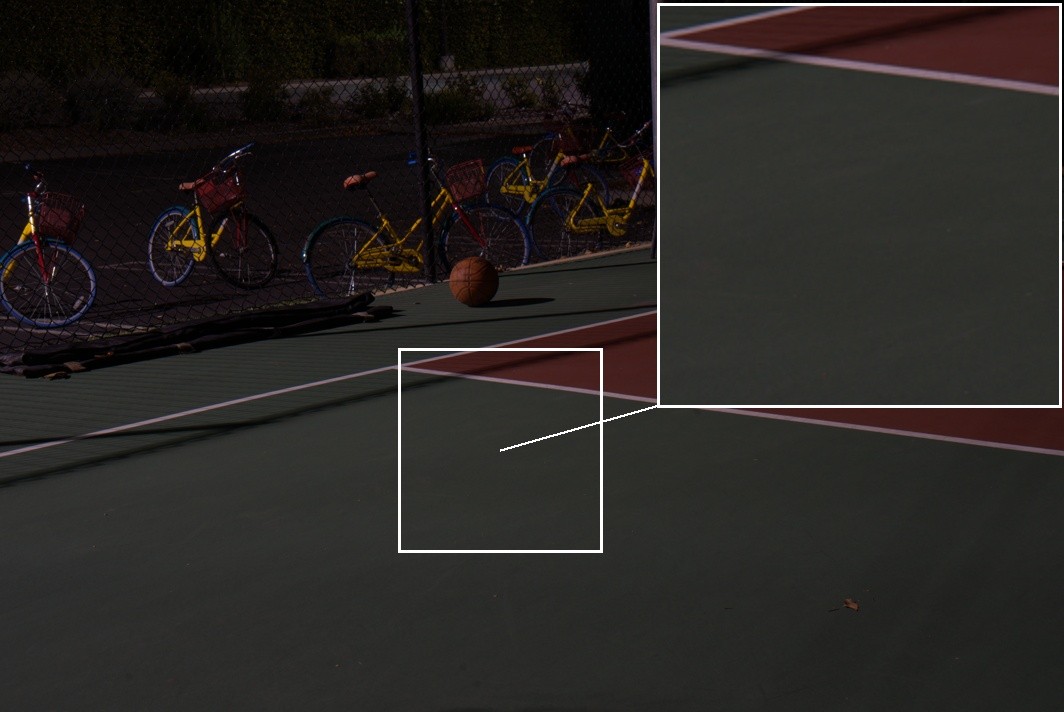}
}\\

\vspace{4pt}

\subfloat[Noisy]{%
    \includegraphics[width=0.24\textwidth]{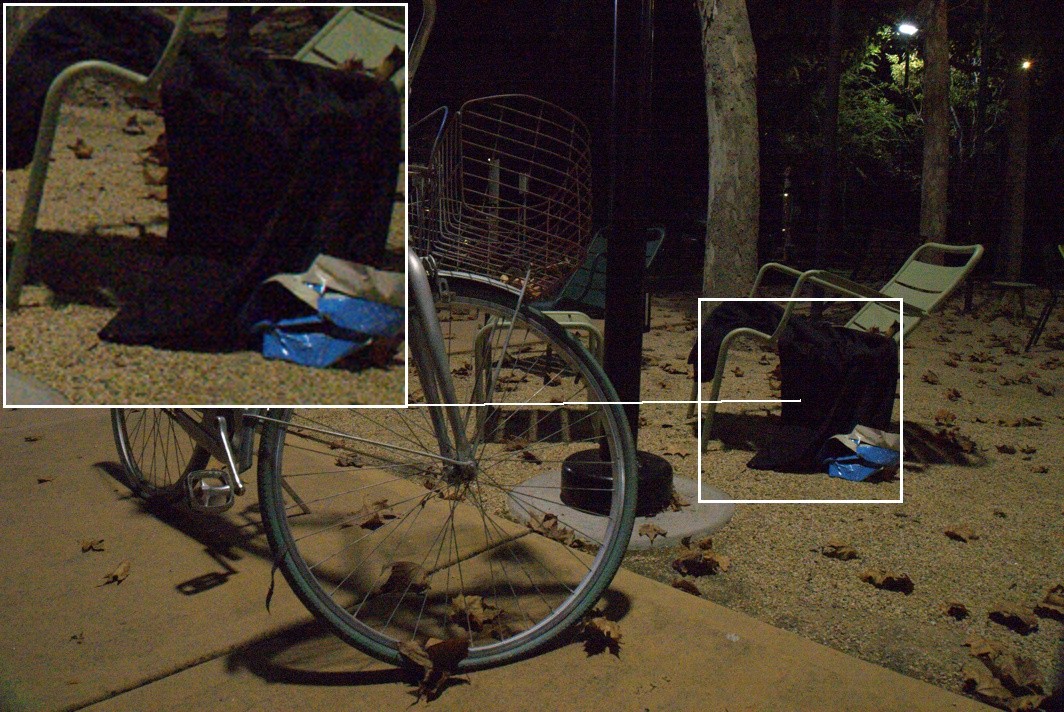}
}
\subfloat[PG]{%
    \includegraphics[width=0.24\textwidth]{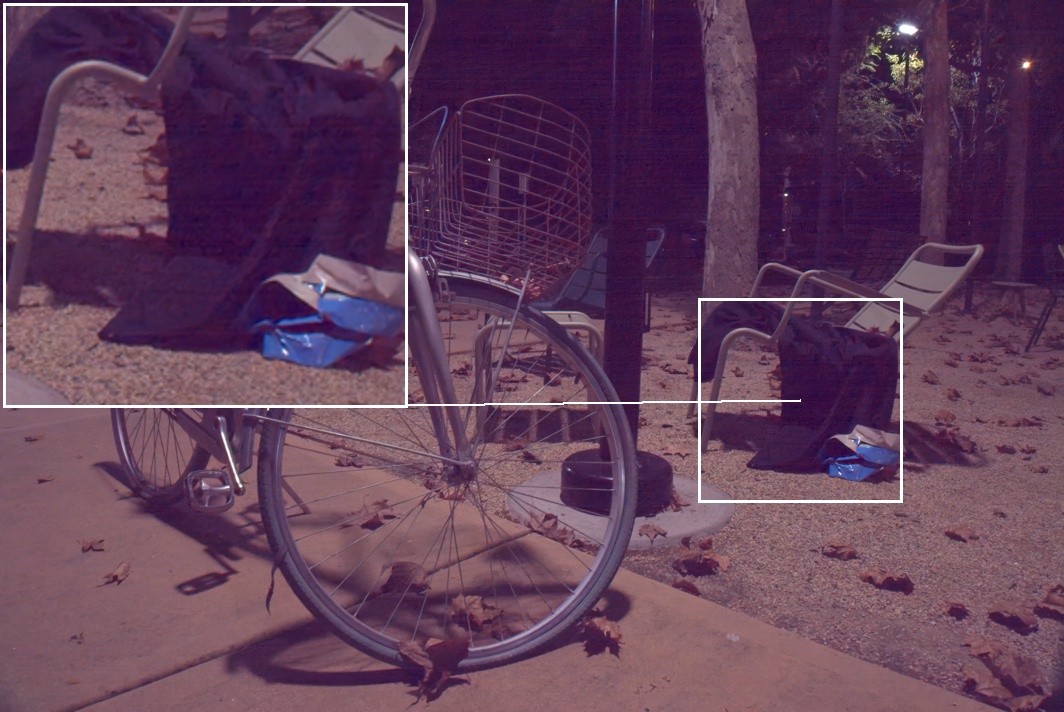}
}
\subfloat[PGRQ]{%
    \includegraphics[width=0.24\textwidth]{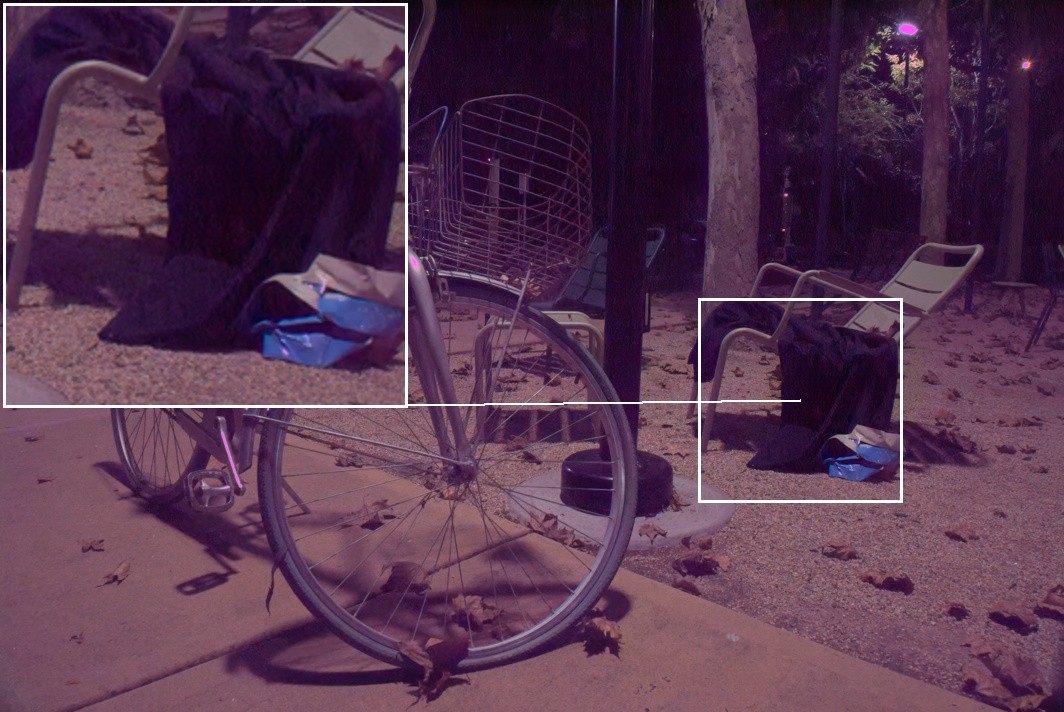}
}
\subfloat[PGRQB]{%
    \includegraphics[width=0.24\textwidth]{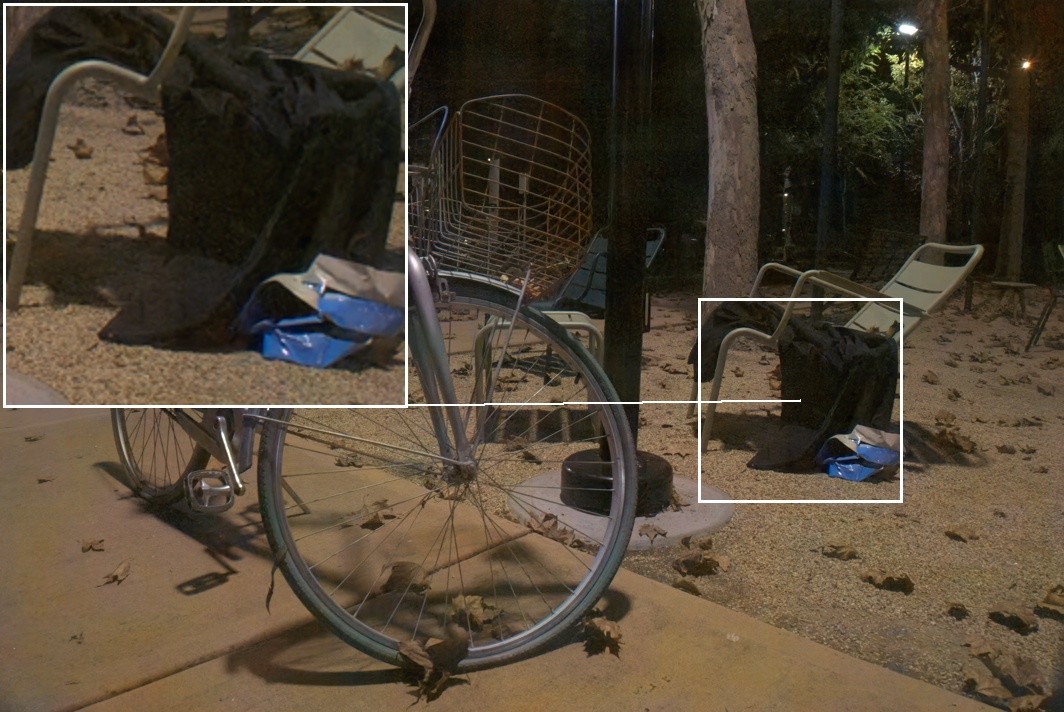}
}\\

\vspace{2pt}

\subfloat[ELD~\cite{wei2021physics}]{%
    \includegraphics[width=0.24\textwidth]{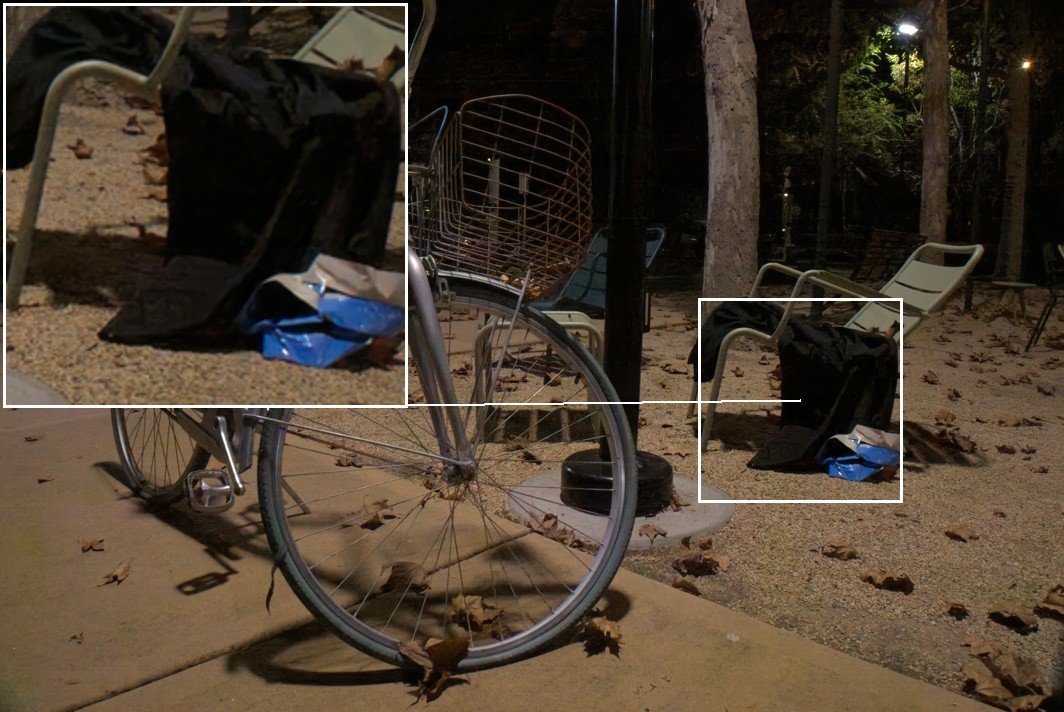}
}
\subfloat[2-Shots~\cite{lu20252}]{%
    \includegraphics[width=0.24\textwidth]{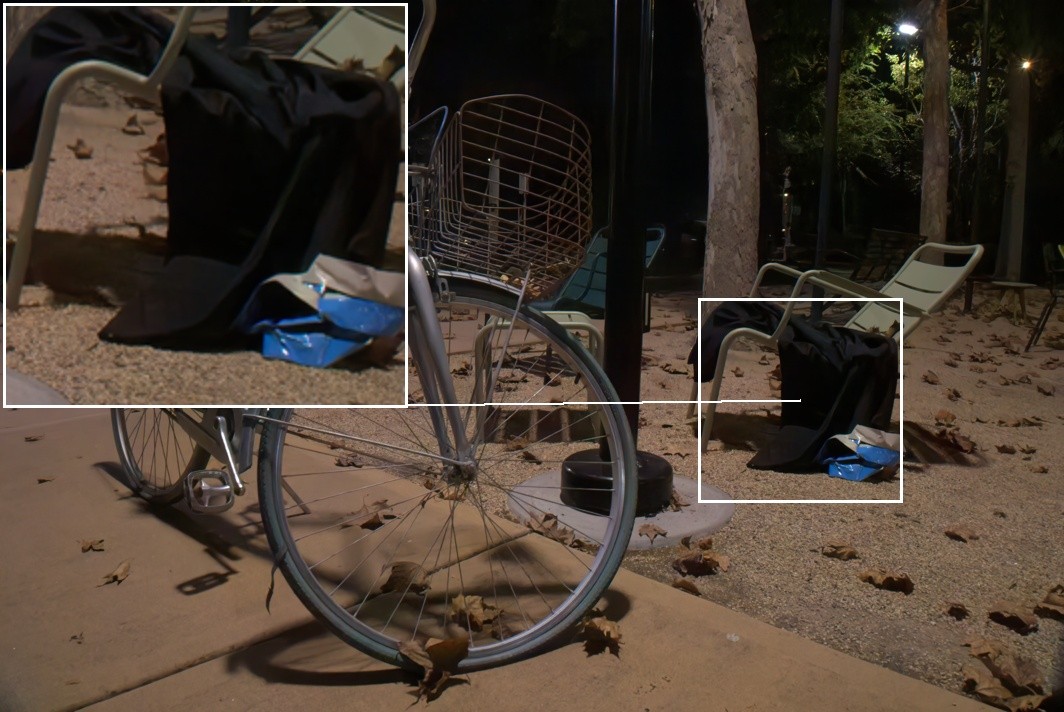}
}
\subfloat[Ours]{%
    \includegraphics[width=0.24\textwidth]{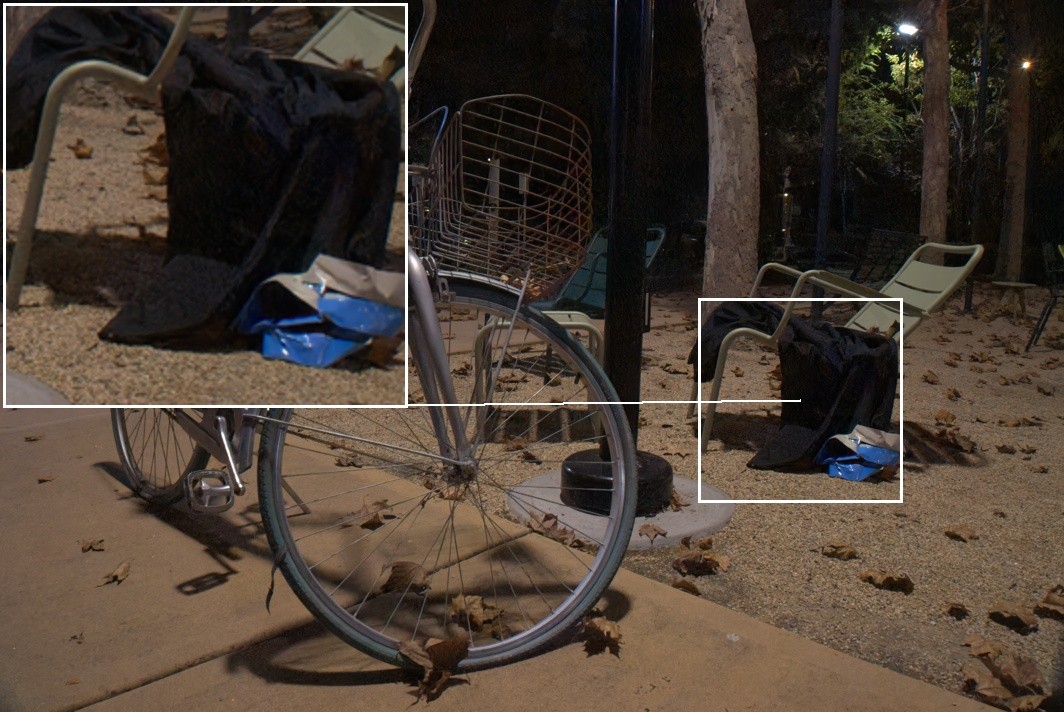}
}
\subfloat[GT]{%
    \includegraphics[width=0.24\textwidth]{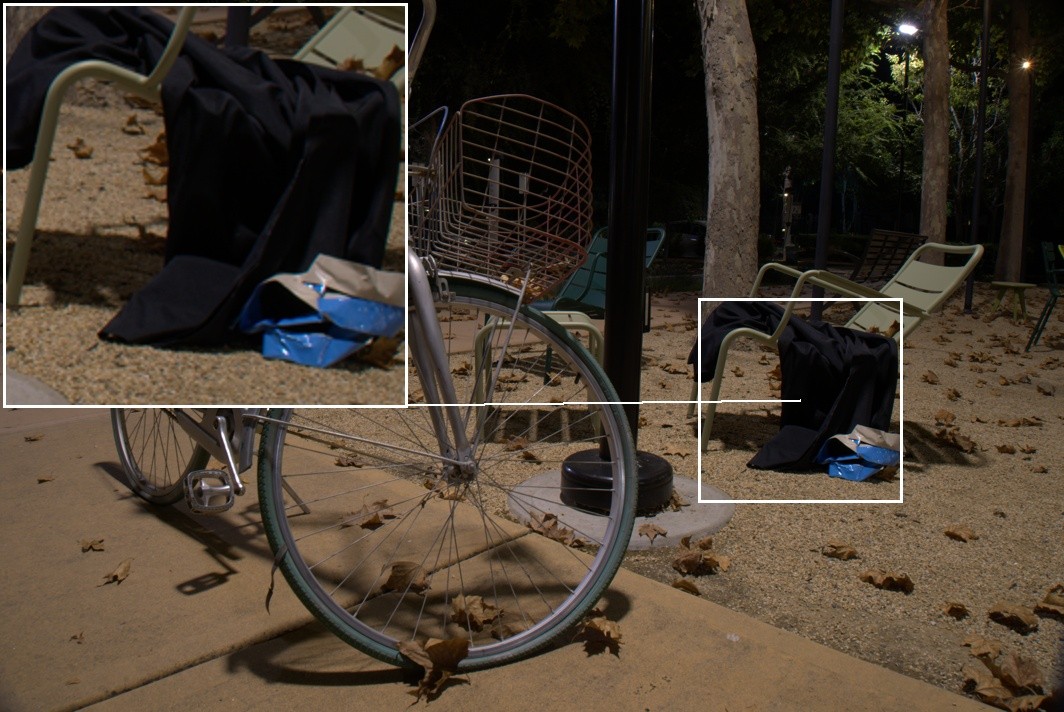}
}\\

\caption{\textbf{Additional qualitative comparisons on SID~\cite{chen2018learning}.} Two scenes across four rows (blind baselines on top, calibrated baselines and ours on the bottom of each scene). PG and PGRQ exhibit pronounced color shifts due to uncompensated black-level error, while our method recovers colors that closely match the ground truth (see zoomed insets).}
\label{fig:supp_results_sid}

\end{figure*}

\begin{figure*}[t]
\centering

\subfloat[Noisy]{%
    \includegraphics[width=0.235\textwidth]{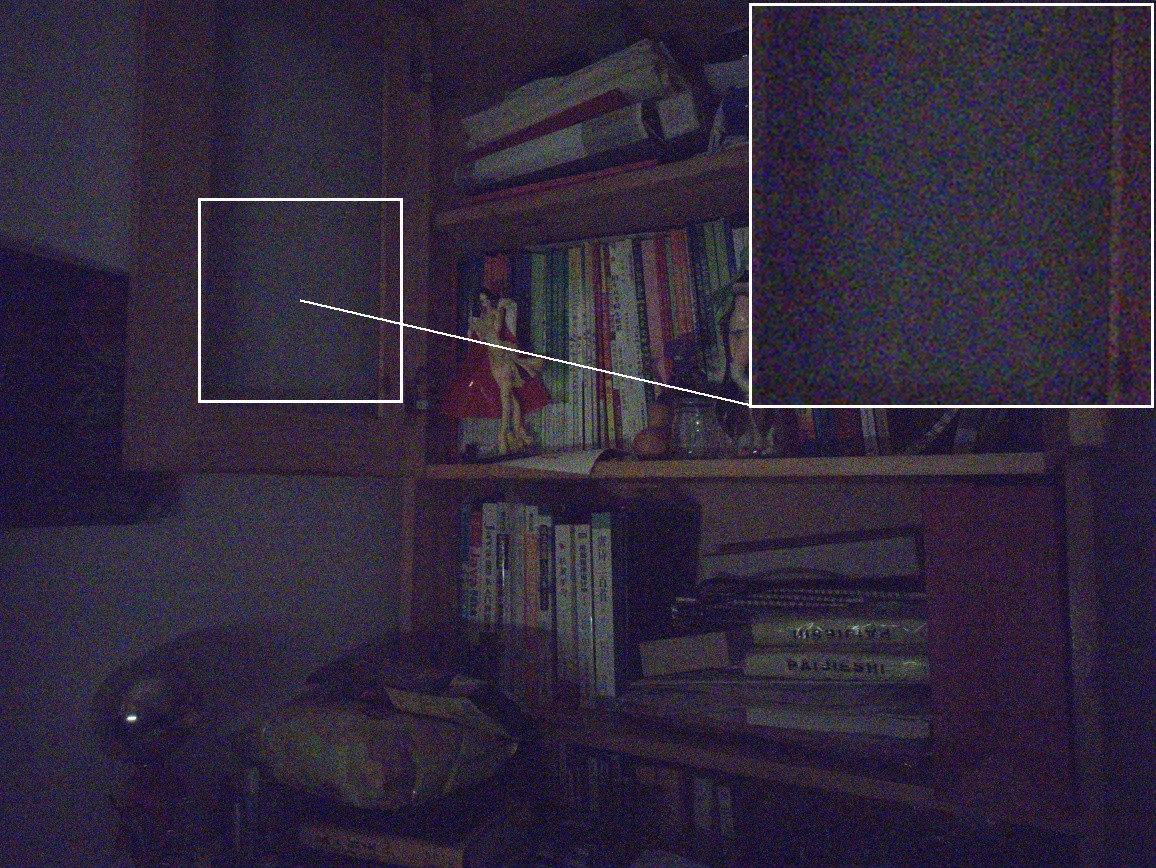}
    \label{fig:results_lrid_noisy}
}
\subfloat[PG]{%
    \includegraphics[width=0.235\textwidth]{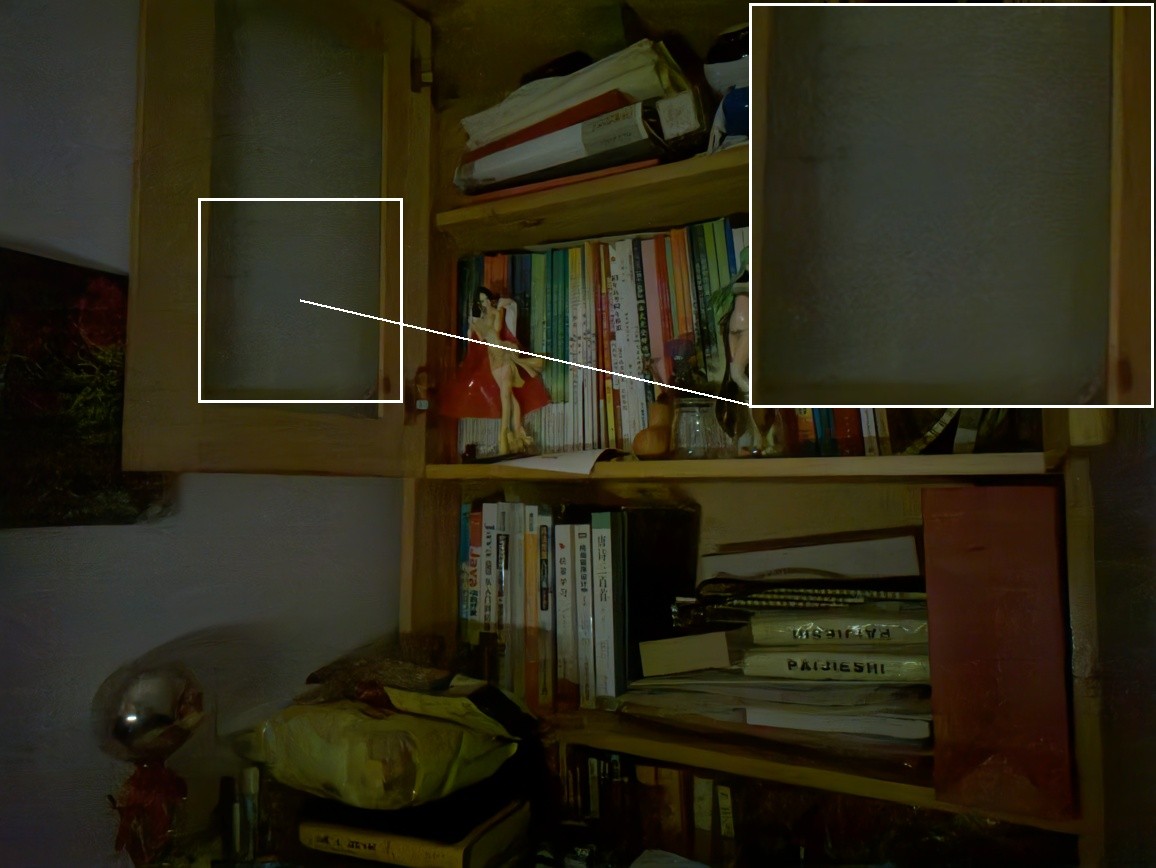}
    \label{fig:results_lrid_pg}
}
\subfloat[PGRQ]{%
    \includegraphics[width=0.235\textwidth]{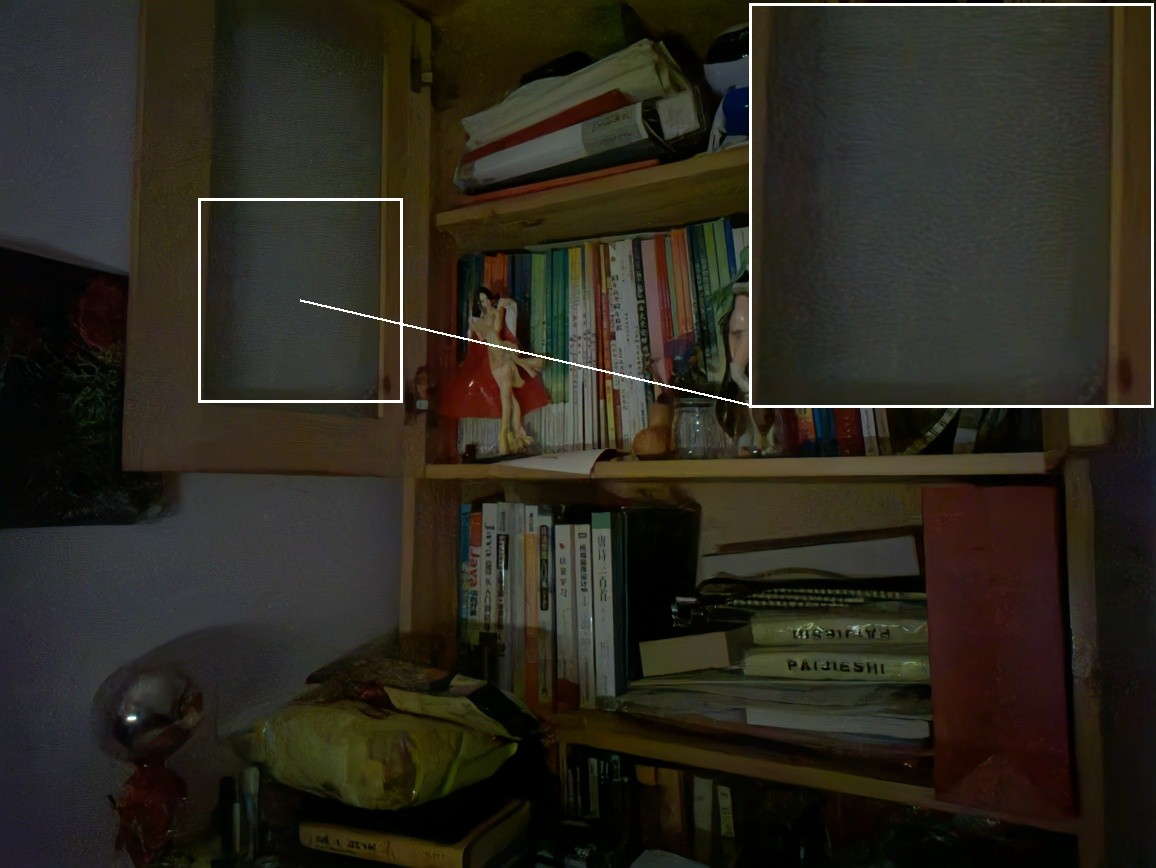}
    \label{fig:results_lrid_pgrq}
}
\subfloat[PGRQB]{%
    \includegraphics[width=0.235\textwidth]{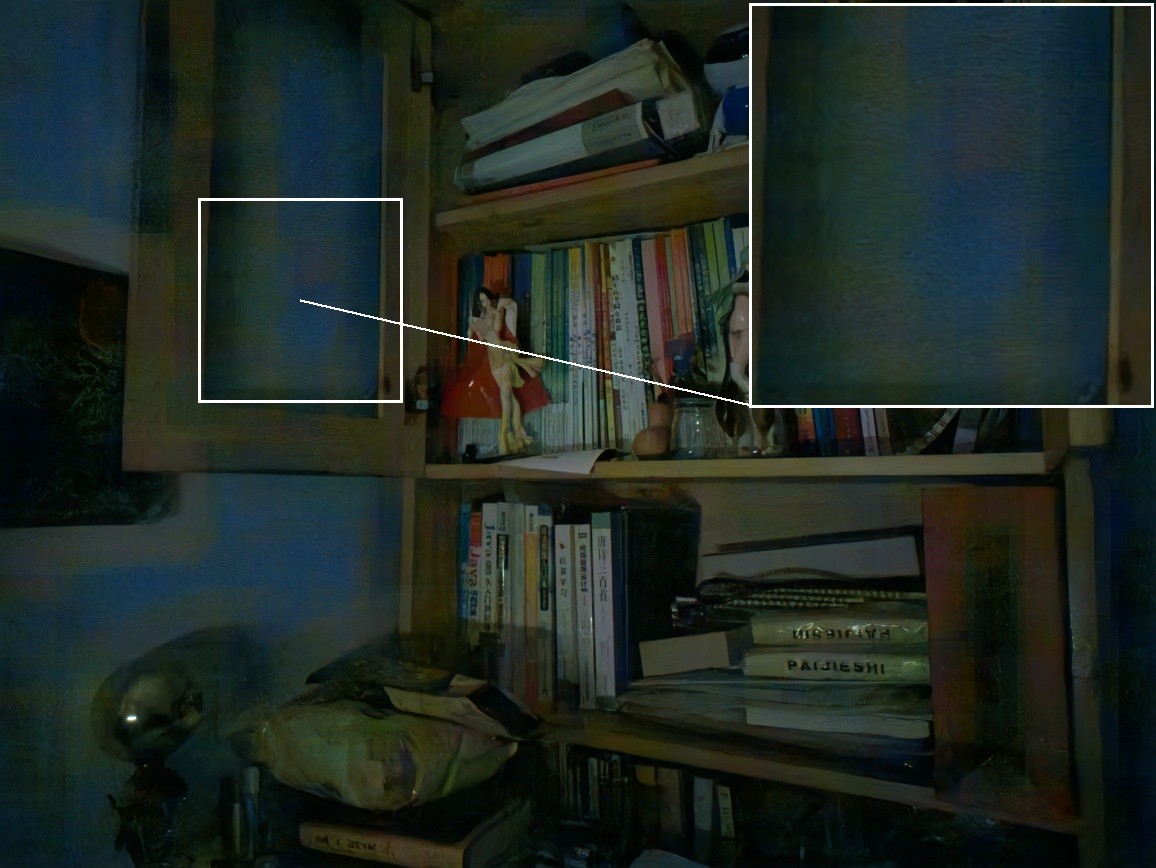}
    \label{fig:results_lrid_pgrqb}
}\\

\vspace{2pt}

\subfloat[ELD~\cite{wei2021physics}]{%
    \includegraphics[width=0.235\textwidth]{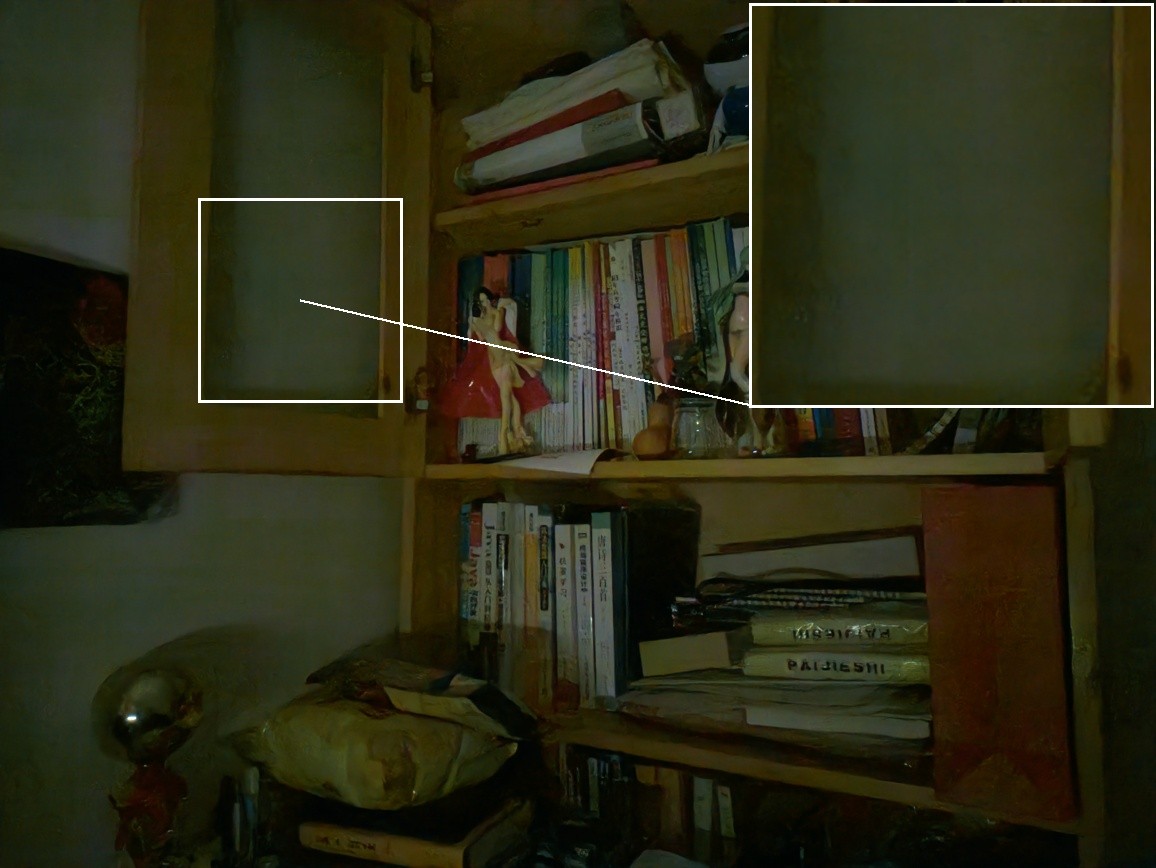}
    \label{fig:results_lrid_eld}
}
\subfloat[2-Shots~\cite{lu20252}]{%
    \includegraphics[width=0.235\textwidth]{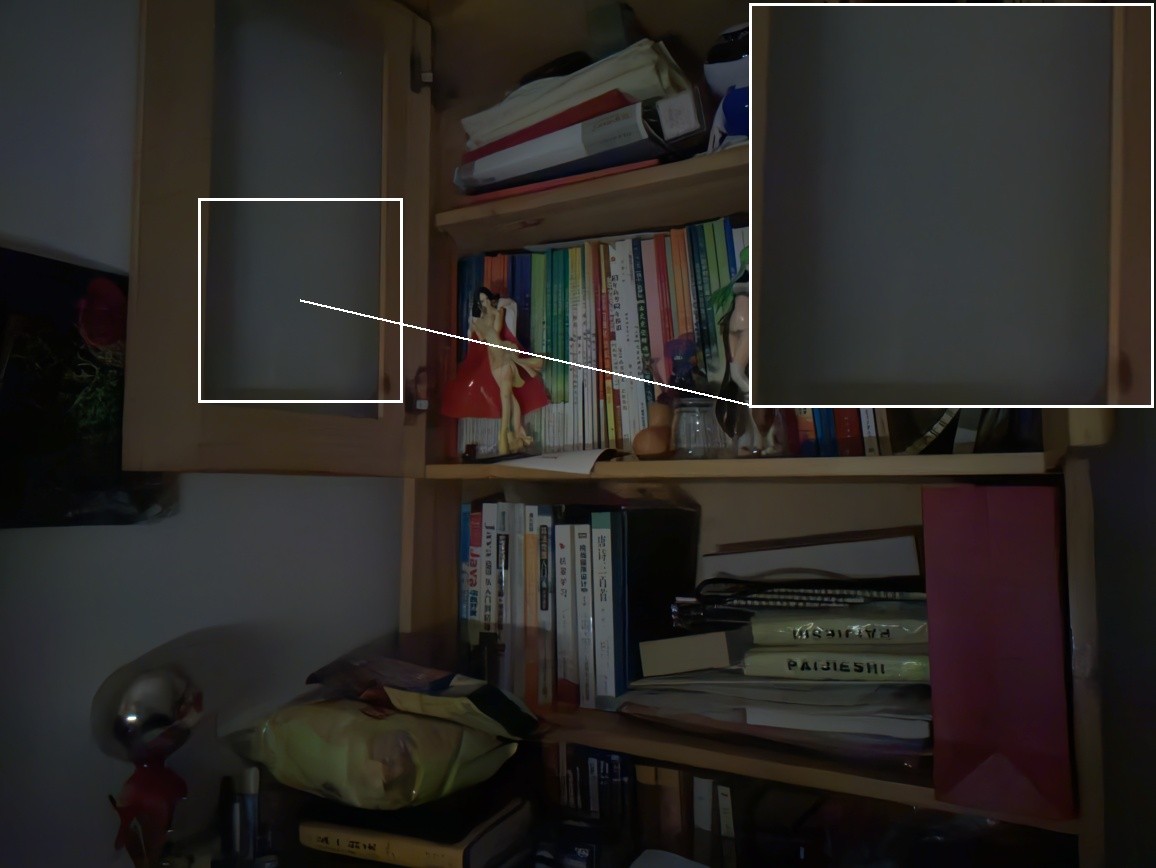}
    \label{fig:results_lrid_2shots}
}
\subfloat[Ours]{%
    \includegraphics[width=0.235\textwidth]{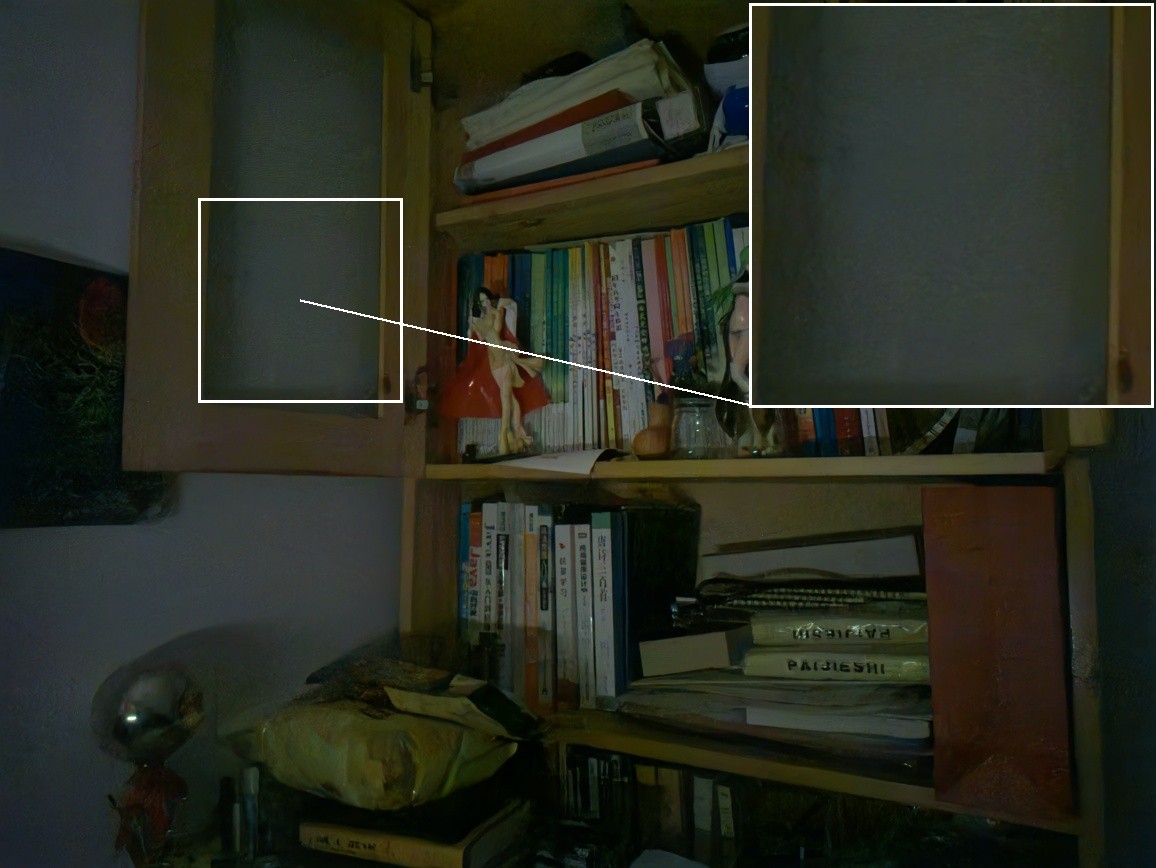}
    \label{fig:results_lrid_ours}
}
\subfloat[GT]{%
    \includegraphics[width=0.235\textwidth]{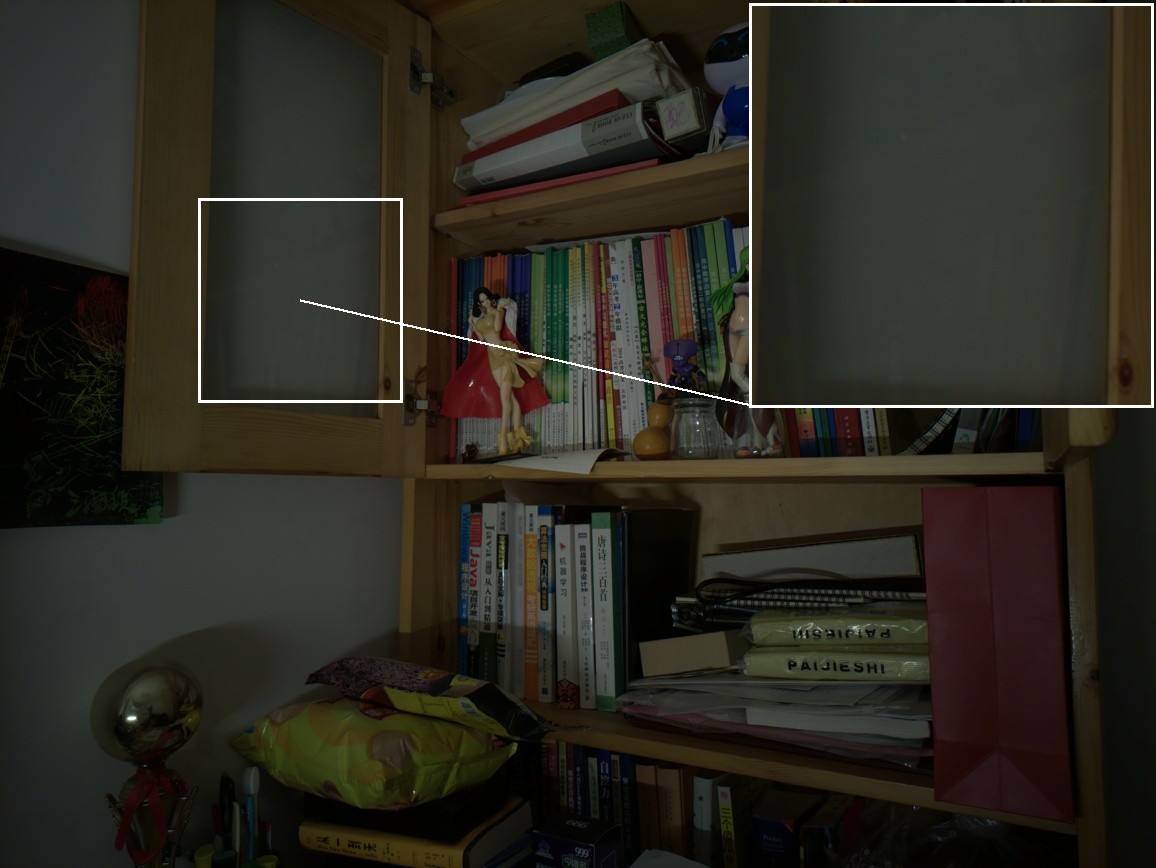}
    \label{fig:results_lrid_gt}
}

\caption{\textbf{Qualitative comparisons on LRID~\cite{feng2022learnability} at the $256\times$ exposure ratio.} As discussed in \cref{sec:supp_ds_discussion}, the dark shading noise---and hence the BLE---is weak in this setting. Our method performs on par with the blind baselines, while PGRQB and ELD introduce noticeable color artifacts.}
\label{fig:supp_results_lrid}

\end{figure*}

We provide qualitative comparisons on ELD~\cite{wei2021physics}, SID~\cite{chen2018learning}, and LRID~\cite{feng2022learnability} in \cref{fig:supp_results_eld_sony,fig:supp_results_eld_nikon,fig:supp_results_sid,fig:supp_results_lrid}. Output images are rendered using the metadata embedded in the RAW files. Overall, PG and PGRQ fail to compensate for black-level error and dark shading noise on ELD-Sony (\cref{fig:supp_results_eld_sony}), ELD-Nikon (\cref{fig:supp_results_eld_nikon}), and SID (\cref{fig:supp_results_sid}), which results in unrealistic colors. PGRQB exhibits limited color recovery and introduces local color distortions that degrade the output, most notably in \cref{fig:supp_results_eld_sony,fig:supp_results_eld_nikon}. ELD~\cite{wei2021physics} likewise fails to restore realistic brightness and color, as observed in \cref{fig:supp_results_eld_nikon,fig:supp_results_lrid}. The calibrated methods (2-Shots~\cite{lu20252} and NoiseDiff~\cite{lu2025dark}) perform well on ELD-Sony, SID, and LRID, where calibrated dark shading noise is subtracted prior to denoising, but fail to produce high-quality results on ELD-Nikon, for which dark shading calibration is unavailable. In contrast, our method effectively suppresses the noise while compensating for the black-level error without access to any target-sensor data, yielding outputs that closely match the ground truth.

\subsection{Additional Ablations}
\label{sec:supp_ablation}
\begin{table}[t]
\centering
\renewcommand{\arraystretch}{1.05}

\begin{tabular*}{\linewidth}{@{\extracolsep{\fill}}lccc}
\toprule
& Restormer & Ours (1-ch) & \textbf{Ours (4-ch)} \\
\midrule
PSNR & $41.33$ & $40.20$ & $\textbf{42.82}$ \\
\bottomrule
\end{tabular*}

\caption{\textbf{Capacity and per-channel BLE ablations} (PSNR in dB, $\uparrow$, on ELD-Nikon~\cite{wei2021physics} $\times200$). Restormer~\cite{zamir2022restormer} is trained with PGRQB noise model under its default configuration. ``1-ch'' predicts a single black-level error (BLE) offset shared across all channels, whereas ``4-ch'' is our full model with one offset per RGBG channel. Despite roughly twice the parameters, Restormer remains below our model, and the per-channel variant clearly outperforms the channel-shared one.}

\label{tab:rebuttal_restormer}

\end{table}

\noindent\textbf{Isolating the effect of BLE correction.}
To verify that our improvement stems from black-level error (BLE) correction rather than from the denoiser itself, we attach our trained BLBE module to an off-the-shelf PGRQ baseline that never observed BLE during training. On ELD-Nikon $\times200$, PGRQ alone achieves $40.81$\,dB, whereas PGRQ+BLBE reaches $42.75$\,dB, close to our full end-to-end model at $42.82$\,dB. This confirms that BLE correction, independent of denoising, is a dominant factor that contributes to the performance. The small additional gain of the full model indicates that end-to-end joint training lets the denoiser compensate for residual inaccuracies in the BLBE estimate.

\noindent\textbf{Comparison with a higher-capacity architecture.}
To rule out model capacity as the source of our gains, we additionally compare against Restormer~\cite{zamir2022restormer}, a transformer-based restoration network trained with our noise model under its default configuration (\cref{tab:rebuttal_restormer}). Despite having roughly twice the parameters of our U-Net backbone, Restormer ($41.33$\,dB) outperforms the U-Net trained on the PGRQB noise model but remains below our full method ($42.82$\,dB) on ELD-Nikon $\times200$. This shows that our improvement arises from explicitly decoupling BLE rather than from added capacity, consistent with the ablation in \cref{tab:ablations_model}.

\noindent\textbf{Per-channel vs.\ shared BLE.}
Finally, we justify modeling BLE independently per RAW channel. Calibrated measurements in ELD~\cite{wei2021physics} show that per-channel black levels are tightly correlated for some sensors (e.g., Sony) but vary substantially for others (e.g., Nikon; see Fig.~4 in~\cite{wei2021physics}). We therefore compare our per-channel (4-ch) model against a variant that predicts a single BLE shared across all channels (1-ch). On ELD-Nikon $\times200$ (\cref{tab:rebuttal_restormer}), the 4-ch model ($42.82$\,dB) substantially outperforms the 1-ch variant ($40.20$\,dB), confirming that per-channel modeling is essential for cross-sensor generalization.

\subsection{Accuracy of the BLE Predictor}
\label{sec:supp_ble_accuracy}
\begin{figure}[t]
\centering
\includegraphics[width=0.87\linewidth]{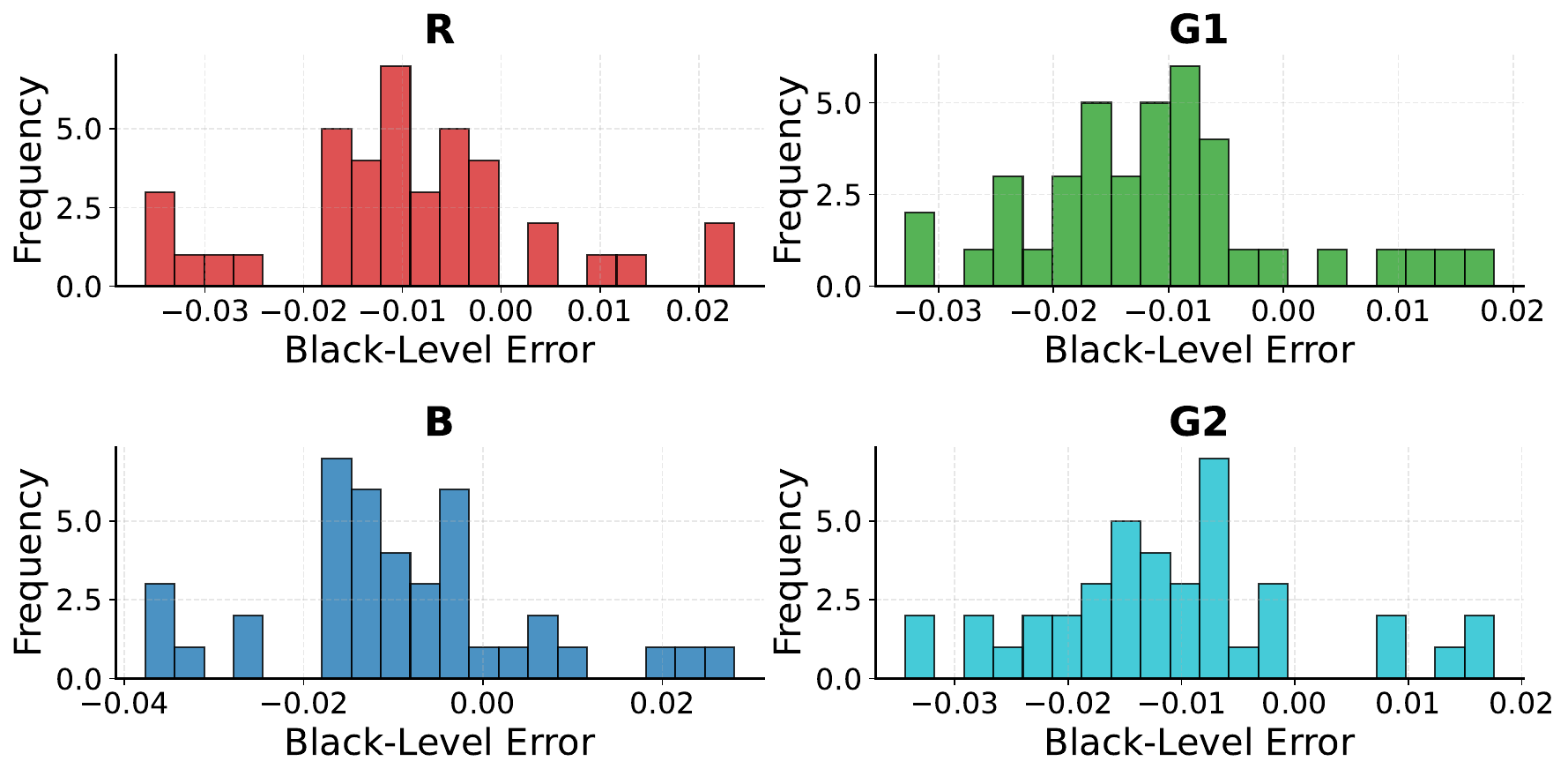}
\caption{\textbf{Distribution of predicted black-level error (BLE) on SID $\times250$~\cite{chen2018learning}.} Per-channel histograms of the BLE offsets estimated by our BLBE module across the RGBG channels of the test set. }
\label{fig:supp_BLBE_hist}
\end{figure}

To directly assess the quality of our black-level error (BLE) estimates, we quantify the prediction error of the BLBE module on a held-out synthetic SID validation set, with intensities normalized to $[0,1]$. Our predictor attains a mean absolute error of $6.4\times10^{-3}$, corresponding to roughly $16\%$ of the maximum injected BLE ($4.0\times10^{-2}$). 
To further illustrate its behavior on real data,  \cref{fig:supp_BLBE_hist} shows the per-channel distribution of predicted BLEs across real SID $\times250$ images.

\section{Dark-Shading Noise Analysis}
\begin{figure*}
    \centering
    \subfloat[][SID sensor, ISO 6400. Per-channel heatmaps for the R, G1, B, and G2 channels.]{%
        \includegraphics[width=0.95\textwidth]{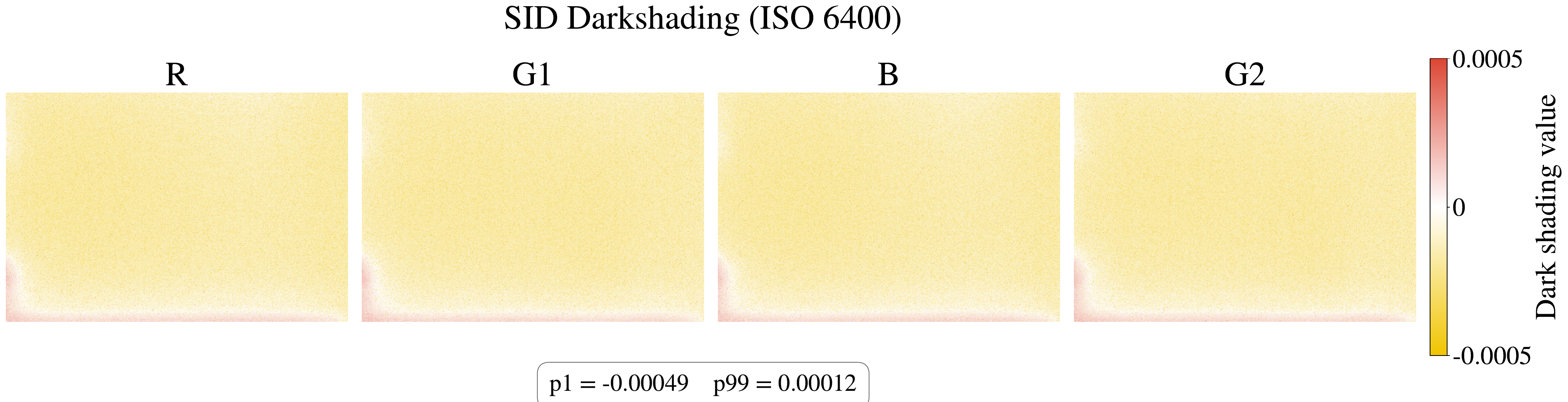}
        \label{fig:supp_SID_ds_heatmap}
    }

    \vspace{4pt}

    \subfloat[][LRID sensor, ISO 6400. Per-channel heatmaps for the R, G1, B, and G2 channels.]{%
        \includegraphics[width=0.95\textwidth]{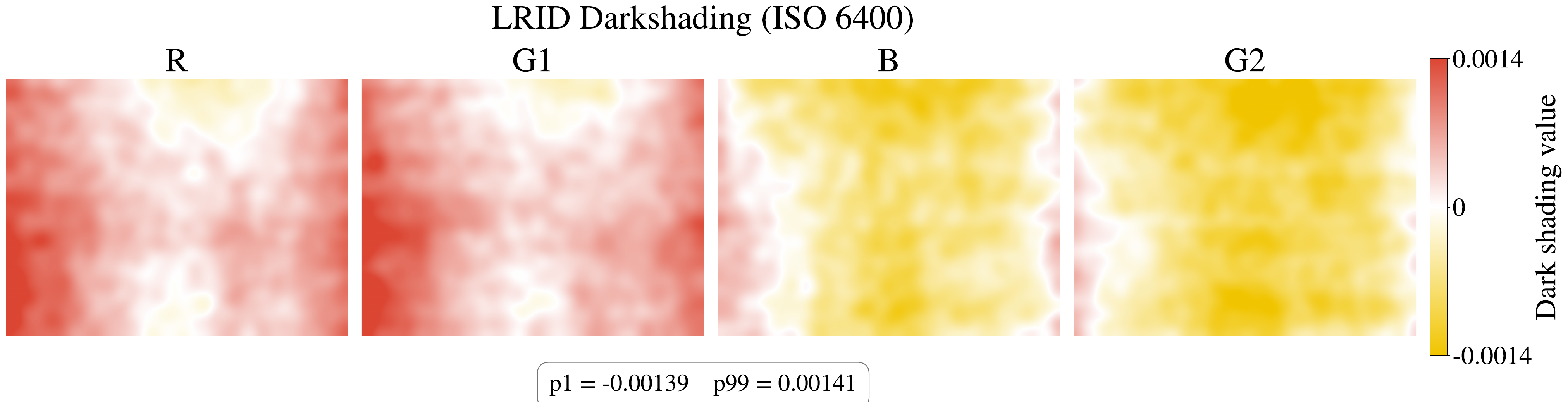}
        \label{fig:supp_LRID_ds_heatmap}
    }
    \caption{\textbf{Dark-shading noise heatmaps for the SID and LRID sensors at ISO 6400.} Each subfigure displays the spatial distribution of dark shading noise for the four Bayer channels (R, G1, B, G2), computed after black-level subtraction and normalization by the white level. The $1$st and $99$th percentile values of each channel are annotated in the corresponding panel (denoted \textit{p1} and \textit{p99}). The LRID sensor exhibits dark shading roughly three times larger than the SID sensor. Although the absolute per-pixel magnitudes are small, amplification by the exposure ratio at inference time scales them into the dominant noise component in short-exposure low-light images (see \cref{sec:supp_ds_discussion}).}
    \label{fig:supp_ds_heatmap}
\end{figure*}

\begin{figure*}[t]
    \centering
    \setlength{\tabcolsep}{2pt}
    \begin{tabular}{cccccc}
        & Noisy & PGRQ & PGRQB & Ours & GT \\
        \rotatebox{90}{\small SID ISO 1600} &
        \includegraphics[width=0.185\textwidth]{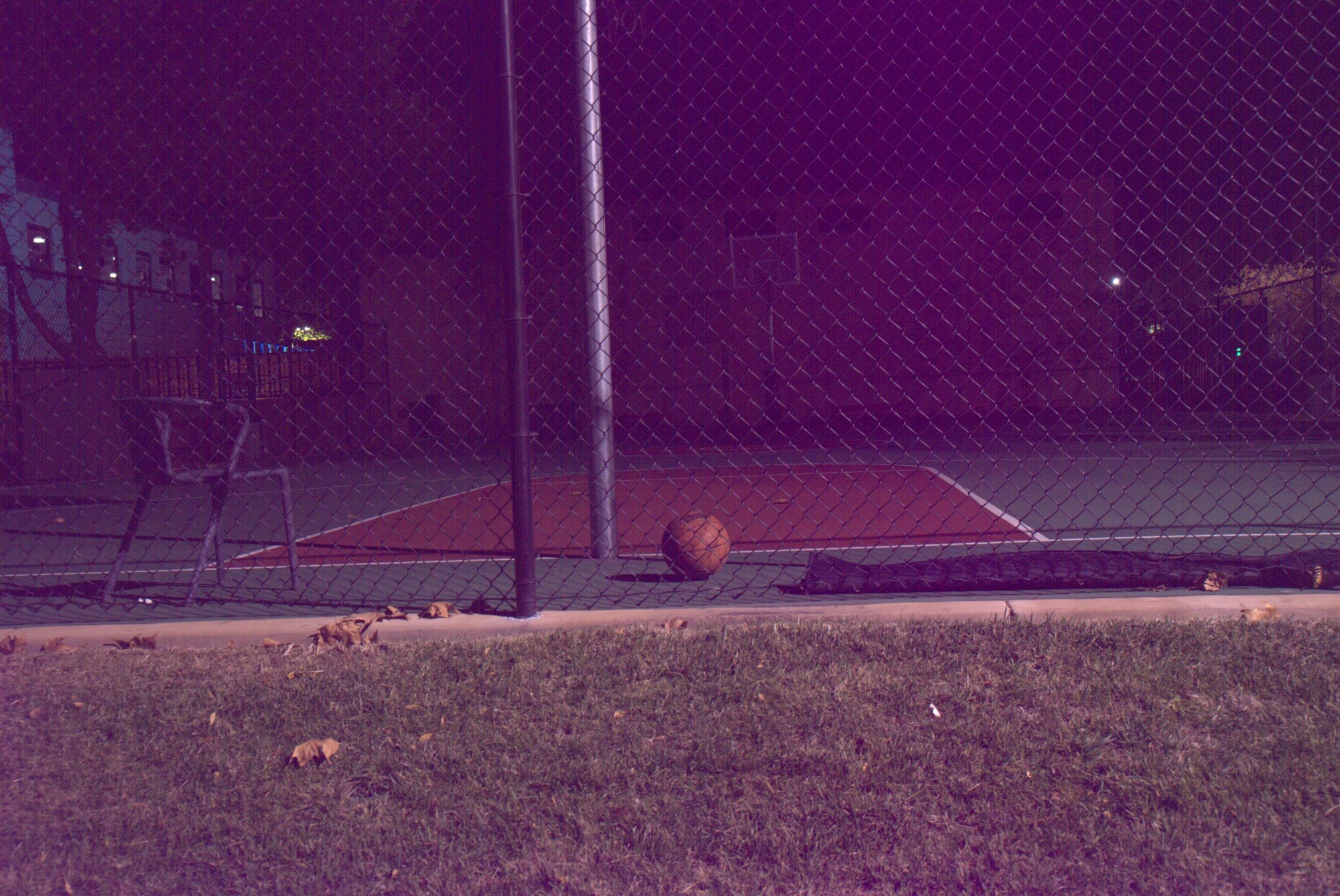} &
        \includegraphics[width=0.185\textwidth]{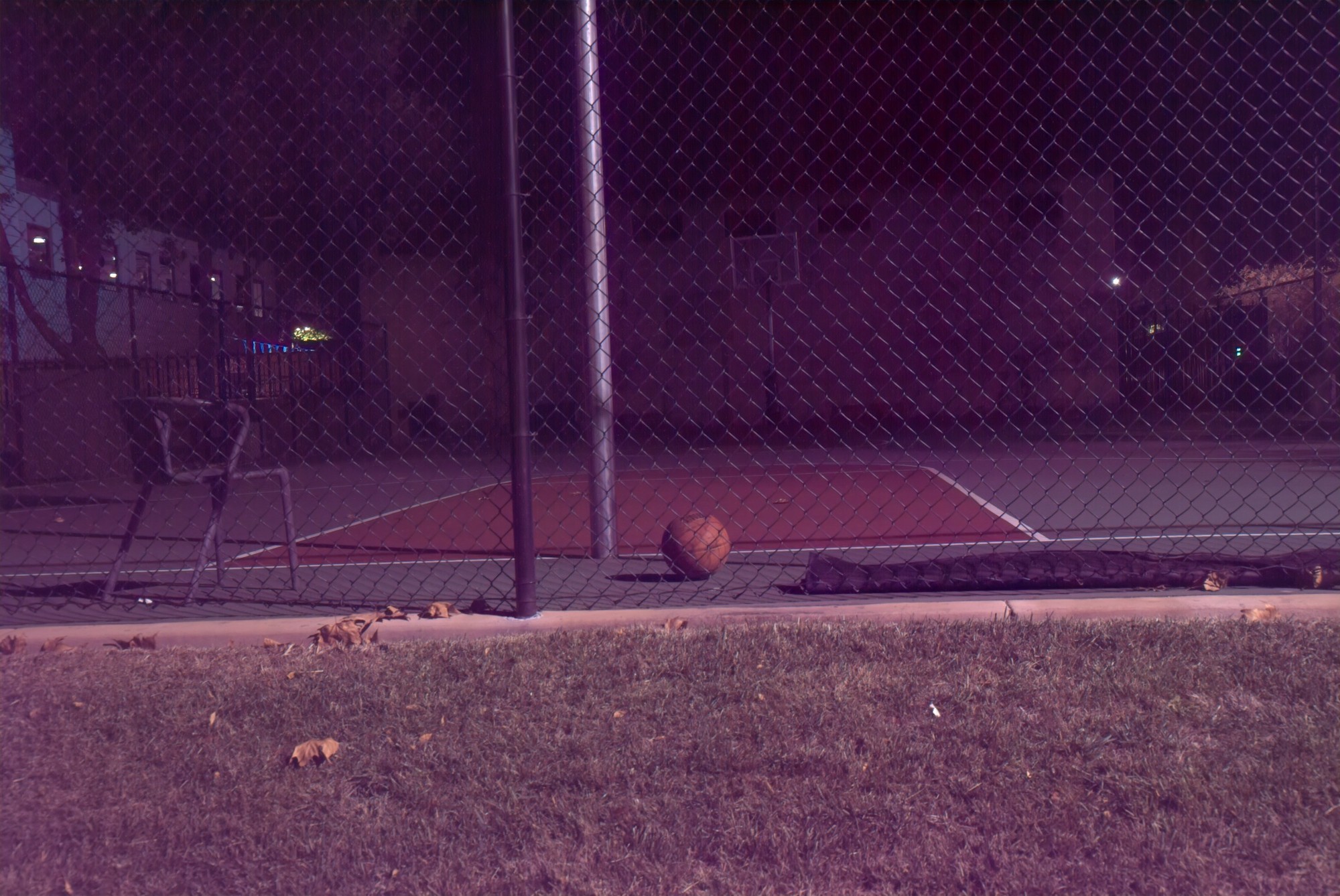} &
        \includegraphics[width=0.185\textwidth]{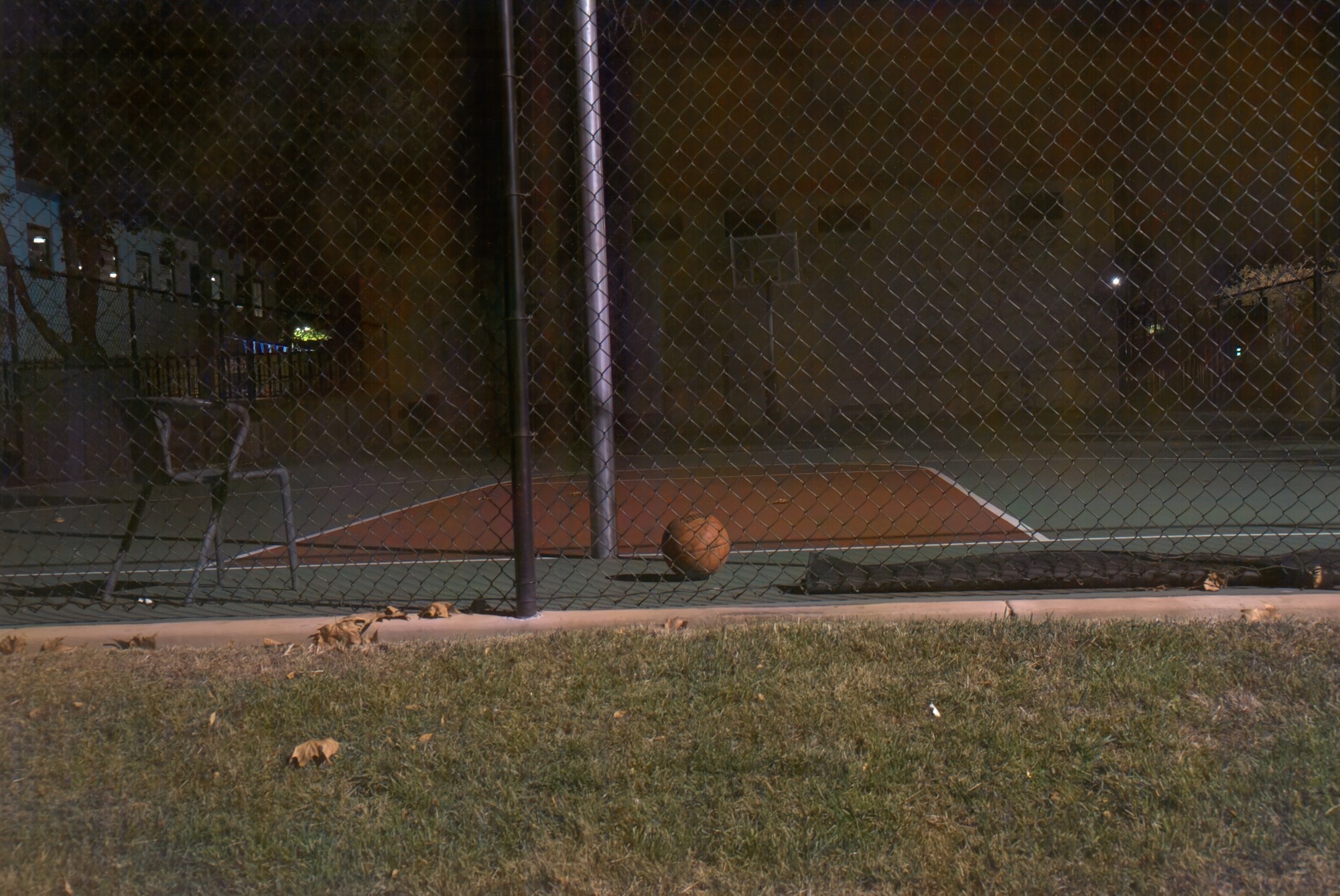} &
        \includegraphics[width=0.185\textwidth]{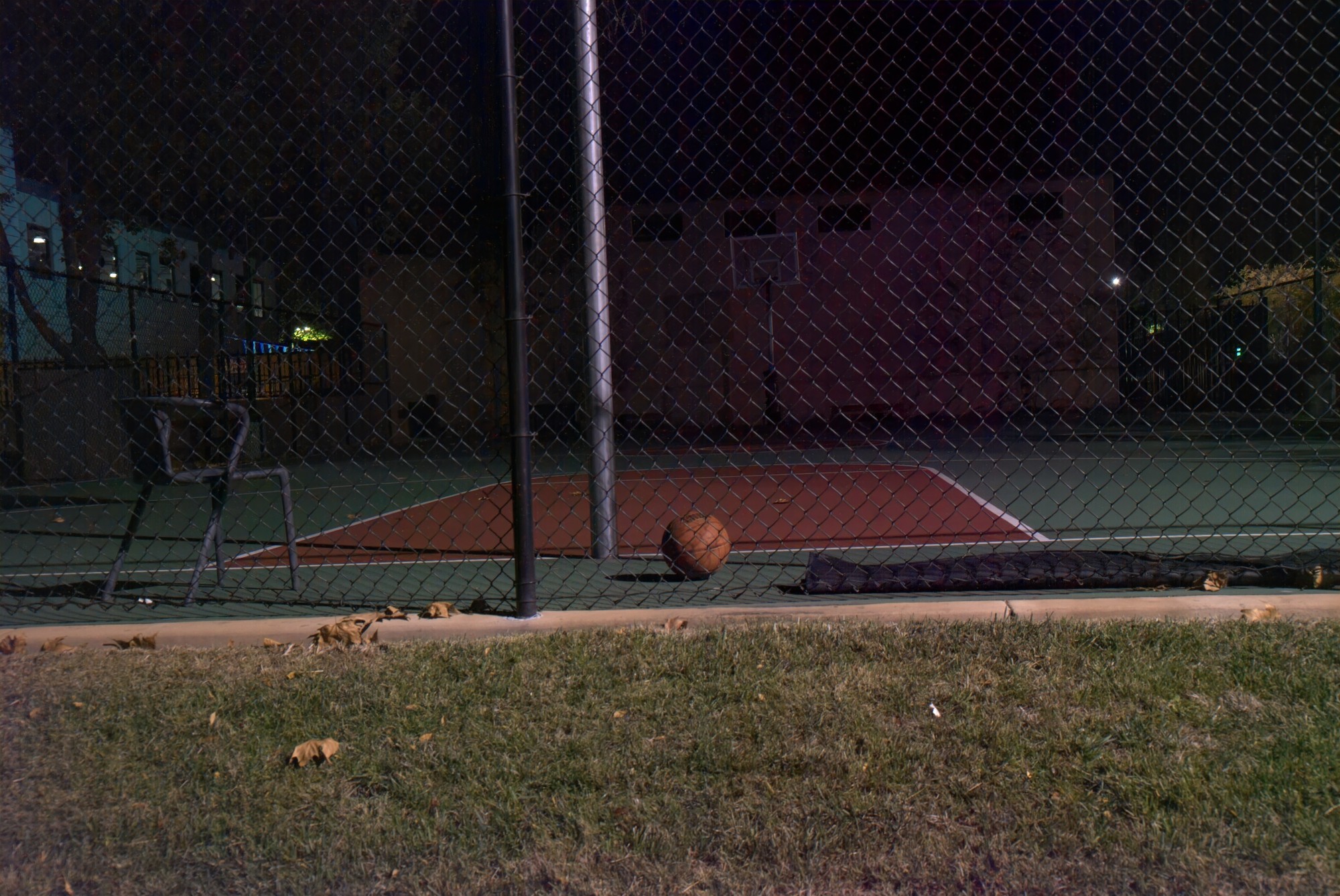} &
        \includegraphics[width=0.185\textwidth]{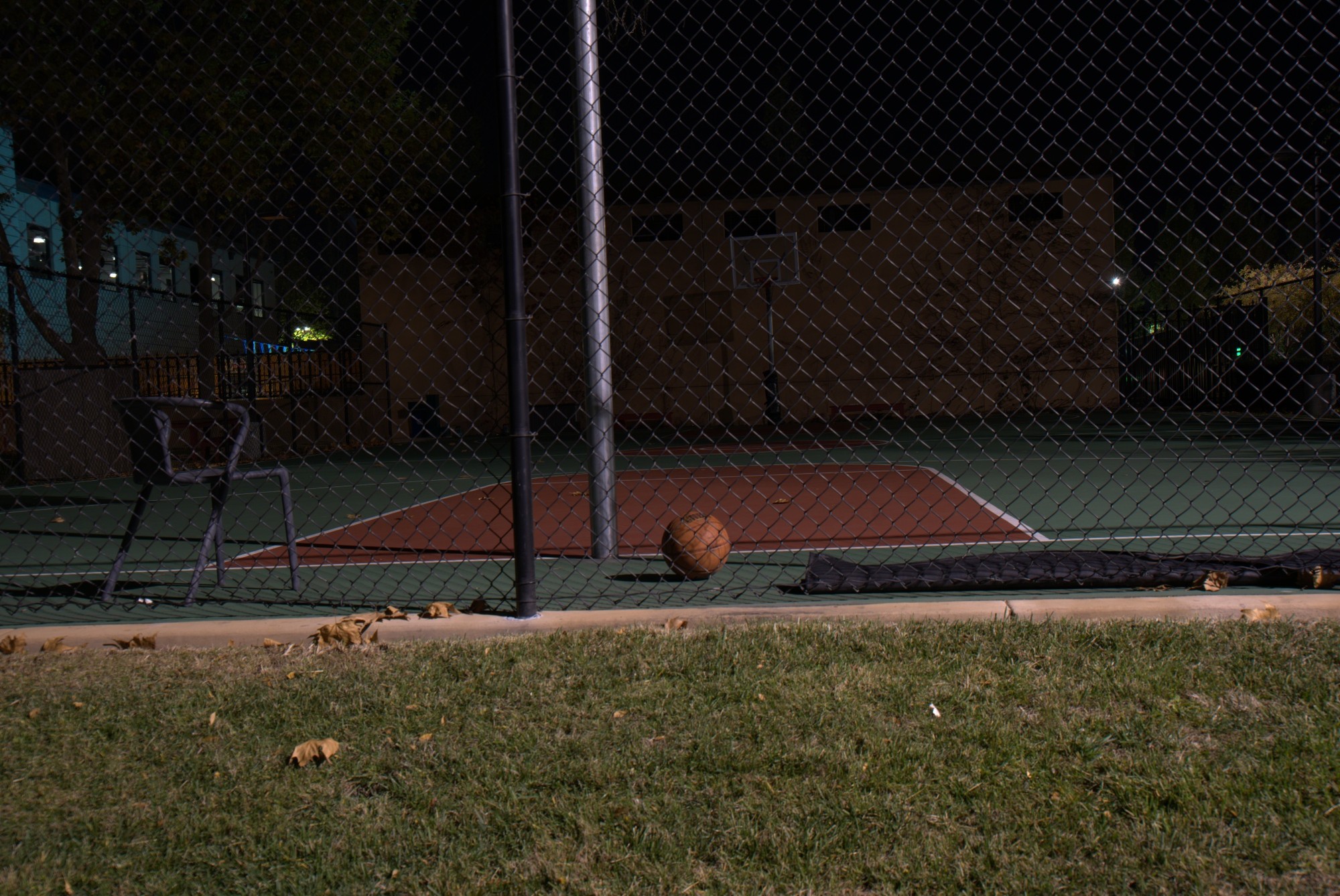} \\
        \rotatebox{90}{\small SID ISO 6400} &
        \includegraphics[width=0.185\textwidth]{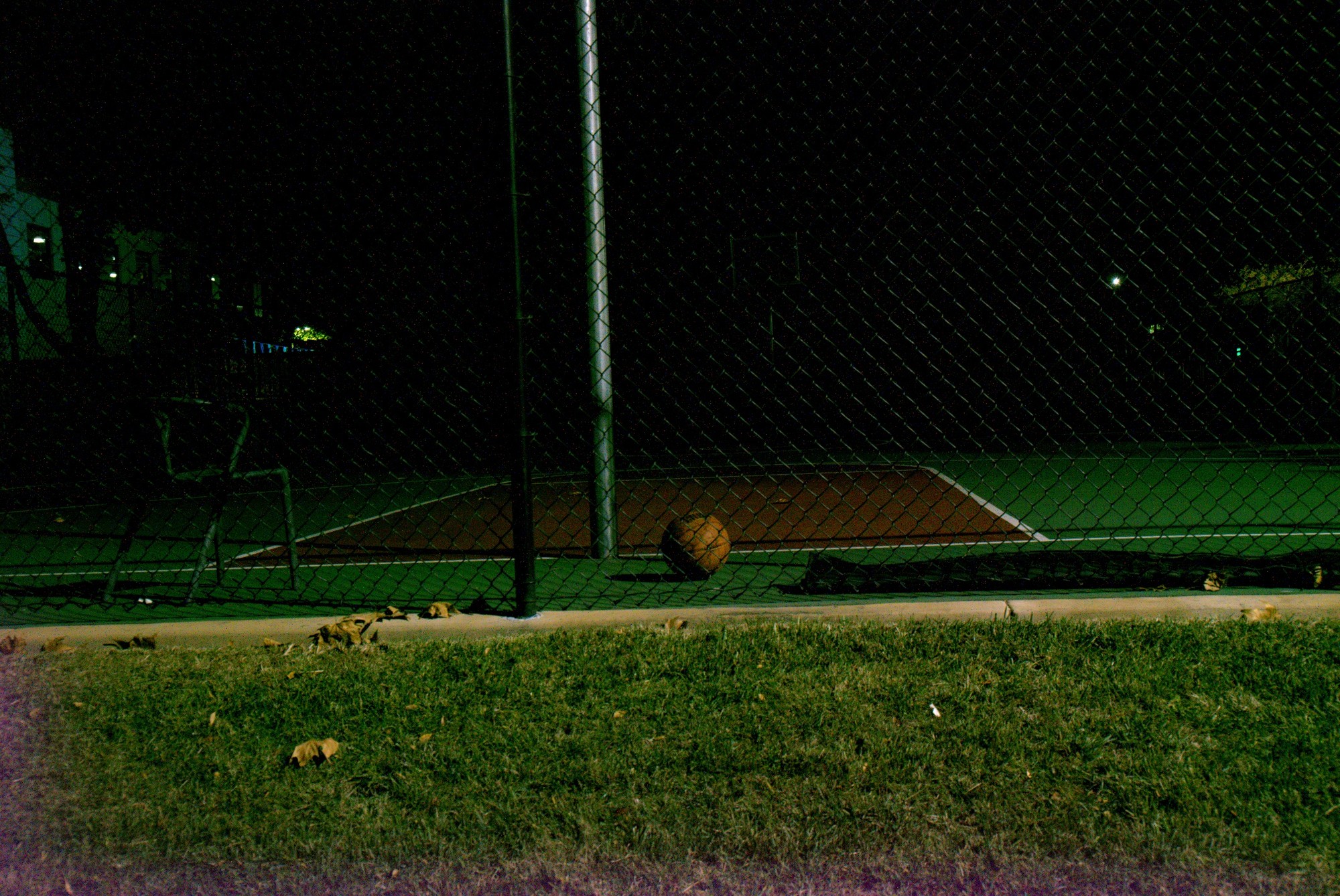} &
        \includegraphics[width=0.185\textwidth]{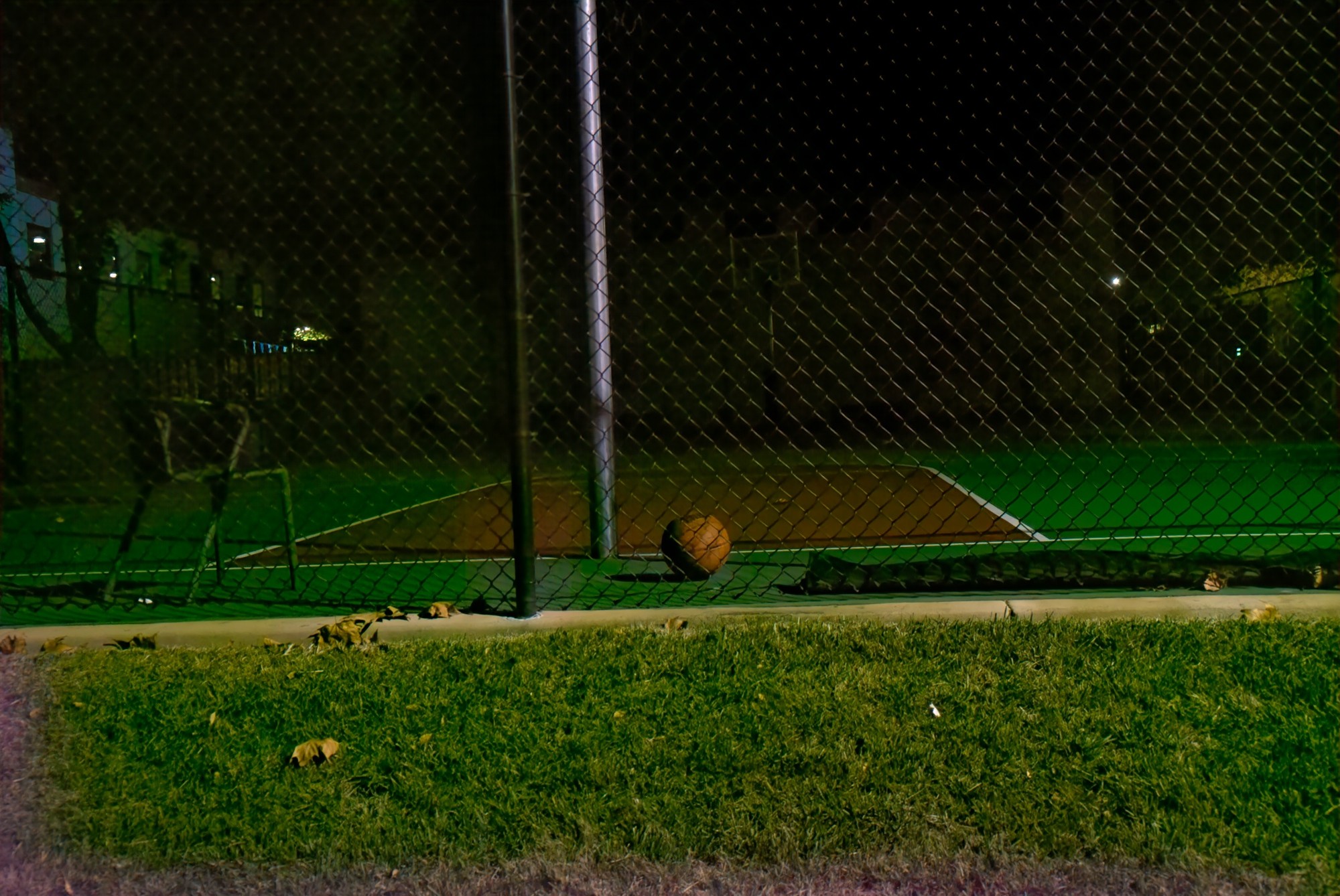} &
        \includegraphics[width=0.185\textwidth]{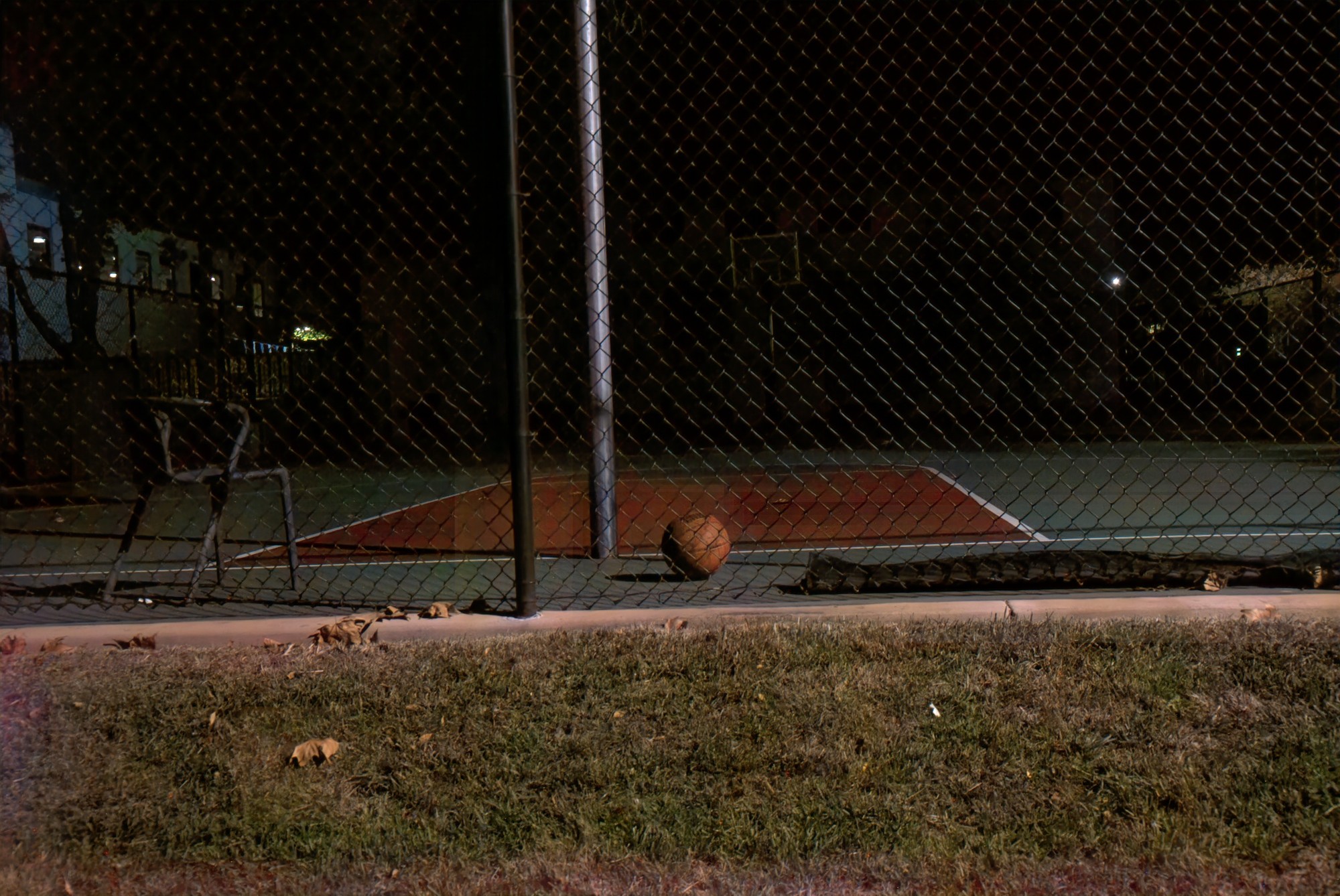} &
        \includegraphics[width=0.185\textwidth]{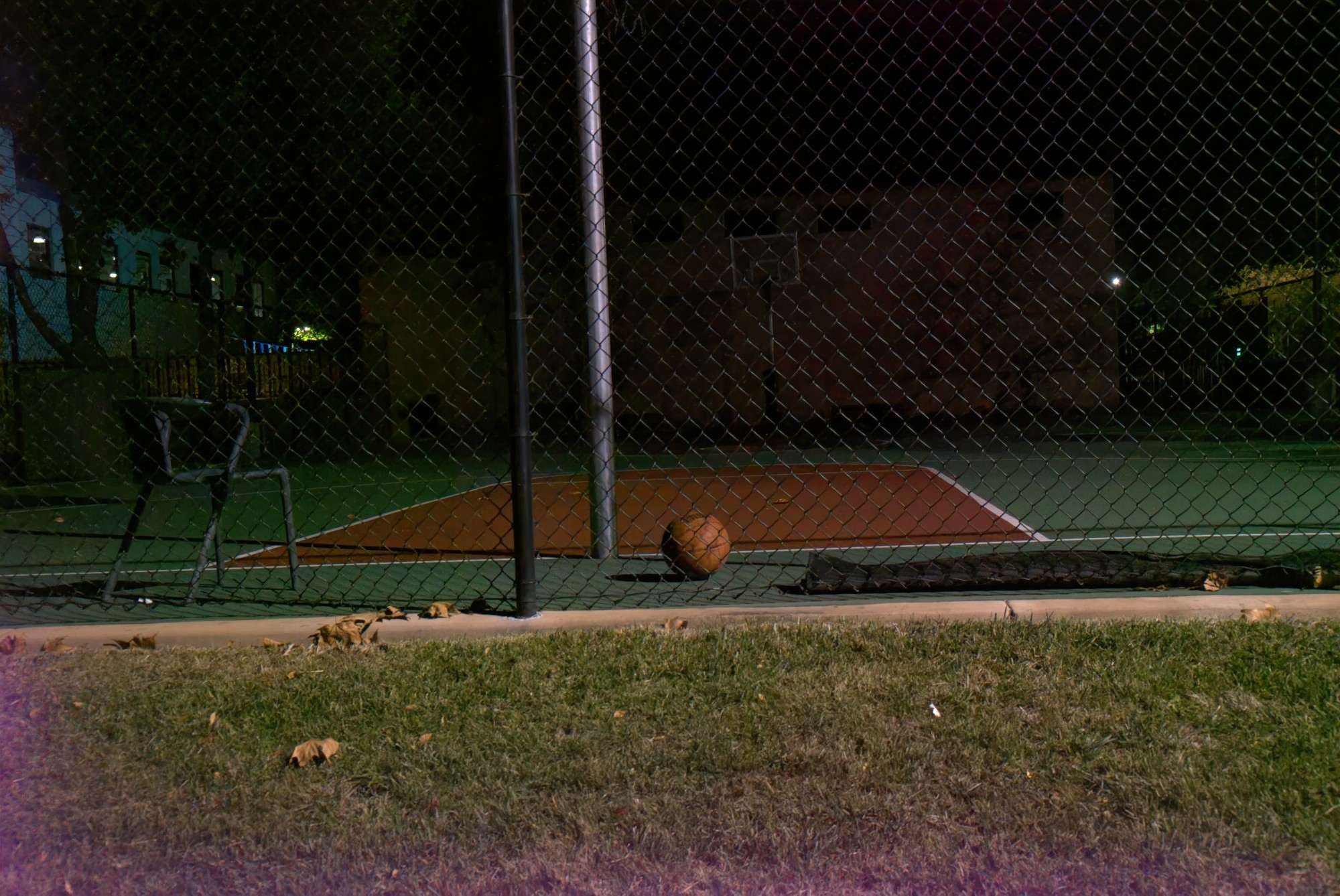} &
        \includegraphics[width=0.185\textwidth]{Figures/supp_darkshading/compare_ds_jpgs/GT.jpg} \\
        \rotatebox{90}{\small SID ISO 12800} &
        \includegraphics[width=0.185\textwidth]{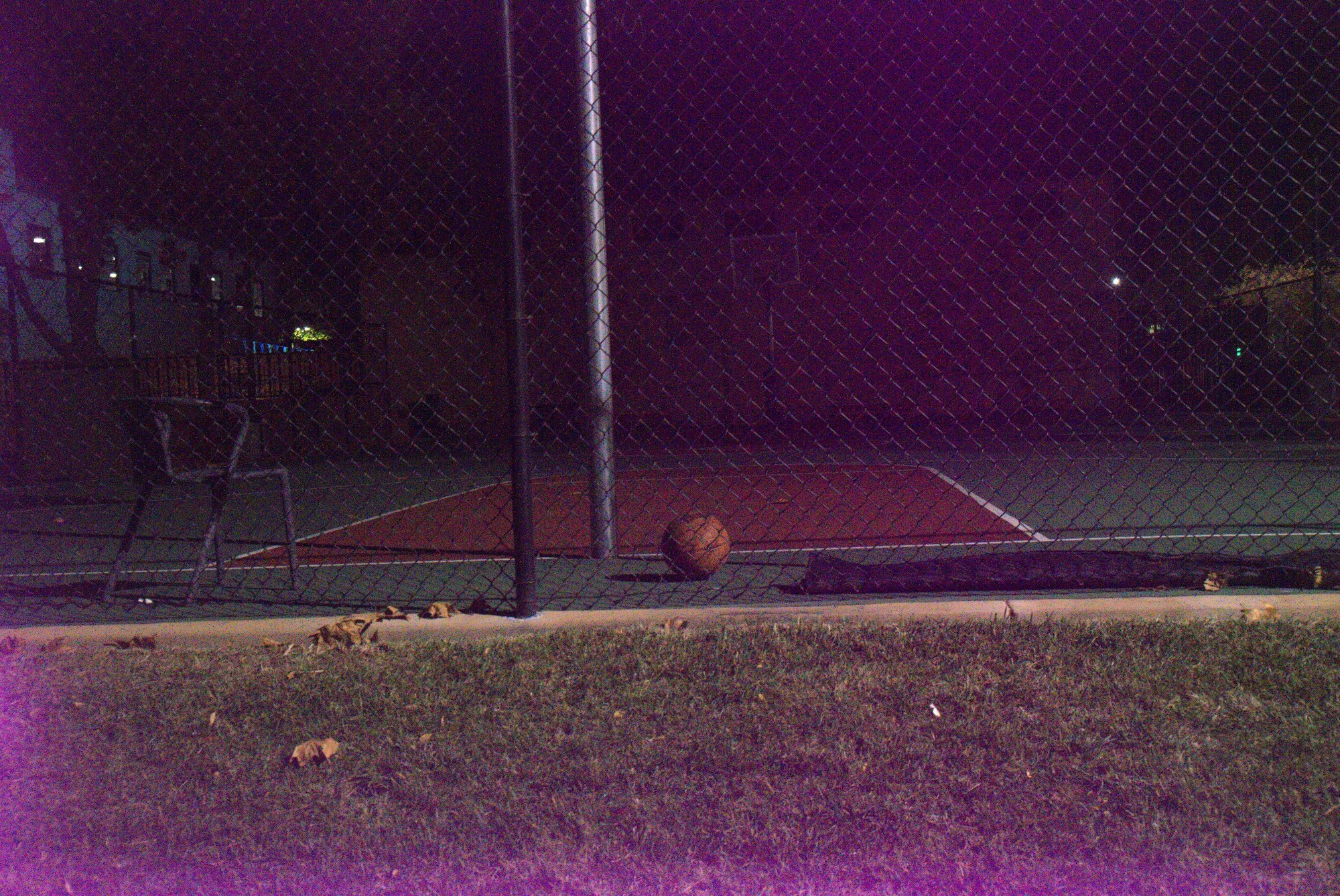} &
        \includegraphics[width=0.185\textwidth]{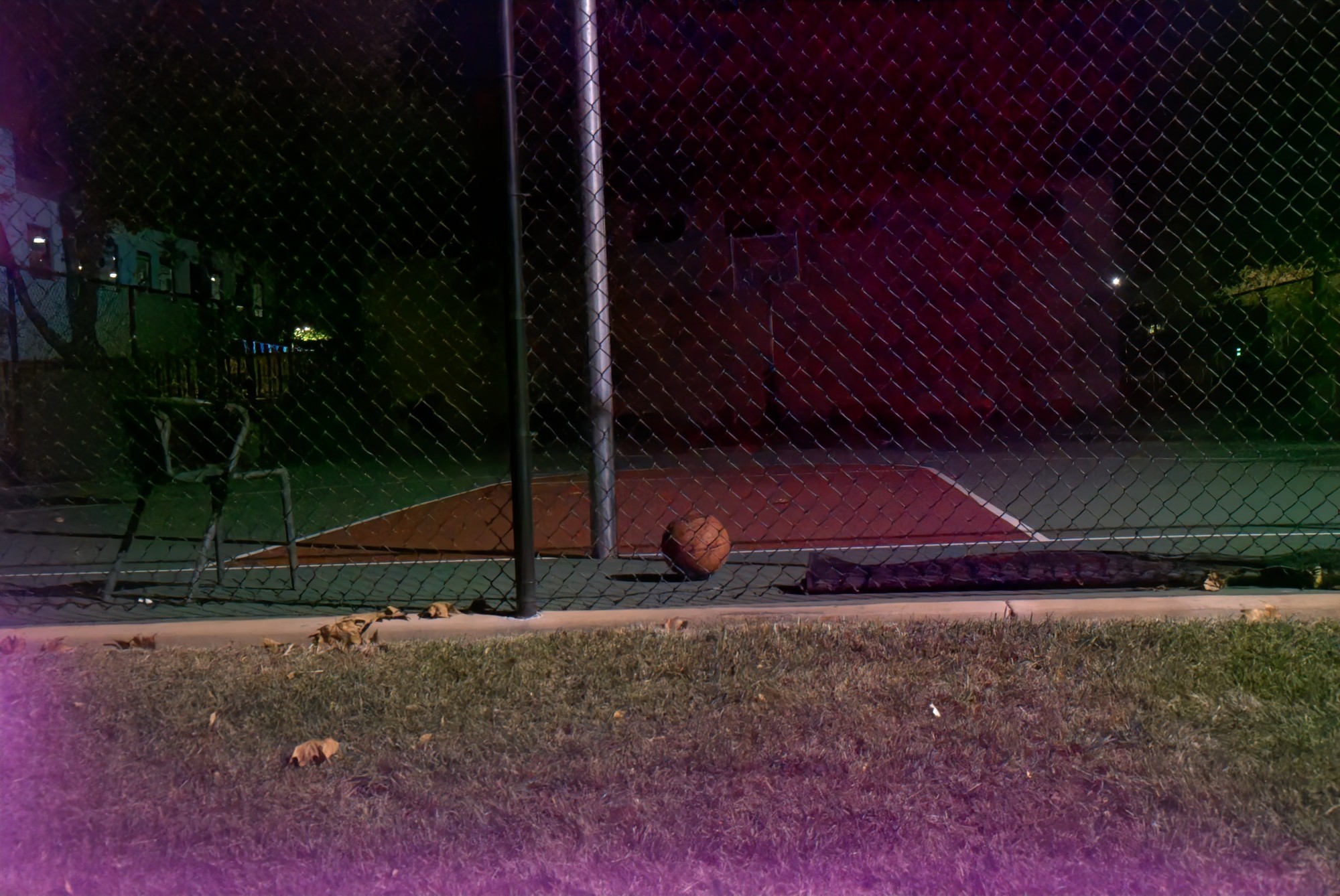} &
        \includegraphics[width=0.185\textwidth]{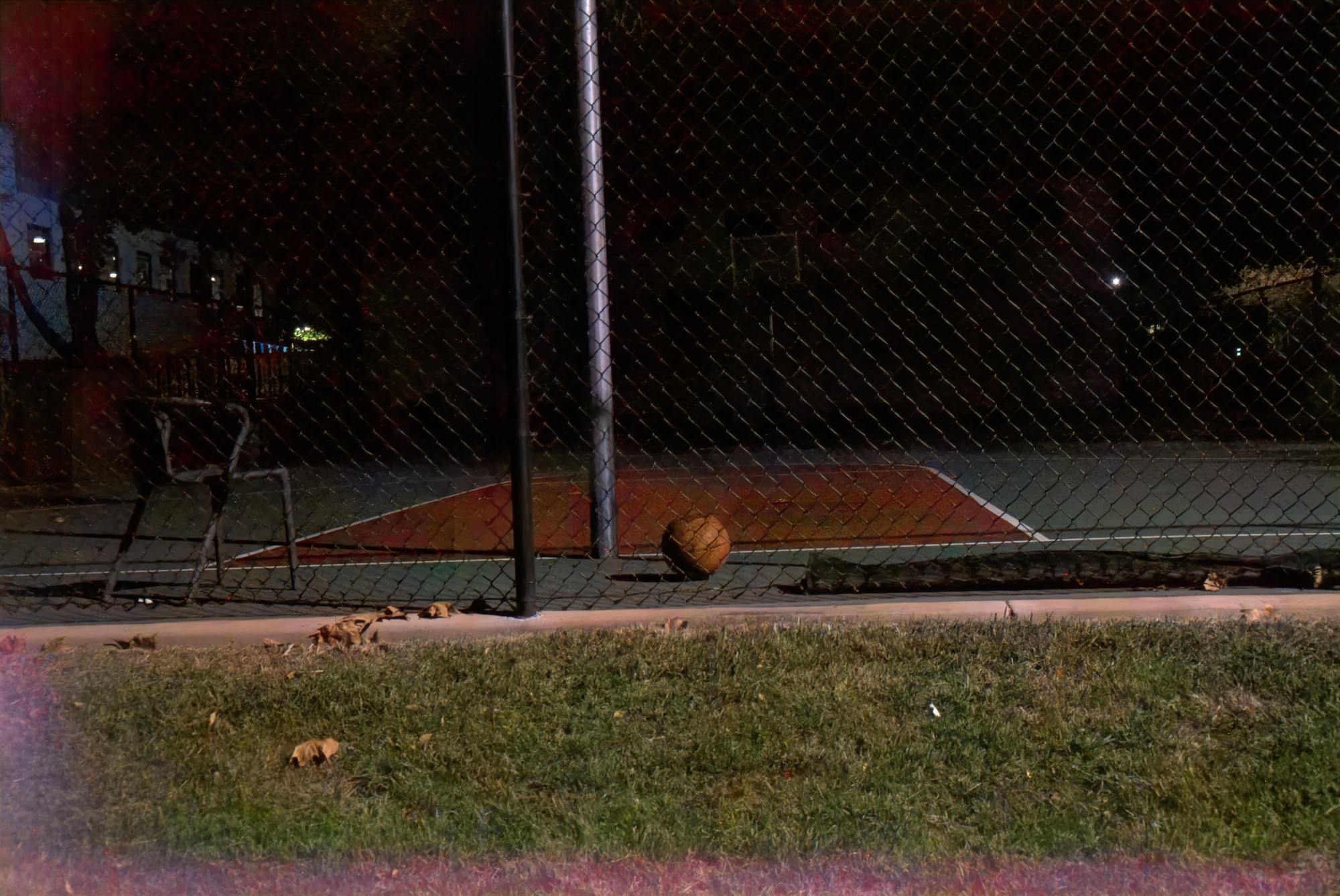} &
        \includegraphics[width=0.185\textwidth]{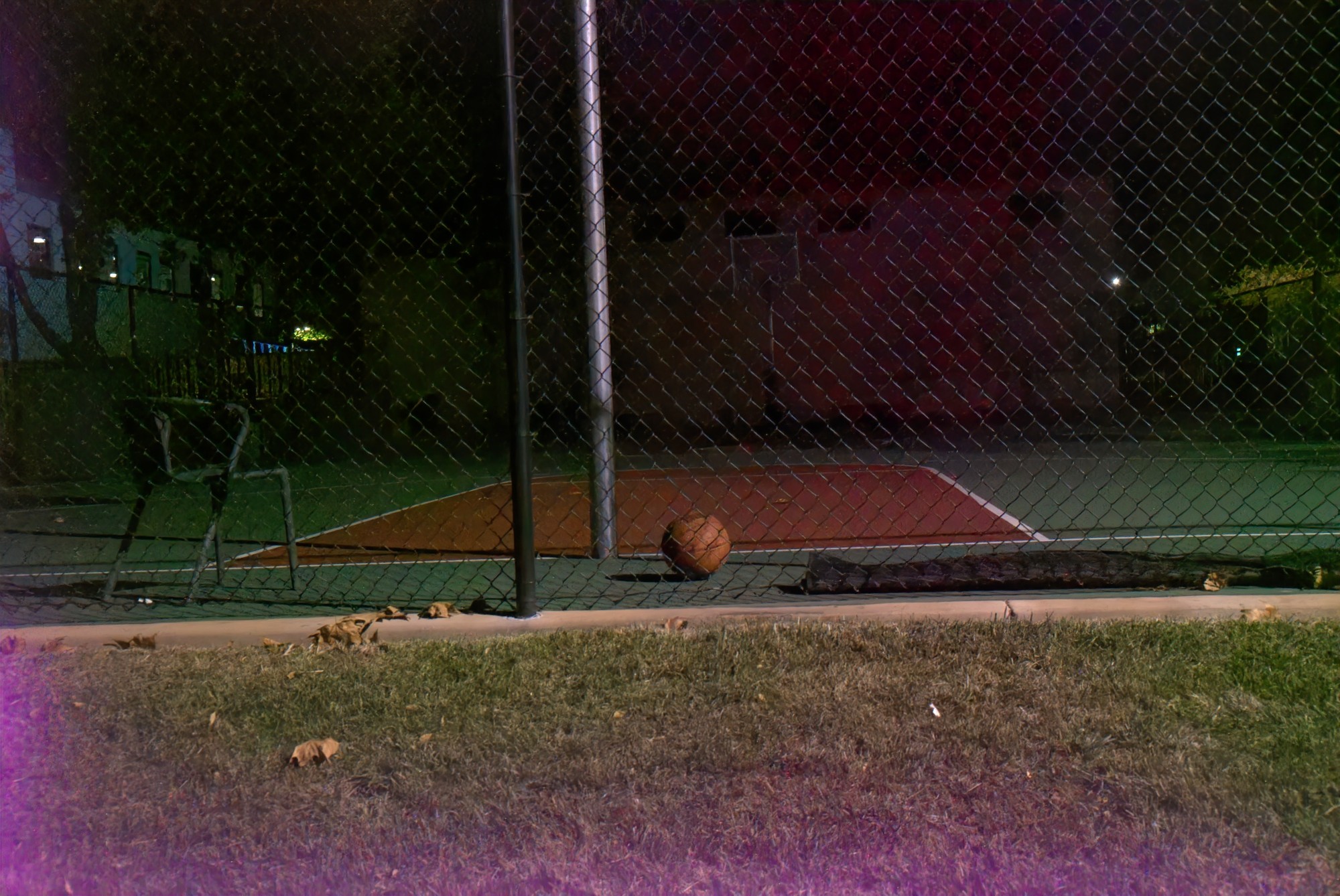} &
        \includegraphics[width=0.185\textwidth]{Figures/supp_darkshading/compare_ds_jpgs/GT.jpg} \\
        \rotatebox{90}{\small LRID ISO 6400} &
        \includegraphics[width=0.185\textwidth]{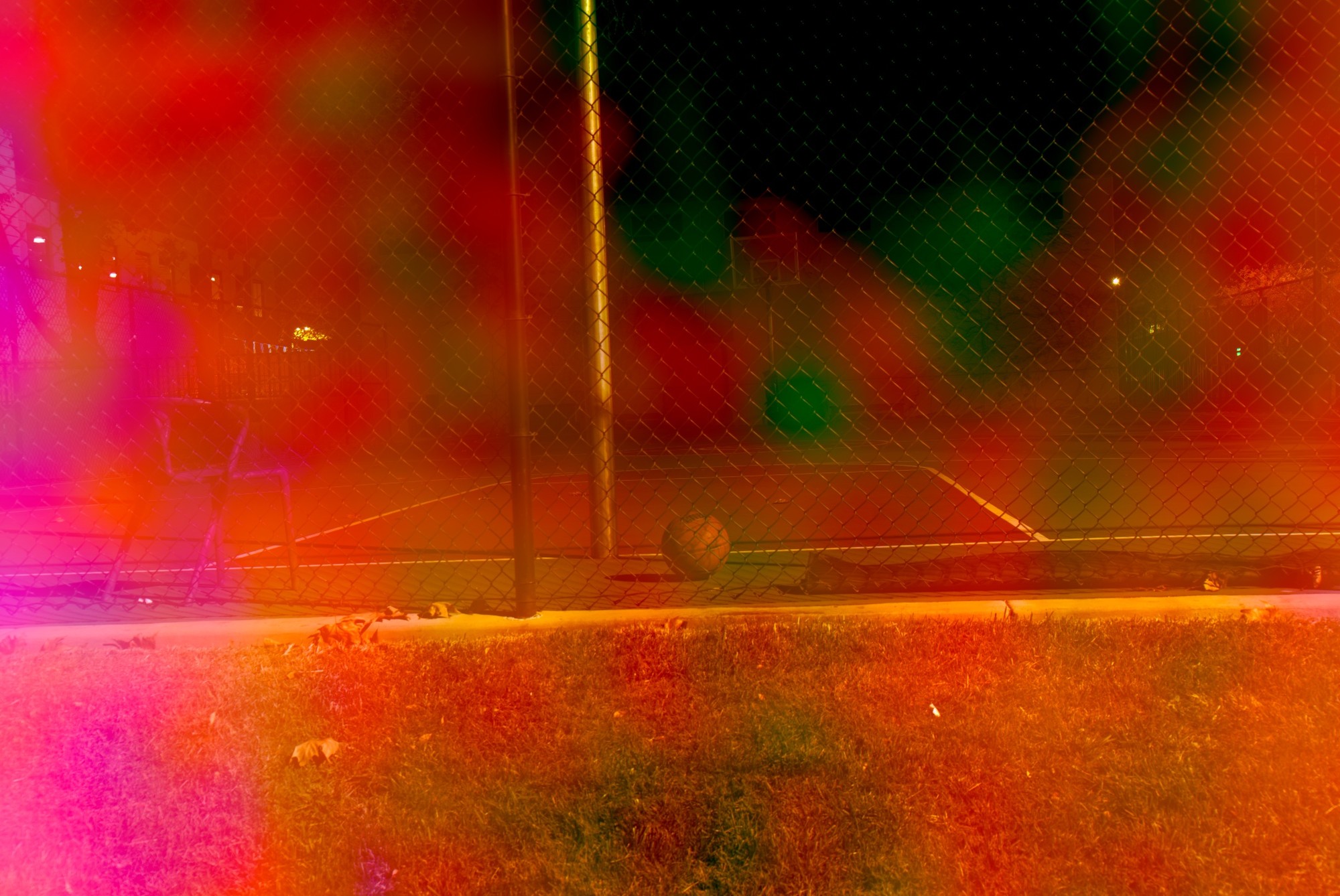} &
        \includegraphics[width=0.185\textwidth]{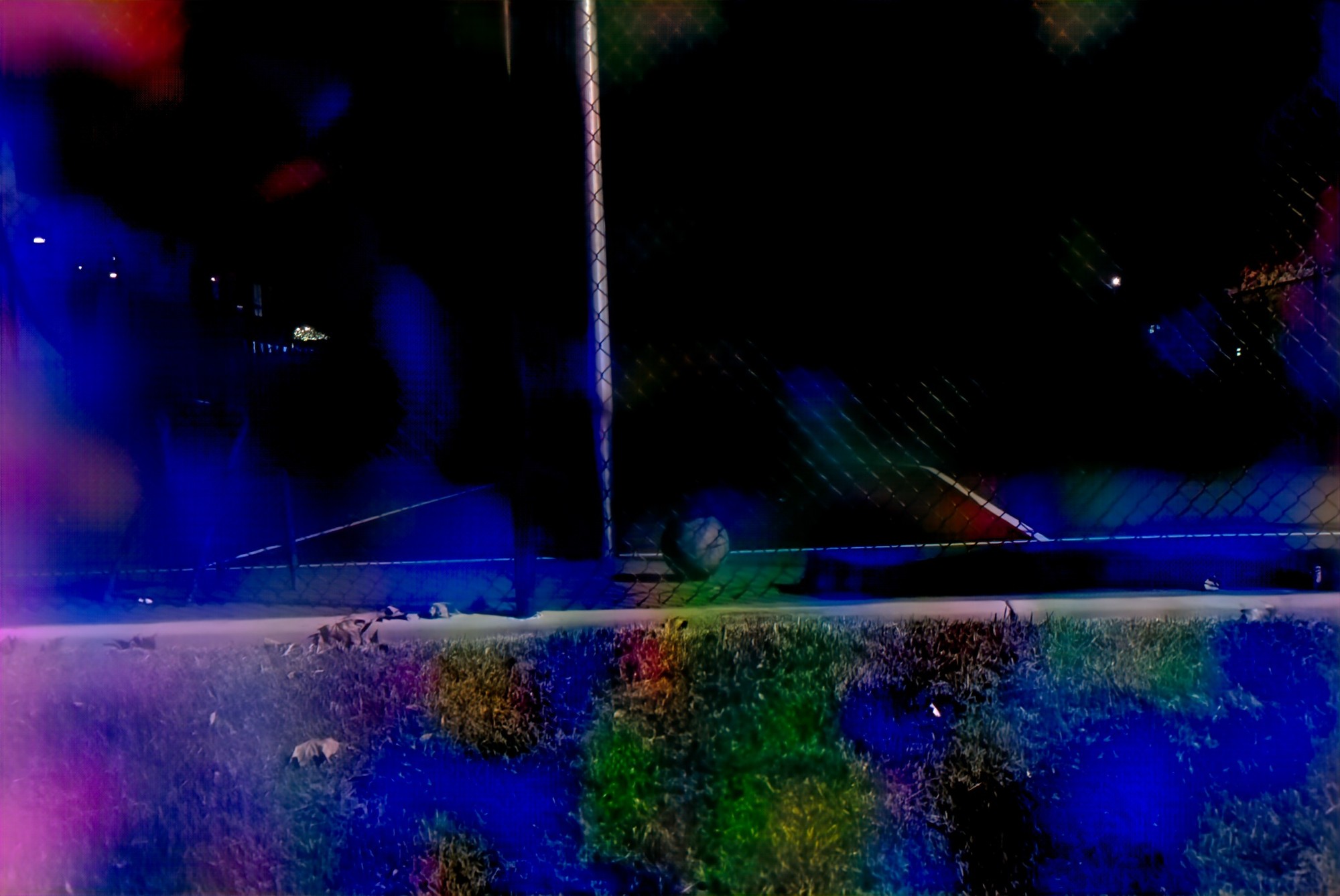} &
        \includegraphics[width=0.185\textwidth]{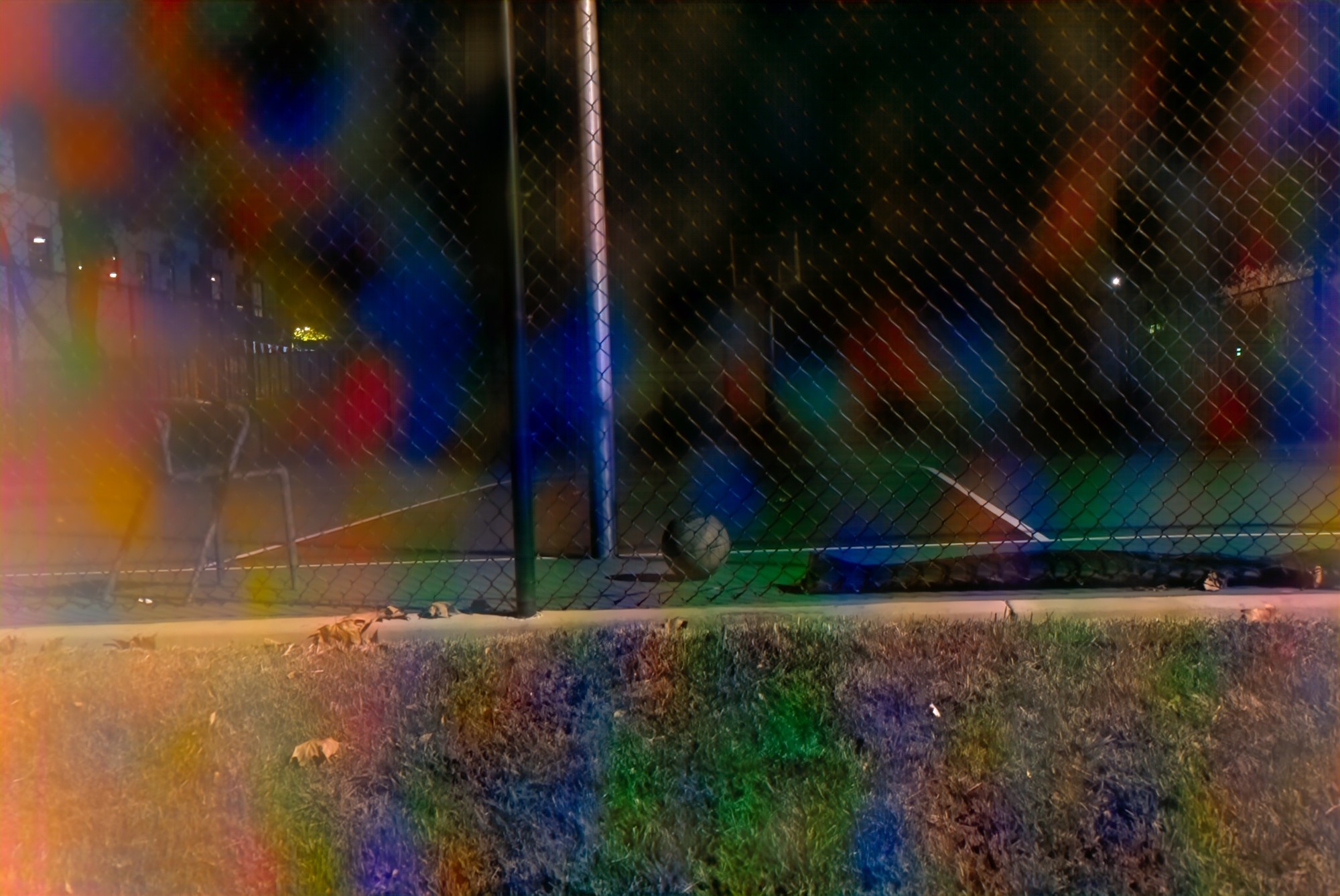} &
        \includegraphics[width=0.185\textwidth]{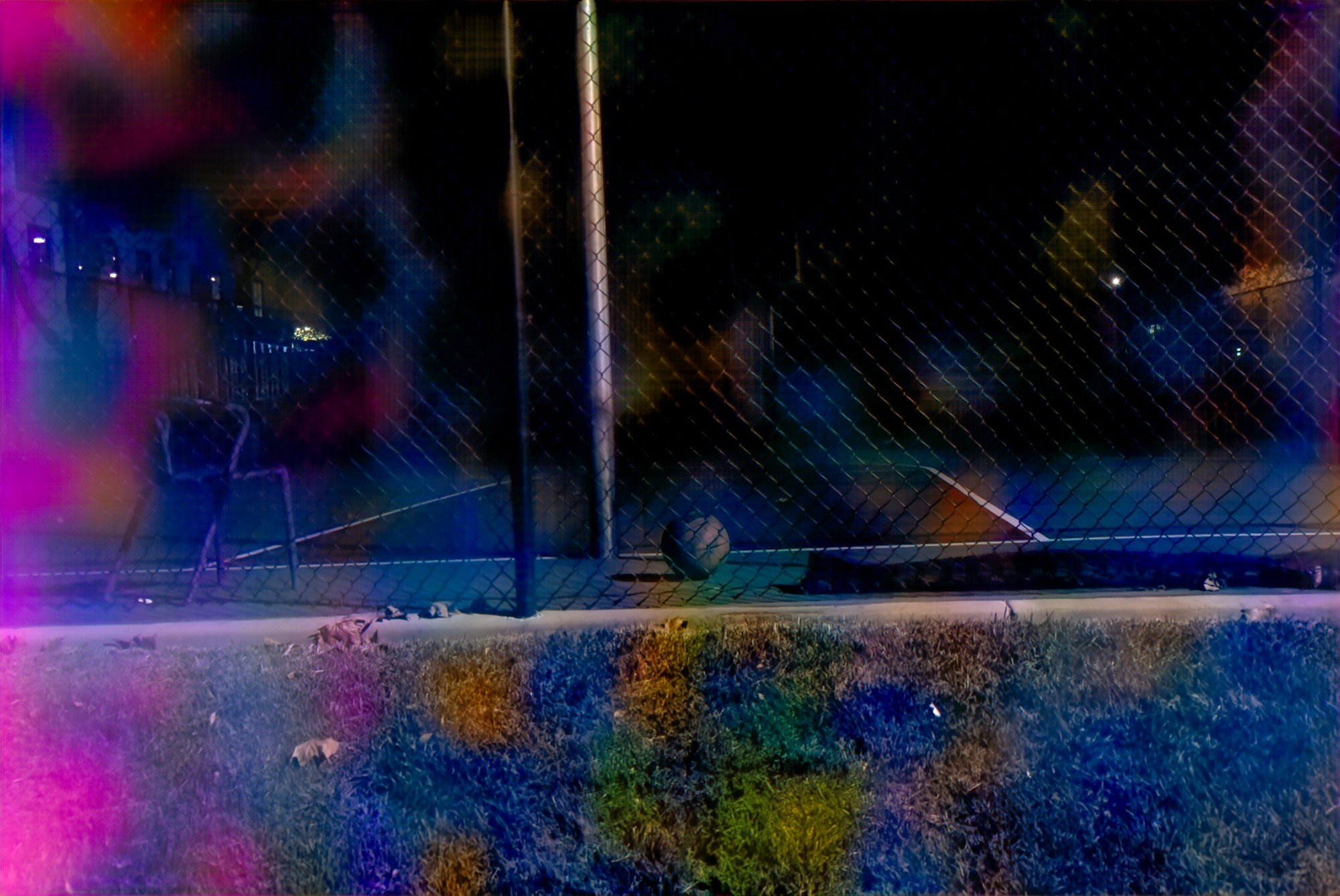} &
        \includegraphics[width=0.185\textwidth]{Figures/supp_darkshading/compare_ds_jpgs/GT.jpg} \\
    \end{tabular}
    \caption{\textbf{Synthetic dark-shading denoising across sensors and ISO settings.} A clean image from SID is corrupted with dark shading noise (amplified by $200\times$) drawn from four different sources: the SID sensor at ISO 1600, 6400, and 12800 (rows 1--3), and the LRID sensor at ISO 6400 (row 4). PGRQ fails to suppress the dark-shading across all configurations. PGRQB attenuates the overall dark-shading magnitude but introduces local color distortions, particularly visible on the LRID variant. Our model effectively corrects the black-level error, while the residual fixed-pattern component---which is not explicitly modeled in our framework---remains visible at higher ISO settings.}
    \label{fig:supp_synthetic_ds}
\end{figure*}

\label{sec:supp_dark_shading_analysis}
In this section, we characterize the dark shading noise of the sensors employed in the ELD-Sony~\cite{wei2021physics}, SID~\cite{chen2018learning}, and LRID~\cite{feng2022learnability} datasets, identify the conditions under which it constitutes a dominant source of image degradation, and assess the capacity of our model to handle a broad spectrum of dark shading patterns through synthetic experiments.

\subsection{Dark-shading Noise in ELD-Sony, SID, and LRID}
\label{sec:supp_dark_shading}
\label{sec:supp_ds_discussion}
Dark shading noise comprises two principal components: fixed-pattern noise (FPN) and black-level error (BLE)~\cite{feng2022learnability, feng2026learning, cao2023physics}. BLE does not follow any known parametric distribution, whereas FPN becomes increasingly stronger at higher ISO settings. In absolute terms, dark shading noise is generally of small magnitude. \cref{fig:supp_ds_heatmap} demonstrates the dark shading noise heatmaps for the SID (and ELD-Sony) and LRID sensors at ISO 6400, computed after black-level subtraction and normalization by the white-level. The values in \cref{fig:supp_SID_ds_heatmap,fig:supp_LRID_ds_heatmap} lie within the ranges $[-5\times10^{-4}, +5\times10^{-4}]$ and $[-1.4\times10^{-3}, +1.4\times10^{-3}]$, respectively, indicating that dark shading noise in LRID is approximately three times larger in magnitude than in SID.

Under sufficiently long exposures, the signal-to-dark-shading-noise ratio is high and dark shading is effectively imperceptible in the captured image. In short-exposure low-light regimes, however, only a small number of photons are integrated per pixel, and the rendered image appears nearly black. A standard strategy is to apply a multiplicative amplification factor that emulates a manual ISO adjustment. While this restores scene visibility, it simultaneously amplifies the dark shading noise by the same factor, rendering it a primary contributor to the residual degradation.

On ELD-Sony, and SID, the evaluated low-light inputs are amplified by factors ranging from $100\times$ to $300\times$, scaling the dark shading noise by an equivalent factor. For ISO 6400 with a $100\times$ amplification, for example, the dark shading range (shown in \cref{fig:supp_SID_ds_heatmap}) extends to $[-5\times10^{-2}, +5\times10^{-2}]$, a magnitude that can no longer be neglected. In contrast, LRID images at ratios $128\times$ and $256\times$ are amplified by only $2\times$ and $4\times$, respectively, owing to the design of the dataset~\cite{wei2021physics, feng2022learnability}, so that dark shading noise remains comparatively weak in the noisy input (e.g., $[-2.8\times10^{-3}, +2.8\times10^{-3}]$ at ISO 6400 with a $128\times$ ratio). This accounts for the observation that, although LRID inputs at ratios $128\times$ and $256\times$ exhibit overall noise levels comparable to those of ELD and SID at similar ratios, the relative contribution of dark shading is substantially smaller, thereby limiting the gains attainable through explicit black-level error correction.

This observation accounts for the comparable performance of our model and the PG and PGRQ baselines on LRID in \cref{tab:main_results,tab:supp_ssim_results}. Notably, even under this regime, PGRQB fails to adequately suppress the noise and produces inaccurate color reproduction (as shown in \cref{fig:supp_results_lrid}), which underscores the limitations of naively incorporating BLE during training and, conversely, the robustness of our approach in estimating BLE when dark shading noise is of low magnitude.

We further discuss a structural characteristic of the dark shading noise in ELD-Sony, SID, and LRID. All noisy images in LRID are captured at ISO 6400, which yields a single, consistent dark shading pattern across the dataset. In contrast, ELD and SID contain images captured at multiple ISO levels, each associated with a distinct dark shading pattern. Since the black-level error can be modeled as a uniform offset applied to the R, G1, B, and G2 channels of a Bayer image, it should appear as a globally constant shift in each per-channel heatmap. \cref{fig:supp_SID_ds_heatmap} exhibits a largely uniform yellow region in the central area of each channel, indicating the presence of a black-level error component, although fixed-pattern noise becomes dominant near the image borders. In contrast, the per-channel heatmaps in \cref{fig:supp_LRID_ds_heatmap} display substantial spatial color variation, which suggests that the fixed-pattern component dominates the dark shading noise for the LRID sensor.

\subsection{Denoising Dark Shading Noise Across ISOs and Sensors}
\label{sec:supp_ds_synthetic}
We design a synthetic experiment to assess the capacity of our model and the baselines to handle dark shading noise. In \cref{fig:supp_synthetic_ds}, a clean image from the SID dataset is corrupted with dark shading noise amplified by a factor of $200\times$, yielding four distinct noisy variants: three using the SID sensor at ISO 1600, 6400, and 12800, and one using the LRID sensor at ISO 6400.

A first observation is that the dark shading characteristics of the SID sensor vary substantially across ISO settings, producing markedly different noisy inputs. Across all SID variants, PGRQ fails to suppress the dark shading noise. PGRQB reduces the overall dark shading magnitude but introduces local color distortions; at ISO 12800, it also attenuates the fixed-pattern component, which suggests that the model treats color-related errors as a local degradation. Our model successfully corrects the black-level error, although the fixed-pattern component remains visible, as it is not explicitly modeled in our framework.

The LRID variant exhibits pronounced color distortions and constitutes a particularly challenging case. PGRQ suppresses the dark shading noise but simultaneously diminishes the overall intensity. PGRQB likewise attenuates the dark shading, but produces spatially inconsistent colors that misrepresent the underlying scene. Our model effectively corrects the black-level error, while the fixed-pattern noise persists due to its considerable magnitude.

\subsection{Additional Results on Higher Exposure Ratios of LRID}
\label{sec:supp_lrid_fail}
\begin{table}[t]
\centering
\footnotesize

\resizebox{\linewidth}{!}{%
\begin{tabular}{ccccccc}
\toprule
\multirow{2}{*}{Dataset} &
\multirow{2}{*}{Ratio} &
\multicolumn{4}{c}{Blind} \\
\cmidrule(lr){3-6}
& &
PG &
PGRQ & 
PGRQB &
Ours & 
Ours$^{*}$ \\
\midrule

\multirow{5}{*}{LRID - Indoor}

&
$\times64$
& \psnrssim{\textbf{48.73}}{\textbf{0.990}}
& \psnrssim{48.47}{\textbf{0.990}}
& \psnrssim{44.93}{0.980}
& \psnrssim{47.99}{0.987}
& \psnrssim{\underline{48.61}}{\textbf{0.990}}
\\

&
$\times128$
& \psnrssim{\underline{46.79}}{\underline{0.983}}
& \psnrssim{46.73}{\textbf{0.984}}
& \psnrssim{43.76}{0.973}
& \psnrssim{46.34}{0.977}
& \psnrssim{\textbf{46.82}}{\underline{0.983}}
\\

&
$\times256$
& \psnrssim{\underline{44.12}}{0.968}
& \psnrssim{44.18}{\textbf{0.971}}
& \psnrssim{42.26}{0.960}
& \psnrssim{43.91}{0.962}
& \psnrssim{\textbf{44.22}}{\underline{0.969}}
\\

&
$\times512$
& \psnrssim{\underline{41.09}}{\underline{0.939}}
& \psnrssim{\textbf{41.34}}{\textbf{0.946}}
& \psnrssim{40.20}{0.936}
& \psnrssim{40.84}{0.928}
& \psnrssim{40.98}{0.931}
\\

&
$\times1024$
& \psnrssim{\underline{37.83}}{\underline{0.887}}
& \psnrssim{\textbf{38.37}}{\underline{0.903}}
& \psnrssim{\textbf{37.83}}{\textbf{0.905}}
& \psnrssim{37.54}{0.880}
& \psnrssim{37.74}{\underline{0.887}}
\\

\bottomrule
\end{tabular}%
}

\caption{\textbf{Quantitative results on LRID-Indoor~\cite{feng2022learnability} under blind settings} (PSNR\,/\,SSIM, both $\uparrow$; PSNR in dB). Ours$^{*}$ averages per-image BLE predictions across captures at the same ISO and exposure (Sec.~C.3). Bold and underline mark the best and second-best results.}
\label{tab:supp_lrid_psnrssim}
\end{table}

\begin{figure*}[t]
\centering

\subfloat[Noisy]{%
    \includegraphics[width=0.235\textwidth]{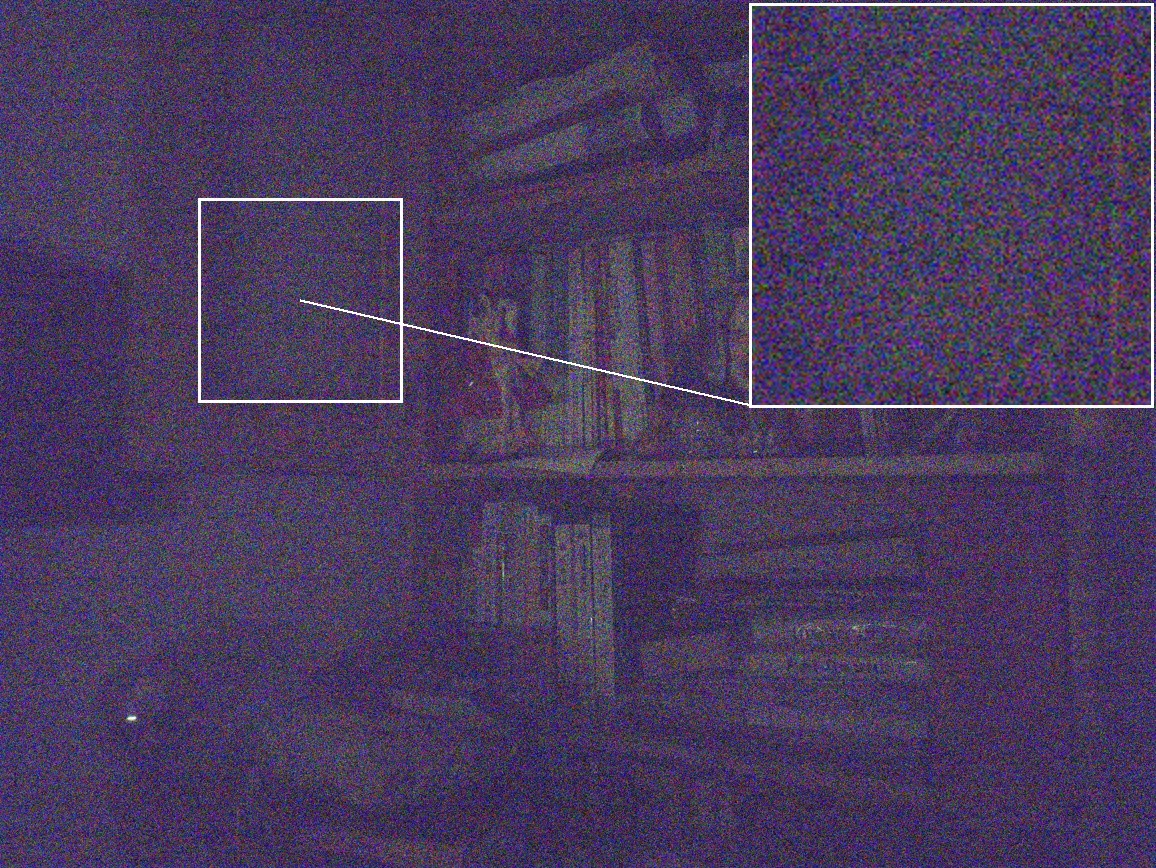}
    \label{fig:results_lrid_noisy_fail}
}
\subfloat[PG]{%
    \includegraphics[width=0.235\textwidth]{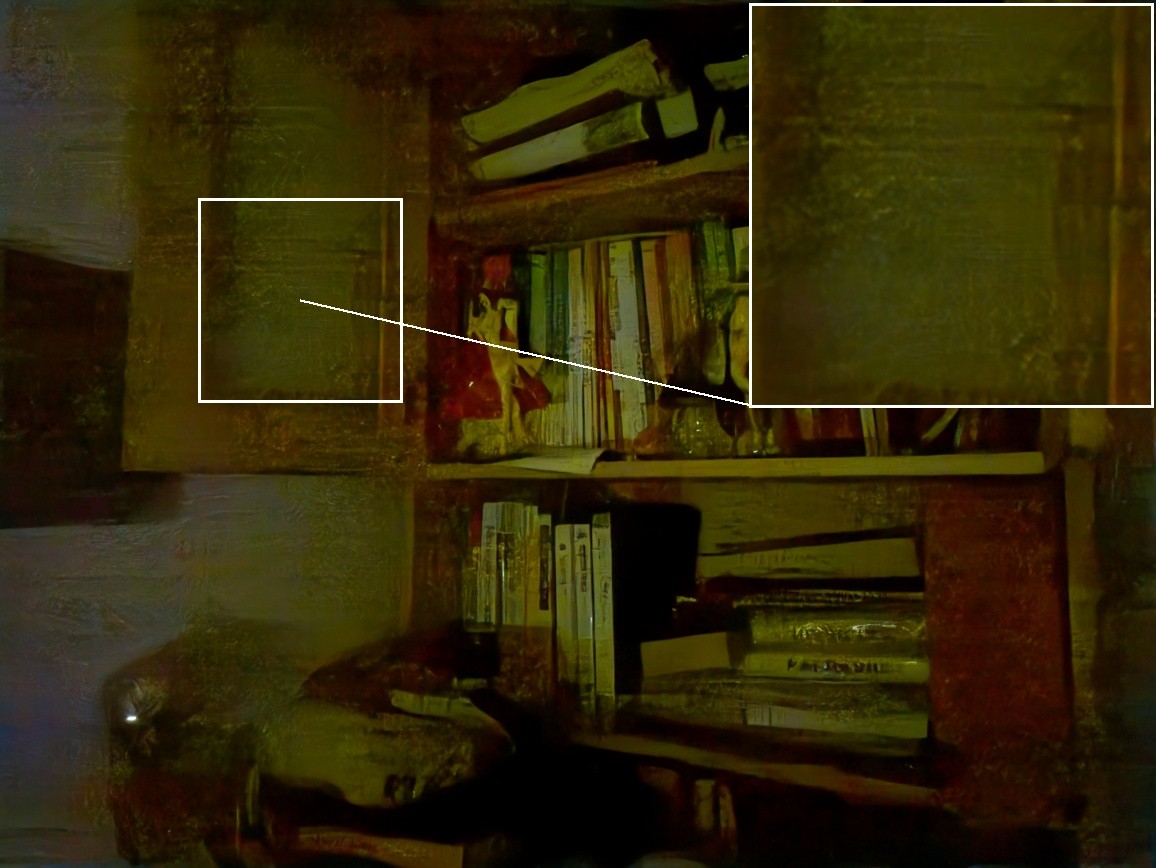}
    \label{fig:results_lrid_pg_fail}
}
\subfloat[PGRQ]{%
    \includegraphics[width=0.235\textwidth]{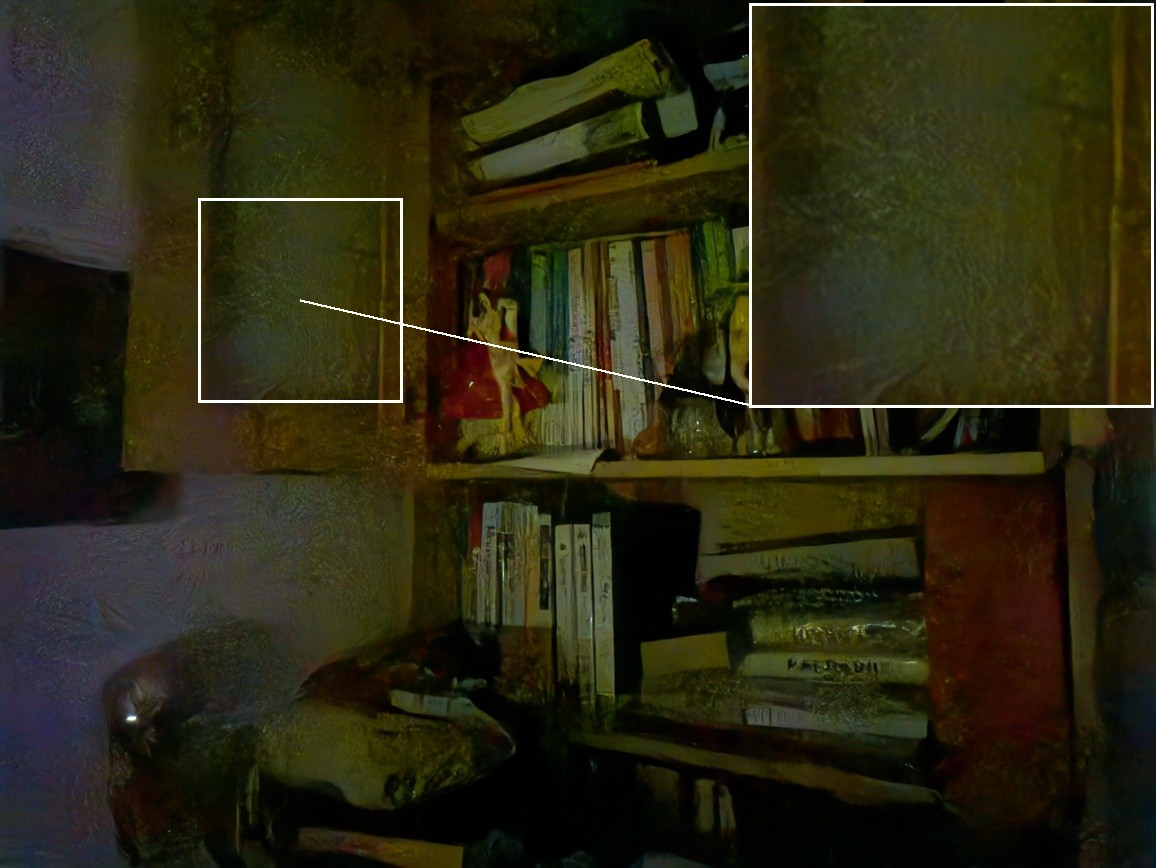}
    \label{fig:results_lrid_pgrq_fail}
}
\subfloat[PGRQB]{%
    \includegraphics[width=0.235\textwidth]{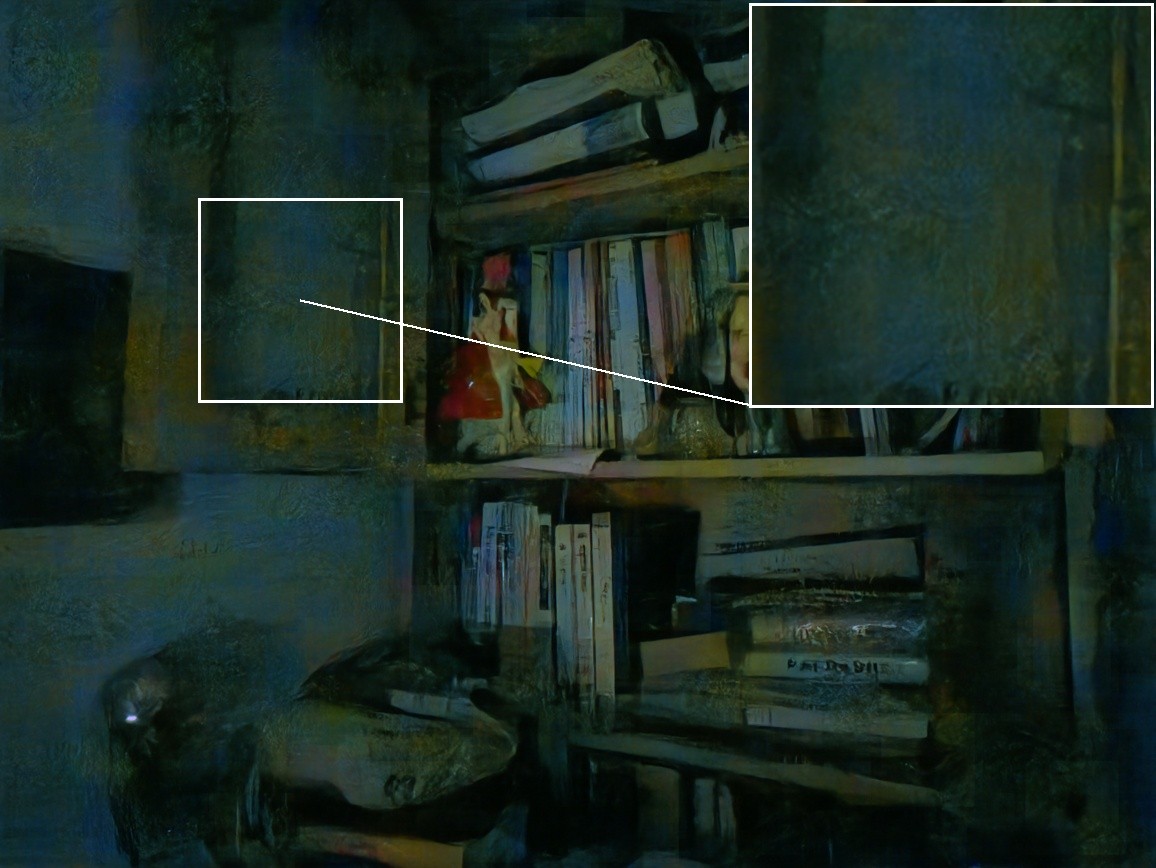}
    \label{fig:results_lrid_pgrqb_fail}
}\\

\vspace{2pt}

\subfloat[ELD~\cite{wei2021physics}]{%
    \includegraphics[width=0.235\textwidth]{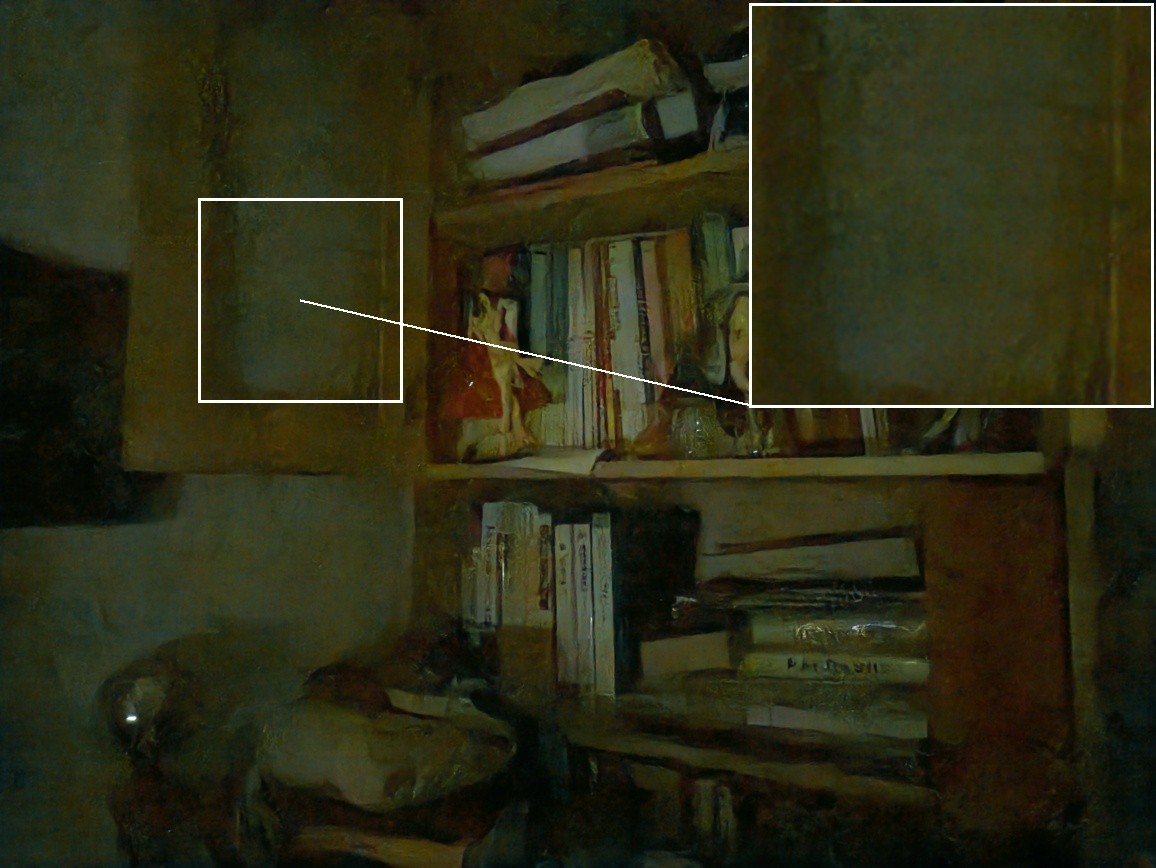}
    \label{fig:results_lrid_eld_fail}
}
\subfloat[2-Shots~\cite{lu20252}]{%
    \includegraphics[width=0.235\textwidth]{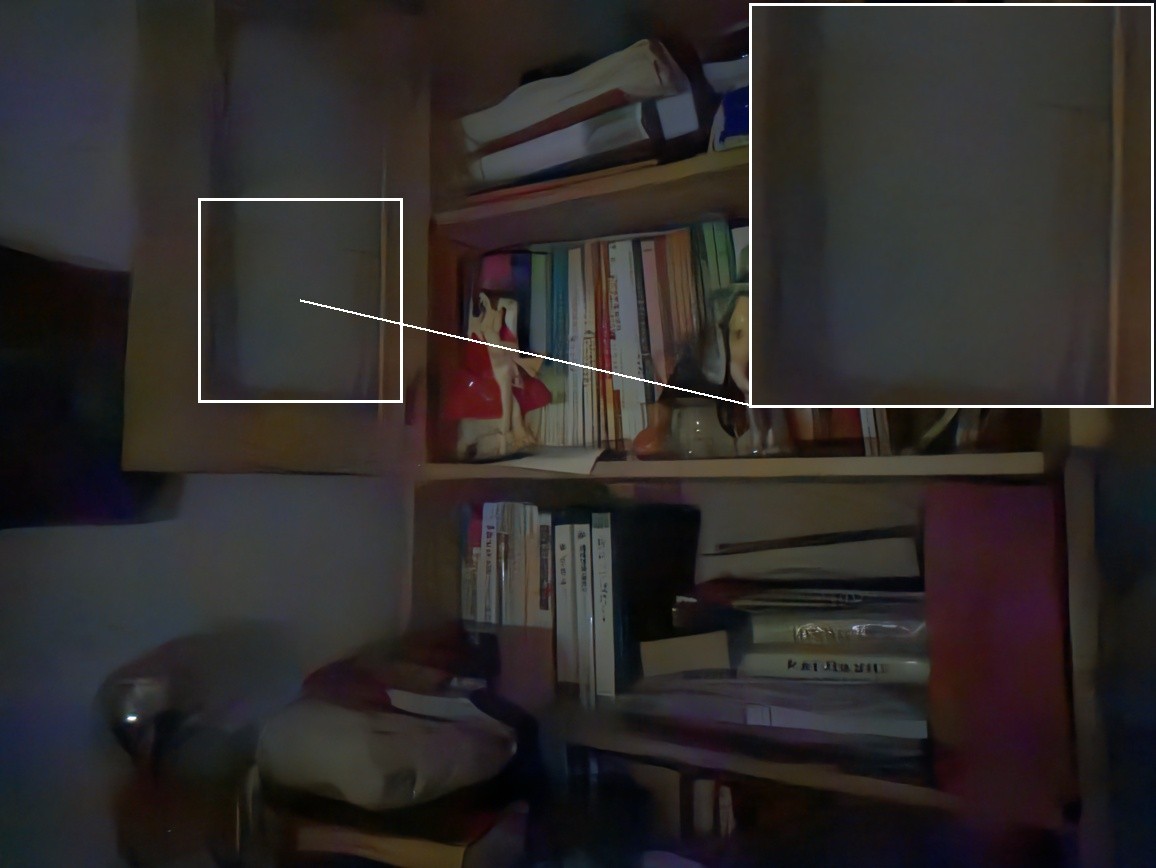}
    \label{fig:results_lrid_2shots_fail}
}
\subfloat[Ours]{%
    \includegraphics[width=0.235\textwidth]{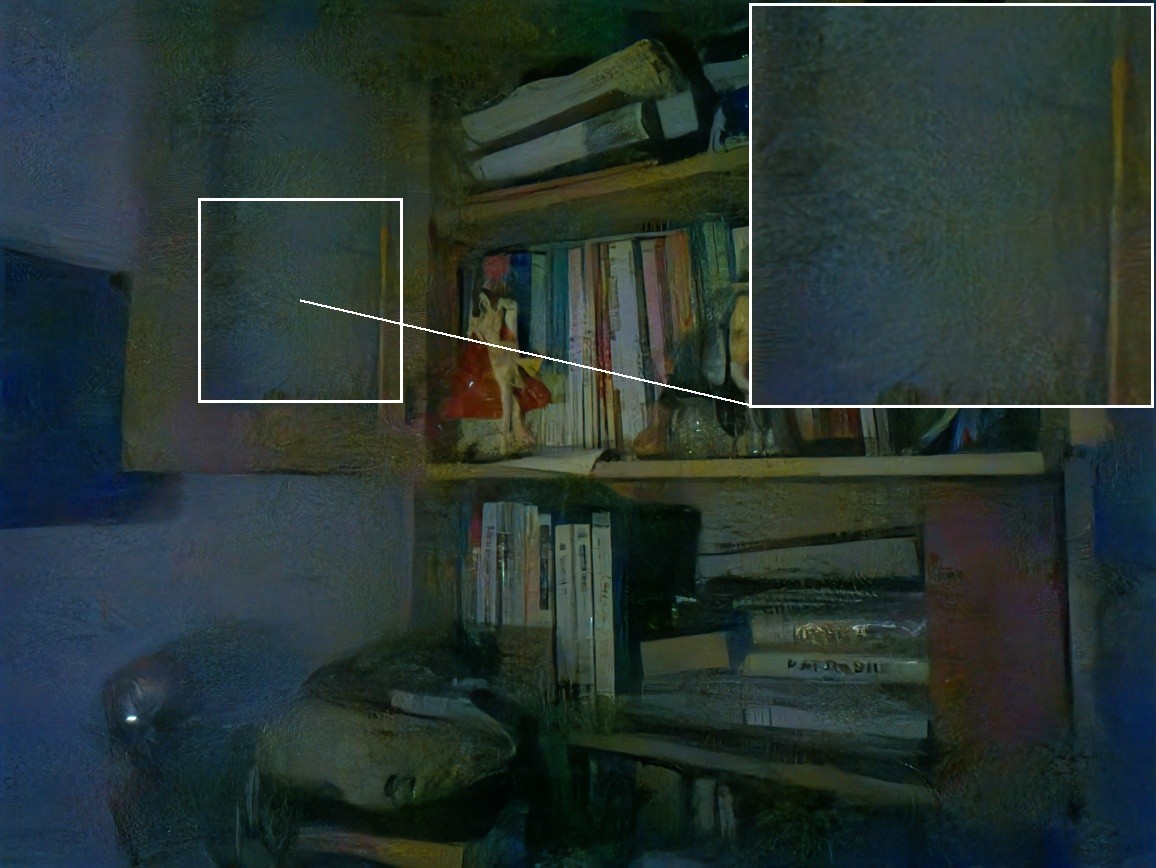}
    \label{fig:results_lrid_ours_fail}
}
\subfloat[GT]{%
    \includegraphics[width=0.235\textwidth]{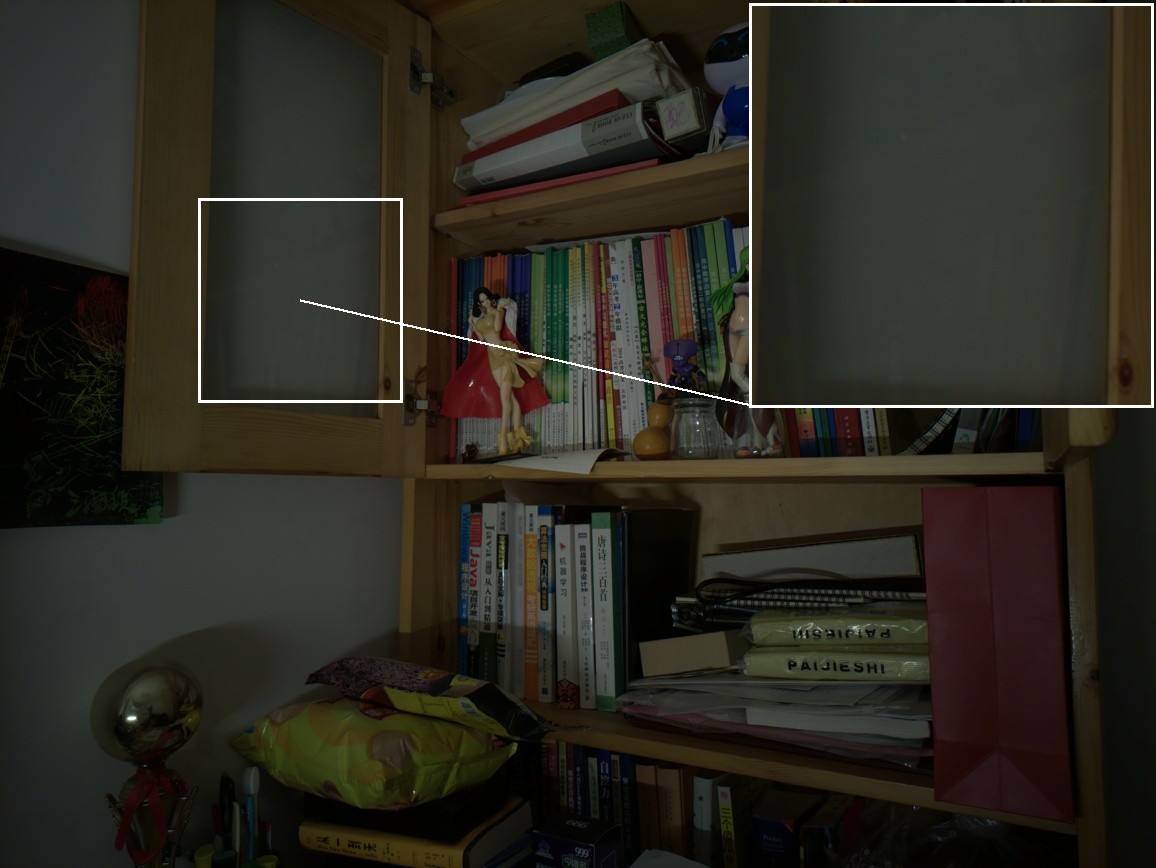}
    \label{fig:results_lrid_gt_fail}
}

\caption{\textbf{Qualitative comparison on LRID~\cite{feng2022learnability} at the $1024\times$ exposure ratio.} A representative failure case for our approach. PG and PGRQ exhibit residual dark shading noise. PGRQB alters the colors but introduces local inconsistencies. ELD, despite applying dark shading correction prior to denoising, fails to recover accurate colors, whereas 2-Shots produces the result closest to the ground truth. Our model partially compensates for the black-level error---the strong green tint visible in the central region of PG and PGRQ is noticeably reduced---but the dominant fixed-pattern component of the LRID dark shading (see \cref{sec:supp_ds_discussion,sec:supp_ds_synthetic}) leads to residual color inaccuracies, illustrating a current limitation of our approach.}
\label{fig:supp_lrid_fail}
\end{figure*}

We further evaluate our method and the baselines on higher-ratio noise settings of the LRID dataset, including $512\times$ and $1024\times$, as reported in \cref{tab:supp_lrid_psnrssim}. We additionally introduce a variant of our model, denoted by $\text{Ours}^{*}$ in the table, in which the black-level error is predicted independently for each image in the test set sharing the same ISO and exposure time, and the resulting predictions are averaged to correct all corresponding noisy inputs. This aggregation stabilizes the estimate when individual images lack sufficient visual content for accurate per-image prediction, and is consistent with the observation in~\cite{cao2023physics} that dark shading noise is approximately constant at a fixed ISO and exposure time. The results in \cref{tab:supp_lrid_psnrssim} confirm that averaging consistently improves the performance of our model; however, it no longer satisfies the blind condition.

\textit{Ratios $128\times$ and $256\times$.} As discussed in \cref{sec:supp_ds_discussion,sec:supp_ds_synthetic}, these ratios do not exhibit a severe fixed-pattern noise component. In this regime, our model achieves performance close to that of PG and PGRQ, which demonstrates its robustness even in the absence of significant black-level error, while consistently outperforming PGRQB.

\textit{Ratios $512\times$ and $1024\times$.} Our model surpasses PGRQB at $512\times$ but falls behind at $1024\times$, and the gap to PGRQ widens. In these settings, the dark shading noise is amplified more strongly and becomes visibly dominant in the noisy input. Since our model is designed to estimate the black-level error and does not explicitly address fixed-pattern noise, which manifests as a local color distortion, a strong fixed-pattern component can mislead the BLE predictor by altering the dominant image statistics. In contrast, the qualitative results on the other datasets (\cref{fig:supp_results_eld_sony,fig:supp_results_eld_nikon}) indicate that PGRQB treats color-related errors as a local degradation, which can be advantageous when the fixed-pattern component dominates, as is the case for the LRID dataset.

\cref{fig:supp_lrid_fail} shows a representative $1024\times$ denoising example. PG and PGRQ clearly suffer from residual dark shading noise. PGRQB modifies the color but introduces local color inconsistencies. ELD~\cite{wei2021physics}, despite applying dark shading correction prior to denoising, fails to recover accurate colors, whereas 2-Shots~\cite{lu20252} produces the result closest to the ground truth. Our model partially compensates for the black-level error---the strong green tint observed in the central region of the PG and PGRQ outputs is noticeably reduced---but the dominant fixed-pattern component in the LRID dark shading noise, as discussed in \cref{sec:supp_ds_discussion,sec:supp_ds_synthetic}, leads to residual color inaccuracies. This case illustrates one of the current limitations of our approach.

\section{SIDD-CC Processing Details}
\label{sec:supp_siddcc}
\begin{figure}[t]
    \centering
    \subfloat[ISO 100, ${\sim}1\%$ initially clipped pixels.]{%
        \includegraphics[width=0.29\linewidth]{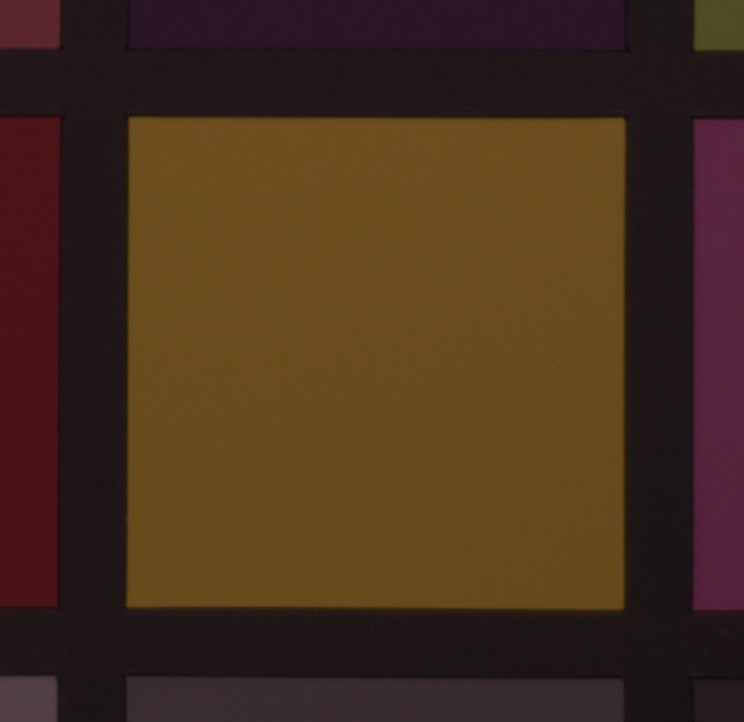}
        \label{fig:sidd_crop_100}
    }
    \hfill
    \subfloat[ISO 800, ${\sim}6\%$ initially clipped pixels.]{%
        \includegraphics[width=0.29\linewidth]{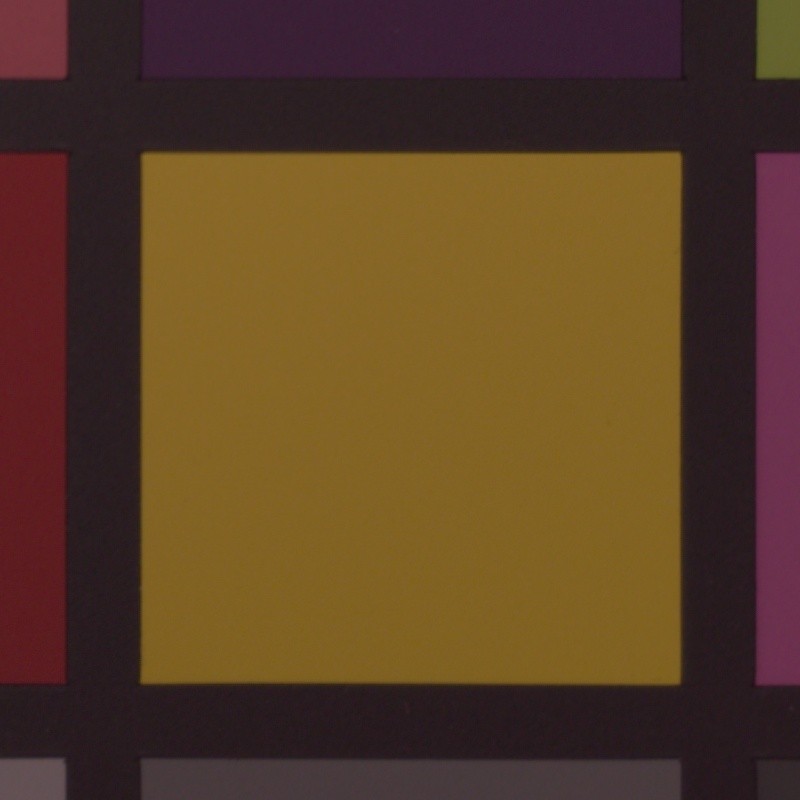}
        \label{fig:sidd_crop_800}
    }
    \hfill
    \subfloat[ISO 3200, ${\sim}47\%$ initially clipped pixels.]{%
        \includegraphics[width=0.29\linewidth]{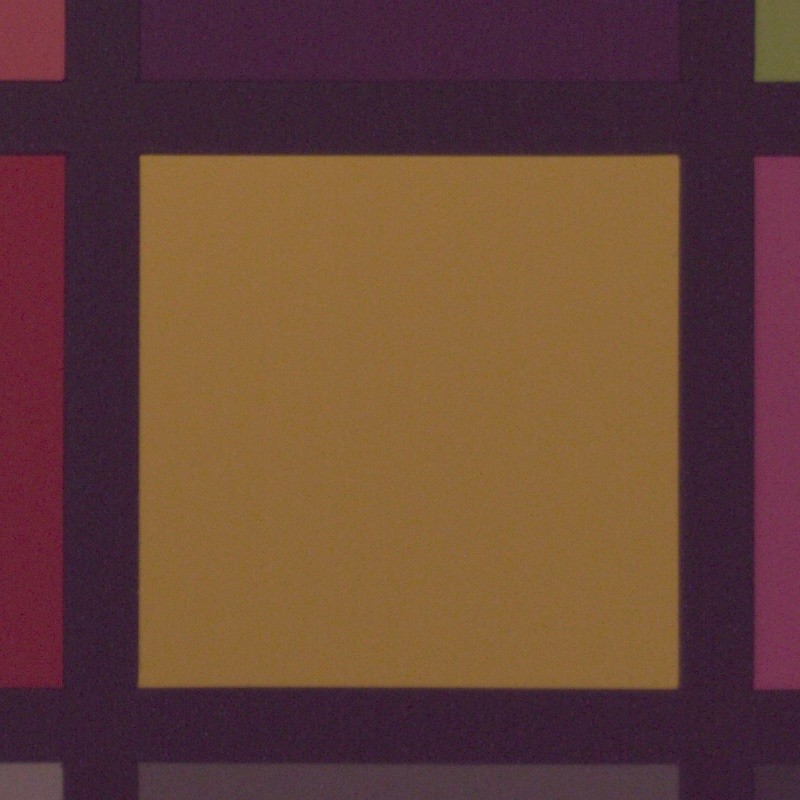}
        \label{fig:sidd_crop_3200}
    }
    \caption{\textbf{Effect of pixel clipping on SIDD ground-truth quality.} Patches cropped from SIDD ground-truth images of a scene containing a color checker, captured at three different ISO levels. As the fraction of initially clipped pixels increases, dark regions develop a pronounced purple tint due to the biased mean estimation. Below our $1\%$ threshold (a), the bias is negligible.}
    \label{fig:supp_sidd_crops}
\end{figure}

\begin{figure}[t]
    \centering
    \includegraphics[width=\linewidth]{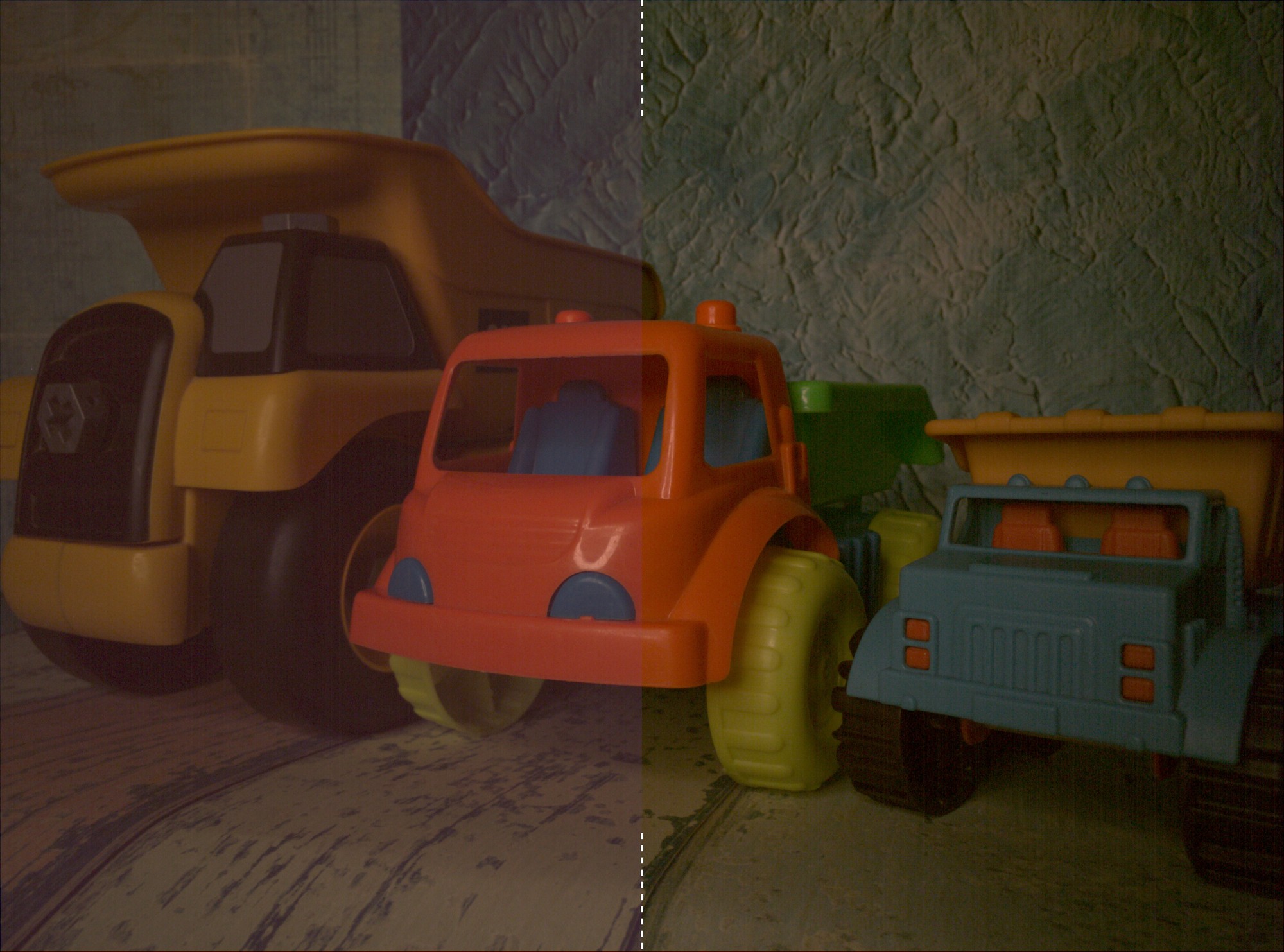}
    \caption{\textbf{Side-by-side comparison of SIDD (left) and \newSIDD{} (right) ground truth on the same scene.} The biased mean estimation in SIDD produces a purple tint that is most pronounced in dark regions but also visibly affects brighter areas such as the scene background.}
        \label{fig:supp_sidcc}
\end{figure}

SIDD~\cite{abdelhamed2018high} is a widely used smartphone denoising dataset with images from five devices (iPhone 7, Google Pixel, Samsung Galaxy S6 Edge, Motorola Nexus 6, LG G4) across 10 scenes. 
Each scene is captured under four different settings, yielding 200 configurations (160 train / 40 validation) with 150 noisy images per setting. 
Ground truth is generated by applying black-level subtraction (BLS) to each noisy image and clipping to $[0,1]$, performing dense local alignment across captures, and estimating the mean image by fitting a cumulative distribution function.

\subsection{Source of Bias}
\label{sec:supp_siddcc_source_of_bias}
In low-light conditions, the noise-to-signal ratio is substantially higher. 
Clipping individual noisy frames to $[0,1]$ after BLS biases the GT estimate: for dark pixels whose true signal lies just above the black-level offset, noisy observations may fall within $[0,\text{black-level}]$. %
While BLS alone is not the issue, the subsequent clipping truncates a large portion of these values, distorting their distribution and overestimating the mean. 
After rendering, this manifests as a purple tone in dark regions, due to unequal amplification of color channels during white balance (see Fig.~\ref{fig:teaser}).

To address this, we propose a more statistically consistent procedure: compute the mean across aligned noisy frames first, then apply BLS and clipping. 
As shown in \cref{fig:sidd_fix}, the original ordering (BLS $+$ clip $\rightarrow$ normalize $\rightarrow$ process $\rightarrow$ average) introduces chromatic bias, whereas our revised pipeline (normalize $\rightarrow$  process $\rightarrow$ average $\rightarrow$ BLS $+$ clip) maintains color accuracy. 
This is particularly relevant when evaluating models trained on external real data or alternative synthetic noise models, as such models do not learn the bias present in the original SIDD ground truth.

\cref{fig:supp_sidcc} provides a further comparison between the SIDD ground truth (left) and the \newSIDD{} ground truth (right) under the same setting. While the purple tint in SIDD is most pronounced in dark regions, as shown in \cref{fig:teaser}, the biased mean estimation also extends to brighter areas such as the scene background, where it introduces noticeable color errors.

\subsection{New Training and Validation Split}
Our revised pipeline resolves the bias for four devices (iPhone 7, Google Pixel, Motorola Nexus 6, LG G4), where fewer than $1\%$ of pixels per scene are truncated during processing. 
However, the Samsung Galaxy S6 Edge exhibits residual color bias because its recorded black-level in the metadata is zero, failing to compensate for the true sensor offset, which leaves many pixels initially truncated. 
We mitigate this by retaining only S6 Edge scenes with less than $1\%$ initially truncated pixels, consistent with the other devices. %

We note that, of the 40 scenes in the SIDD validation split, only five are included in our dataset. Reconstructing ground truth under our pipeline requires the full stack of 150 unprocessed RAW captures per scene; however, upon contacting the corresponding author of SIDD, we found that this raw data remains available for only five scenes, while the rest have been lost.

Maintaining the original ratio of the training and validation subsets, we obtain a split of 115 training and 29 validation scenes. Following the original protocol, we extract 32 patches per validation scene (928 total). 
We refer to this dataset as \textbf{SIDD-Color Corrected (\newSIDD{})}.

\subsection{Samsung Galaxy S6 Edge}
Unlike the other sensors in SIDD, the Samsung Galaxy S6 Edge reports a black-level of $0$ in its metadata. In low-light conditions, many pixels in dark regions fall below the true black-level of the sensor; since negative values are not permitted in RAW images, a reported black-level of $0$ causes these dark pixels to be clipped. As explained in \cref{sec:supp_siddcc_source_of_bias}, clipping dark pixels biases the estimated intensity at those locations. We therefore exclude all S6 Edge scenes with more than $1\%$ of initially clipped pixels in \raw{} space (i.e., pixels equal to $0$). This threshold is chosen empirically; below, we examine how the fraction of initially clipped pixels affects the extracted ground truth.

\cref{fig:supp_sidd_crops} shows patches cropped from three SIDD ground-truth images of a scene containing a color checker. The image with a high fraction of initially clipped pixels (\cref{fig:sidd_crop_3200}) exhibits a pronounced purple tint in dark regions, whereas those with a low fraction (\cref{fig:sidd_crop_100,fig:sidd_crop_800}) preserve accurate dark colors, indicating that the bias is negligible below our threshold.

\subsection{Benchmarking}
We evaluate state-of-the-art methods on \newSIDD{} for denoising in both \raw{} and sRGB domains. In sRGB, we consider BM3D~\cite{bm3d} (non-learning baseline), AT-BSN~\cite{atbsn} (self-supervised), and NAFNet~\cite{chen2022simple} (supervised on \newSIDD{}). AT-BSN and NAFNet are retrained on the new split with default settings; results are in \cref{tab:siddcc_results_rgb}. In \raw{}, we again use BM3D as a non-learning baseline, retrain B2U~\cite{wang2022blind2unblind} (self-supervised) on \newSIDD{}, and train NAFNet adapted to 4-channel \raw{} inputs on the new split. We also evaluate YOND~\cite{feng2025yond} without retraining, as it is originally trained on external datasets. Results are in \cref{tab:rgb_raw}.

\section{Memory and Runtime Efficiency}
\label{sec:supp_efficiency}
\begin{table}[t]
\centering
\scriptsize
\renewcommand{\arraystretch}{0.9}
\setlength{\tabcolsep}{3pt}

\begin{tabular*}{\linewidth}{@{\extracolsep{\fill}}lcccc}
\toprule
& \multicolumn{3}{c}{Ours} & PGRQB \\
\cmidrule(lr){2-4}\cmidrule(lr){5-5}
& (a) & (b) & (c) & (d) \\
\midrule
BLBE input $\downarrow 4$       & \xmark & \cmark & \cmark & -- \\
Denoiser res.                   & Full & Full & $512^2$ & $512^2$ \\
\midrule
PSNR (dB) $\uparrow$            & $42.82$ & $42.85$ & $42.69$ & $39.93$ \\
Mem.\ (GB) $\downarrow$         & $2.0$+$4.4$ & $0.3$+$4.4$ & $0.3$+$0.4$ & $0.4$ \\
Time (ms) $\downarrow$          & $16.8$+$46.1$ & $1.1$+$46.1$ & $1.1$+$91.8$ & $91.8$ \\
\bottomrule
\end{tabular*}

\caption{\textbf{Memory and runtime trade-offs} on ELD-Nikon~\cite{wei2021physics} $\times200$, reported as (BLE predictor)\,+\,(denoiser). ``BLBE input $\downarrow4$'' downsamples the BLBE input by a factor of four; the denoiser runs on the full image or tiled at $512^2$. Downsampling the BLBE input cuts its cost by an order of magnitude with no loss in accuracy, and tiling the denoiser trades runtime for memory. Combined (c), our method reaches a footprint comparable to the patch-based PGRQB (d) while keeping a $\sim$2.8\,dB PSNR advantage.}
\label{tab:supp_memory}
\end{table}

Both the denoiser and the BLBE module benefit from larger inputs (\cref{tab:ablations_input}), but to very different degrees: the denoiser gains only marginally from additional spatial context, whereas BLBE relies on it to estimate an accurate global black-level error (BLE). The denoiser can therefore run on small patches with little loss in quality, while BLBE requires larger inputs and raises the peak memory footprint relative to a fully patch-based pipeline. This cost can be reduced substantially in two complementary ways, with no significant loss in accuracy (\cref{tab:supp_memory}).

First, while BLBE requires diverse visual information from the input to estimate the BLE accurately, this information is preserved at lower input resolutions, so the BLBE input can be substantially downsampled. Reducing the input resolution by a factor of four lowers BLBE runtime from $16.8$\,ms to $1.1$\,ms and its memory from $2.0$\,GB to $0.3$\,GB, while PSNR is unaffected ($42.82 \rightarrow 42.85$\,dB). Second, the denoiser itself can be run on patches: tiling the input at $512^2$ reduces denoiser memory by roughly an order of magnitude ($4.4 \rightarrow 0.4$\,GB) at the cost of approximately doubling its runtime ($46.1 \rightarrow 91.8$\,ms), with only a marginal drop in PSNR ($42.69$\,dB).

Applied together, the two strategies reduce total peak memory from $6.4$\,GB to $0.7$\,GB, comparable to the patch-based PGRQB baseline ($0.4$\,GB), while retaining a $\sim2.8$\,dB PSNR advantage over it ($42.69$ vs.\ $39.93$\,dB). The two options are independent, offering flexibility under either runtime- or memory-constrained deployment.

\section{Additional Real Experiments}
\begin{figure*}[t]
\centering
\subfloat[Noisy (ISO 1600, $1/400$s)]{%
    \includegraphics[width=0.24\textwidth]{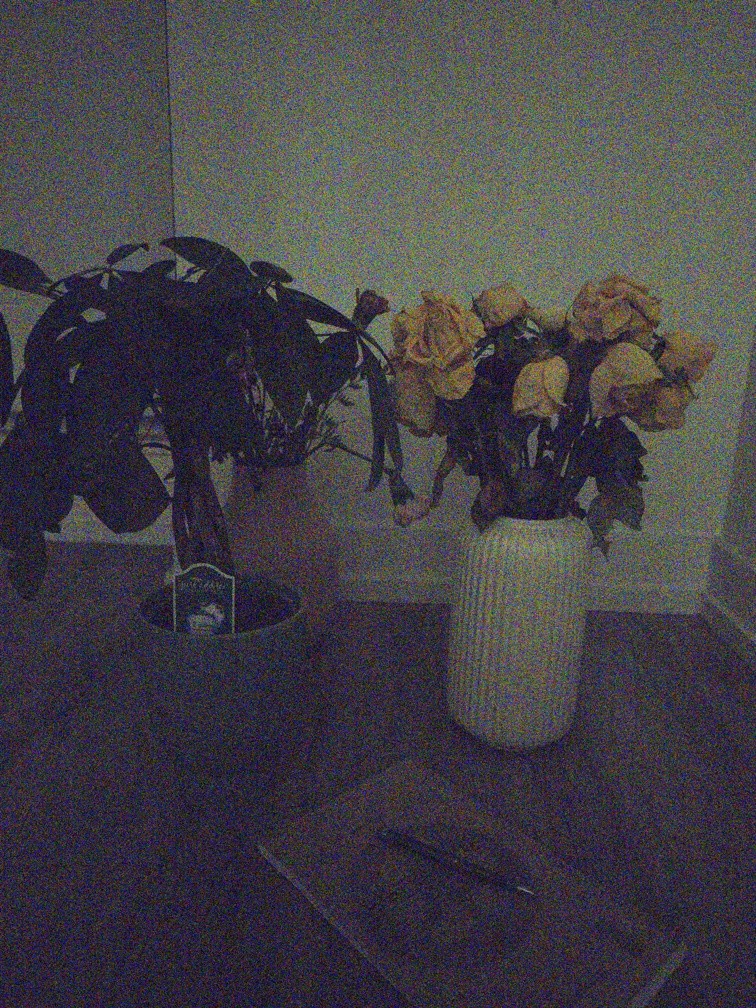}
}
\subfloat[PGRQ]{%
    \includegraphics[width=0.24\textwidth]{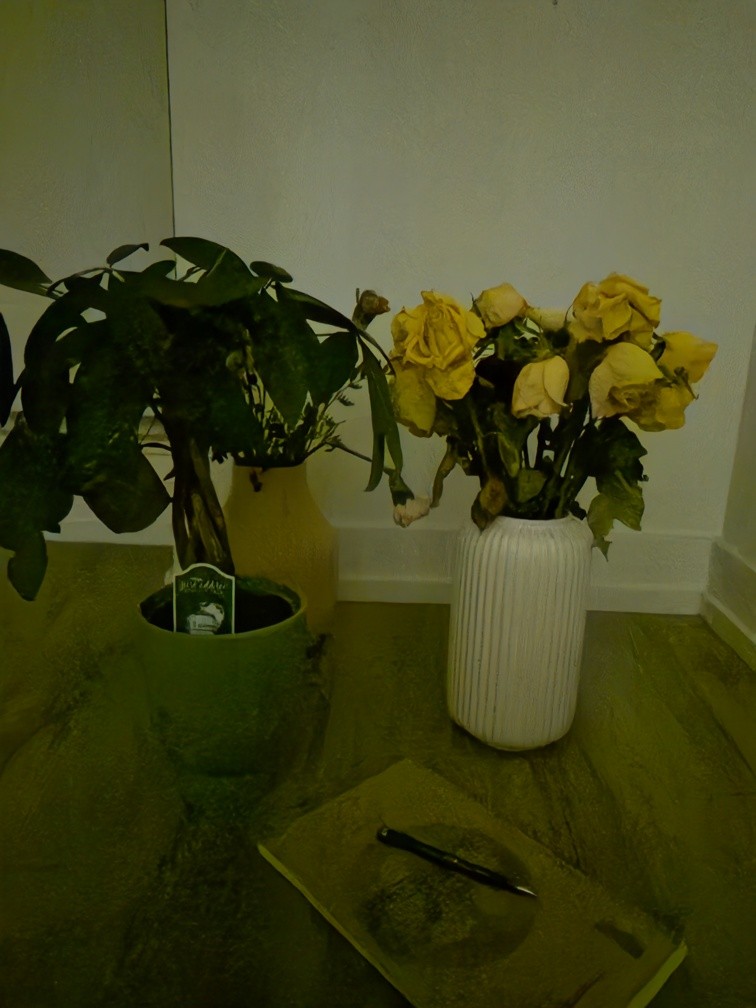}
}
\subfloat[PGRQB]{%
    \includegraphics[width=0.24\textwidth]{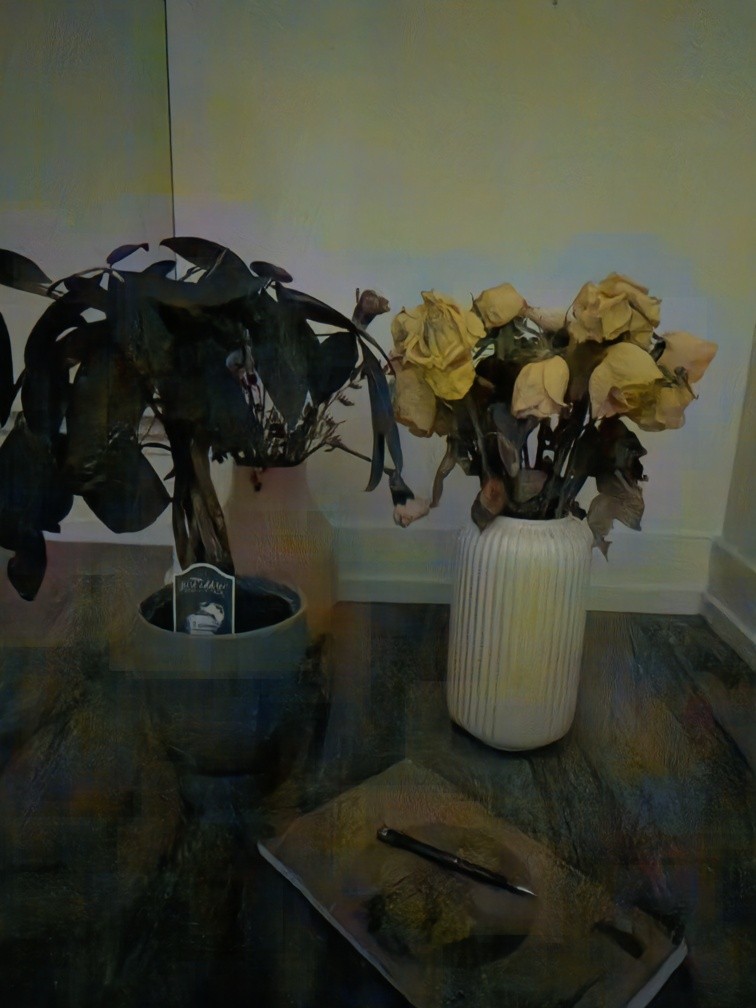}
}
\subfloat[Ours]{%
    \includegraphics[width=0.24\textwidth]{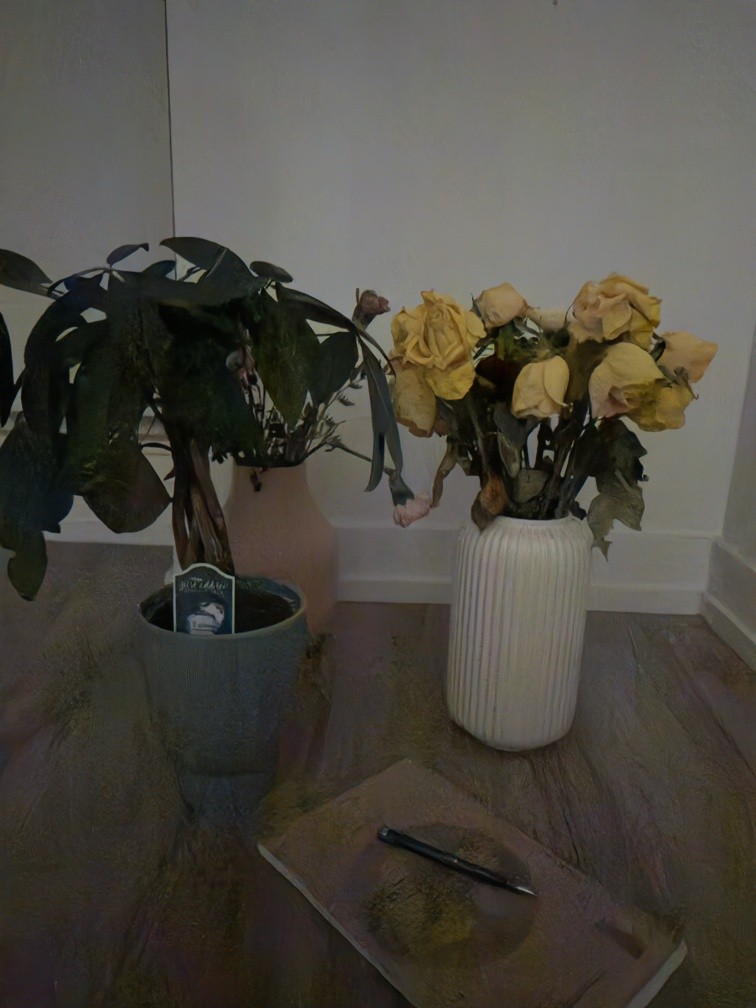}
} \\
\subfloat[Noisy (ISO 800, $1/200$s)]{%
    \includegraphics[width=0.24\textwidth]{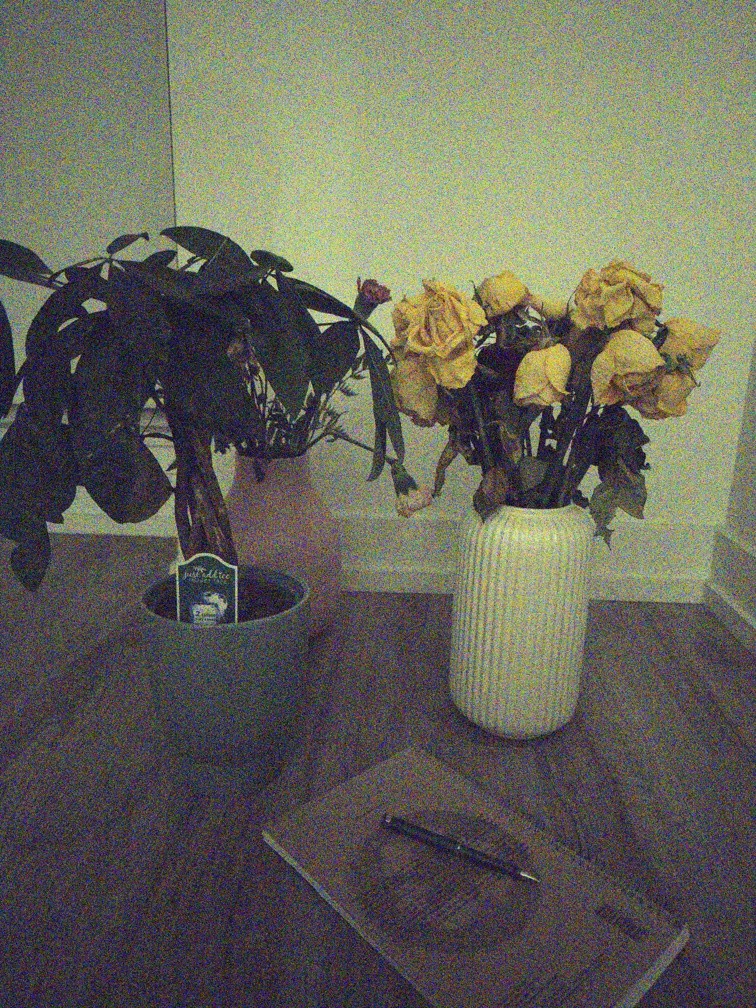}
}
\subfloat[PGRQ]{%
    \includegraphics[width=0.24\textwidth]{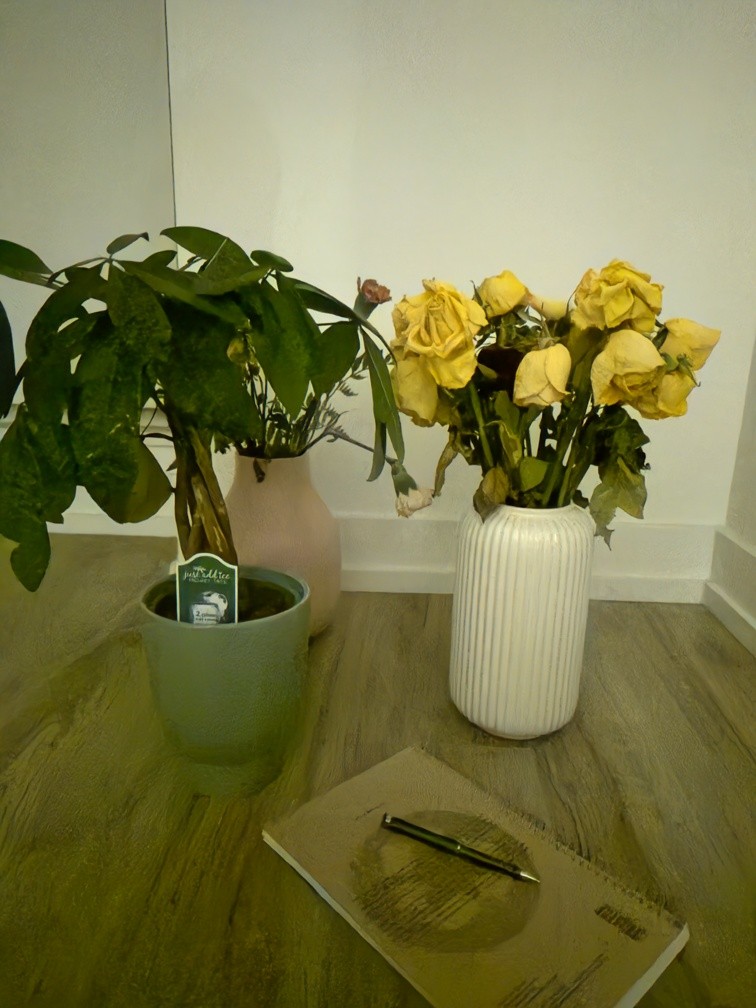}
}
\subfloat[PGRQB]{%
    \includegraphics[width=0.24\textwidth]{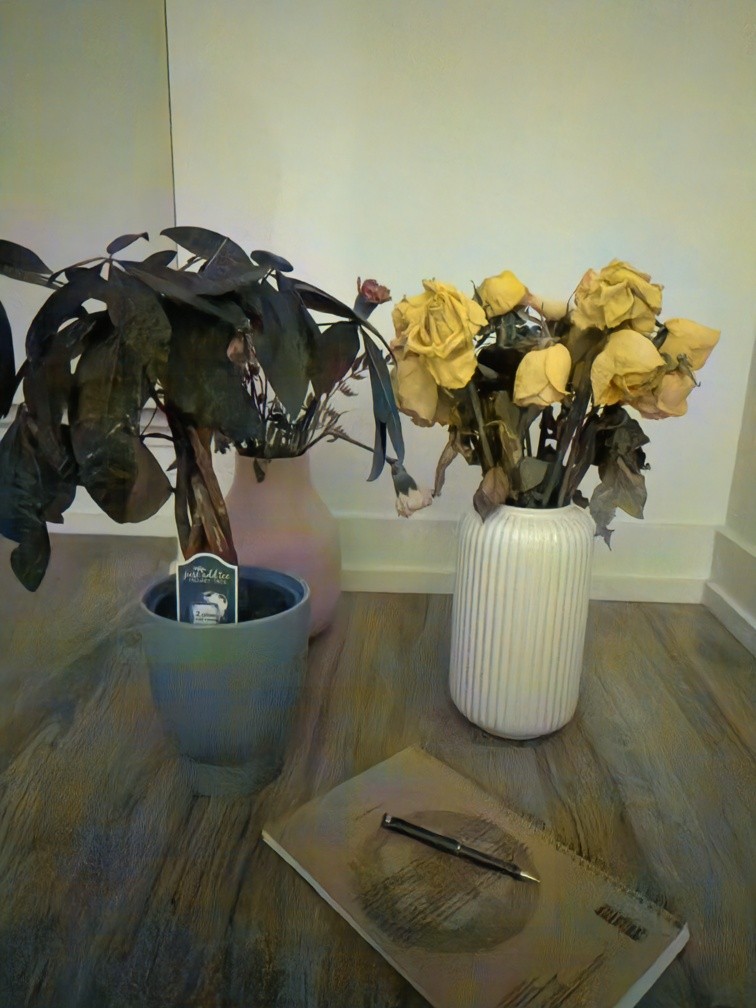}
}
\subfloat[Ours]{%
    \includegraphics[width=0.24\textwidth]{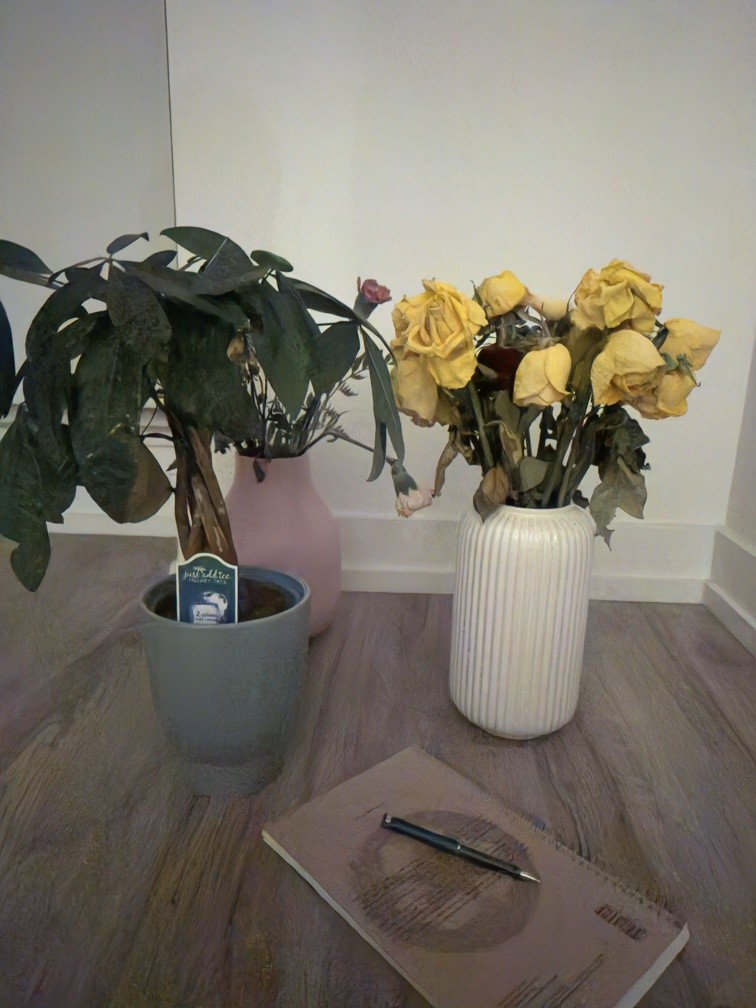}
} \\
\subfloat[Noisy (ISO 400, $1/100$s)]{%
    \includegraphics[width=0.24\textwidth]{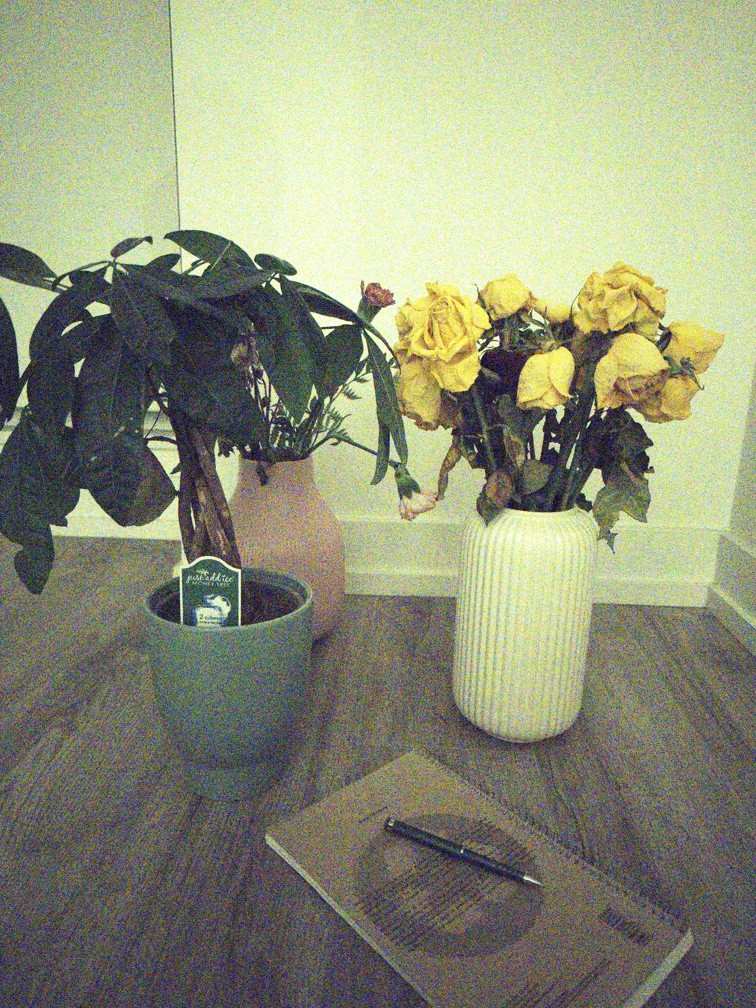}
}
\subfloat[PGRQ]{%
    \includegraphics[width=0.24\textwidth]{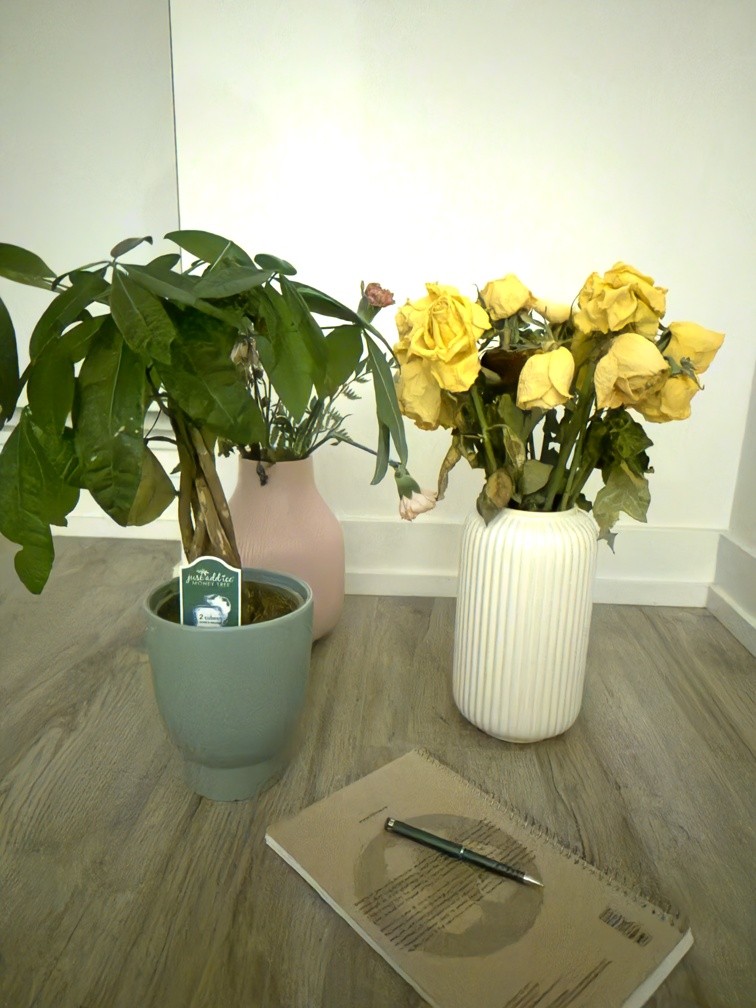}
}
\subfloat[PGRQB]{%
    \includegraphics[width=0.24\textwidth]{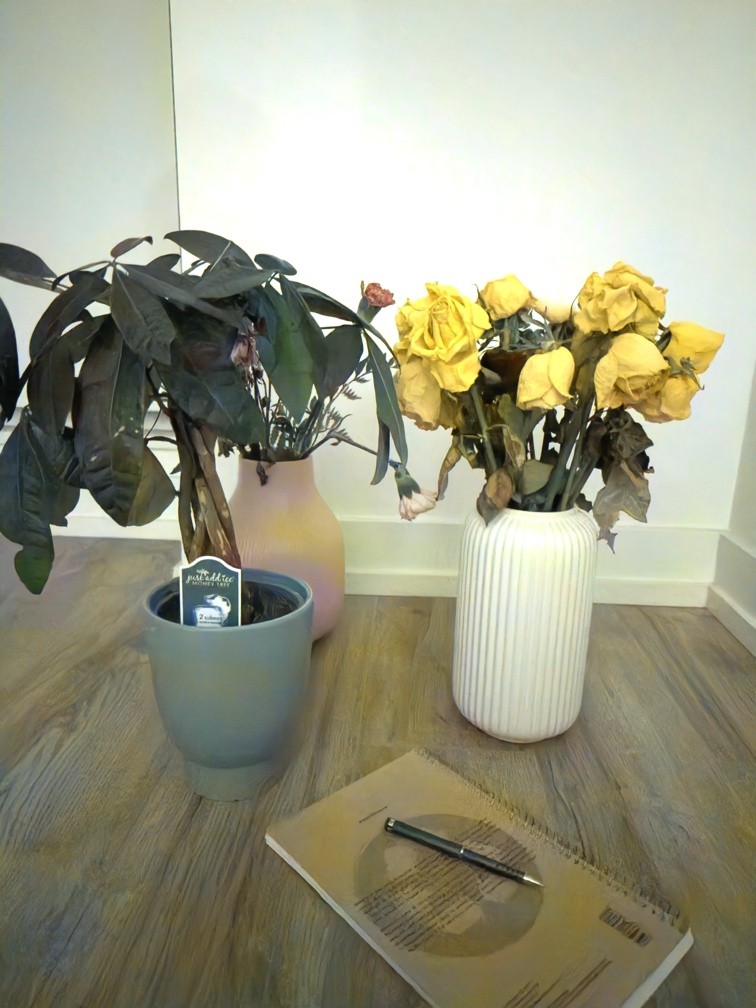}
}
\subfloat[Ours]{%
    \includegraphics[width=0.24\textwidth]{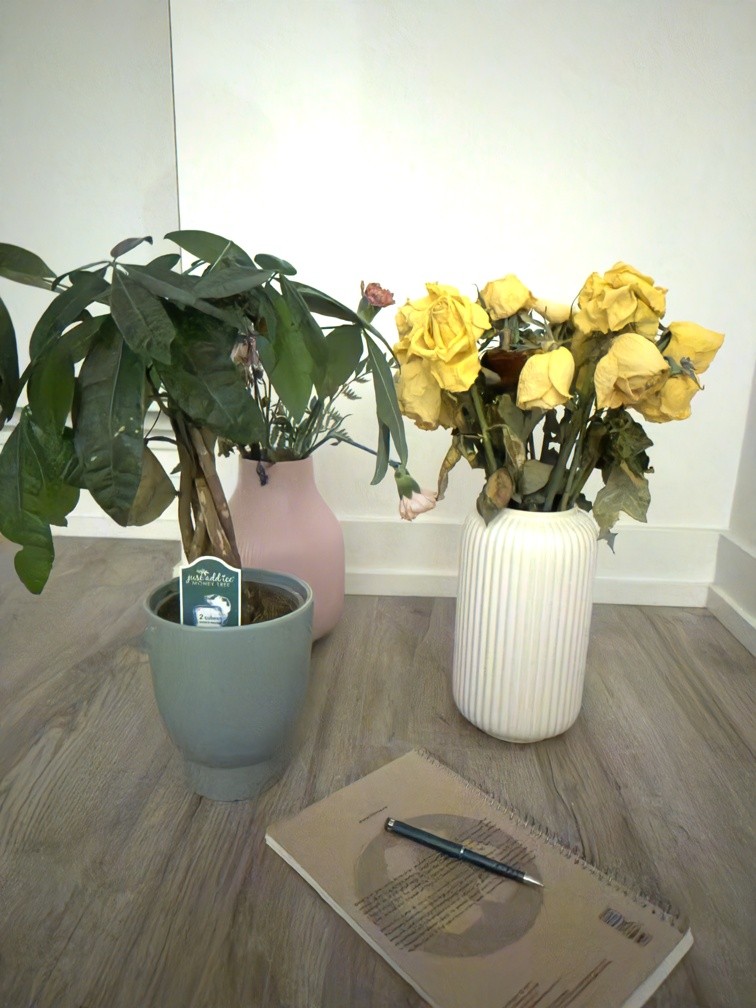}
} \\

\caption{\textbf{Qualitative results on real-world low-light captures from an iPhone 16 Pro.} All images are captured under the same lighting condition, and noisy inputs are amplified by $100\times$ for visualization. As ISO increases, dark shading noise becomes increasingly dominant. PGRQ exhibits a yellowish tint caused by uncompensated black-level error, while PGRQB introduces local color artifacts, most notably in the ISO 1600 case. Our method consistently recovers accurate colors across all examples.}
\label{fig:real_result}

\end{figure*}

We further evaluate our model on images captured in the wild with an iPhone 16 Pro under low-light conditions and varying ISO levels (\cref{fig:real_result}). Each noisy input is amplified by $100\times$ for visualization. PGRQ exhibits a yellow tint that becomes more pronounced at higher ISO levels due to severe black-level error, and PGRQB introduces local color distortions, whereas our model successfully recovers accurate colors by predicting the black-level error.

\section{Extending to Fixed-Pattern Noise}
\label{sec:supp_fpn}
\begin{table}[t]
\centering
\renewcommand{\arraystretch}{1.1}
\setlength{\tabcolsep}{6pt}

\begin{tabular*}{\linewidth}{@{\extracolsep{\fill}}lcc}
\toprule
& PGRQB & Ours \\
\midrule
w/o FPN subtraction & $33.75$ & $35.23$ \\
w/\ FPN subtraction & $35.12$ & $\mathbf{37.20}$ \\
\midrule
$\Delta$ & $+1.37$ & $+1.97$ \\
\bottomrule
\end{tabular*}

\caption{\textbf{Effect of fixed-pattern noise (FPN) subtraction} on SID~\cite{chen2018learning} $\times300$ (PSNR in dB, $\uparrow$). Subtracting the calibrated fixed-pattern component prior to denoising improves both PGRQB and our method, showing that explicit FPN handling is complementary to black-level correction and offers a promising direction for further gains.}

\label{tab:supp_fpn}
\end{table}

Dark shading noise comprises two components: the black-level error (BLE), which acts as a global per-channel offset on the RGBG channels of the RAW image, and the fixed-pattern noise (FPN), which manifests as local distortions without a known structure (\cref{sec:supp_dark_shading_analysis}). Our method explicitly estimates and corrects the global BLE, but does not model the local FPN component. As shown in \cref{fig:supp_synthetic_ds,fig:supp_lrid_fail}, when the dark shading noise is severe the fixed-pattern residual can remain visible in the output, and when the fixed-pattern component is dominant it may be interpreted as scene content, which in turn affects the BLE estimate. A natural extension of our framework is therefore to handle FPN explicitly.

One option is to place a learned FPN-estimation module before the BLBE predictor, so that the FPN-suppressed image provides a cleaner input from which the global black-level offset can be estimated more reliably. As a proof of concept, on SID~\cite{chen2018learning} $\times300$, where the calibrated fixed-pattern noise is available, subtracting it prior to denoising improves both our method and PGRQB (\cref{tab:supp_fpn}), which shows that explicit FPN handling complements our black-level correction and offers substantial headroom for further gains. Alternatively, FPN estimation can be deferred to after denoising: because the fixed pattern is temporally stable, it can be recovered from multiple frames of the same sensor. A complementary direction is to separate the fixed pattern from scene content in a zero-shot manner, potentially through semantic knowledge of the scene. Together, these directions would extend the method toward fully addressing dark-shading noise in the blind setting.

\section{Limitations and Future Work}
\label{sec:supp_limitations}

\begin{figure*}[t]
\centering

\subfloat[Noisy]{%
    \includegraphics[width=0.235\textwidth]{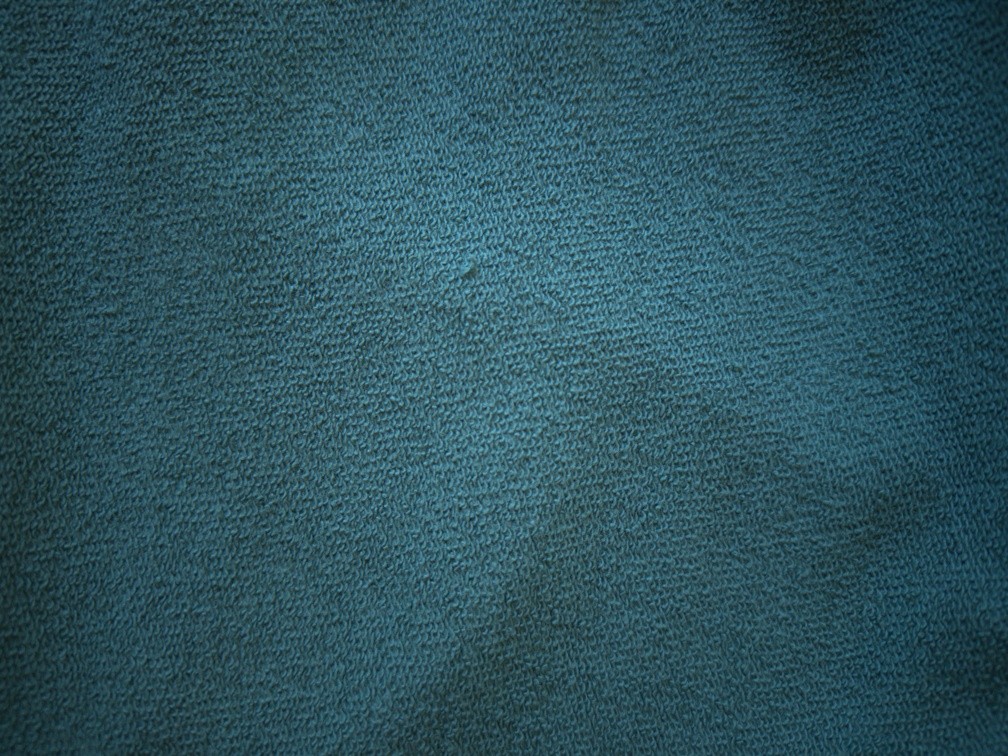}
    \label{fig:failure_clean}
}
\subfloat[PGRQ]{%
    \includegraphics[width=0.235\textwidth]{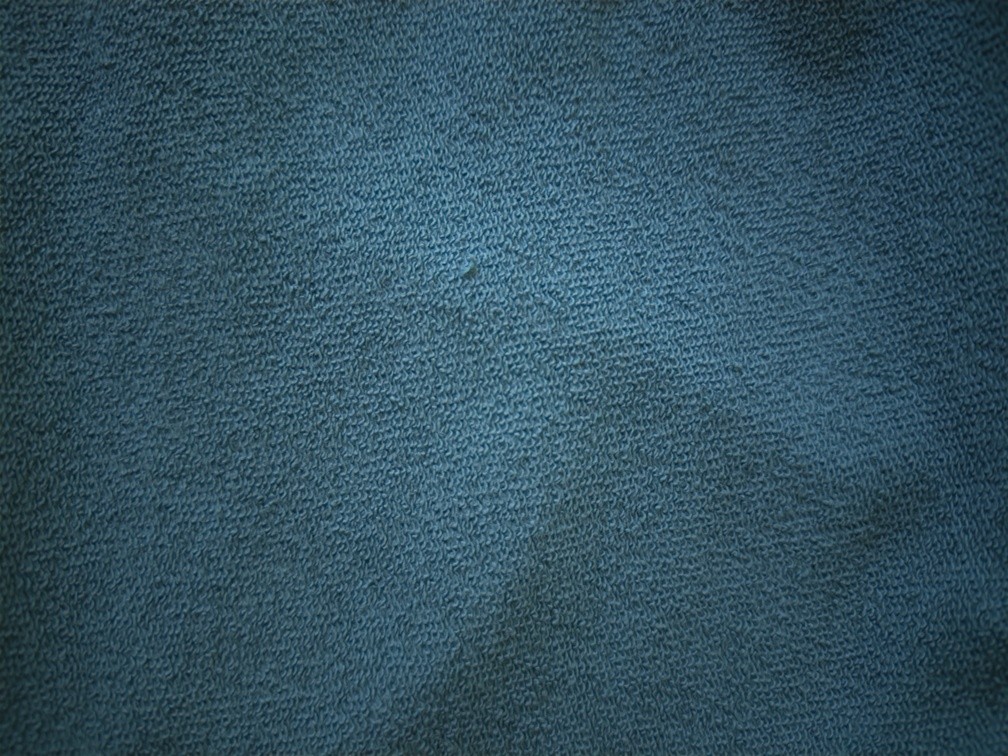}
    \label{fig:failure_pgrq}
}
\subfloat[PGRQB]{%
    \includegraphics[width=0.235\textwidth]{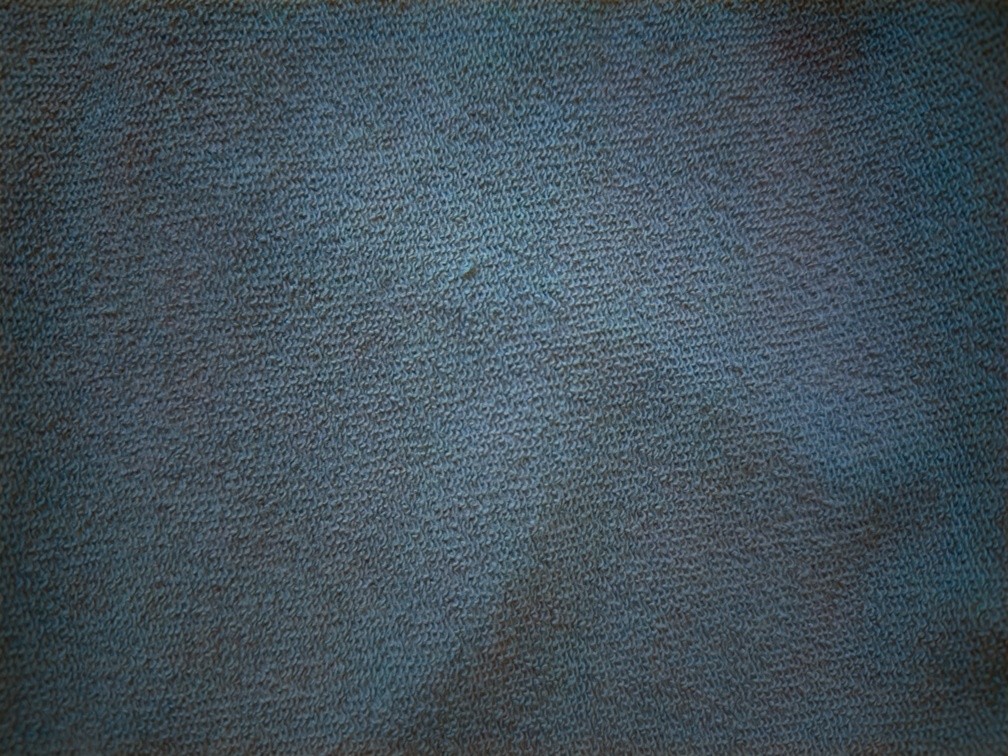}
    \label{fig:failure_pgrqb}
}
\subfloat[Ours]{%
    \includegraphics[width=0.235\textwidth]{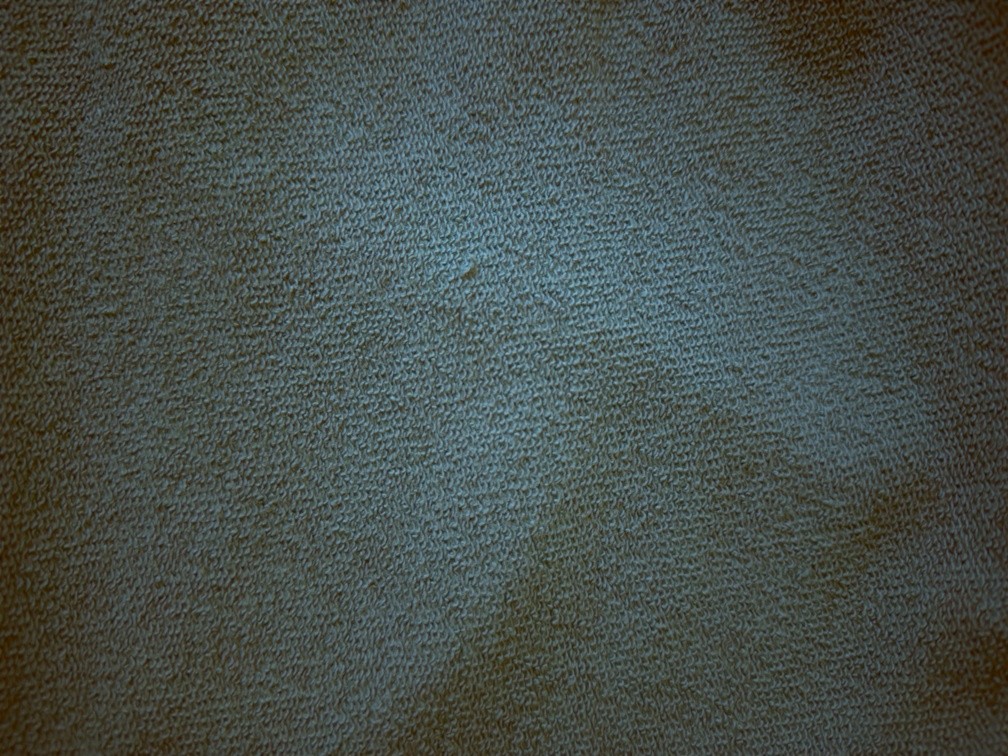}
    \label{fig:failure_ours}
}\\

\caption{\textbf{Failure case on a well-exposed capture from an iPhone 16 Pro.} The scene is captured under normal lighting with a low ISO setting, so the input is essentially noise-free and free of black-level error. PGRQ preserves the true colors. PGRQB introduces local color distortions, while our model predicts a spurious black-level error because the image content does not provide a sufficient signal for accurate BLE estimation.}
\label{fig:supp_failure}

\end{figure*}

The main limitation of our method concerns the dependence of the BLE prediction on the visual content of the input image. Our predictor requires sufficient scene diversity to estimate a reliable black-level offset. When the input is dominated by a uniform region with limited visual content, the predictor can produce a spurious BLE estimate that introduces a color shift in the denoised output. \cref{fig:supp_failure} illustrates such a case, where a normal-light image captured at low ISO with an iPhone 16 Pro, containing no actual black-level error, exhibits a color shift after denoising. A practical remedy is to obtain a small set of additional images from the target sensor across diverse scenes, which can be used to stabilize the BLE estimate. This requires access to the target sensor but does not impose any constraints on the captured content, making it accessible to end users, in contrast to dark frame or sensor calibration procedures that require dedicated equipment and technical expertise. We also note that our method does not explicitly model the fixed-pattern component of dark shading noise; we discuss this together with a proof-of-concept extension in \cref{sec:supp_fpn}.

\end{document}